\documentclass[10pt,journal,compsoc]{IEEEtran}
\usepackage{graphicx}
\usepackage{amsmath,amssymb} 
\usepackage{color}
\usepackage{todonotes}

\usepackage{algpseudocode}  
\usepackage{multirow}
\usepackage{booktabs}
\usepackage{subcaption}
\usepackage{hyperref}
\usepackage{amsmath}
\usepackage{cases}
\usepackage{makecell}
\usepackage{pifont}
\usepackage{xcolor}
\usepackage{rotating}
\usepackage{enumitem} 
\usepackage{utfsym}
\usepackage{hhline}
\usepackage{colortbl}
\usepackage{nicematrix}
\usepackage{tikz}
\newcommand{\groupdoubleline}{\hhline{=:=:===:======:=}}

\usepackage[ruled,linesnumbered]{algorithm2e}

\ifCLASSOPTIONcompsoc
  \usepackage[nocompress]{cite}
\else
  \usepackage{cite}
\fi
\ifCLASSINFOpdf
\else
\fi
\begin{document}
%

\title{OpenAHBU: Open-sourced Automatic Human Behaviour Understanding Toolkit Through Facial Behaviour Analysis}
%
%
%
%

\author{Zijian Wu,
        Yupeng Huo,
        Cheng Luo,
        Yichen Huang,
        Junan Long,
        Haijun Lin,
        Lu Liu,
        Hatice Gunes,
        Bernard Ghanem,
        and Siyang Song$^*$
        

\IEEEcompsocitemizethanks{
\IEEEcompsocthanksitem Zijian Wu, Yichen Huang, and Haijun Lin are with
Hunan Normal University, Changsha, Hunan, China (e-mail: wuzijian@njust.edu.cn,
202670183783@hunnu.edu.cn, and linhaijun@hunnu.edu.cn, respectively).
Zijian Wu is also with Nanjing University of Science and Technology,
Nanjing, Jiangsu, China.
\IEEEcompsocthanksitem Yupeng Huo is with PAttern Learning and Mining (PALM) Lab, School of Computer Science and Engineering, Southeast University, Nanjing,
Jiangsu, China (e-mail: yupeng@seu.edu.cn).
\IEEEcompsocthanksitem Cheng Luo and Bernard Ghanem are with King Abdullah
University of Science and Technology (KAUST), Thuwal 23955-6900, Kingdom of
Saudi Arabia (e-mail: cheng.luo@kaust.edu.sa and bernard.ghanem@kaust.edu.sa,
respectively).
\IEEEcompsocthanksitem Junan Long is with School of Computer Science and Technology, Tongji University, Shanghai, China
(e-mail: junanlong@tongji.edu.cn).
\IEEEcompsocthanksitem Hatice Gunes is with the Affective Intelligence and
Robotics (AFAR) Lab, Department of Computer Science and Technology,
University of Cambridge, Cambridge, United Kingdom (e-mail: hg410@cam.ac.uk).
\IEEEcompsocthanksitem Lu Liu and Siyang Song are with the HBUG Lab,
Department of Computer Science, University of Exeter, Exeter, United Kingdom
(e-mail: L.Liu3@exeter.ac.uk and S.Song@exeter.ac.uk, respectively).
\textsuperscript{*}\textit{Corresponding author: Siyang Song.}
}

\thanks{Manuscript received November 19, 2026.}}

%
%

\markboth{Journal of \LaTeX\ Class Files,~Vol.~14, No.~8, August~2015}%
{Shell \MakeLowercase{\textit{et al.}}: Bare Advanced Demo of IEEEtran.cls for IEEE Computer Society Journals}
\IEEEtitleabstractindextext{%
\begin{abstract}

Automatic facial behaviour understanding (AFBU) is fundamental to various human-centred applications, yet existing academically available toolkits offer limited functionality, closed training pipelines, and weak generalisation. This paper presents OpenAHBU, a fully open-sourced, generalisable toolkit integrating 11 AFBU functions across three semantic levels: (i) short-term external facial attributes, including landmarks, blendshapes, Action Units, and expressions; (ii) short-term facial behaviour-related human psychological states, including valence/arousal and pain; and (iii) long-term facial behaviour-related psychological traits, including depression and personality. Particularly, we re-developed eight functions, where at least four deep learning models covering different computational complexities are developed for each function,  providing flexible deployment on resource-constrained and high-throughput computing scenarios. More importantly, models of most functions are trained on a comprehensive corpus aggregating multiple academically available datasets, ensuring robust cross-dataset generalisability.
In addition, we include a Synergistic Multi-task Enhancement (SMTE) module to efficiently and adaptively enhance each OpenAHBU function by leveraging the complementary cues contained in its most-related support function's predictions.
Extensive evaluations demonstrate that our OpenAHBU outperformed existing open-source toolkits not only in functional coverage but also the predictive performances, while providing superior deployment flexibility across diverse computational scenarios.
With fully publicly available training and inference scripts, pre-trained weights, user-friendly GUI, Python library, and cross-platform compatibility, 
our OpenAHBU provides a reproducible, scalable and extensible toolkit at \url{https://github.com/OpenAHBU/OpenAHBU}.

\end{abstract}

}

\maketitle

\IEEEdisplaynontitleabstractindextext

%
\IEEEpeerreviewmaketitle

\section{Introduction}
\label{sec:introduction}


\noindent \IEEEPARstart{H}{uman} facial behaviours provide rich spatio-temporal cues reflecting hierarchical human psychological and physiological status \cite{ellgring2007non}, which play an essential role in conveying intentions and messages during human-human interactions \cite{cohn2006foundations}. Thus, facial behaviour analysis are fundamental for various real-world human-centred applications such as emotion recognition \cite{toisoul2021estimation,kossaifi2019sewa}, pain state recognition \cite{egede2020emopain}, mental health assessment \cite{song2022spectral,wu2025depmgnn,jaiswal2019automatic}, personality recognition \cite{palmero2021context,song2022learning} and humanoid robot \cite{chen2023coupled,shangguan2025learning}.

To objectively and comprehensively describe human static facial expressions, psychologists have established various in-depth and practical strategies such as facial expressions of emotions theories \cite{ekman2000basic}  
and Facial Action Coding System (FACS) \cite{ekman1978facial}, 
While emotional facial expression theories defining only a few fundamental and universally recognised human facial displays (e.g., neutral, anger, disgust, fear, happiness, sadness, and surprise) expressed by people across diverse demographic backgrounds, the FACS defines a set of facial muscle movements called Action Units (AUs) to objectively characterise a wider range of plausible facial expressions. Given the high accessibility of human facial behaviours in daily life and their time-consuming/costly natures for manual annotation, establishing an automatic, objective, fast and comprehensive facial behaviour understanding toolkit is essential for further advancing various real-world human-centred applications.

Early automatic facial behaviour understanding (AFBU) methods primarily focus on designing hand-crafted visual descriptors to extract facial textural patterns (e.g., Histogram of Oriented Gradients (HOG) \cite{orrite2009hog} and Local Binary Patterns (LBP) \cite{zhao2007dynamic}) or geometrical patterns (e.g., Active Shape Models (ASM) \cite{cootes1995active} and Active Appearance Models (AAM) \cite{cootes2001active}), 
followed by traditional machine learning (ML) predictors (e.g., Support Vector Machines (SVM) and Hidden Markov Models (HMM)) to assess different observable facial status and their related human internal psychological states. 
While these hand-crafted methods are limited by complexity and subjectivity of manual design,  
recent deep learning (DL)-based AFBU solutions can automatically learn task-specific but human uninterpretable cues 
from the given face images/videos through task-oriented optimised networks, making them not only experience substantial performance improvements compared to hand-crafted competitors in handling almost all AFBU tasks. 

\begin{table*}[htbp]
\centering
\small
\setlength{\doublerulesep}{0.3em}
\renewcommand{\arraystretch}{1.3}
\caption{Comparison of automatic facial behaviour understanding tools, where AFFEDX 2.0 \cite{bishay2023affdex}, FACET, and FaceReader are commercial AFBU platforms without open-sourced code and developed by private training data. \textbf{Bbox}: Face Bounding Box. The meaning of FLD, FBSE, SFER, DFER, FAUD, FAUE, VAIE, FPLE, FDA, FAPR, and FRPR are defined in Sec. 2.1. Problems P(1)-P(4) are defined in Introduction. Comcl denotes commercial toolkit.}
\resizebox{1.0\linewidth}{!}{
\begin{tabular}{ll|cccccccccccc|cccc}
\Xhline{1.5px}
\rowcolor{gray!30} 
\multicolumn{2}{c|}{\cellcolor{gray!30}} &
  \multicolumn{12}{c|}{\cellcolor{gray!30}\textbf{Facial Behaviour Understanding Functions}} &
  \multicolumn{4}{c}{\cellcolor{gray!30}\textbf{Problems}} \\ \hhline{~~||------------||----}
\rowcolor{gray!30}
\multicolumn{2}{c|}{\multirow{-2}{*}{\cellcolor{gray!30}\textbf{Toolkit}}}    
& \textbf{Bbox} 
& \textbf{FLD} 
& \textbf{FBSE} 
& \textbf{SFER} 
& \textbf{DFER} 
& \textbf{FAUD} 
& \textbf{FAUE} 
& \textbf{VAIE} 
& \textbf{FPLE}
& \textbf{FDA} 
& \textbf{FAPR} 
& \textbf{FRPR} 
& \textbf{P(1)}
& \textbf{P(2)}
& \textbf{P(3)}
& \textbf{P(4)}\\ 
\hhline{==================}
& AFFEDX 2.0 \cite{bishay2023affdex}       
& \ding{51} 
& \ding{51} 
&
& \ding{51} 
&      
& \ding{51}    
&      
&     
&     
&    
&     
&     
& \ding{51} 
& -
& \ding{51} 
& -\\ 
& FACET         
& \ding{51} 
& \ding{51} 
&
& \ding{51} 
&      
& \ding{51}   
&      
&     
&     
&    
&     
&     
& \ding{51} 
& -
& \ding{51} 
& -\\ 
\multirow{-3}{*}{\rotatebox[origin=c]{90}{Comcl.}}
& FaceReader          
& \ding{51} 
& \ding{51} 
&
& \ding{51} 
&      
& \ding{51}   
&     
& \ding{51}   
&     
&    
&     
&     
& \ding{51} 
& -
& \ding{51} 
& -\\ 
\hhline{==================}
& OpenFace 2.0 \cite{baltruvsaitis2016openface}     
& \ding{51}
& \ding{51} 
&
& 
&  
& \ding{51}    
& \ding{51}   
&    
&  
&   
&    
&     
& \ding{55}          
& \ding{55}          
& \ding{51}          
& \ding{51} \\ 
& Py-Feat \cite{cheong2023py}   
& \ding{51}
& \ding{51}   
&
& \ding{51}
&    
& \ding{51}  
&    
&   
&   
&   
&   
&     
& \ding{51}          
& \ding{51}          
& \ding{51}          
& \ding{55} \\ 
& LibreFace \cite{chang2024libreface}    
&        
& \ding{51} 
&
& \ding{51}
& 
& \ding{51}   
& \ding{51}  
&    
&     
&  
&  
&     
& \ding{55}           
& \ding{51}          
& \ding{51}          
& \ding{51} \\ 
& LibreFace 2.0 \cite{guan2026libreface}    
&        
& \ding{51} 
&
& \ding{51}
& 
& \ding{51}   
& \ding{51}  
&    
&     
&  
&  
&     
& \ding{55}           
& \ding{51}          
& \ding{51}          
& \ding{51} \\ 
\multirow{-3}{*}{\rotatebox[origin=c]{90}{Academic}}
& Behaviour4All \cite{kollias2024behaviour4all}    
& \ding{51} 
& \ding{51}  
&
& \ding{51} 
&  
& \ding{51}   
& 
& \ding{51}   
&    
&   
&  
&     
& \ding{51}          
& \ding{51}         
& \ding{51}         
& \ding{51} \\ 
& OpenFace 3.0 \cite{hu2025openface30lightweightmultitask}    
& \ding{51}  
& \ding{51}  
&
& \ding{51}  
&  
& \ding{51}      
& 
&  
&   
&  
&   
&     
& \ding{55}         
& \ding{51}          
& \ding{51}          
& \ding{51}  \\
\hhline{~=================} \rowcolor{gray!10}
& \textbf{OpenAHBU}     
& \ding{51}  
& \ding{51}  
& \ding{51}  
& \ding{51} 
& \ding{51} 
& \ding{51}  
& \ding{51}  
& \ding{51}  
& \ding{51} 
& \ding{51} 
& \ding{51} 
& \ding{51}     
&  \ding{55}         
&  \ding{55}          
&  \ding{55}          
&  \ding{55} \\ 
\Xhline{1.5px}
\end{tabular}
}
\label{tab:compare}
\end{table*}

However, only a minority of existing AFBU methods release their codebases, which are further fragmented across not only various programming languages (e.g., Python \cite{zhang2023dual,luo2022learning,kowalski2017deep}, C++ \cite{zhang2016joint,cao2018vggface2}, and Matlab \cite{song2018human,jaiswal2016deep}), but also DL frameworks (e.g., PyTorch \cite{zhang2023dual,wang2024multi}, TensorFlow \cite{kowalski2017deep}, and Caffe \cite{zhang2016joint,cao2018vggface2}), where some repositories are even incomplete or non-executable. Such situations create substantial barriers preventing these methods' application, reproduction, and extension. Besides, these models were generally evaluated on a limited number of datasets (typically one to three task-oriented datasets), with each model being trained and tested on an individual dataset. For example, AU recognition models \cite{luo2022learning,ning2024representation,jin2021multi} typically use BP4D \cite{zhang2014bp4d}, DISFA \cite{mavadati2013disfa}, or Aff-Wild2 \cite{kollias2018aff}, raising concerns about overfitting and cross-dataset generalisability. Although single-task open-sourced AFBU benchmarks \cite{chen2021cross,zhao2023dfme,alharbawee2021benchmark,liao2024open} provide standardised and reproducible evaluation frameworks, they are not plug-and-play solutions and require expertise in DL, coding, and environment configuration. Additionally, each benchmark targets an isolated task, such as facial expression recognition \cite{chen2021cross}, AU detection \cite{alharbawee2021benchmark}, or personality computing \cite{liao2024open}, limiting its applicability to comprehensive multi-task AFBU scenarios.

Alternatively, several freely accessible, ready-to-use AFBU toolkits, including OpenFace \cite{zadeh2018openface,hu2025openface30lightweightmultitask}, Py-Feat \cite{cheong2023py}, LibreFace \cite{chang2024libreface}, LibreFace 2.0 \cite{guan2026libreface}, and Behaviour4All \cite{kollias2024behaviour4all}, have been released in the past decade. They typically either integrate multiple task-specific models \cite{zadeh2018openface,cheong2023py,chang2024libreface, guan2026libreface} or employ multi-task learning to jointly analyse multiple facial attributes \cite{kollias2024behaviour4all,hu2025openface30lightweightmultitask}. However, as summarised in Table \ref{tab:compare}, they remain limited by \textbf{following problems:} (1) unavailable training code, hindering their reproduction and extension \cite{cheong2023py,kollias2024behaviour4all}; (2) each model is trained on only one or a few datasets, restricting their real-world generalisability \cite{cheong2023py,chang2024libreface,kollias2024behaviour4all,hu2025openface30lightweightmultitask,guan2026libreface}; (3) all these toolkits only focus on recognise frame-level observable attributes, without modelling complex or long-term latent facial behaviour-related human states \cite{zadeh2018openface,cheong2023py,chang2024libreface,kollias2024behaviour4all,hu2025openface30lightweightmultitask,guan2026libreface}; and (4) suboptimal performance due to their deployed lightweight models \cite{zadeh2018openface,chang2024libreface,kollias2024behaviour4all,hu2025openface30lightweightmultitask,guan2026libreface}, with some still utilised  outdated traditional ML models \cite{zadeh2018openface,cheong2023py}.

In this paper, we propose a comprehensive, generalisable and fully open-sourced AFBU toolkit called OpenAHBU. As demonstrated in Table \ref{tab:compare}, it can jointly understand 11 types of human facial behaviours or their related psychological states at three different levels: (i)  short-term observable facial behaviours including facial landmarks, blendshapes, AU occurrences, AU intensities, static facial expressions and dynamic facial expressions; (ii) short-term facial behaviour-related human psychological states including valence/arousal intensities and pain states; as well as (iii) long-term facial behaviour-related human psychological states including depression severity, apparent and real personality traits (\textbf{addressing Problem (3)}). Importantly, most AFBU functions include at least four varying-scale DL models, including lightweight models prioritising memory-efficiency/inference speed and large models prioritising effectiveness (\textbf{addressing Problem (4)}), where each is developed on a combination of multiple function-oriented datasets (\textbf{addressing Problem (2)}). Their training scripts, model architectures, pre-trained weights and other implementation details are fully open-sourced (\textbf{addressing Problem (1)}).
Furthermore, our OpenAUBH is engineered for ubiquitous accessibility across mainstream operating systems (Linux, MacOS, and Windows) while offering versatile interaction modes from Graphical User Interface (GUI) for direct visualisation to python library for flexible programming, facilitating the easy usage for non-expert users. 
The main contributions of this paper are summarised as follows:
\begin{itemize}

\item We propose a fully open-sourced, easy-usage, generalisable and cross-operating system OpenAHBU toolkit which comprehensively understand 11 types of human facial behaviours or their related psychological states at three different levels, and can be flexibly employed as a GUI-based toolkit or a python library. To the best of our knowledge, it provides the most extensive functional coverage among existing fully open-sourced academically available AFBU toolkit.

\item We introduce a Synergistic Multi-Task Enhancement (SMTE) module that adaptively refines each OpenAHBU function using its most relevant facial behaviour predictions from other functions, resulting in consistent performance gains with minimal computational overhead.

\item Extensive evaluation on 28 public academically available datasets demonstrated not only the superior performances of the proposed OpenAHBU toolkit over existing academically available toolkit, but also its robust cross-dataset generalisation capability.


\end{itemize}

\section{Related Work}


\subsection{Automatic facial behaviour understanding}
\label{subsec:AHBU}


\noindent \textit{\textbf{Automatic short-term observable facial behaviour understanding}} analyses observable, momentary facial patterns from visual signals. Specifically, \textbf{Facial landmark detection (FLD)} localises semantically meaningful spatial points in face images.  It has evolved from hand-crafted methods based on statistical shape priors and iterative optimisation \cite{saragih2011deformable,cao2014face,kazemi2014one}, which are vulnerable to extreme poses and occlusions, to more accurate and generalisable DL-based approaches \cite{kowalski2017deep,chiang2025orformer}. \textbf{Facial Blendshape Estimation (FBSE)} predicts geometric deformation coefficients (e.g., JawOpen and MouthSmile) for synthesising diverse, high-fidelity expressions, where a lightweight MediaPipe \cite{lugaresi2019mediapipe} efficiently estimates 52 frame-level blendshapes. Based on FACS \cite{ekman1978facial}, \textbf{Facial Action Unit Recognition (FAUR)} detects facial muscle activations (FAUD) or estimates their intensities on a 0-5 scale (FAUE) from a face image. Existing methods typically model AU-specific local regions \cite{zhao2015joint,chen2021region}, inter-AU dependencies \cite{luo2022learning,yang2023fan,wang2024multi,fan2020facial}, and spatio-temporal facial patterns \cite{li2023knowledge,song2021dynamic,sanchez2021affective}. \textbf{Facial Expression Recognition (FER)} identifies emotion-induced categorical expressions \cite{ekman2000basic} from images (static FER) or videos (dynamic FER). While early methods combined hand-crafted appearance and geometric features with traditional ML classifiers \cite{zhao2007dynamic,orrite2009hog}, recent DL solutions fine-tune Convolution Neural Networks (CNNs) \cite{toisoul2021estimation,zhang2023dual} or Transformers \cite{zheng2023poster,mao2025poster++} pre-trained on large face datasets such as VGGFace2 \cite{cao2018vggface2} for static FER, with temporal models (e.g., 2D CNNs/Transformers  \cite{chen2024static,liu2025robust}) further processing them for dynamic FER. Besides, spatio-temporal models (e.g., 3D CNNs or Video Transformers) also have been employed to jointly capture appearance and motion cues \cite{sun2023mae, sun2024svfap, chen2025static}.

\textit{\textbf{Automatic short-term facial behaviour-related internal state understanding}} infers human affective or physiological states from short-term facial dynamics. \textbf{Valence and Arousal Intensity Estimation (VAIE)} quantifies continuous valence and arousal intensities (typically ranging from -1 to 1) from facial images based on Russell’s emotion theory \cite{russell1980circumplex}. Recent competitions such as AVEC \cite{ringeval2018avec,ringeval2019avec} and ABAW \cite{kollias20246th,kollias2025advancements} have witnessed a proliferation of VAIE solutions, with top-performing ones employing: (i) 2D facial encoders with temporal models \cite{zhang2023facial,dresvyanskiy2024multi,liang2025mamba} or 3D CNNs \cite{praveen2022joint}; (ii) multi-task learning to explore emotion-related auxiliary facial cues  \cite{kollias2021affect,zhang2021prior,savchenko2023emotieffnet}; and (iii) pre-training VAIE models on large-scale face datasets \cite{yu2024exploring,dresvyanskiy2024multi}.
\textbf{Facial Pain Level Estimation (FPLE)} predicts observers' perceived pain states from face video, where 3D CNNs \cite{bellantonio2016spatio,tavakolian2018deep}, transformer \cite{gkikas2023full, gkikas2024multimodal}, and GNNs \cite{patania2022deep,alhamdoosh2025pain} have been widely developed to capture short-/long-term spatio-temporal facial appearance or geometrical patterns throughout the given face video for estimating pain intensities or binary pain-free / maximum-pain conditions. 

\textit{\textbf{Automatic long-term facial behaviour-related internal state understanding}} estimates relatively stable psychological traits or mental conditions from facial behaviours over extended periods. \textbf{Facial Depression Assessment (FDA)} assesses individual-level depression severity (e.g., BDI-II or PHQ-9 score) from a face video typically through three main paradigms: (1) inference from a single facial display \cite{liu2024multimodal,zhao2023novel}; (2) frame- or segment-level facial cues modelling followed by feature \cite{xu2024two} or prediction aggregation \cite{al2018video,niu2022dual}; and (3) compact video representation through sparse sampling or key-frame selection \cite{pan2024spatial,wu2025depmgnn}. Similar to FDA, \textbf{Facial Apparent Personality Recognition (FAPR)} predicts observer-perceived personality traits from facial videos using static facial appearance \cite{wei2017deep,kampman2018investigating}, frame/segment modelling with video-level aggregation \cite{li2020cr,subramaniam2016bi}, or direct video-level modelling \cite{song2021self,sun2024svfap}. \textbf{Facial Real Personality Recognition (FRPR)} instead infers individual-level real personality traits from face videos. Since real personality is associated with human cognition, recent breakthroughs \cite{song2022learning,kong2026learning,shao2021personality} represent personalised cognition using individual-specific network topologies or weights, encoding them as an individual-level graph representation to infer real personality traits with GNNs.

\subsection{AFBU toolkits}
\label{subsec:related-toolkit}

\noindent Several libraries package existing AFBU models into accessible interfaces. For example, Dlib \cite{king2009dlib} provides a HOG-SVM face detection, an ensemble of regression trees (ERT)-based 68-point landmark localisation, and a ResNet-based face recognition. Based on facial geometrics only, MediaPipe \cite{lugaresi2019mediapipe} delivers frame-level 468-point face meshes and 52 blendshape coefficients. Besides, DeepFace \cite{serengil2020lightface} offers lightweight models for age, gender, race, and static static expression recognition, while InsightFace \cite{deng2018arcface} integrating high performance DL models for 2D/3D face detection \cite{deng2020retinaface,guo2021sample}, recognition (ArcFace \cite{deng2018arcface}), alignment (SDUNet \cite{guo2018stacked}), landmark detection \cite{deng2018menpo}, as well as gender and age estimation, face quality assessment and anti-spoofing. However, these libraries generally require programming expertise, limiting their accessibility to non-specialists.

Alternatively, Table \ref{tab:compare} summarises existing widely-used and academically available AFBU toolkits with intuitive GUIs for non-technical users, which are developed using academically available datasets. Among them, OpenFace 2.0 \cite{Baltrusaitis2018openface}, Py-Feat \cite{cheong2023py}, LibreFace \cite{chang2024libreface}, and Libreface 2.0 \cite{guan2026libreface}, deploy independent function-specific models, while Behaviour4All \cite{kollias2024behaviour4all} and OpenFace 3.0 \cite{hu2025openface30lightweightmultitask} employing shared backbones with task-specific heads, which offer superior deployment efficiency at the cost of greater training challenges. Specifically, \textbf{OpenFace 2.0} combines MTCNN face detection \cite{zhang2016joint} with CE-CLM FLD\cite{zadeh2017convolutional}, orthographic head-pose estimation, CLNF-based gaze estimation \cite{baltrusaitis2013constrained,wood2015rendering}, and HOG/geometric feature-based recognition of 18 AU occurrences and 17 AU intensities. \textbf{Py-Feat} provides lightweight FLD, Img2Pose-based head pose estimation \cite{albiero2021img2pose} fine-tuned on 300W-LP dataset \cite{zhu2016face}, HOG-based FAUD using SVM or XGBoost \cite{chen2016xgboost} trained on DISFA+ \cite{mavadati2016extended}, and seven-class SFER using ResMaskNet \cite{pham2021facial} and linear SVMs \cite{chang2011libsvm} optimised on AffectNet \cite{mollahosseini2017affectnet}. \textbf{LibreFace} combines MediaPipe-based FLD \cite{lugaresi2019mediapipe} with separately distilled models for FAUD, FAUE, and seven-class SFER, where ViT-B teacher pre-trained with MAE self-supervision on EmotioNet \cite{fabian2016emotionet} is fine-tuned on AffectNet \cite{mollahosseini2017affectnet} and FFHQ \cite{karras2019style}, distilling expression knowledge to a ResNet-18 student for SFER. For joint FAUD and FAUE, a teacher is adapted on DISFA and distils knowledge to another ResNet-18 student. LibreFace 2.0 \cite{guan2026libreface} further combines real and synthetic AU data (from BP4D and DISFA) for training, distils knowledge to RepVGG \cite{ding2021repvgg} students.


The recent \textbf{OpenFace 3.0} \cite{hu2025openface30lightweightmultitask} extends OpenFace 2.0 into a deep multi-task framework to improve the efficiency–accuracy trade-off. It removes FAUE, but adds SFER while replacing face detection and FLD with RetinaFace \cite{deng2020retinaface} and Hourglass network \cite{yang2017stacked}. It also leverages EfficientNet \cite{tan2019efficientnet} as a shared backbone 
with function-specific heads for FAUD (ME-GraphAU \cite{luo2022learning} trained on DISFA and BP4D), 8-class SFER (MLP fine-tuned on AffectNet \cite{mollahosseini2017affectnet}), and gaze estimation (MLP trained on MPII Gaze \cite{zhang2015appearance} and Gaze360 \cite{kellnhofer2019gaze360}). Similarly, \textbf{Behaviour4All} \cite{kollias2024behaviour4all} combines RetinaFace-based \cite{deng2020retinaface} face detection and alignment with a shared ResNet backbone and function-specific MLP heads for: VA estimation trained on Aff-Wild \cite{kollias2018aff} and AffectNet, 7-class SFER trained on AffectNet and RAF-DB \cite{shan2018reliable}, as well as FAUD trained on EmotioNet \cite{fabian2016emotionet}, DISFA \cite{mavadati2013disfa}, BP4D \cite{zhang2014bp4d} and BP4D+ \cite{zhang2016multimodal}. Such unified designs educes inference latency while maintaining competitive performance.

However, the unreleased training code (OpenFace 2.0, Py-Feat, and Behaviour4all) and reliance on limited training data (only one or two datasets used per function in Py-Feat and OpenFace 3.0) severely restrict their reproducibility, extensibility and generalisation. Besides, outdated models (e.g., SVMs in OpenFace 2.0 and Py-Feat) and resource-oriented lightweight models (e.g., LibreFace and Libreface 2.0) also compromise accuracy in precision-critical applications. Furthermore, these toolkits generally only understand static facial displays (e.g., landmarks, AUs, and expressions), offering limited support for spatio-temporal behaviour understanding (e.g., DFER) and long-term facial behaviour-related psychological implications (e.g., depression and personality).

\begin{figure*}[htbp]
    \centering
    \includegraphics[width=1.0\linewidth]{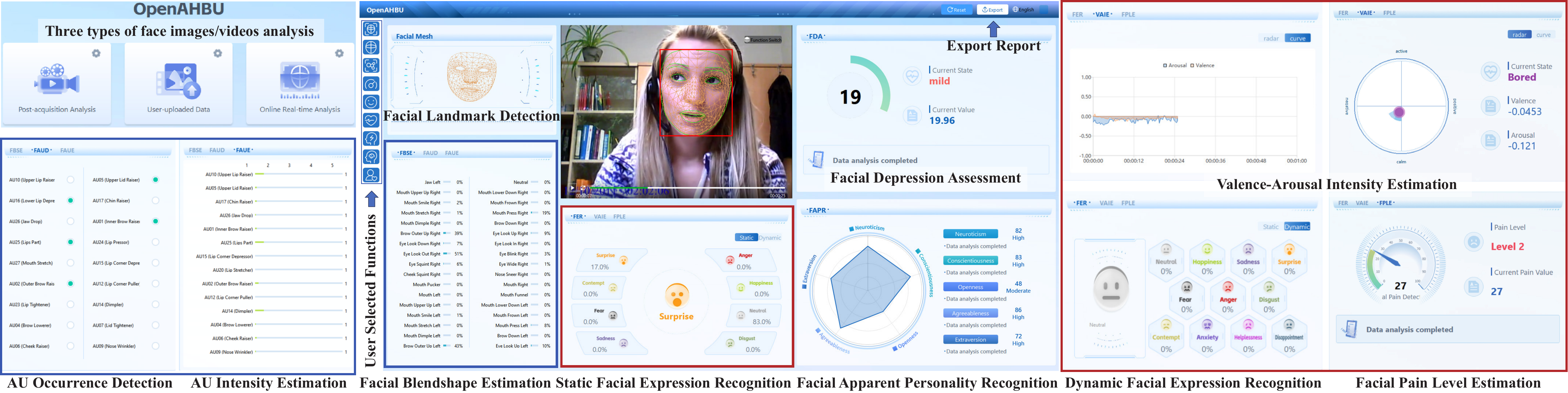}
    \caption{Interface of the proposed OpenAHBU.}
    \label{fig:main}
\end{figure*}

\section{OpenAHBU Toolkit}

\begin{figure*}[htbp]
    \centering
    \includegraphics[width=1.0\linewidth]{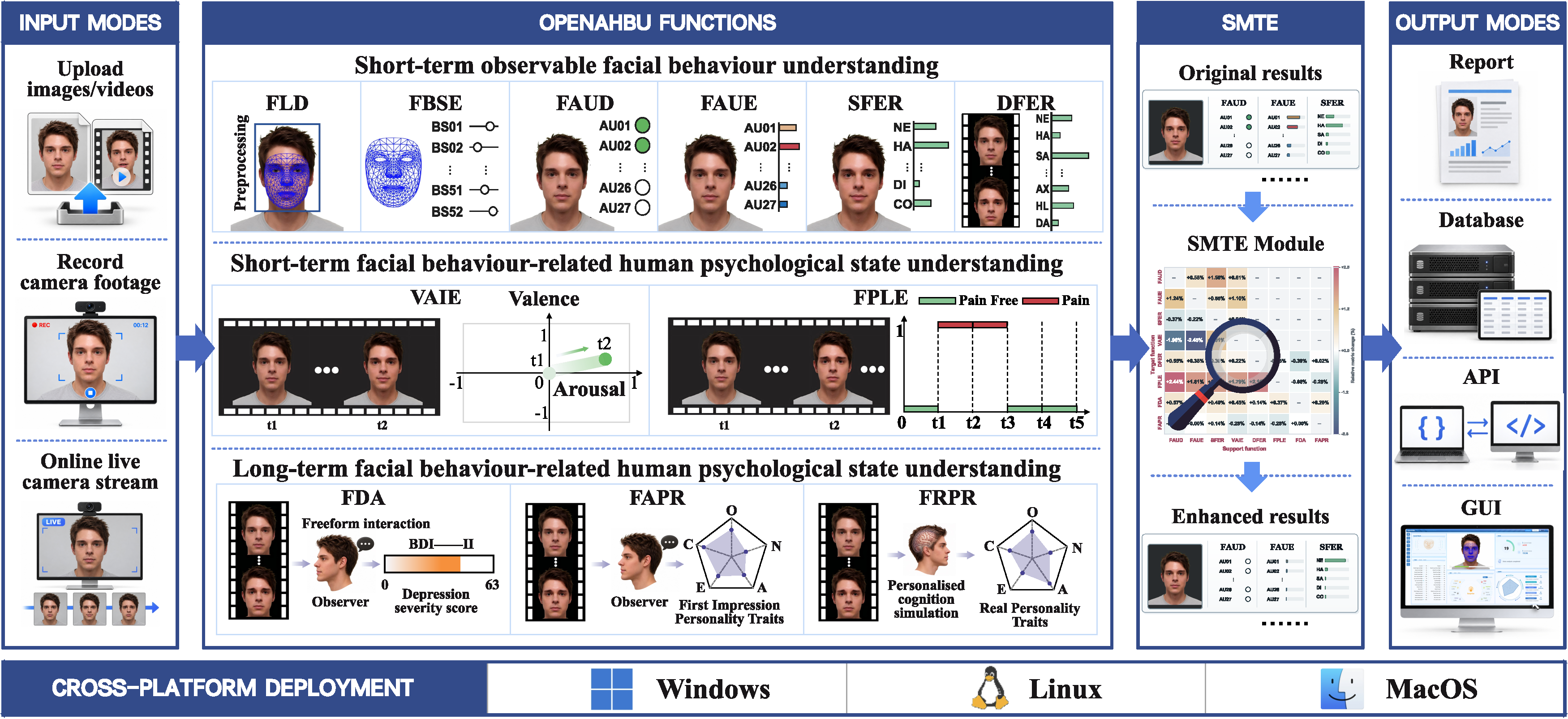}
    \caption{Overview of the proposed OpenAHBU toolkit which supports: (a) three types of face images/videos analysis; (b) 11 multi-level human facial behaviour and psychological state understanding; (c) enhancing function performances by modelling inter-dependencies between them; (d) four types of result visualisation/output; and (e) flexible deployment on Windows, Linux and MacOS.}
    \label{fig:main}
\end{figure*}

\noindent This section presents the architectural design, functional taxonomy, and model development details of OpenAHBU toolkit, whose pipeline is illustrated in Fig. \ref{fig:main}.


\subsection{Toolkit overview}
\label{sec: overview}

\noindent \textbf{Flexible Usage.} The proposed OpenAHBU toolkit supports \textbf{three types of face images/videos analysis in practical:} (1) analysis of user-uploaded face videos/images; (2) live camera face videos/image recording and post-capture analysis; and (3) online real-time analysis of live camera recordings. Subsequently, four prediction output types are provided: (1) storing them in local disk as a structured CSV or PDF report for further usage; (2) storing them in a database system for persistent access; (3) online prediction streaming for downstream applications or services; or (4) displaying them in an integrated user interface for real-time visualisation.

\textbf{Multi-level human facial behaviour and psychological state understanding.} Human facial behaviours are intrinsically multi-layered: observable motor patterns (e.g., AUs and expressions) also convey short-term affective or physiological states, such as valence–arousal and pain \cite{russell1980circumplex,toisoul2021estimation,egede2020emopain}, and long-term psychological conditions and traits, e.g., depression and personality \cite{cohn2009detecting,song2022spectral,vinciarelli2014survey}. Accordingly, the proposed OpenAHBU automatically understands 11 types of human facial behaviours and their related psychological states at three semantic levels: \textbf{(i) short-term observable facial behaviour understanding} including facial landmark detection (FLD), facial blend shape estimation (FBSE), static facial expression recognition (SFER), dynamic facial expression recognition (DFER), facial AU occurrence detection (FAUD) and intensity estimation (FAUE); \textbf{(ii) short-term facial behaviour-related human psychological state understanding} including valence-arousal intensity estimation (VAIE) and facial pain level estimation (FPLE); as well as \textbf{(iii) long-term facial behaviour-related human psychological state understanding} encompassing facial depression assessment (FDA), facial apparent personality recognition (FAPR) and real personality recognition (FRPR). 

\textbf{Adaptive Synergistic Inference.} To enhance the performance of the deployed functions, OpenAHBU enhance function-specific models using a Synergistic Multi-task Enhancement (SMTE) module. This is inspired by the inter-dependencies among facial behaviours and their associated psychological states (Sec. \ref{sec: SMTE}). For single function inference, it directly applies the user-selected model of the corresponding function. When multiple functions are activated, SMTE identifies complementary relationships among them, and accordingly refines their predictions to enhance their performances.

\textbf{Flexible and Efficient Deployment.} From an engineering perspective, the proposed OpenAHBU is designed to facilitate flexible and efficient deployment across diverse computational constraints. It not only offers a scalable range of models (i.e., ranging from lightweight architectures for real-time edge processing to high-capacity models for precision-critical analysis) alongside the adaptive synergistic inference pipeline, but also a stand-alone GUI for code-free deployment across diverse mainstream operating systems (i.e., Windows, Linux, and MacOS) as well as Python library (installable via standard package managers like pip) for customised extension.

\subsection{OpenAHBU functions}
\label{sec:funcs}

\subsubsection{OpenAHBU function and method selection criteria}

\textbf{Function selection criteria:} The 11 functions deployed in our OpenAHBU are carefully selected according to the following four criteria:
\begin{itemize}[leftmargin=1em, itemindent=0pt, labelsep=0.5em]

\item \textbf{Relevant to face-based affective computing.} A function is considered only if its output is directly or indirectly supports tasks along the pathway from facial behaviour understanding to the inference of associated affective/psychological. This includes: (i) canonical facial behaviours with established affective interpretation, e.g., AUs and categorical facial expressions; (ii) short-term internal affective or physiological states for which facial behaviour provides key visual evidence, e.g., valence/arousal and pain states; and (iii) psychological states or traits can be reliably inferred from long-term facial behaviour patterns, e.g., depression and personality. In contrast, generic/non-affective facial analysis tasks, such as identity, age, gender, race, head pose, gaze, face quality, and anti-spoofing, are not included as they neither characterise facial behaviours with established affective relevance nor infer human internal psychological states from such behaviours targeted by OpenAHBU.

\item \textbf{Availability of function-specific standardised and academically accessible datasets and benchmarks.} Except for the two pre-developed base functions (FLD and FBSE), a function is considered only if it is supported by at least one academically accessible dataset and an opensoruced benchmark that jointly provide: (i) function-specific labels; (ii) a documented annotation protocol; (iii) a pre-defined train/validation/test splits or a clear cross-validation protocol; and (iv) evaluation metrics that are commonly used in peer-reviewed method comparisons. Such criteria ensures researchers to inspect label definitions, reproduce data pre-processing and model training, and compare methods under identical data partitions. We therefore exclude candidates that does not have a public benchmark satisfying these requirements, such as facial stress or anxiety recognition.


\item \textbf{Facial data captured via standard RGB cameras.} Our OpenAHBU is designed for processing face images or videos acquired with standard RGB cameras under ordinary frame rates (20-30 FPS) and single-camera settings. This excludes functions developed on videos acquired by special recording protocol (e.g., high-speed capture or infrared thermal camera), e.g.,  micro-expression analysis.

\item \textbf{Availability of two open and reproducible DL literature.} Except for the two pre-developed base functions (FLD and FBSE) discussed below, each selected function is included only if the literature already contains at least two fully reproducible DL implementation. Please see method selection criteria for the detailed definition of `fully reproducible DL implementation'.


\end{itemize}
Here, FLD and FBSE are included as pre-developed functions: FLD provides essential facial alignment and geometric normalisation required by downstream OpenAHBU functions while FBSE providing frame-level facial blendshapes for high-fidelity expression synthesis. 

\noindent\textbf{Function-specific methods selection criteria:} While there are typically many methods have been proposed for each chosen OpenAHBU function, we only include and re-develop models for the AFBU methods meeting the following two criteria, ensuring the proposed OpenAHBU to be accurate, reproducible, and adaptable to diverse computational demands.
\begin{itemize} [leftmargin=1em, itemindent=0pt, labelsep=0.5em]

\item \textbf{Method openness and full reproducibility.} The methods should clearly present its model architecture, pre-processing pipeline, training/validation strategy, evaluation protocol and executable code with sufficient details for independent reproduction under the original setting. In this paper, full reproducibility means that the released materials are sufficient for an independent researcher to directly re-develop the method under the same protocol while fully reproducing its reported results with less than $1\%$ relative performance difference. Methods with paper-only descriptions, partial code releases, or inference-only checkpoints are not considered.

\item \textbf{Decent performances under unified evaluation.} The method selection also depends on their performances under a unified evaluation setting (detailed in Sec. \ref{sec:4.1.1}). For functions having multiple standardised benchmark datasets, we prioritise their methods with strong average performance. For functions with only one benchmark dataset, selection is based on performance of primary metric on that dataset. 
For each function, at least four methods are included, with parameter scales ranging from lightweight to high-capacity (see Sec. 3 of the Supplementary Material). This design supports fair comparison while allowing users to match method choice to specific application scenarios and computational resources.

\end{itemize}

\subsubsection{Hierarchical OpenAHBU functions and their methods}

\noindent Based on the above selection criteria, OpenAHBU includes 11 automatic facial behaviour understanding functions at three hierarchical levels, progressing from directly observable, transient facial attributes to latent, enduring psychological states. Here, FLD and FBSE are achieved via deploying existing well-developed models while we directly include the state-of-the-art (SOTA) opensourced model for the research-oriented FRPR function (they are not included in the later quantitative comparison). 
The number of selected methods varies across functions depending on the availability of methods meeting the above criteria.



\noindent \textbf{Automatic short-term observable human facial behaviour understanding:}
\begin{itemize}[leftmargin=1em, itemindent=0pt, labelsep=0.5em]

    \item \textbf{Facial Landmark Detection (FLD):} Since head pose and face scale variations within images can significantly degrade the subsequent facial analysis, our OpenAHBU employs the lightweight Dilb \cite{king2009dlib} and MediaPipe \cite{lugaresi2019mediapipe} to detect 68 2D facial landmarks and a dense mesh of 468 3D facial landmarks, respectively. These fiducial points enables OpenAHBU to perform canonical spatial normalisation, allowing the downstream models to receive spatially invariant inputs (i.e., face frames of a fixed scale, orientation, and position) under rigid head movements.

    \item \textbf{Facial Blendshape Estimation (FBSE):} 
    OpenAHBU incorporates FBSE provided by MediaPipe \cite{lugaresi2019mediapipe} to compute 52 specific blendshapes that quantify the activation levels of pre-defined facial deformations, enabling a direct and interpretable projection from input face video frames to 3D character animation systems.

    \item \textbf{Facial AU Occurrence Detection (FAUD):} By breaking down the complex human facial expression into a set of indivisible AUs, our OpenAHBU provides local and objective activation status for a set of facial muscles. This is achieved by re-training eight varying-scale models of four widely-used FAUD methods including ME-GraphAU (ResNet-50, ViT-S, ViT-B, ViT-L) \cite{luo2022learning}, MDHR (ResNet-50) \cite{wang2024multi}, FMAE (ViT-B, ViT-L) \cite{ning2024representation} and MAE-Face (ViT-B) \cite{ma2024facial}, where each FAUD model detects occurrences of 18 distinct AUs, including AU01, AU02, AU04, AU05, AU06, AU07, AU09, AU10, AU12, AU14, AU15, AU16, AU17, AU23, AU24, AU25, AU26, and AU27.

    \item \textbf{Facial AU Intensity Estimation (FAUE):} OpenAHBU includes five varying-scale FAUE models based on the four methods used in FAUD, each of which quantifies the continuous magnitudes/intensities of 14 AUs' activations including AU01, AU02, AU04, AU05, AU06, AU09, AU10, AU12, AU14, AU15, AU17, AU20, AU25, and AU26, allowing to distinguish between subtle and exaggerated AU activations. The selections of AUs detected by FAUD and FAUE functions are determined by the available labels provided by the employed academically available dataset.

    \item \textbf{Static Facial Expression Recognition (SFER):} To interpret human categorical expressions from static facial displays, OpenAHBU develops seven varying-scale models based on six top-performing methods (EMOFAN \cite{toisoul2021estimation}, DDAMFN \cite{zhang2023dual}, Poster++ \cite{mao2025poster++}, MHAN \cite{wang2026mhan}, MAE-Face \cite{ma2024facial} and FMAE (ViT-B/ViT-L) \cite{ning2024representation}), with each model recognising the target face image as one of eight facial expressions (e.g., Neutral, Happiness, Sadness, Surprise, Fear, Anger, Disgust, and Contempt). 

    \item \textbf{Dynamic Facial Expression Recognition (DFER):} Our OpenAHBU follows existing mainstream DFER benchmarks (DFEW \cite{jiang2020dfew}, FERV39k \cite{wang2022ferv39k} and MAFW \cite{liu2022mafw}) to process 2-second face videos for DFER, where five methods (TimeSFormer \cite{bertasius2021spacetime}, MAE4DFER \cite{sun2023mae}, SVFAP \cite{sun2024svfap}, FRU-Adapter (ViT-B/ViT-L) \cite{her2025fru} and MGNN \cite{wu2025depmgnn}) encompassing six varying-scale models are developed, with each detecting 11 categorical expressions: neutral, happiness, sadness, surprise, fear, anger, disgust, contempt, anxiety, helplessness, and disappointment. 

\end{itemize}

\noindent \textbf{Automatic short-term facial behaviour-related human psychological state understanding:} 
\begin{itemize} [leftmargin=1em, itemindent=0pt, labelsep=0.5em]

    \item \textbf{VA Intensity Estimation (VAIE):} 
    To capture continuous emotional transitions, OpenAHBU incorporates both static (DDAMFN \cite{zhang2023dual} and FMAE \cite{ning2024representation}) and temporal sequence modelling (MAVENC \cite{dresvyanskiy2024multi}, MAE4DFER \cite{sun2023mae} and SVFAP \cite{sun2024svfap}) for face frame-level continuous valence and arousal intensity estimation (ranging from -1 to 1).
    

    \item \textbf{Facial Pain Level Estimation (FPLE):} 
    OpenAHBU follows prominent pain benchmarks such as BioVid \cite{walter2013biovid} to adopt a 4-second temporal window for non-invasive pain estimation. This function facilitates analysis binary classification for pain and pain-free states by four different models, i.e., MGNN \cite{wu2025depmgnn}, TimeSFormer \cite{bertasius2021spacetime}, MAE4DFER \cite{sun2023mae}, and SVFAP \cite{sun2024svfap}. 
    
\end{itemize}

\noindent \textbf{Automatic long-term facial behaviour-related human state understanding:}
\begin{itemize} [leftmargin=1em, itemindent=0pt, labelsep=0.5em]

    \item \textbf{Facial Depression Assessment (FDA):} 
    Our OpenAHBU follows the `Freeform' interaction format consistent with AVEC 2013/2014 datasets, allowing users to freely interact with the toolkit, i.e., talking for a favoured topic from 30s to 4 minutes.
    Aligning with the AVEC 2013/2014, this function regresses a continuous BDI-II depression severity score within [0, 63] from the recorded user face video, and additionally map it to four severity levels: minimal (0–13), mild (14–19), moderate (20–28), and severe (29–63). We provide 
    four varying-scale models deprived from three methods for the FDA function (i.e., MGNN \cite{wu2025depmgnn}, MAE4DFER \cite{sun2023mae}, and SVFAP \cite{sun2024svfap}). 

    \item \textbf{Facial Apparent Personality Recognition (FAPR):} According to the widely-used ChaLearn First Impression benchmark \cite{ponce2016chalearn},  OpenAHBU regresses five intensity scores (from 0 to 1) as the individual-level Big-Five apparent personality traits (i.e., Openness, Conscientiousness, Extraversion, Agreeableness, and Neuroticism) from each 15s face video input. This function is achieved by four models based on four FAPR methods, i.e., MGNN \cite{wu2025depmgnn}, TimeSFormer \cite{bertasius2021spacetime}, MAE4DFER \cite{sun2023mae}, and SVFAP \cite{sun2024svfap}.

    \item \textbf{Facial Real Personality Recognition (FRPR):} 
    As the FRPR remains underexplored with limited solutions \cite{song2022learning,shao2021personality,kong2026learning}, OpenAHBU incorporates the SOTA FRPR model \cite{kong2026learning} to infer real personality traits reflecting an individual's real internal dispositions. Acknowledging that current FRPR performances have yet to meet the requirements for real-world deployment, this function is positioned for the research purpose only. 

\end{itemize} 


\subsection{Synergistic Multi-task Enhancement}
\label{sec: SMTE}

\begin{algorithm}[t]
\caption{OpenAHBU inference with Synergistic Multi-task Enhancement (SMTE)}
\label{alg:smte_inference}
\KwIn{Sample $x$; User-selected AFBU function set $\mathcal{A}$}
\KwOut{Prediction set $\mathcal{Y}$ for all functions in $\mathcal{A}$}
\BlankLine
$\mathcal{Y} \leftarrow \varnothing$\;
\tcp{$x=I^{(t)}$ for a frame input, or $x=v=\{I^{(t)}\}_{t=1}^{T}$ for a video input.}
\ForEach{$\mathrm{AFBU}_i \in \mathcal{A}$ \tcp*[f]{target function}}{
    Load the pretrained target model for $\mathrm{AFBU}_i$\;
    $p_i(x) \leftarrow \mathrm{AFBU}_i(x)$
    \tcp*[f]{original target prediction}\;
    \tcp{$p_i(x)$ denotes $\mathrm{AFBU}_i(I^{(t)})$ for $x=I^{(t)}$, and
    $\mathrm{AFBU}_i(v)$ for $x=v$.}
    \tcp{$\mathcal{C}_i$ is the validated support-function set;
    $\Delta_i(j)$ is the validation improvement obtained by using
    $\mathrm{AFBU}_j$ to support $\mathrm{AFBU}_i$.}
    $j_i^{\ast} \leftarrow \mathrm{ASFS}_i(\mathcal{A})$\;
    \tcp{$j_i^{\ast}=
    \arg\max_{j\in\mathcal{A}\cap\mathcal{C}_i}\Delta_i(j)$
    if $\mathcal{A}\cap\mathcal{C}_i\neq\varnothing$; otherwise,
    $j_i^{\ast}=\varnothing$.}
    \eIf{$j_i^{\ast}=\varnothing$}{
        $\mathrm{AFBU}_i^{\mathrm{SMTE}}(x) \leftarrow p_i(x)$
        \tcp*[f]{retain the original prediction}\;
    }{
        Set $\mathrm{AFBU}_{j_i^{\ast}}$ as the support function and
        load its pretrained model\;
        $p_{j_i^{\ast}}(x) \leftarrow \mathrm{AFBU}_{j_i^{\ast}}(x)$\
        \tcp*[f]{obtain the support function prediction}\;
        Load the trained pair-specific SMTE module
        $\mathrm{SMTE}_{i\leftarrow j_i^{\ast}}$\;
        $\mathrm{AFBU}_i^{\mathrm{SMTE}}(x) \leftarrow
        \mathrm{SMTE}_{i\leftarrow j_i^{\ast}}
        \bigl(x;p_{j_i^{\ast}}(x)\bigr)$\;
    }
    $\mathcal{Y} \leftarrow \mathcal{Y} \cup
    \mathrm{AFBU}_i^{\mathrm{SMTE}}(x)$\;
}
\KwRet{$\mathcal{Y}$}\;
\end{algorithm}

\begin{figure}[htbp]
    \centering
    \includegraphics[width=1.0\linewidth]{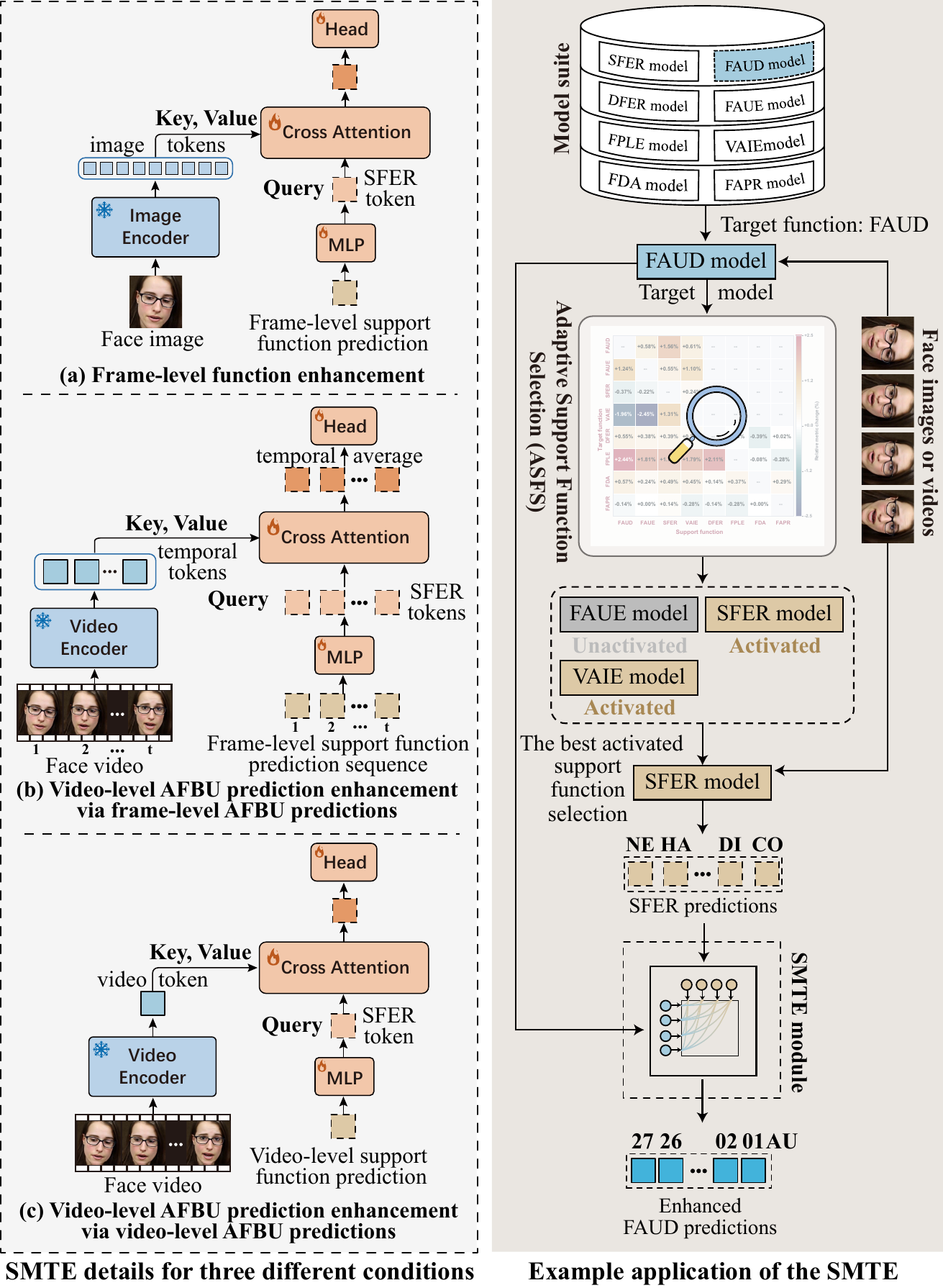}
    \caption{Schematic illustration of the proposed Synergistic Multi-Task Enhancement (SMTE) module. \textbf{Left:} SMTE details for three different conditions. The "snowflake" and "flame" icons denote frozen parameters to preserve pre-trained representations and learnable modules, respectively. \textbf{Right:} Example application of the SMTE. The user selects the FAUD as the target function, while the SMTE selects its best activated support function SFER. Then, the given image/video is processed by both FAUD and SFER, followed by SMTE refining the FAUD predictions using the SFER predictions.}
    \label{fig:second_stage}
\end{figure}

\noindent Empirical evidence shows that short-term facial behaviours are mutual-related, and are informative for reflecting long-term psychological states, e.g., AU activations and valence-arousal trajectories are closely associated with categorical facial expressions~\cite{ekman1978facial,russell1980circumplex}, while depression~\cite{song2022spectral} and personality traits~\cite{vinciarelli2014survey} are reflected in various facial temporal patterns. In contrast, long-term psychological states provide limited cues for understanding short-term facial behaviours as individuals with different psychological profiles may exhibit similar short-term expressions.
Please see Sec. 6.2.1  and Secs. 6.2.2–6.2.9 of the Supplementary Material for a summary of such relationships and experimental validation.



Inspired by this, the proposed OpenAHBU introduces a Synergistic Multi-task Enhancement (SMTE) module to explicitly model the relationship between: (1) frame-level facial behaviour predictions; (2) frame-level and video-level facial behaviour predictions; and (3) video-level facial behaviour predictions. Specifically, it adaptively and efficiently enhances each of the following OpenAHBU functions by attending informative facial behaviour predictions provided by its most related function, with less than 2\% additional parameters as:
\begin{itemize} [leftmargin=1em, itemindent=0pt, labelsep=0.5em]

    \item \textbf{SFER:} OpenAHBU applies frame-level VAIE predictions to support frame-level SFER function.

    \item \textbf{VAIE:} OpenAHBU applies frame-level SFER predictions to support frame-level VAIE function.

    \item \textbf{FAUD:} OpenAHBU applies frame-level FAUE, SFER or VAIE predictions to support frame-level FAUD function.

    \item \textbf{FAUE:} OpenAHBU applies frame-level SFER, FAUD or VAIE predictions to support frame-level FAUE function.

    \item \textbf{DFER:} OpenAHBU applies frame-level FAUD, FAUE, SFER or VAIE predictions to support short video-level DFER function.

    \item \textbf{FPLE:} OpenAHBU applies frame-level FAUD, FAUE, SFER, VAIE or short video-level DFER predictions to support short video-level FPLE function.

    \item \textbf{FDA:} OpenAHBU applies frame-level FAUD, FAUE, SFER, VAIE, short video-level FPLE or long video-level FAPR predictions to support long video-level FDA function. 

    \item \textbf{FAPR:} OpenAHBU applies frame-level SFER predictions to support long video-level FAPR function.

\end{itemize}

\textbf{Frame-level function enhancement:} As illustrated in Fig.~\ref{fig:second_stage}a, the proposed SMTE enhances the target image/frame-level function $\text{AFBU}_i$ by integrating image/frame-level facial behaviour predictions achieved by its most associated function $\text{AFBU}_{j}$. For the $t_\text{th}$ face frame $I^{(t)}$, let $X^{(t)}_{i} \in \mathbb{R}^{L \times C}$ and $p^{(t)}_{i} = \text{AFBU}_i(I^{(t)})$ denote its latent facial representation extracted by the frozen encoder and the original prediction of the target function $\text{AFBU}_i$, while $p_j^{(t)} = \text{AFBU}_j(I^{(t)})$ be the prediction produced by the associated support function $\text{AFBU}_j$, the SMTE first projects the support prediction $p_j^{(t)}$ as a representation of the same dimension as the $X^{(t)}_{i}$ via a lightweight MLP $\phi_{i,j}(\cdot)$. It is then used as a query to further enhance task-relevant facial cues in $X_{i}^{(t)}$, resulting in a $z_{i\leftarrow j}^{(t)}$ as:
\begin{equation}
\begin{split}
z_{i\leftarrow j}^{(t)}
&=
\operatorname{CA}\left(f(p_{j}^{(t)}), X_{i}^{(t)}, X_{i}^{(t)}\right) \\
&=\operatorname{Softmax}
\left(
\frac{\phi_{i,j}(p_{j}^{(t)})(X_{i}^{(t)})^{\top}}{\sqrt{C}}
\right)X_{i}^{(t)}
\end{split}
\end{equation}
where $\operatorname{CA}(\cdot)$ denotes the cross-attention operation. This way, the complementary target-related cues learned by the support function $\text{AFBU}_j(I^{(t)})$ is incorporated to enhance the target function $\text{AFBU}_i(I^{(t)})$. When \textbf{multiple support functions are available} for the target function, the SMTE includes an \textbf{Adaptive Support Function Selection (ASFS)} block that adaptively selects the most effective available support function providing most improvement for the target function as:
\begin{equation}
\label{eq:asfs}
j_i^{\ast}
= \operatorname{ASFS}_i(\mathcal{A}) =
\begin{cases} 
\arg\max_{j\in \mathcal{A}\cap\mathcal{C}_i}\Delta_i(j), & 
    \mathcal{A}\cap\mathcal{C}_i \neq \varnothing, \\
\varnothing, &
     \text{otherwise}.
\end{cases}
\end{equation}
where $\mathcal{A}$ denotes all AFBU functions activated by the user; $\mathcal{C}_i$ denotes the candidate support functions empirically validated for the target function $\text{AFBU}_i$, and $\Delta_i(j)$ denotes the validation improvement achieved by $\text{AFBU}_j$ for supporting $\text{AFBU}_i$. As a result, the final enhanced frame-level prediction is obtained as:
\begin{equation}
\operatorname{AFBU}_i^{\operatorname{SMTE}}(I^{(t)})
=
\begin{cases}
\operatorname{AFBU}_i(I^{(t)}), & j_i^{\ast}=\varnothing,\\
h_i\left(z_{i\leftarrow j_i^{\ast}}^{(t)}\right), & \text{otherwise}.
\end{cases}
\end{equation} 
where $h_i(\cdot)$ is a MLP prediction head. The proposed SMTE does not jointly apply multiple support functions as our results (detailed in Supplementary Material) reveal that compared to using the most informative single support function to complement the target AFBU function, introducing multiple support functions does not show consistent performance improvement but large additional memory usage and computational complexity.

\textbf{Video-level AFBU prediction enhancement via video-level AFBU predictions:} As shown in Fig.~\ref{fig:second_stage}c, SMTE can further enhance a video-level target function $\text{AFBU}_i$ using a video-level support function $\text{AFBU}_j$. Specifically, ASFS first selects the most effective available support function according to Eq.~\ref{eq:asfs}. Its prediction $\text{AFBU}_j(v)$ is then projected by $\phi_{i,j}(\cdot)$ and used as a query to attend to the target video representation $F_i(v)$:
\begin{equation}
R_{i\leftarrow j}(v)
=
\operatorname{CA}\left(
\phi_{i,j}(\text{AFBU}_j(v)),
F_i(v),
F_i(v)
\right).
\end{equation}
The final enhanced prediction is produced by applying temporal aggregation and the target prediction head as:
\begin{equation}
\operatorname{AFBU}_i^{\operatorname{SMTE}}(v)=
\begin{cases}
\operatorname{AFBU}_i(v), & j_i^\ast=\varnothing,\\
h_i\left(\rho\left(R_{i\leftarrow j_i^\ast}(v)\right)\right), & \text{otherwise}.
\end{cases}
\end{equation}

\textbf{Video-level AFBU prediction enhancement via video-level AFBU predictions:} As shown in Fig.~\ref{fig:second_stage}c, the SMTE also enhances the target video-level AFBU function $\text{AFBU}_i$ by integrating the video-level prediction $\text{AFBU}_j(v)$ of its associated support video-level AFBU function. Following the similar process as frame-level AFBU prediction enhancement, the SMTE adaptively selects the most effective available video-level support function of $\text{AFBU}_i$ and conducts cross-attention to obtain a video-level enhanced representation from video-level representation learned by $\text{AFBU}_i$ and the video-level prediction $\text{AFBU}_j(v)$. Finally, an MLP-based head further projects the enhanced video-level representation to the final enhanced video-level prediction.

\subsection{OpenAHBU model development}
\label{sec:3.4}

\ExplSyntaxOn
\NewDocumentCommand { \OpenAHBUCDashLine }
  { > { \SplitArgument { 1 } { - } } m }
  {
    \tl_gput_right:Nx \g_nicematrix_code_after_tl
      {
        \exp_not:N \__openahbu_cdash_line:nnn
        { \int_use:c { c@iRow } } #1
      }
  }
\cs_new_protected:Nn \__openahbu_cdash_line:nnn
  {
    \tikz \draw
      [
        line~width = \arrayrulewidth,
        dash~pattern = on~4pt~off~2pt,
        line~cap = butt
      ]
      (#1-|#2) -- (#1-|\int_eval:n { #3 + 1 }) ;
  }
\ExplSyntaxOff

\begin{table*}[htbp]
\centering
\Huge
\setlength{\doublerulesep}{0.3em}
\renewcommand{\arraystretch}{1.5}
\caption{Datasets used for developing OpenAHBU. Example numbers represent the numbers of images/videos for image-/video-level functions. NE: Neutral, HA: Happiness, SA: Sadness, SU: Surprise, FE: Fear, AN: Anger, DI: Disgust, CO: Contempt, AX: Anxiety, HL: Helplessness, DA: Disappointment, W: in-the-wild, C: controlled environment.}
\resizebox{1.0\linewidth}{!}{
\begin{NiceTabular}{cccccc}
\Xhline{5pt}
\rowcolor{gray!15}
\textbf{Function} &
  \textbf{Datasets} &
  \textbf{Spilt (Train/Val/Test)} &
  \textbf{Example number (Train/Val/Test)} &
  \textbf{Label} &
  \textbf{Source} \\ \Hline\Hline
 &
  AffectNet-8 \cite{mollahosseini2017affectnet} &
  80\% Training set/20\% Training set/Validation set &
  230120/57531/4000 &
  NE, HA, SA, SU, FE, AN, DI, CO &
  W \\
 &
  ExpW \cite{zhang2018facial} &
  All data/-/- &
  87305/-/- &
  NE, HA, SA, SU, FE, AN, DI &
  W \\
 &
  RAF-DB \cite{shan2018reliable} &
  80\% Training set/20\% Training set/Test set &
  9817/2454/3068 &
  NE, HA, SA, SU, FE, AN, DI &
  W \\
 &
  CK+ \cite{lucey2010extended} &
  All data/-/- &
  920/-/- &
  NE, HA, SA, SU, FE, AN, DI, CO &
  C \\
 &
  RAFD \cite{langner2010presentation} &
  All data/-/- &
  1608/-/- &
  NE, HA, SA, SU, FE, AN, DI, CO &
  C \\
 &
  SFEW \cite{dhall2015video} &
  80\% Training set/20\% Training set/Validation set &
  682/171/404 &
  NE, HA, SA, SU, FE, AN, DI &
  W \\
 &
  FER+ \cite{BarsoumICMI2016} &
  Training set/Validation set/Test set &
  25060/3191/3153 &
  NE, HA, SA, SU, FE, AN, DI, CO &
  W \\
 &
  KDEF \cite{lundqvist1998karolinska} &
  All data/-/- &
  4888/-/- &
  NE, HA, SA, SU, FE, AN, DI &
  C \\
\multirow{-8}{*}{SFER} &
  AFFW-FER \cite{kollias2018aff} &
  80\% Training set/20\% Training set/Validation set &
  332758/83190/173305 &
  NE, HA, SA, SU, FE, AN, DI &
  W \\ \OpenAHBUCDashLine{2-6}
  &
  Function-level &
  Training set/Validation set/Test set &
  693158/146537/183930 &
  NE, HA, SA, SU, FE, AN, DI, CO &
  - \\
  \Hline
 &
  BP4D \cite{zhang2014bp4d} &
  80\% Fold 2,3/20\% Fold 2,3/Fold 1 &
  77332/19333/50182 &
  AU 01, 02, 04, 06, 07, 10, 12, 14, 15, 17, 23, 24 &
  C \\
 &
  DISFA \cite{mavadati2013disfa} &
  80\% Fold 2,3/20\% Fold 2,3/Fold 1 &
  69768/17442/43604 &
  AU 01, 02, 04, 06, 09, 12, 25, 26 &
  C \\
 &
  RAF-AU \cite{yan2020raf} &
  Training set/Validation set/Test set &
  3031/433/863 &
  AU 01, 02, 04, 05, 06, 09, 10, 12, 16, 17, 25, 26, 27 &
  W \\
 &
  CK+ \cite{lucey2010extended} &
  Training set/Validation set/Test set &
  384/59/145 &
  AU 01, 02, 04, 05, 06, 07, 12, 17, 25 &
  C \\
 &
  CASME2 \cite{yan2014casme} &
  Training set/Validation set/Test set &
  8729/1995/5995 &
  AU 01, 02, 04, 07, 12, 14, 17 &
  C \\
\multirow{-5}{*}{FAUD} &
  AFFW-AU \cite{kollias2018aff} &
  80\% Training set/20\% Training set/Validation set &
  1056304/264076/445845 &
  AU 01, 02, 04, 06, 07, 10, 12, 15, 23, 24, 25, 26 &
  W \\ \OpenAHBUCDashLine{2-6}
  &
  Function-level &
  Training set/Validation set/Test set &
  1215548/303338/546634 &
  AU 01, 02, 04, 05, 06, 07, 09, 10, 12, 14, 15, 16, 17, 23, 24, 25, 26, 27 &
  - \\
  \Hline
 &
  BP4D \cite{zhang2014bp4d} &
  80\% Fold 2,3/20\% Fold 2,3/Fold 1 &
  77332/19333/50182 &
  AU 06, 10, 12, 14, 17 &
  C \\
\multirow{-1}{*}{FAUE} &
  DISFA \cite{mavadati2013disfa} &
  80\% Fold 2,3/20\% Fold 2,3/Fold 1 &
  69768/17442/43604 &
  AU 01, 02, 04, 05, 06, 09, 12, 15, 17, 20, 25, 26 &
  C \\ \OpenAHBUCDashLine{2-6}
  &
  Function-level &
  Training set/Validation set/Test set &
  147100/36775/93786 &
  AU 01, 02, 04, 05, 06, 09, 10, 12, 14, 15, 17, 20, 25, 26 &
  C \\
  \Hline
 &
  DFEW \cite{jiang2020dfew} &
  80\% Fold 2,3,4,5/20\% Fold 2,3,4,5/Fold 1 &
  7485/1871/2341 &
  NE, HA, SA, SU, FE, AN, DI &
  W \\
 &
  FERV39k \cite{wang2022ferv39k} &
  80\% Training set/20\% Training set/Test set &
  24870/6218/7847 &
  NE, HA, SA, SU, FE, AN, DI &
  W \\
\multirow{-2}{*}{DFER} &
  MAFW \cite{liu2022mafw} &
  80\% Fold 2,3,4,5/20\% Fold 2,3,4,5/Fold 1 &
  5866/1467/1839 &
  NE, HA, SA, SU, FE, AN, DI, CO, AX, HL, DA &
  W \\ \OpenAHBUCDashLine{2-6}
  &
  Function-level &
  Training set/Validation set/Test set &
  38221/9556/12027 &
  NE, HA, SA, SU, FE, AN, DI, CO, AX, HL, DA &
  W \\
  \Hline
 &
  AFEW-VA \cite{kossaifi2017afew} &
  Training set/Validation set/Test set &
  17736/4434/7881 &
  Valence, Arousal &
  W \\
 &
  SEWA \cite{kossaifi2019sewa} &
  Training set/Validation set/Test set &
  322102/80526/94775 &
  Valence, Arousal &
  C \\
 &
  AffectNet \cite{mollahosseini2017affectnet} &
  80\% Training set/20\% Training set/Validation set &
  230121/57530/4000 &
  Valence, Arousal &
  W \\
\multirow{-3}{*}{VAIE} &
  AFFW-VA \cite{kollias2018aff} &
  80\% Training set/20\% Training set/Validation set &
  1323354/330839/376334 &
  Valence, Arousal &
  W \\ \OpenAHBUCDashLine{2-6}
  &
  Function-level &
  Training set/Validation set/Test set &
  1893313/473329/482990 &
  Valence, Arousal &
  - \\
  \Hline
 &
  BioVid \cite{walter2013biovid} &
  80\% Training set/20\% Training set/Test set &
  11200/2780/3400 &
  Pain Free, Pain &
  C \\
\multirow{-1}{*}{FPLE} &
  MIntPAIN \cite{haque2018deep} &
  80\% Training set/20\% Training set/Test set &
  1993/498/630 &
  Pain Free, Pain &
  C \\ \OpenAHBUCDashLine{2-6}
  &
  Function-level &
  Training set/Validation set/Test set &
  13193/3278/4030 &
  Pain Free, Pain &
  C \\
  \Hline
FDA &
  AVEC2014 \cite{valstar2014avec} &
  Training set/Validation set/Test set &
  50/50/50 &
  BDI-II depression severity score &
  C \\ \Hline
FAPR &
  First Impression \cite{ponce2016chalearn} &
  Training set/Validation set/Test set &
  6000/2000/2000 &
  Big five personality traits &
  W \\
  \Xhline{5pt}
\end{NiceTabular}
}
\label{tab:datasets}
\end{table*}

\subsubsection{Model development datasets} 
\label{sec:3.4.1}

\noindent To ensure robust cross-domain generalisation, each trainable OpenAFBU model is developed using one or more academically available datasets collected under different contexts, with total 28 datasets employed for developing eight AFBU functions (i.e., FAUD, FAUE, SFER, DFER, VAIE, FPLE, FDA, and FAPR). Besides, FLD and FBSE are adopted as off-the-shelf supporting functions while FRPR is included as a research-oriented function, all of which are directly inherited from their open-sourced models. Table. \ref{tab:datasets} provides the details of the training, validation and testing subsets for all developed OpenAFBU models. For datasets providing official training, validation, and test sets, we directly follow the official partitions. For datasets with only publicly available official training and test/validation sets (e.g., AffectNet in SFER and VAIE), we randomly divide the released training set into 80\% for training and 20\% for validation, and report the results on the publicly evaluable official test/validation set.For each function, these defined splits are fixed and used for the development of all corresponding models to ensure their fair comparison. For datasets defined with cross-validation protocols, we employ the pre-defined first fold as the evaluation split, and randomly divide the remaining folds into 80\% for training and 20\% for validation. 
For datasets without official partitions, we either include all available samples in training or establish fixed train/validation/test splits. 
Under these dataset-specific protocols, the training, validation, and test subsets of all datasets belonging to the same function are separately merged to form the function-level partitions. The validation partition is used for model selection, whereas the evaluation partition is used only for final performance reporting. More details of the dataset are provided in the Supplementary Material.

\subsubsection{Multi-dataset function-level OpenAHBU model development}
\label{sec:3.4.2}

\noindent All core OpenAHBU models (excluding the off-the-shelf FLD and FBSE functions) are developed based on a unified two-stage protocol. Specifically, \textbf{Stage I} independently trains each function-specific model under the dataset configurations described in Sec. \ref{sec:3.4.1}. For each function, the best-performing model from Stage I is selected according to its validation performance, which is then enhanced by the SMTE module developed in Stage II.

\textbf{Stage I: Multi-dataset function-specific model training:} The \textbf{General Principles} for developing function-specific baseline models are introduced below.
\begin{itemize}[leftmargin=1em, itemindent=0pt, labelsep=0.5em]

\item \textbf{Data pre-processing:} All face images and videos undergo face detection, cropping, and alignment using the FLD function. During training, we apply unified data augmentations, including random horizontal flipping, resized cropping, colour jittering, and random erasing. For face video-based functions, identical spatial transformations are applied to all sampled frames within a video clip to preserve temporal coherence. Input resolutions, temporal lengths, and sampling strategies follow the corresponding original models unless otherwise specified.

\item \textbf{Model training:} To improve generalisation across heterogeneous datasets, we employ Sharpness-aware Minimisation (SAM) \cite{foret2021sharpnessaware,kwon2021asam} to enhance the optimisation for all trainable models. Beyond this unified setting, model-specific configurations are kept consistent with the implementations described in their original papers, including pre-trained weights, auxiliary loss functions, loss weighting ratios, etc,  ensuring reproducible and fair comparisons within a unified training pipeline.

\item \textbf{Model validation and selection:} Model checkpoints are evaluated on the function-level validation partitions defined in Sec. \ref{sec:3.4.1}. For each model, the best checkpoint is selected according to the primary validation metric of the function 
(detailed in Sec. 5 of the Supplementary Material), while the test partitions are used only for reporting the final results. Early stopping is applied when validation performance does not improve for a pre-defined patience, reducing over-fitting during prolonged training.

\item \textbf{Other settings:} For functions affected by both dataset-size imbalance (unequal numbers of examples across datasets) and label imbalance (unequal occurrence frequencies across label categories), 
we adopt a dual-level resampling strategy (detailed in the Supplementary Material). At the dataset level, sampling probabilities are inversely related to dataset sizes to prevent large datasets from dominating training. At the label level, samples from under-represented categories or AU labels are assigned higher sampling probabilities within each dataset. This strategy provides a more balanced exposure to different datasets and labels during training.

\end{itemize}
In addition to such general principles, \textbf{function-specific configurations} for developing our OpenAHBU models are described as follows:
\begin{itemize}[leftmargin=1em, itemindent=0pt, labelsep=0.5em]

\item \textbf{FAUD:} All FAUD models are trained using the asymmetric loss~\cite{ridnik2021asymmetric} that can reduce the dominance of easy negative samples caused by the imbalanced labels.

\item \textbf{SFER and DFER:} Models of both functions are trained using label-smoothing cross-entropy loss to reduce over-confident predictions and improve the model robustness under noisy expression labels.

\item \textbf{FAUE, VAIE, FDA and FAPR:} All continuous regression FAUE, VAIE, FDA and FAPR models are trained using the Concordance Correlation Coefficient (CCC) loss that jointly measures correlation and bias between predictions and labels over the temporal dimension.

\item \textbf{FPLE:} For binary pain state classification, all models are trained using the standard Binary Cross-Entropy loss.

\item \textbf{FDA:} Since AVEC2014 is relatively small, we additionally employ a cosine annealing learning rate schedule with warm restarts \cite{loshchilov2017sgdr} to improve the training stability.

\end{itemize}

\textbf{Stage II: SMTE Development:} Building upon the best function-specific models developed in Stage I, 
During \textbf{Stage II}, the selected models developed in Stage I are frozen. Only the parameters (i.e., the projection MLP layers $\phi_{i,j}(\cdot)$, the cross-attention block $\operatorname{CA}(\cdot)$,  and the function-specific prediction head $h_i(\cdot)$) introduced by SMTE are trainable, i.e., this stage exploits cross-function facial behaviour cues without changing the visual or spatio-temporal representations learned in Stage I. This decoupled design ensures that Stage I models remain independently deployable as robust baselines, while Stage II providing optional performance enhancement when beneficial support functions are available.

\section{Experiments}
\label{sec:exp}

This section comprehensively evaluates the proposed OpenAHBU from three perspectives: cross-dataset generalisation, accuracy-efficiency trade-off, and the effectiveness of the SMTE module. While our OpenAHBU supporting 11 AFBU functions, the employed FLD, FBSE, and FRPR models are directly inherited from previous papers. Thus, we only quantitatively evaluate the eight re-developed functions (SFER, DFER, FAUD, FAUE, VAIE, FPLE, FDA, and FAPR). Note that it is time-consuming and impractical to conduct all experiments for all re-developed models (totally 43 models) on all OpenAHBU functions. 
More comprehensive analysis, including dataset-wise model performances, detailed SMTE results, and statistical significance analysis are provided in the \textbf{Supplementary Material}.

\subsection{Experimental settings}
\label{sec:4.1.1}

\textbf{Evaluation and comparison protocol:} All re-developed models of each function are evaluated under the same test partitions defined in Sec. \ref{sec:3.4.1}. For functions developed on multiple datasets (SFER, FAUD, FAUE, VAIE, DFER, and FPLE), results are reported on their corresponding function-level test sets. Otherwise, results are reported on the official test set of the corresponding dataset (FDA and FAPR). Existing toolkits, including DeepFace, OpenFace 2.0/3.0, Py-Feat, LibreFace, and LibreFace 2.0 (their details are provided in the Supplementary Material), are evaluated using their official inference pipelines on the same partitions as OpenAHBU. Results for recent SOTA methods are taken directly from their original papers.

\noindent \textbf{Evaluation metrics:} For each function, we adopt the primary metric commonly used in its research community to rank re-developed models and their competitors: \textbf{FAUD:} F1 score that balances precision and recall for multi-label AU occurrence detection; \textbf{FAUE:} Intra-class Correlation Coefficient (ICC) for AU intensity estimation; \textbf{SFER and DFER:} classification Accuracy (ACC) and Weighted Average Recall (WAR), respectively, for single-label expression recognition; \textbf{VAIE and FAPR:} Concordance Correlation Coefficient (CCC) that jointly measures correlation and differences between predictions and ground-truth values; \textbf{FPLE:} accuracy (ACC) for binary pain classification; and \textbf{FDA:} Root Mean Square Error (RMSE) for depression estimation. Results of other primary metrics are additionally provided in the Supplementary Material.


\begin{table*}[htbp]
\small
\centering
\setlength{\doublerulesep}{0.3em}
\renewcommand{\arraystretch}{1.4}
\caption{Comparison among OpenAHBU, SOTA competitors, and existing open-sourced AFBU toolkits. DF: DeepFace\cite{serengil2020lightface}, OF2: OpenFace 2.0 \cite{Baltrusaitis2018openface}, PF: Py-Feat \cite{cheong2023py}, LF: LibreFace \cite{chang2024libreface}, LF2: LibreFace 2.0 \cite{guan2026libreface}, OF3: OpenFace 3.0 \cite{hu2025openface30lightweightmultitask}.}
\resizebox{0.9\linewidth}{!}{
\begin{tabular}{cc|ccc|cccccc|c}
\Xhline{1.5px}
\multicolumn{2}{c|}{\cellcolor{gray!15}\textbf{Functions}} &
  \multicolumn{3}{c|}{\cellcolor{gray!15}\textbf{SOTA Methods}} &
  \multicolumn{7}{c}{\cellcolor{gray!15}\textbf{Toolkits/Libraries}} \\ 
  \groupdoubleline
\multicolumn{1}{c|}{} &
  Datasets &
  MDHR \cite{wang2024multi} &
  MAE-Face \cite{ma2024facial} &
  FMAE \cite{ning2024representation} &
  DF &
  OF2 &
  PF &
  LF &
  LF2 &
  OF3 &
  \textbf{OpenAHBU} \\ \hhline{~-||---||-------} 
\multicolumn{1}{c|}{} &
  BP4D &
  66.10 &
  67.40 &
  67.10 &
  - &
  \textbf{69.52} &
  53.02 &
  60.73 &
  66.77 &
  27.82 &
  \underline{67.38} \\
\multicolumn{1}{c|}{} &
  DISFA &
  66.20 &
  70.80 &
  70.10 &
  - &
  48.01 &
  56.88 &
  57.58 &
  33.74 &
  \underline{73.10} &
  \textbf{73.53} \\
\multicolumn{1}{c|}{} &
  RAF-AU &
  - &
  - &
  - &
  - &
  40.31 &
  29.32 &
  40.39 &
  27.46 &
  \underline{58.09} &
  \textbf{76.74} \\
\multicolumn{1}{c|}{} &
  CK+ &
  - &
  - &
  - &
  - &
  22.59 &
  \underline{24.51} &
  20.95 &
  17.17 &
  23.88 &
  \textbf{30.22} \\
\multicolumn{1}{c|}{} &
  CASME2 &
  - &
  - &
  - &
  - &
  18.86 &
  14.43 &
  \underline{26.97} &
  20.36 &
  6.51 &
  \textbf{32.06} \\
\multicolumn{1}{c|}{\multirow{-6}{*}{\begin{tabular}[c]{@{}c@{}}FAUD\\ (F1 score↑)\end{tabular}}} &
  AFFW-AU &
  - &
  - &
  - &
  - &
  32.82 &
  32.28 &
  32.92 &
  33.53 &
  \underline{40.03} &
  \textbf{52.62} \\ \groupdoubleline
\multicolumn{1}{c|}{} &
   &
  APs \cite{sanchez2021affective} &
  FRL \cite{bulat2022pre} &
  MAE-Face \cite{ma2024facial} &
  &
  &
  &
  &
  &
  & 
  \\ \hhline{~~||---||}

\multicolumn{1}{c|}{} &
  BP4D &
  \textbf{0.740} &
  0.719 &
  \textbf{0.740} &
  - &
  0.685 &
  - &
  0.546 &
  0.535 &
  - &
  \underline{0.707} \\
\multicolumn{1}{c|}{\multirow{-3}{*}{\begin{tabular}[c]{@{}c@{}}FAUE\\ (ICC↑)\end{tabular}}} &
  DISFA &
  0.580 &
  0.598 &
  0.674 &
  - &
  0.674 &
  - &
  \underline{0.753} &
  \textbf{0.885} &
  - &
  0.729 \\ \groupdoubleline
\multicolumn{1}{c|}{} &
  &
  DDAMFN \cite{zhang2023dual} &
  Poster++ \cite{mao2025poster++} &
  MHAN \cite{wang2026mhan} &
  &
  &
  &
  &
  & 
  \\ \hhline{~~||---||}

\multicolumn{1}{c|}{} &
  AffectNet-8 &
  \underline{64.24} &
  63.76 &
  \textbf{65.08} &
  30.35 &
  - &
  40.40 &
  42.15 &
  48.40 &
  53.83 &
  63.72 \\
\multicolumn{1}{c|}{} &
  RAF-DB &
  91.35 &
  92.21 &
  \underline{92.57} &
  52.67 &
  - &
  62.68 &
  82.79 &
  82.67 &
  55.22 &
  \textbf{93.22} \\
\multicolumn{1}{c|}{} &
  FER+ &
  90.74 &
  - &
  \underline{90.90} &
  56.77 &
  - &
  54.49 &
  66.10 &
  68.60 &
  45.35 &
  \textbf{92.23} \\
\multicolumn{1}{c|}{} &
  SFEW &
  - &
  - &
  - &
  27.48 &
  - &
  \underline{46.53} &
  36.88 &
  33.17 &
  34.41 &
  \textbf{62.38} \\
\multicolumn{1}{c|}{\multirow{-5}{*}{\begin{tabular}[c]{@{}c@{}}SFER\\ (ACC↑)\end{tabular}}} &
  AFFW-FER &
  - &
  - &
  - &
  32.03 &
  - &
  39.73 &
  53.21 &
  \underline{62.55} &
  32.55 &
  \textbf{70.92} \\ \groupdoubleline
\multicolumn{1}{c|}{} &
  &
  MAE-DFER \cite{sun2023mae} &
  SVFAP \cite{sun2024svfap} &
  S4D \cite{chen2025static} &
  &
  &
  &
  &
  & 
  \\ \hhline{~~||---||}

\multicolumn{1}{c|}{} &
  DFEW &
  74.43 &
  74.81 &
  \textbf{76.68} &
  - &
  - &
  - &
  - &
  - &
  - &
  \underline{76.51} \\
\multicolumn{1}{c|}{} &
  FERV39k &
  52.07 &
  52.29 &
  \underline{53.65} &
  - &
  - &
  - &
  - &
  - &
  - &
  \textbf{54.10} \\
\multicolumn{1}{c|}{\multirow{-4}{*}{\begin{tabular}[c]{@{}c@{}}DFER\\ (WAR↑)\end{tabular}}} &
  MAFW &
  \underline{54.31} &
  54.28 &
  \textbf{58.44} &
  - &
  - &
  - &
  - &
  - &
  - &
  48.45 \\ \groupdoubleline
\multicolumn{1}{c|}{} &
  &
  EMOFAN \cite{toisoul2021estimation} &
  TGCT \cite{li2025temporal} &
  MAVENC \cite{dresvyanskiy2024multi} &
  &
  &
  &
  &
  & 
  \\ \hhline{~~||---||}
\multicolumn{1}{c|}{} &
  AFFW-VA &
  - &
  - &
  \textbf{0.574} &
  - &
  - &
  - &
  - &
  - &
  - &
  \underline{0.469} \\
\multicolumn{1}{c|}{} &
  AFEW-VA &
  \textbf{0.675} &
  - &
  - &
  - &
  - &
  - &
  - &
  - &
  - &
  \underline{0.646} \\
\multicolumn{1}{c|}{} &
  SEWA &
  0.630 &
  \textbf{0.739} &
  - &
  - &
  - &
  - &
  - &
  - &
  - &
  \underline{0.735} \\ 
  
  \multicolumn{1}{c|}{\multirow{-4}{*}{\begin{tabular}[c]{@{}c@{}}VAIE\\ (CCC↑)\end{tabular}}} &
  AffectNet &
  \textbf{0.620} &
  0.565 &
  - &
  - &
  - &
  - &
  - &
  - &
  - &
  \underline{0.601} \\ \groupdoubleline
\multicolumn{1}{c|}{} &
  &
  FTF \cite{gkikas2023full} &
  MAAP \cite{gkikas2024multimodal} &
  PainFormer \cite{gkikas2025painformer} &
  &
  &
  &
  &
  & 
  \\ \hhline{~~||---||}

\multicolumn{1}{c|}{} &
  MIntPain &
  - &
  - &
  - &
  - &
  - &
  - &
  - &
  - &
  - &
  59.78 \\
  
  \multicolumn{1}{c|}{\multirow{-3}{*}{\begin{tabular}[c]{@{}c@{}}FPLE\\ (ACC↑)\end{tabular}}} &
  BioVid &
  72.75 &
  \textbf{77.10} &
  \underline{76.55} &
  - &
  - &
  - &
  - &
  - &
  - &
  76.40 \\ \groupdoubleline
\multicolumn{1}{c|}{} &
  &
  VAT$^\dagger$ \cite{liao2024open} &
  SVFAP \cite{sun2024svfap} &
  MGNN \cite{wu2025depmgnn}  &
  &
  &
  &
  &
  & 
  \\ \hhline{~~||---||}
\multicolumn{1}{c|}{\multirow{-2}{*}{\begin{tabular}[c]{@{}c@{}}FAPR\\ (CCC↑)\end{tabular}}} &
  First Impression &
  0.630 &
  0.665 &
  \underline{0.705} &
  - &
  - &
  - &
  - &
  - &
  - &
  \textbf{0.728} \\ \groupdoubleline
\multicolumn{1}{c|}{} &
  &
  MTB-DFE \cite{xu2024two} &
  MGNN \cite{wu2025depmgnn} &
  MTE-DFS \cite{liao2026explainable} &
  &
  &
  &
  &
  & 
  \\ \hhline{~~||---||}
\multicolumn{1}{c|}{\multirow{-2}{*}{\begin{tabular}[c]{@{}c@{}}FDA\\ (RMSE↓)\end{tabular}}} &
  AVEC 2014 &
  7.180 &
  \textbf{6.623} &
  6.890 &
  - &
  - &
  - &
  - &
  - &
  - &
  \textbf{6.623} \\ \Xhline{1.5px}
\end{tabular}
}
\label{tab:compare1}
\end{table*}

\subsection{Function-level comparison with SOTA methods and open-sourced toolkits}

Table \ref{tab:compare1} compares OpenAHBU with dataset-specific SOTA models and existing open-sourced AFBU toolkits. Unlike SOTA models, which are trained and optimised separately for each dataset, OpenAHBU models (except FAPR and FDA) are trained on heterogeneous datasets without dataset-specific fine-tuning. Despite this inherently favourable setting for \textbf{dataset-specific SOTA models}, OpenAHBU outperformed them on at least one dataset for FAUD, FAUE, SFER, and DFER tasks, achieving maximum relative performance gains of 3.9\% F1 on DISFA (FAUD), 8.2\% ICC on DISFA (FAUE), 1.5\% ACC on FER+ (SFER), and 0.8\% WAR on FERV39k (DFER). On most remaining datasets, it remains competitive, with relative performance drops below 0.1\% F1 for FAUD and 0.3\% WAR for DFER, where MAFW and AFFW-VA are the main exceptions, likely due to differences in annotation protocols, data distributions, acquisition conditions, and label noise. Nevertheless, our generic OpenAHBU models still generally achieved decent performances across most datasets, even compared to dataset-specific SOTA models. For the single-dataset FAPR and FDA models, OpenAHBU improved CCC by 3.26\% on First Impressions and matched the best reported RMSE on AVEC 2014, respectively.

\begin{figure}[htbp]
    \centering
    \includegraphics[width=1.0\linewidth]{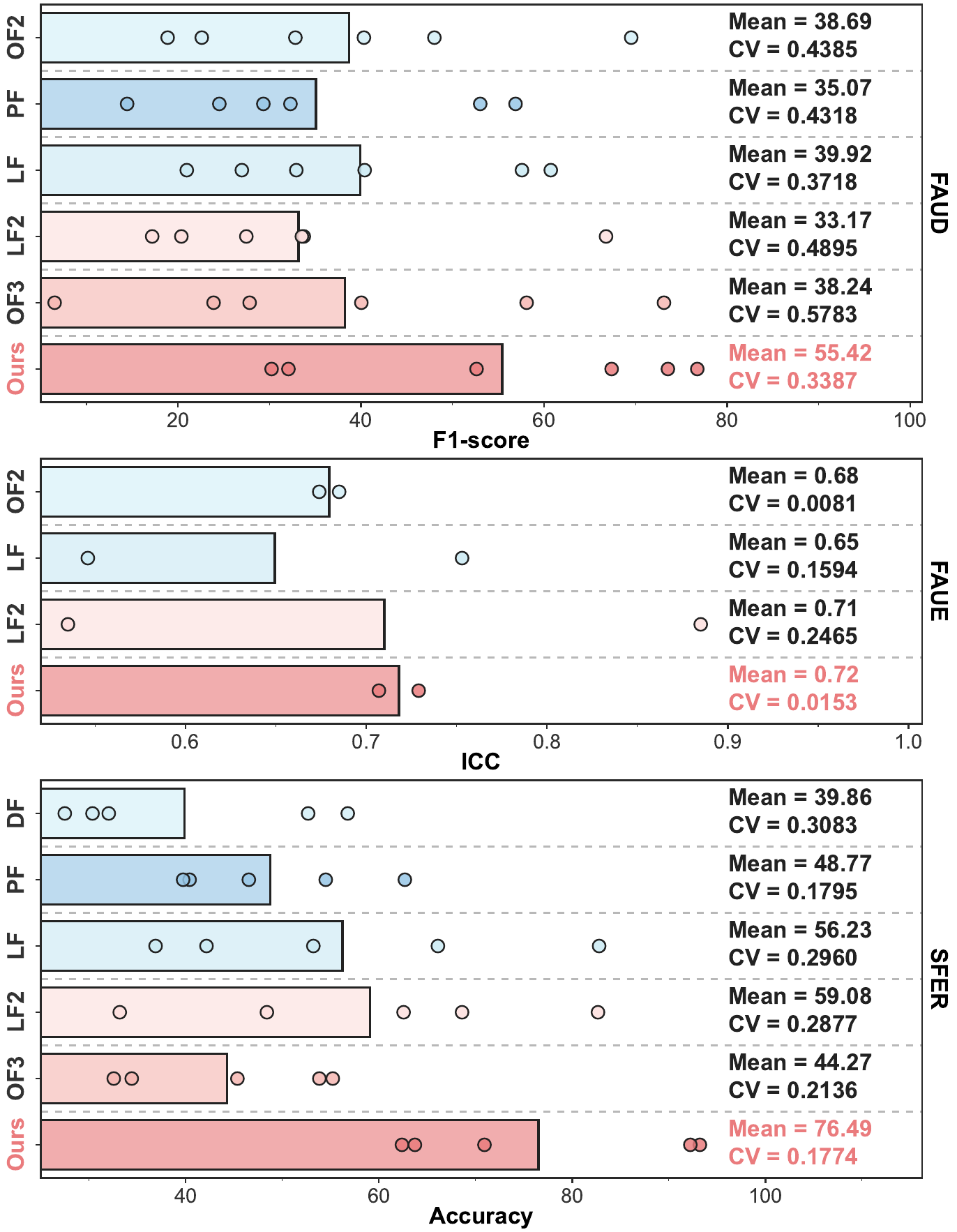}
    \caption{Cross-dataset performance distributions of OpenAHBU and existing open-sourced AFBU toolkits on FAUD, FAUE, and SFER, measured by F1 score, ICC, and accuracy, respectively. The coefficient of variation (CV) is defined as the ratio of the standard deviation to the mean performance across datasets, with a lower CV indicating more stable cross-dataset performance. Mean: average performance. Toolkit abbreviations are defined in Table \ref{tab:compare1}.}
    \label{fig:stas}
\end{figure}

As illustrated in Fig. \ref{fig:stas} and Table \ref{tab:compare1}, compared to existing \textbf{open-sourced AFBU toolkits} that can only understand frame-level/short-term observable facial behaviours, our OpenAHBU \textit{additionally covers FAPR, FRPR and FDA functions for understanding various human internal states reflected by relatively long-term facial behaviours, as well as additional frame-level VAIE and short-term DFER and FPLE functions.} 
Specifically, for the overlapping functions, the proposed OpenAHBU achieved superior performances on the majority of evaluated datasets. For SFER, it outperformed all compared toolkits throughout all five datasets, with large relative improvements ranging from 12.6\%–34.4\% to the second best toolkits, despite that the second best toolkit changes across datasets. For FAUD task, OpenAHBU lead on five out of the six datasets, with particularly large relative performance gains ranging from 18.9\% to 32.1\% on RAF-AU, CK+, CASME2, and AFFW-AU datasets, respectively. For FAUE task, it outperformed OpenFace 2.0 on both datasets by 3.2\% and 8.2\%, respectively, while exceeding LibreFace and LibreFace 2.0 by 29.5\% and 32.1\% on BP4D.
According to the unweighted function-level average performance, and the coefficient of variation (CV) measuring the standard deviation relative to the average performance, there is a superior balance between overall performance and cross-dataset stability for our OpenAHBU compared to other toolkits. Compared with the strongest toolkit's  mean performance, the proposed OpenAHBU improved the mean value by 38.8\% for FAUD and 36.0\% for SFER, while reducing the CV from 0.372 to 0.339 and from 0.296 to 0.177, respectively. For FAUE, OpenAHBU increased the best competing mean ICC by 5.7\%. Although its CV is higher than that of OpenFace 2.0, it remains approximately $1/10$ that of LibreFace, indicating that this gain is achieved without the clear cross-dataset fluctuation observed in LibreFace.

In summary, the proposed generic OpenAHBU delivered: (i) comparable or even better performances over the dataset-specific SOTA models on their corresponding datasets; and (ii)
broader functional coverage and higher average performances over all existing open-sourced toolkits on all shared functions, with the lowest CV for FAUD and SFER and the second-lowest for FAUE. These demonstrate not only its strong AFBU performance but also superior balance between functional accuracy and cross-dataset stability.

\subsection{Function-level OpenAHBU model performances}
\label{sec:4.1.3}

\begin{figure*}[htbp]
    \centering
    \includegraphics[width=1.0\linewidth]{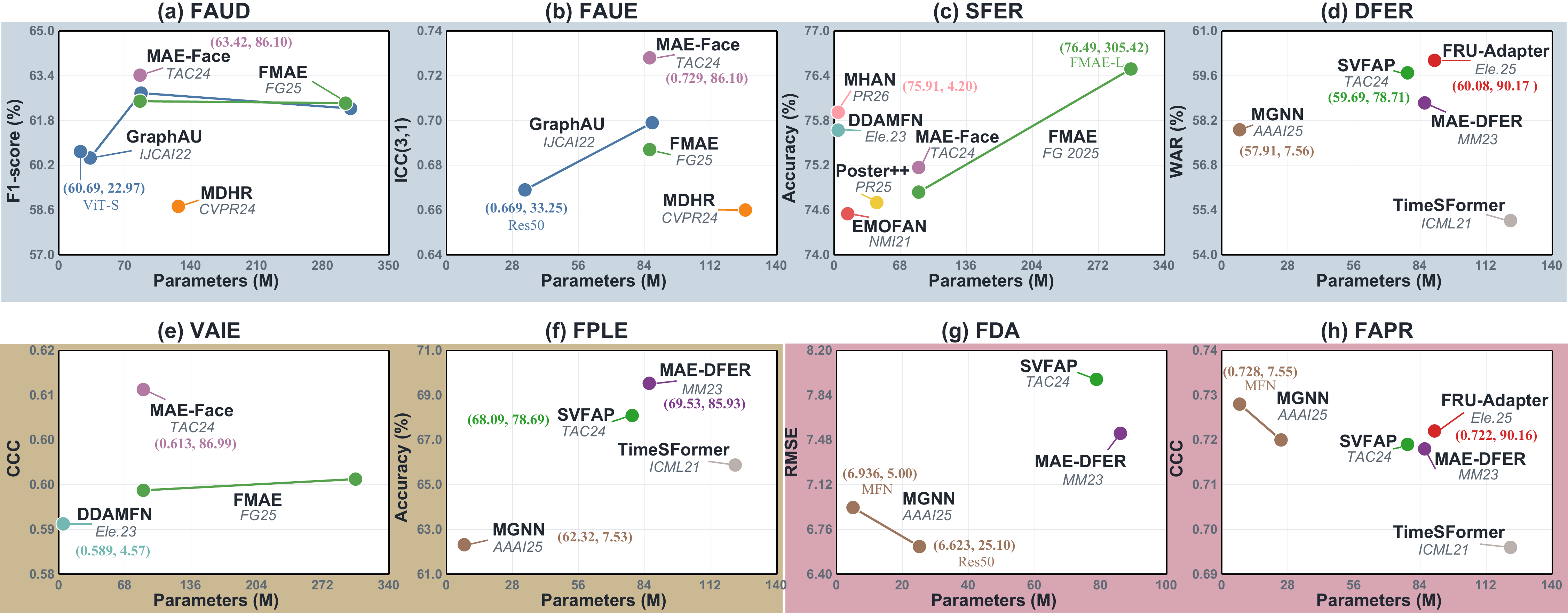}
    \caption{Function-level performances and parameter counts for OpenAHBU Stage I models. The primary metrics are F1 score for FAUD, ICC(3,1) for FAUE, accuracy for SFER and binary FPLE, WAR for DFER, CCC for VAIE and FAPR, and RMSE for FDA. Higher values indicate better performance except for FDA.}
    \label{fig:baseline_results}
\end{figure*}


Fig. \ref{fig:baseline_results} compares the effectiveness and efficiency (parameter counts) of all function-level models of every OpenAHBU function on the corresponding function-level test set. For the models developed on multiple datasets, their results on each individual dataset, together with FLOPs and inference throughput, are provided in the Supplementary Material.

\textbf{Short-term observable facial behaviour understanding.} Fig. \ref{fig:baseline_results}(a)-(d) reveal a trade-off in terms of model parameters and performances. The performances of SFER and DFER models are generally improved with their model scale increased. For SFER models, FMAE-L reached the highest average accuracy of 76.49\% with 305.42M parameters, whereas MHAN obtains a closely comparable 75.91\% accuracy using only 4.20M parameters, reducing the model size by approximately 73 times for a only 0.76\% accuracy drop. Regarding DFER models, FRU-Adapter leads with an average WAR of 60.08\% using 90.17M parameters, while SVFAP reducing the parameter count by 11.46M with only 0.65\% performance drop. MGNN (MFN) achieved 57.91\% WAR with merely 7.56M parameters, making it the most economical DFER configuration. 
In contrast, increasing the number of parameters yields no consistent benefits to FAUD and FAUE models or even reduces their performances, possibly due to the fact that these models are optimised to focus on patterns describing more frequently activated AUs rather than less appeared AUs, i.e., the unbalanced AU activation in the training set leads the trained FAUD and FAUE models to overfit to the more frequently activated AUs but underfit to the less frequently activated AUs, which is more pronounced in models with larger parameters 
(detailed analysis is provided in Secs. 6.1.1 and 6.1.2 of the Supplementary Material). In this sense, large models for FAUE such as GraphAU (ViT-L) and FMAE-L are thus not retained in the OpenAHBU model suite.
More specifically, MAE-Face model achieved both the highest average FAUD F1 score of 63.42\% and the highest average FAUE ICC of 0.729 with 86.10M parameters, respectively, while GraphAU (ViT-S) and GraphAU (ResNet-50) provided the smallest configuration with 22.97M and 33.25M parameters for FAUD and FAUE, reporting 60.69\% F1 score and 0.669 ICC. Here, the largest FAUD model (GraphAU (ViT-L) of 309.11M parameters) and FAUE model (MDHR (ResNet-50) of 126.66M parameters) in our OpenAHBU only yielded 62.23\% F1 score and 0.660 ICC, respectively.

\textbf{Short-term facial behaviour-related human internal state understanding.} Fig. \ref{fig:baseline_results}(e) and (f) show a broadly positive relationship between parameters and performances for VAIE and FPLE models, with FPLE models benefiting more clearly from the increased parameters while VAIE models exhibiting only modest performance gains. Specifically, for FPLE models, MAE-DFER obtained the highest average accuracy of 69.53\% with 85.93M parameters, while MGNN (MFN) provided the most parameter-efficient configuration, achieving 62.32\% with only 7.53M parameters. For VAIE, MAE-Face achieved the highest average CCC of 0.613 with 86.99M parameters, whereas DDAMFN achieved a competitive CCC of 0.589 using only 4.57M parameters, approximately $1/20$ of MAE-Face's parameters. These modest benefits brought by parameter scaling for VAIE models may caused by the different model convergence rates in valence and arousal learning, which leads models with larger parameters to quickly overfit to one affective states before the other is adequately fit.

\textbf{Long-term facial behaviour-related human psychological state understanding.} Fig. \ref{fig:baseline_results}(g) and (h) show that the increased model parameters does not consistently improve FDA and FAPR's performances, where MGNN models providing the most favourable accuracy-efficiency trade-offs for both functions. In terms of FDA models, MGNN (ResNet-50) achieved the lowest RMSE of 6.623 with 25.10M parameters, while MGNN (MFN) reduced the number of parameters by 80.1\% (5.00M) while only increasing 0.313 RMSE, making it a more balanced model between parameters and performance. For FAPR models, MGNN (MFN) is not only the smallest model but also the best-performing model, achieving a CCC of 0.728 with only 7.55M parameters, while TimeSFormer with the most parameters (122.34M) only reported 0.696 CCC. Here, the slight difference between the number of parameters in the MGNN (MFN) models used for FDA and FAPR arises from function-specific configurations (e.g., the regression head).

\subsection{Synergistic multi-task enhancement (SMTE) results}

\begin{figure}[htbp]
    \centering
    \includegraphics[width=1.0\linewidth]{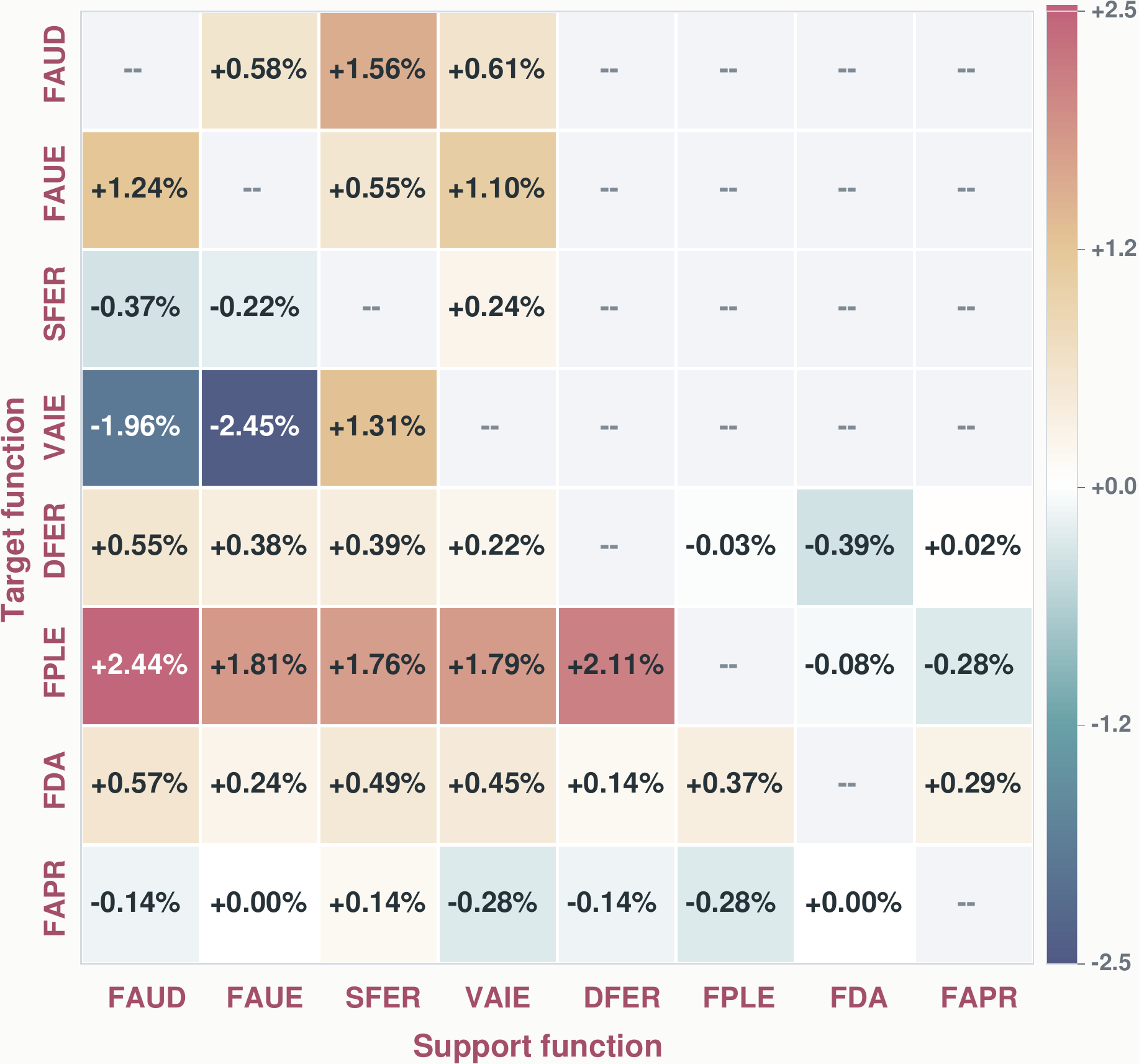}
    \caption{Relative performance change (\%) caused by SMTE. Metrics employerd here are consistent with Fig. \ref{fig:baseline_results}.}
    \label{fig:au_stas}
\end{figure}

Fig. \ref{fig:au_stas} evaluates the SMTE module across the eight re-developed OpenAHBU functions, where SMTE effectively exploits complementary facial behaviours across AFBU functions to refine 7 out of 8 OpenAHBU functions, with minimal additional computational overhead. Additional analysis and results are detailed in Sec.6.2 of the Supplementary Material.


\textbf{Frame-level AFBU prediction enhancement.} Every frame-level predictions provided by other frame-level OpenAHBU functions can benefit the best FAUD and FAUE models. In particular, predictions of the best SFER model provided the largest performance gain for the best FAUD model (+1.56\%), highlighting global facial contexts of the predicted categorical facial expressions in complementing AU occurrence detection. Meanwhile, FAUD and VAIE predictions produced the two largest performance gains (+1.24\% and +1.10\%) for supporting the best FAUE model. This indicates that FAUD predictions are effective guidance for FAUE as they explicitly indicate which AUs are activated, while global facial affect intensities supplied by VAIE predictions also providing complementary cues for AU intensity estimation.
In contrast, the best-performing SFER and VAIE models only benefit from each other's predictions, with SFER predictions improving the best VAIE model by 1.31\% while VAIE predictions improving the best SFER model by 0.24\%, showing that they are strongly associated and contain informative complementary cues.
Interestingly, predictions from both the best FAUD and FAUE models reduce the performance of the best SFER and VAIE models, possibly because these strong target models already capture most AU-related facial patterns, leaving limited complementary information for further enhancement. 
Further experimental evidence is provided in Sec. 6.2.4 and 6.2.5 of the Supplementary Material.

\textbf{Video-level AFBU prediction enhancement via frame-level AFBU predictions.} The best-performing DFER, FPLE, and FDA models all benefit, to varying degrees, from incorporating prediction sequences generated by every frame-level OpenAHBU function, indicating that the frame-level local AU occurrences/intensities, categorical facial expressions, and continuous valence-arousal states provided by OpenAHBU can supply complementary and informative fine-grained local facial cues that are ignored by such video-level models. In particular, FAUD predictions provide the largest improvements for the best DFER, FPLE, and FDA models, increasing WAR by 0.55\% and accuracy by 2.44\%, while reducing RMSE by 0.57\%.
In contrast, FAPR benefits only marginally from SFER, with a CCC improvement of 0.14\%, possibly because its MGNN (MFN) baseline already captures sufficiently rich behavioural representations, leaving limited complementary information to be provided by additional frame-level predictions.

\textbf{Video-level AFBU prediction enhancement via video-level AFBU predictions.} As observed in Fig. \ref{fig:au_stas}, DFER predictions improved the accuracy of the best FPLE model by 2.11\%, while binary FPLE predictions failed to enhance the best DFER model. This indicates that temporal facial expression predictions are informative indicator for pain states but pain states not necessarily indicate specific facial expressions. Long video-based FDA and FAPR predictions likewise provide little or no benefit to both short video-based AFBU functions, as long-term psychological states may contain various short-term facial behaviours. 
In contrast, the best FDA model is enhanced by every other video-level predictions, with DFER, FPLE, and FAPR predictions reducing its RMSE by 0.14\%, 0.37\%, and 0.29\%, respectively, possibly due to the fact that depression can be reflected by sustained changes in dynamic expressions \cite{fisher2024facing}, pain-related behaviours \cite{bair2003depression}, and is associated with personality \cite{jaiswal2019automatic}. In contrast, none of DFER, FPLE, or FDA improved FAPR, as people of different personalities can show similar short-term facial expression, pain states and long-term depression states.

\section{Conclusion and Discussion}


\textbf{Conclusion.} This paper presents an open-sourced and generalisable OpenAHBU toolkit which includes 11 functions for comprehensive understanding six types of observable frame-level facial behaviours, two short-term facial affective states, and three long-term facial behaviour-related psychological states, where eight functions are systematically re-developed.
To achieve them, the proposed OpenAHBU followed a unified protocol to train at least four models of different parameter scales on a combination of multiple heterogeneous academically available datasets for each function (except FDA and FAPR), 
enabling users to flexibly select models under different computational resources and application scenarios. Extensive evaluations demonstrate that the proposed OpenAHBU not only provides broader AFBU functional coverage over existing open-sourced AFBU toolkits but also achieved superior average performances on all shared functions. Its generic models reports comparable or even enhanced performances over existing dataset-specific SOTA models on their corresponding datasets without dataset-specific fine-tuning. Moreover, the proposed SMTE module further adaptively exploits complementary cues across functions to consistently enhance their performances with minimal computational overhead. We not only fully release training and inference code as well as all pre-trained model weights of OpenAHBU to support further research and development, but also offer a stand-alone GUI for code-free deployment on mainstream operating systems (i.e., Windows, Linux, and MacOS) as well as a Python library installable via pip for customised integration and extension.

\textbf{Limitations and future work.} A key limitation of our OpenAHBU is that it does not include AFBU functions without standardised and academically available datasets or reproducible DL methods. Moreover, FRPR is currently included only as a research-oriented function and has not been re-developed under the unified OpenAHBU protocol. Given the rapid and continuous emergence of new models while many lack sufficient details for reliable reproduction, another limitation is that it does not include all recently developed function-specific models. In addition, the proposed OpenAHBU only includes functions for facial behaviour understanding, ignoring behaviours of other modalities.
Moreover, the SMTE module currently limits each target function to its most effective single support function, as jointly incorporating multiple support functions yielded no consistent performance improvements in our experiments but increasing memory use and computational cost. Furthermore, evaluating SMTE across all models would require a prohibitively large number of experiments, and thus it has not been exhaustively validated across all re-developed model configurations, leaving its effectiveness across the remaining models unverified.
Future work will extend OpenAHBU to include not only additional AFBU functions but also functions for automatically understanding human behaviours of other modalities as well as multi-modal human behaviour understanding, conditioned on the availability of standardised datasets. For current functions, we will continuously develop, update, and integrate more advanced function-level models, while investigating more efficient and effective mechanisms for SMTE and validating their effectiveness across a broader range of model configurations. We further aim to build an open-sourced community around the proposed OpenAHBU through standardised model interfaces, contribution guidelines, and evaluation protocols, enabling researchers from all over the world to contribute new functions, datasets, and models while collaboratively improving the codebase.

\appendix[Supplementary Material for OpenAHBU]

\section{Datasets Details}
\label{datasets}

\subsection{Facial AU Occurrence Detection (FAUD)}
Six datasets containing AUs recorded under controlled conditions and wild conditions have been used to develop the FAUD function of the proposed OpenAHBU, all of which provide per image or frame-level Facial Action Coding System (FACS) annotations. Their details are provided below:


\textbf{BP4D \cite{zhang2014bp4d}.} The BP4D spontaneous dataset was collected under controlled laboratory conditions using a 3D face-capturing system. Each participant completed eight tasks designed to elicit spontaneous facial behaviours associated with amusement, sadness, surprise, embarrassment, fear, pain, anger, and disgust. The dataset contains synchronised 2D texture videos and dynamic 3D facial data from 41 participants, including 23 women and 18 men, resulting in 328 recorded sequences and approximately $140,000$ AU-annotated frames. Selected expressive segments were manually coded frame by frame by two certified FACS annotators. For FAUD, OpenAHBU uses the 2D facial frames and the occurrence annotations of AUs 1, 2, 4, 6, 7, 10, 12, 14, 15, 17, 23, and 24. Following the subject-independent three-fold partition listed in Table \ref{tab:au_partition} \cite{luo2022learning}, OpenAHBU uses Fold 1 exclusively for evaluation, while the samples from Folds 2 and 3 are randomly divided into training and validation subsets at an 8:2 ratio.

\textbf{DISFA \cite{mavadati2013disfa}.} The Denver Intensity of Spontaneous Facial Action (DISFA) dataset was collected under uniform illumination while 27 adults, including 12 women and 15 men, watched a four-minute video assembled to elicit spontaneous facial expressions. The stereo facial recordings were captured at 20 frames per second with a resolution of $1024\times768$, producing $4,845$ frames per participant and approximately $130,000$ frames in total. A certified FACS annotator manually assigned each frame an intensity score from 0 to 5 for 12 AUs. A second certified FACS annotator additionally coded recordings from ten randomly selected participants to assess annotation agreement. Following the commonly adopted FAUD protocol \cite{luo2022learning, wang2024multi, ma2024facial}, OpenAHBU treats intensity levels 0 and 1 as absence and levels 2–5 as occurrence, and evaluates AUs 1, 2, 4, 6, 9, 12, 25, and 26. OpenAHBU adopts the subject-independent three-fold partition listed in Table \ref{tab:au_partition} \cite{luo2022learning}, using Fold 1 exclusively for evaluation and randomly dividing the samples from Folds 2 and 3 into training and validation subsets at an 8:2 ratio.

\textbf{RAF-AU \cite{yan2020raf}.} RAF-AU was constructed from facial images collected from the Internet. It extends the Real-world Affective Faces Multi-Label (RAF-ML) dataset \cite{li2019blended} and contains $4,601$ facial images. These images are accompanied by subjective emotion judgements obtained through crowdsourcing. Two experienced FACS annotators independently coded the facial images and resolved disagreements through discussion. OpenAHBU uses the occurrence annotations of AUs 1, 2, 4, 5, 6, 9, 10, 12, 16, 17, 25, 26, and 27, with fixed training, validation, and test partitions containing $3,031$, $433$, and $863$ images, respectively. The images exhibit substantial variations in subject identity, head pose, illumination, occlusion, and facial behaviour.

\textbf{CK+ \cite{lucey2010extended}.} The Extended Cohn-Kanade (CK+) dataset was collected under controlled conditions, with participants instructed to perform a series of facial displays involving individual AUs and their combinations. It contains 593 posed facial expression sequences from 123 participants. Each image sequence consists of 10 to 60 frames and records the progression from a neutral onset to a peak facial display. The peak frame of every posed sequence is fully coded according to FACS, with approximately 15\% of the sequences additionally coded by a second certified FACS annotator. For FAUD, OpenAHBU uses the available peak frames and the occurrence annotations of AUs 1, 2, 4, 5, 6, 7, 12, 17, and 25. The fixed training, validation, and test partitions used by OpenAHBU contain 384, 59, and 145 images, respectively.

\textbf{CASME2 \cite{yan2014casme}.} CASME II was collected under controlled illumination while participants watched emotion-eliciting video clips and were instructed to suppress their facial expressions. It contains 247 spontaneous micro-expression sequences from 26 participants, selected from nearly $3,000$ elicited facial movements. The sequences were captured at 200 frames per second, with the facial region occupying approximately $280\times340$ pixels, enabling subtle and short-duration facial movements to be observed. Each sequence is provided with onset, apex, and offset frames together with AU and emotion annotations. Two coders marked AUs according to FACS and assigned emotion labels based on the AUs, participants’ self-reports, and the contents of the video clips, with disagreements resolved through discussion. OpenAHBU uses seven annotated AU categories: AUs 1, 2, 4, 7, 12, 14, and 17. Its fixed training, validation, and test partitions contain $8,729$, $1,995$, and $5,995$ facial frames, respectively.

\textbf{AFFW-AU \cite{kollias2018aff}.} AFFW-AU denotes the AU-annotated subset of Aff-Wild2, a large-scale audiovisual database assembled from YouTube videos recorded in unconstrained environments. For FAUD, OpenAHBU uses $1,766,225$ AU-annotated facial frames. The release adopted by OpenAHBU provides frame-level occurrence annotations for 12 AUs, namely AUs 1, 2, 4, 6, 7, 10, 12, 15, 23, 24, 25, and 26. These annotations are obtained through a semi-automatic procedure combining manual and automatic annotations. OpenAHBU divides the released training portion into $1,056,304$ training and $264,076$ validation frames and retains the $445,845$ frames from the official validation portion only for final evaluation. The videos cover substantial variations in subject age and ethnicity, head pose, illumination, occlusion, recording environment, and facial behaviour.


\begin{table*}[htbp]
\large
\centering
\setlength{\doublerulesep}{0.3em}
\renewcommand{\arraystretch}{1.5}
\caption{Subject composition of the three-fold partitions adopted for BP4D and DISFA.}
\resizebox{1.0\linewidth}{!}{
\begin{tabular}{c|ccccccc||c|ccccccc}
\Xhline{2px}
\rowcolor{gray!15}
BP4D & \multicolumn{7}{c||}{\cellcolor{gray!15}Three-Fold Data Partition} & DISFA & \multicolumn{7}{c}{\cellcolor{gray!15}Three-Fold Data Partition} \\ \hhline{--------||--------}
& F003 & F005 & F011 & F013 & F020 & F022 & M002 &                    & SN001 & SN002 & SN009 & SN010 & SN016 & SN026 & SN027 \\
\multirow{-2}{*}{Fold-1} 
& M005 & M010 & M011 & M013 & M016 & M017 & M018 
& \multirow{-2}{*}{Fold-1} 
& SN030 & SN032 &       &       &       &       &       \\
\hhline{--------||--------}
& F004 & F006 & F007 & F012 & F014 & F015 & F017 &                    & SN003 & SN004 & SN005 & SN007 & SN008 & SN017 & SN023 \\
\multirow{-2}{*}{Fold-2} 
& F019 & F021 & M003 & M006 & M009 & M015 &      
& \multirow{-2}{*}{Fold-2} 
& SN025 & SN029 &       &       &       &       &       \\
\hhline{--------||--------}
& F001 & F002 & F008 & F009 & F010 & F016 & F018 &                    & SN006 & SN011 & SN012 & SN013 & SN018 & SN021 & SN024 \\
\multirow{-2}{*}{Fold-3} 
& F023 & M001 & M004 & M007 & M008 & M012 & M014 
& \multirow{-2}{*}{Fold-3} 
& SN028 & SN031 &       &       &       &       &       \\ \Xhline{2px}
\end{tabular}
}
\label{tab:au_partition}
\end{table*}

\begin{figure}[htbp]
    \centering
    \includegraphics[width=1.0\linewidth]{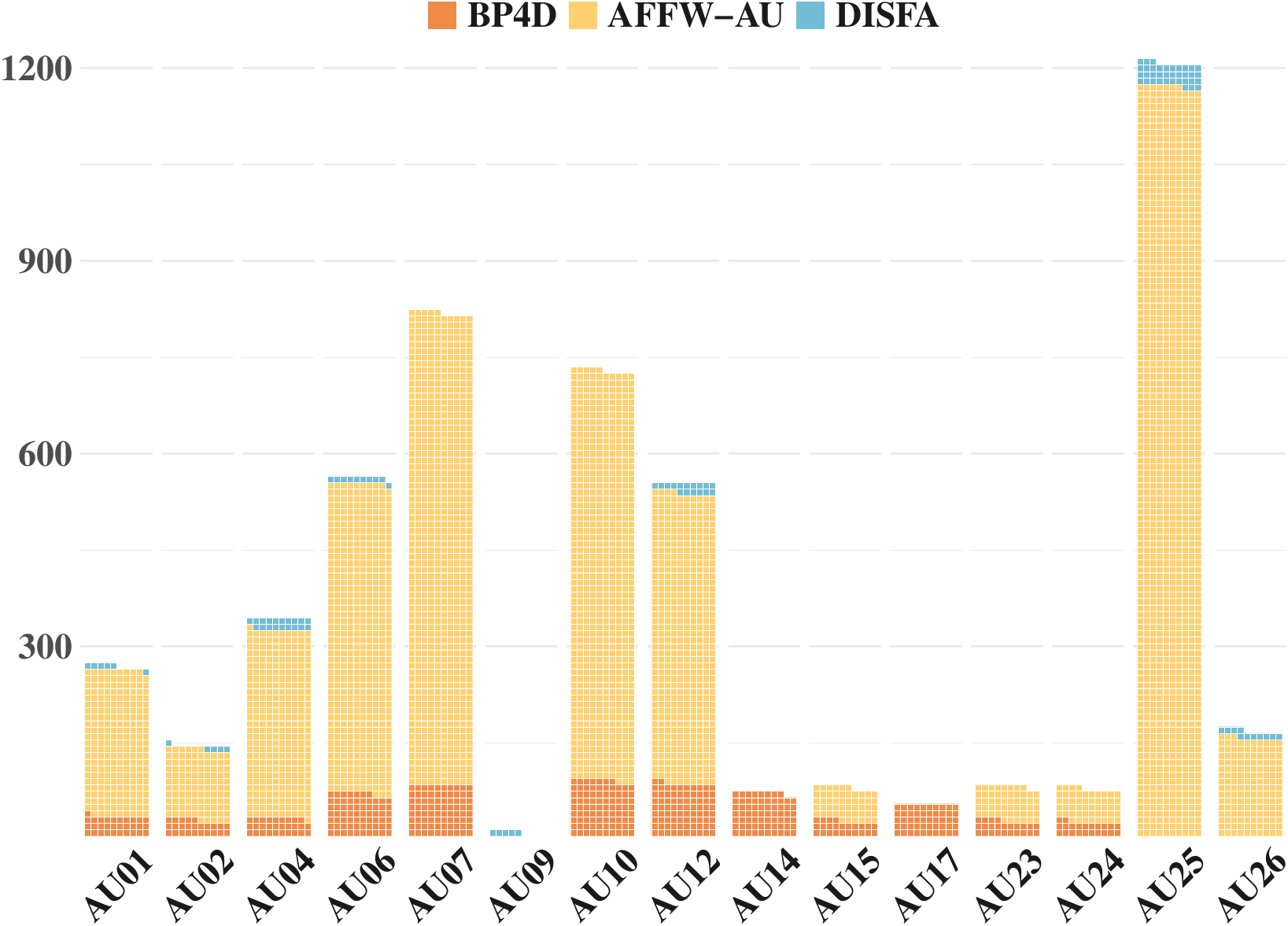}
    \caption{Distribution of positive samples for individual AUs across BP4D, AFFW-AU, and DISFA. Example counts are reported in thousands.}
    \label{fig:au_stas}
\end{figure}

\begin{figure}[htbp]
    \centering
    \includegraphics[width=1.0\linewidth]{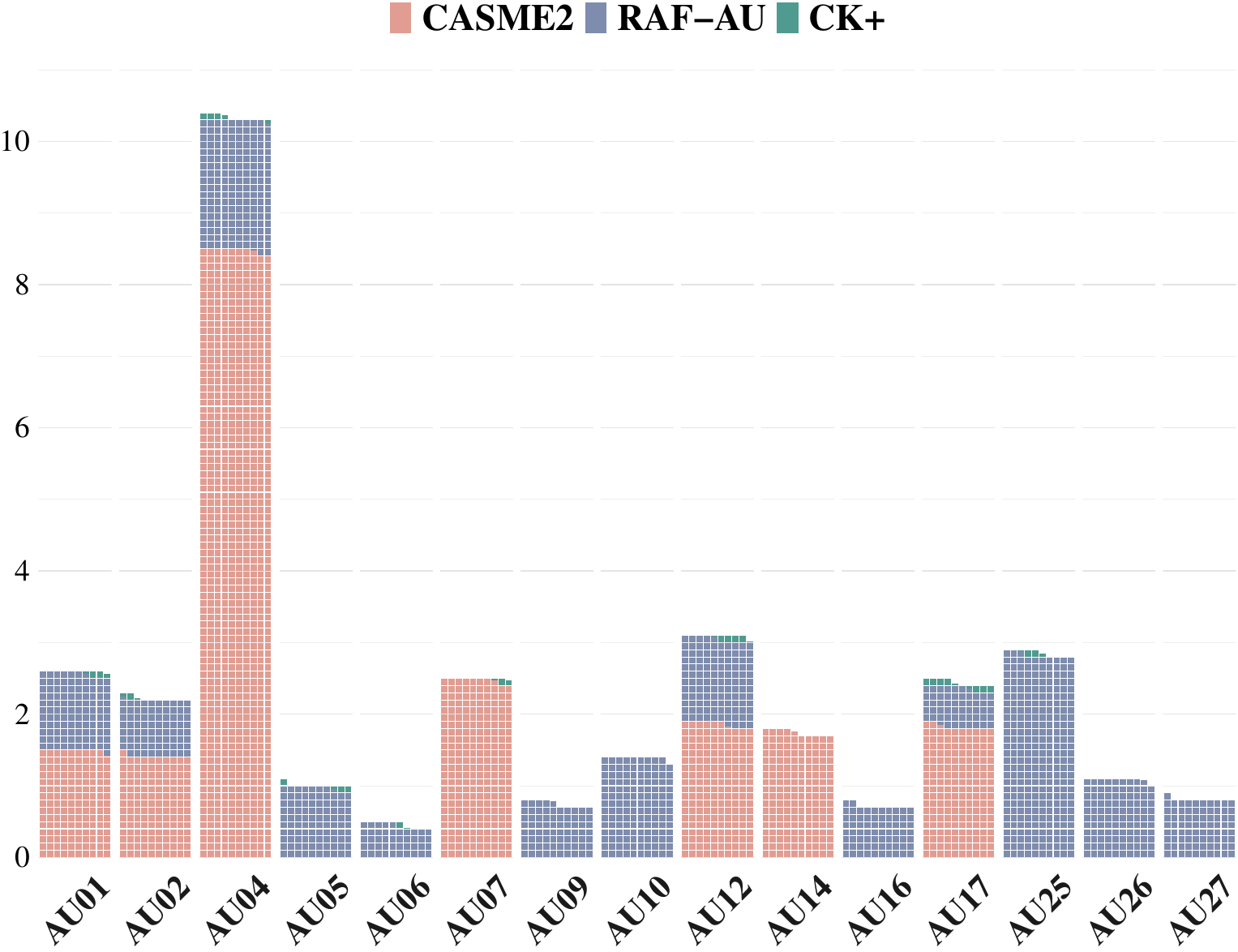}
    \caption{Distribution of positive samples for individual AUs across CASME2, RAF-AU, and CK+. Example counts are reported in thousands.}
    \label{fig:au_stas2}
\end{figure}

Fig. \ref{fig:au_stas} and Fig. \ref{fig:au_stas2} further compare the numbers of positive examples for individual AUs. The datasets are presented in two groups because their substantial differences in data example numbers would otherwise obscure the AU activation distributions of the smaller datasets when displayed on a common scale. Both distributions reveal substantial imbalance at both the AU and dataset levels: AU25 has approximately 220 times as many positive samples as AU9, while AFFW-AU contributes substantially more positive samples than BP4D and DISFA for most shared AUs. These imbalances are addressed through the dual-level resampling strategy (Sec. \ref{sec: implementation details}) adopted during model development.

\subsection{Facial AU Intensity Estimation (FAUE)}

Only two AU datasets, BP4D and DISFA, includes AU intensity labels for developing FAUE function of the OpenAHBU. These two controlled datasets contain spontaneous facial behaviours with frame-level FACS intensity annotations. While both datasets represent AU intensity using ordered levels, they differ in their elicitation procedures and annotated AU sets, as detailed below:


\textbf{BP4D \cite{zhang2014bp4d}.} The BP4D spontaneous dataset was collected under controlled laboratory conditions using a 3D face-capturing system, with each participant completing eight tasks designed to elicit spontaneous facial behaviours. It contains synchronised 2D texture videos and dynamic 3D facial data from 41 participants, including 23 women and 18 men, resulting in 328 recorded sequences. In addition to AU occurrence labels, BP4D provides frame-level FACS intensity annotations for selected AUs, which were manually assigned by trained FACS annotators to selected expressive segments. The five increasing FACS intensity levels are represented numerically from 1 to 5 in OpenAHBU, while 0 indicates AU absence. Following the commonly adopted five-AU intensity estimation setting \cite{sanchez2021affective, bulat2022pre, ma2024facial}, OpenAHBU uses the 2D facial frames and the intensity annotations of AUs 6, 10, 12, 14, and 17. Following the subject-independent three-fold partition listed in Table \ref{tab:au_partition} \cite{luo2022learning}, OpenAHBU uses Fold 1 exclusively for evaluation, while the samples from Folds 2 and 3 are randomly divided into training and validation subsets at an 8:2 ratio. The resulting training, validation, and test partitions contain $77,332$, $19,333$, and $50,182$ frames, respectively.

\textbf{DISFA \cite{mavadati2013disfa}.} The Denver Intensity of Spontaneous Facial Action (DISFA) dataset was collected under uniform illumination while 27 adults, including 12 women and 15 men, watched a four-minute video assembled to elicit spontaneous facial expressions. The stereo facial recordings were captured at 20 frames per second with a resolution of $1024 \times 768$, producing $4,845$ frames per participant and approximately $130,000$ frames in total. DISFA provides frame-level intensity annotations for 12 AUs, with 0 indicating absence and values from 1 to 5 representing progressively stronger activation. These annotations were manually assigned by a certified FACS annotator throughout all recorded frames. A second certified FACS annotator additionally coded recordings from ten randomly selected participants to assess annotation agreement. OpenAHBU uses the complete set of intensity-labelled AUs: AUs 1, 2, 4, 5, 6, 9, 12, 15, 17, 20, 25, and 26. OpenAHBU adopts the subject-independent three-fold partition listed in Table \ref{tab:au_partition} \cite{luo2022learning}, using Fold 1 exclusively for evaluation and randomly dividing the samples from Folds 2 and 3 into training and validation subsets at an 8:2 ratio. The resulting training, validation, and test partitions contain $69,768$, $17,442$, and $43,604$ frames, respectively.

\subsection{Static Facial Expression Recognition (SFER)}

Nine datasets are employed for developing the SFER function of our OpenAHBU, including six in-the-wild datasets (AffectNet-8 \cite{mollahosseini2017affectnet}, ExpW \cite{zhang2018facial}, RAF-DB \cite{shan2018reliable}, SFEW \cite{dhall2015video}, FER+ \cite{BarsoumICMI2016}, and AFFW-FER \cite{kollias2018aff}) and three controlled datasets (CK+ \cite{lucey2010extended}, RAFD \cite{langner2010presentation}, and KDEF \cite{lundqvist1998karolinska}). These datasets provide categorical facial expression annotations for neutral, happiness, sadness, surprise, fear, anger, disgust, and, where available, contempt, despite that the annotated category set varies across datasets. Their details are provided below:

\textbf{AffectNet-8 \cite{mollahosseini2017affectnet}.} AffectNet was collected from the Internet through Google, Bing, and Yahoo using $1,250$ emotion-related keywords in six languages. It contains more than one million facial images, approximately $450,000$ of which were manually annotated by a team of twelve trained annotators. The annotations include categorical expressions and continuous valence and arousal values, with each image initially assigned to one annotator. AffectNet-8 uses eight expression categories: neutral, happiness, sadness, surprise, fear, anger, disgust, and contempt. It contains $287,651$ training images and a balanced validation set of $4,000$ images, with 500 images per category. OpenAHBU divides the official training set into 230,120 training and 57,531 validation images and uses the official validation set only for final evaluation.

\textbf{ExpW \cite{zhang2018facial}.} The Expression in-the-Wild (ExpW) dataset was collected through Google image searches using combinations of emotion-related terms and occupation nouns. Face regions were extracted from the retrieved images using a face detector, resulting in $91,793$ facial images. Each detected face was manually assigned to one of seven categories: neutral, happiness, sadness, surprise, fear, anger, or disgust, while non-face images were removed during annotation. Following the unified face-preprocessing pipeline, OpenAHBU retains $87,305$ images and uses them only for training to increase the diversity of in-the-wild facial expressions.

\textbf{RAF-DB \cite{shan2018reliable}.} The Real-world Affective Faces Database (RAF-DB) was constructed from facial images collected from the Internet and contains approximately $30,000$ images. Each image was independently annotated by approximately 40 crowd-sourced annotators, with an expectation-maximisation algorithm used to estimate the expression labels from their judgements. The resulting annotations were organised into subsets for seven basic expressions and 11 compound expressions. OpenAHBU uses the seven-class basic-expression subset, which contains an official training set of $12,271$ images and a test set of $3,068$ images. The official training set is further divided into $9,817$ training and $2,454$ validation images, while the official test set is retained for final evaluation. The images cover substantial variations in subject age and ethnicity, head pose, and illumination.

\textbf{CK+ \cite{lucey2010extended}.} The Extended Cohn-Kanade (CK+) dataset was collected under controlled conditions, with participants instructed to perform facial displays involving individual AUs and their combinations. It contains 593 posed facial-expression sequences from 123 participants, with each sequence progressing from a neutral onset to a peak facial display. The peak frames were fully coded according to FACS, and expression labels were assigned by comparing the AU annotations with FACS-based emotion criteria and visually inspecting the corresponding sequences. Among these sequences, 327 satisfy the criteria for one of seven discrete expressions: happiness, sadness, surprise, fear, anger, disgust, or contempt. For SFER, OpenAHBU uses the neutral onset and expression-labelled peak frames, yielding 920 images across neutral and the seven expression categories, all of which are included only in training.

\textbf{RAFD \cite{langner2010presentation}.} The Radboud Faces Database (RaFD) was collected under controlled conditions, with subjects trained by FACS coders to display prototypical facial expressions. Each expression was displayed with three gaze directions and photographed simultaneously from five camera angles. The released database contains $8,040$ images of 67 subjects. Its expression labels correspond to eight instructed categories: neutral, happiness, sadness, surprise, fear, anger, disgust, and contempt. OpenAHBU uses the $1,608$ frontal-view images covering all eight expressions and three gaze directions and includes them only in training.

\textbf{SFEW \cite{dhall2015video}.} Static Facial Expressions in the Wild 2.0 (SFEW 2.0) was constructed from AFEW 5.0 movie clips by selecting representative frames through facial-landmark clustering. After unified face preprocessing, OpenAHBU retains $1,257$ facial images from the released training and validation sets. These images are annotated with neutral, happiness, sadness, surprise, fear, anger, or disgust. The 853 images retained from the released training set are divided into 682 training and 171 validation images, while the 404 images retained from the released validation set are used only for final evaluation. SFEW 2.0 serves as the image-based benchmark of the EmotiW 2015 challenge. 

\textbf{FER+ \cite{BarsoumICMI2016}.} FER+ was constructed by re-annotating the in-the-wild facial images of FER2013, originally collected through web image searches using emotion-related keywords. The source dataset contains $35,887$ greyscale facial images with a resolution of $48 \times 48$ pixels. Each image was independently labelled by ten annotators using eight expression categories: neutral, happiness, sadness, surprise, fear, anger, disgust, and contempt, resulting in a distribution of expression votes for each image. The released annotations additionally contain vote counts for unknown expressions and non-face images. OpenAHBU follows the released training, validation, and test organisation, retaining $25,060$, $3,191$, and $3,153$ images, respectively.

\textbf{KDEF \cite{lundqvist1998karolinska}.} The Karolinska Directed Emotional Faces (KDEF) dataset was collected under controlled lighting, with participants receiving written instructions and rehearsing the expressions before photography. Each individual displayed every expression twice, with each display photographed from five viewing angles. The dataset contains $4,900$ images of 70 individuals, comprising 35 women and 35 men. Its expression labels correspond to seven instructed categories: neutral, happiness, sadness, surprise, fear, anger, and disgust. OpenAHBU retains $4,888$ images after face preprocessing and includes all of them only in training.

\textbf{AFFW-FER \cite{kollias2018aff}.} AFFW-FER denotes the expression-annotated portion of Aff-Wild2, a large-scale audiovisual database assembled from YouTube videos recorded under unconstrained conditions. For SFER, OpenAHBU uses $589,253$ facial frames. The database provides frame-level annotations for neutral, happiness, sadness, surprise, fear, anger, and disgust. These expressions were manually annotated by human experts, with agreement among annotators used to select the retained labels. OpenAHBU divides the released training portion into $332,758$ training and $83,190$ validation frames and retains the $173,305$ frames from the official validation portion only for final evaluation.

\begin{figure}[htbp]
    \centering
    \includegraphics[width=0.9\linewidth]{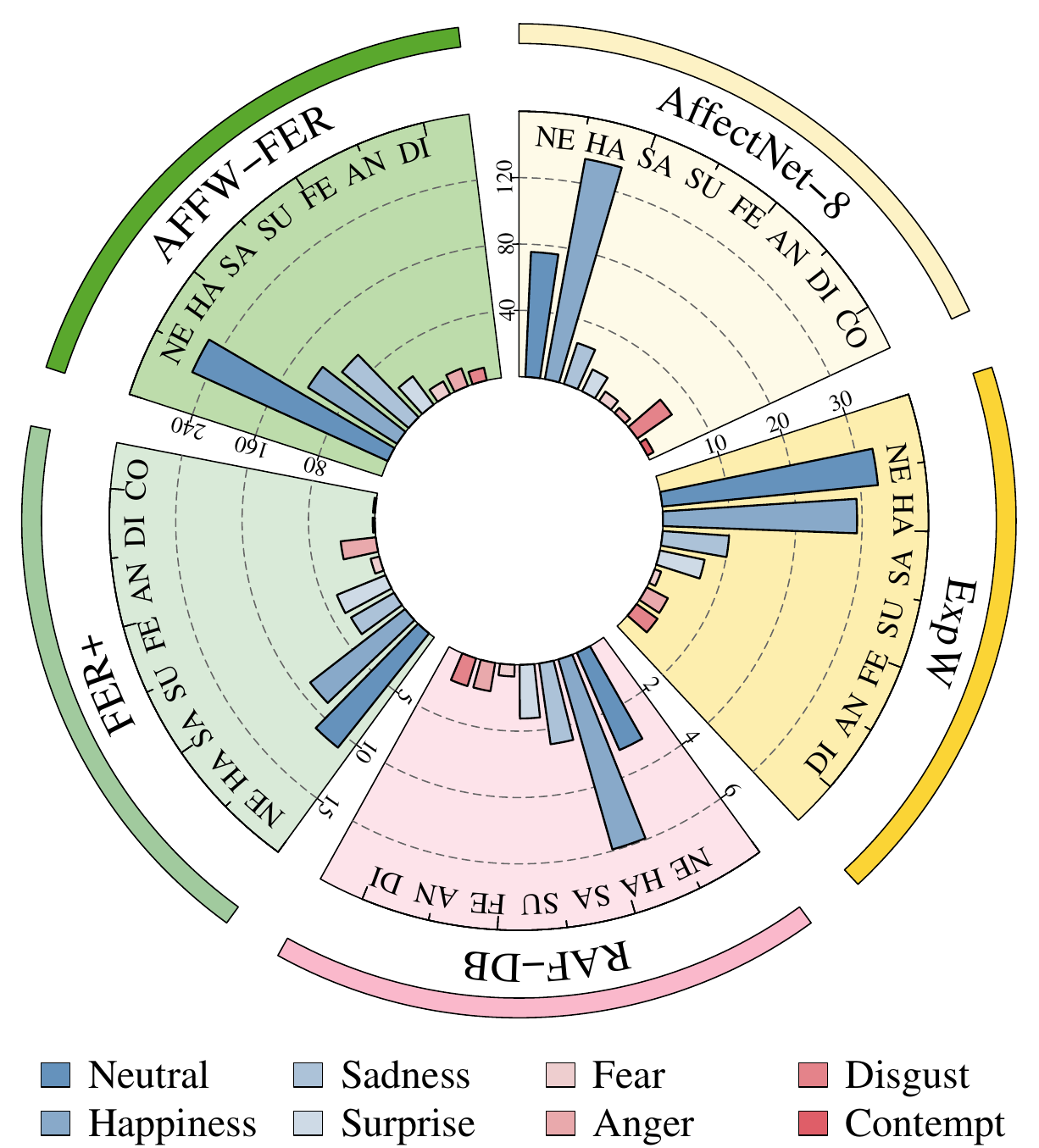}
    \caption{Facial expression-category distributions of the five largest SFER datasets used by OpenAHBU. The radial axes report sample counts in thousands. The absence of a bar indicates that the corresponding expression category is not annotated in that dataset rather than that it contains zero observed samples. ('NA': Neutral, 'HA': Happiness, 'SA': Sadness, 'SU': Surprise, 'FE': Fear, 'AN': Anger, 'DI': Disgust, 'CO': Contempt)}
    \label{fig:fer_data}
\end{figure}

Fig. \ref{fig:fer_data} compares the facial expression category distributions of the five largest SFER datasets used by OpenAHBU, revealing substantial imbalance both across datasets and among expression categories within each dataset. In particular, neutral and happiness generally contain more samples, whereas fear,  disgust, and contempt are comparatively under-represented. These dataset-level and label-level imbalances are addressed by the dual-level resampling strategy.

\subsection{Dynamic Facial Expression Recognition (DFER)}

Three large-scale in-the-wild datasets: DFEW, FERV39k, and MAFW are employed for developing the DFER function. Specifically, DFEW and FERV39k annotating seven basic facial expression categories (i.e., neutral, sadness, fear, disgust, happiness, surprise, and anger), have been employed to develop the DFER function of our OpenAHBU, whereas MAFW additionally annotates contempt, anxiety, helplessness, and disappointment, expanding the DFER label to 11 categories. Their details are provided below:


\textbf{DFEW \cite{jiang2020dfew}.} The Dynamic Facial Expression in-the-Wild (DFEW) dataset was constructed from more than $1,500$ films collected from the Internet, with facial expression clips manually extracted using video editing software. It contains $16,372$ video clips. Twelve trained annotators participated in the annotation process, and each clip was independently annotated ten times for neutral, happiness, sadness, surprise, fear, anger, or disgust. The annotation votes form a seven-dimensional expression distribution, from which a single-expression label is assigned when at least seven annotators agree on the same category. The standard single-label benchmark used by OpenAHBU contains $11,697$ facial video clips and provides five predefined folds. OpenAHBU uses fold 1, containing $2,341$ clips, for final evaluation and divides the remaining four folds into $7,485$ training and $1,871$ validation clips. The clips exhibit substantial variations in illumination, occlusion, and head pose.

\textbf{FERV39k \cite{wang2022ferv39k}.} FERV39k was collected from online videos, television shows, and movies using scene-specific search keywords. The collection covers four broad scenarios, namely Daily Life, Weak-Interactive Shows, Strong-Interactive Activities, and Anomaly Issues, which are further divided into 22 fine-grained scenes. A four-stage automatic segmentation and selection process generated approximately $86,000$ candidate clips from $4,000$ source videos, with $38,935$ clips ultimately retained. Each clip is assigned one of seven expressions: neutral, happiness, sadness, surprise, fear, anger, or disgust, through a quality-controlled annotation process involving crowd-sourced labelling, statistical checking, professional review, and weighted winner-take-all voting. This process involved 20 crowd-sourced annotators and ten professional researchers. The official benchmark randomly divides the clips from all scenes into $31,088$ training and $7,847$ test clips without overlap. OpenAHBU further divides the official training set into $24,870$ training and $6,218$ validation clips while retaining the official test set only for final evaluation.

\textbf{MAFW \cite{liu2022mafw}.} The Multi-modal Affective Faces in the Wild (MAFW) database was collected from movies, television dramas, and short videos on BiliBili and YouTube, supplemented with clips from DFEW, AFEW 7.0, and EmoVoxCeleb. Face detection and manual screening were used to select $10,045$ audio-video clips. Each clip was independently annotated by 11 trained annotators, who assigned one or more expression categories and corresponding self-confidence scores, producing an 11-dimensional expression score vector. An expectation-maximisation algorithm was used to filter annotations before deriving single- and multiple-expression labels. Bilingual textual descriptions of the displayed affective behaviour are additionally provided for non-neutral clips. OpenAHBU uses the $9,172$ clips belonging to the 11-class single-expression subset, whose categories comprise neutral, happiness, sadness, surprise, fear, anger, disgust, contempt, anxiety, helplessness, and disappointment. Following its predefined five-fold protocol, OpenAHBU uses the $1,839$ clips in fold 1 for final evaluation and divides the remaining folds into $5,866$ training and $1,467$ validation clips. Only the visual data are used for developing the DFER function of OpenAHBU.

\subsection{Valence/Arousal Intensity Estimation (VAIE)}

As illustrated in Fig. \ref{fig:va}, VAIE represents facial affect in a two-dimensional continuous space, where valence describes how negative or positive an affective state is and arousal describes its degree of activation. Our OpenAHBU employs four datasets: AFEW-VA, SEWA, AffectNet, and AFFW-VA, for developing this function, which provide image-level or frame-level annotations for both valence and arousal dimensions under varied in-the-wild conditions. Their details are provided below:

 \begin{figure}[htbp]
    \centering
    \includegraphics[width=1.0\linewidth]{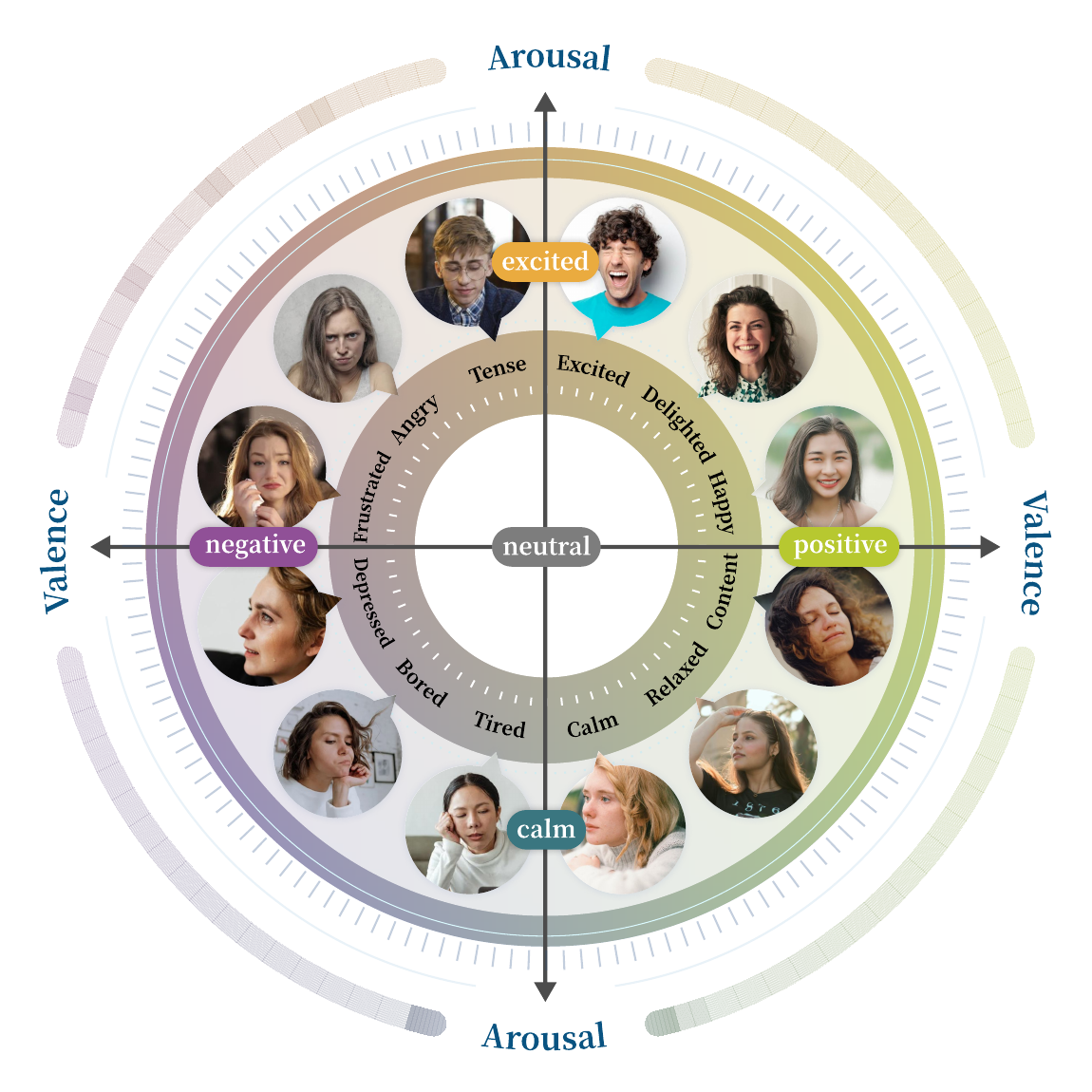}
    \caption{Illustration of the two-dimensional valence-arousal space used for VAIE. Valence ranges from negative to positive affect along the horizontal axis, while arousal ranges from low to high activation along the vertical axis. Representative facial displays and affective terms illustrate example states across this space.}
    \label{fig:va}
\end{figure}

\textbf{AFEW-VA \cite{kossaifi2017afew}.} The AFEW-VA dataset was constructed by extracting facial video clips from feature films recorded under challenging indoor and outdoor conditions. It comprises 600 video clips containing more than $30,000$ frames, with some clips overlapping the AFEW dataset \cite{dhall2012collecting}. It provides per-frame valence and arousal annotations on a 21-level integer scale ranging from -10 to 10. Two FACS-certified experts annotated all videos together using an online frame-by-frame annotation tool and resolved disagreements through discussion. OpenAHBU uses $17,736$ frames for training, $4,434$ frames for model validation, and $7,881$ frames for final evaluation.

\textbf{SEWA \cite{kossaifi2019sewa}.} The SEWA dataset was collected in unconstrained environments using the webcams and microphones on participants’ own computers, while they watched advertisements and discussed them with another participant through video chat. It contains $1,990$ audiovisual recordings totalling more than $2,000$ minutes from 398 participants across six cultures, with approximately equal proportions of women and men and an age range of 18 to 65 years. The database provides continuously valued valence and arousal annotations alongside facial landmarks, facial AUs, vocalisations, and other behavioural annotations. Five annotators from each culture annotated selected video-chat segments using a joystick-based tool, with valence and arousal recorded in separate passes. The original annotation procedure combines the individual annotation tracks using Canonical Time Warping and normalises the resulting ground truth to [0, 1]. In OpenAHBU, $322,102$ frames are used for model training, $80,526$ frames are used for model validation, and $94,775$ frames are reserved for final evaluation. Only the visual facial frames and their corresponding valence and arousal annotations are used for developing the VAIE function.

\textbf{AffectNet \cite{mollahosseini2017affectnet}.} AffectNet was collected from the Internet through Google, Bing, and Yahoo using $1,250$ emotion-related keywords in six languages. It contains more than one million facial images, approximately $450,000$ of which were manually annotated with categorical expressions and continuous valence and arousal values. A team of twelve trained annotators participated in the annotation process, with each image initially assigned to one annotator. Using a dedicated annotation tool, annotators selected a point in the two-dimensional valence-arousal space, with both coordinates ranging from -1 to 1. For VAIE, OpenAHBU uses the $287,651$ images in the released training set and the $4,000$ images in the released validation set. The released training set is divided into $230,121$ training and $57,530$ validation images, while the released validation set is retained only for final evaluation.

\textbf{AFFW-VA \cite{kollias2018aff}.} AFFW-VA denotes the valence- and arousal-annotated portion of Aff-Wild2, a large-scale audiovisual database assembled from YouTube videos recorded under unconstrained conditions. For VAIE, OpenAHBU uses $2,030,527$ facial frames from the released training and validation portions. The database provides frame-level valence and arousal annotations with values ranging from -1 to 1. Annotators continuously rated both dimensions using a joystick-based tool, and the annotation tracks were subsequently reviewed and combined to obtain the final labels. OpenAHBU divides the $1,654,193$ frames from the released training portion into $1,323,354$ training and $330,839$ validation frames, while the $376,334$ frames from the released validation portion are used only for final evaluation. Its videos exhibit substantial variations in subject age and ethnicity, head pose, illumination, recording environment, and facial behaviour.

\subsection{Facial Pain Level Estimation (FPLE)}

Two datasets collected under controlled pain-induction protocols: BioVid (induced via heat stimulation) and MIntPAIN (induced via electrical stimulation) are employed for developing the FPLE function. Both datasets define five stimulus-intensity levels, ranging from level 0 (no stimulus) to level 4 (the maximum tolerable pain level). The details of these datasets are provided below:

\textbf{BioVid \cite{walter2013biovid}.} The BioVid Heat Pain dataset was collected under controlled laboratory conditions, with healthy participants exposed to individually calibrated thermal stimuli applied to the right arm using a thermode. It contains frontal facial videos and physiological recordings. OpenAHBU uses Parts A and B, which comprise $8,700$ clips from 87 participants recorded without facial electromyography (EMG) electrodes and 8,600 clips from 86 participants recorded with facial EMG electrodes, respectively. Each clip lasts 5.5 seconds, and each participant contributes 20 clips for the no-pain baseline and each of the four pain levels in each part. The clip labels correspond to the applied thermal stimulation, with level 0 representing the baseline temperature of 32°C. Participants’ pain-threshold and pain-tolerance temperatures were determined before recording and used for levels 1 and 4, respectively, while levels 2 and 3 were equally spaced intermediate temperatures.For binary pain recognition, only clips from levels 0 and 4 are retained, giving a total of 6,920 video clips. OpenAHBU divides the training portion into training and validation subsets at an 8:2 ratio, while the test partition is reserved for final evaluation.

\textbf{MIntPAIN \cite{haque2018deep}.} The Multimodal Intensity Pain (MIntPAIN) dataset was collected under controlled laboratory conditions using electrical stimulation, with RGB and depth data captured using a Microsoft Kinect v2 sensor and thermal data captured using an Axis Q1922 camera. Each participant completed two trials consisting of 40 stimulation sweeps per trial, with each sweep producing one no-pain sequence and one sequence corresponding to one of four pain-intensity levels. The dataset provides synchronised RGB, depth, and thermal facial sequences from 20 participants, with $9,366$ video sequences in total across the three modalities. The resulting sequences vary between 7 and 51 frames in length. The frames were synchronised using their recording timestamps and organised into sequences with stimulus-intensity labels from 0 to 4 based on the associated EMG data. Participants additionally rated their perceived pain intensity and unpleasantness using electronic visual analogue scales (VAS) ranging from 0 to 10. OpenAHBU divides the training portion into training and validation subsets at an 8:2 ratio, while the test partition is reserved for final evaluation. Because every sweep contains a level-0 sequence but only one of the four pain-level sequences, level 0 contains more samples than each individual pain level.

\begin{figure}[htbp]
\centering
\captionsetup[subfigure]{skip=2pt}
\begin{subfigure}{1.0\linewidth}
    \includegraphics[width=\linewidth]{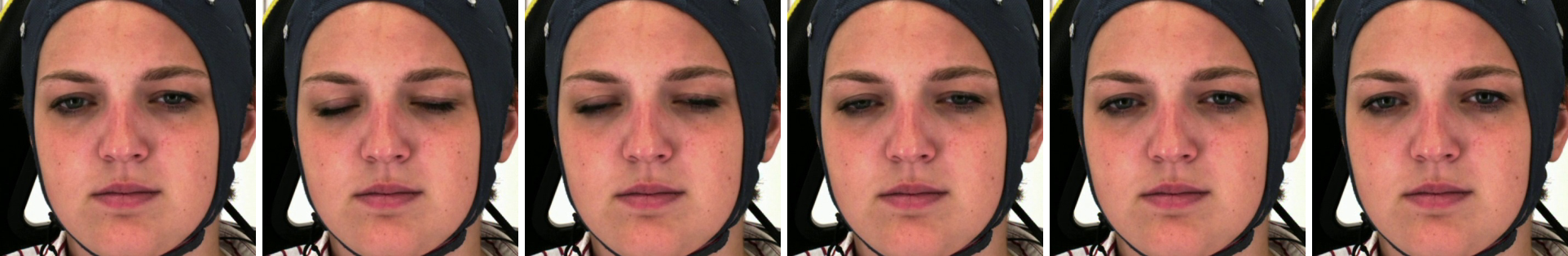}
    \caption{An example of Pain Level 0 in BioVid (No Pain).}
    \label{fig:pain_a}
\end{subfigure}

\begin{subfigure}{1.0\linewidth}
    \includegraphics[width=\linewidth]{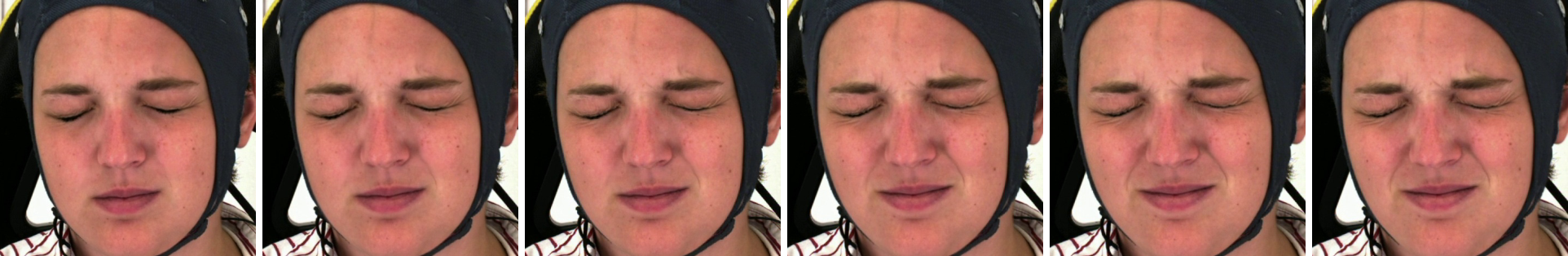}
    \caption{An example of Pain Level 4 in BioVid (Heat pain).}
    \label{fig:pain_b}
\end{subfigure}

\begin{subfigure}{1.0\linewidth}
    \includegraphics[width=\linewidth]{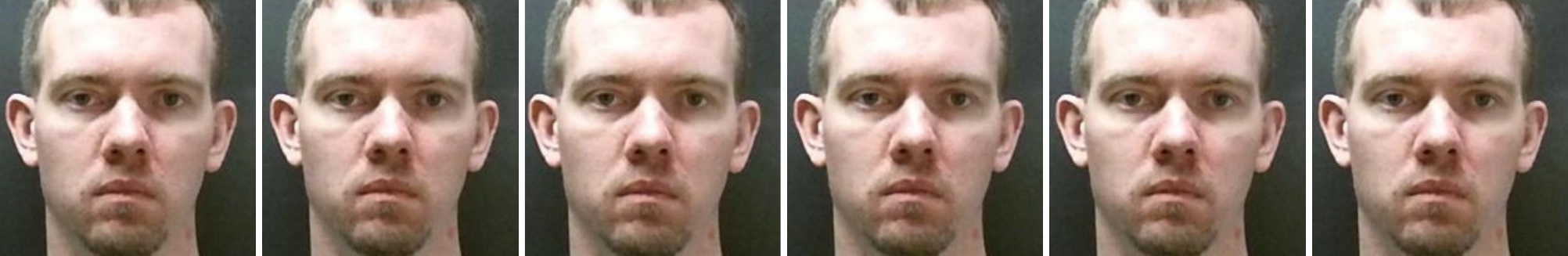}
    \caption{An example of Pain Level 0 in MIntPAIN (No Pain).}
    \label{fig:pain_c}
\end{subfigure}

\begin{subfigure}{1.0\linewidth}
    \includegraphics[width=\linewidth]{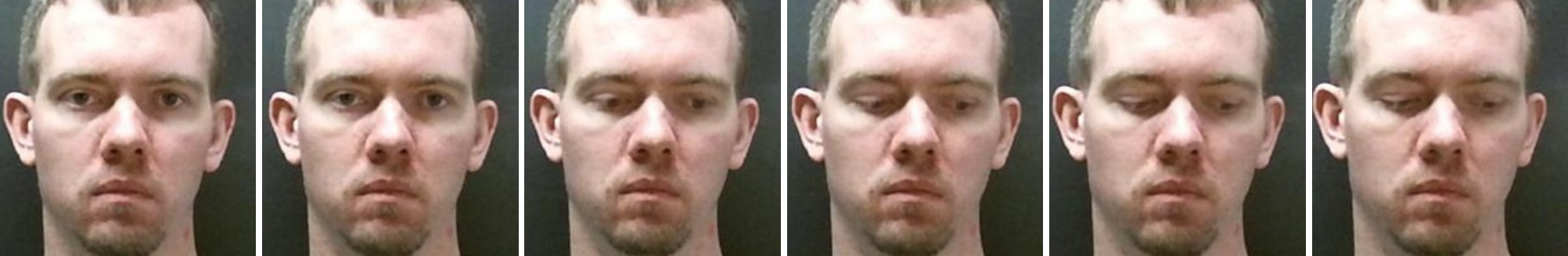}
    \caption{An example of Pain Level 4 in MIntPAIN (Electrical Pain).}
    \label{fig:pain_d}
\end{subfigure}

\caption{Example comparison of facial expressions at pain level 0 (no stimulus) vs. pain level 4 (maximum pain level) in subjects from the BioVid \cite{walter2013biovid} and MIntPAIN \cite{haque2018deep} datasets, with BioVid applying stimulated heat pain and MIntPAIN applying stimulated electrical pain.}
\label{fig:pain}
\end{figure}

Representative facial sequences at pain levels 0 and 4 from both datasets are shown in Fig. \ref{fig:pain}.

\subsection{Facial Depression Assessment (FDA)}

AVEC2014 depression corpus is employed for developing the FDA function, which provides facial videos recorded during two speaking tasks, together with depression-severity annotations. It was collected in a task-oriented human-computer interaction setting. During the Northwind task, participants read aloud an excerpt from the German fable “Die Sonne und der Wind” (The North Wind and the Sun), whereas the Freeform task requires them to respond spontaneously to one of a set of questions in German. It comprises 150 paired recording sessions, each containing a Northwind recording and a Freeform recording, resulting in 300 videos in total. Individual task recordings range from 6 seconds to 4 minutes 8 seconds. The released videos were resampled to a uniform frame rate of 30 frames per second at a resolution of $640 \times 480$ pixels. Each recording pair is associated with a Beck Depression Inventory-II (BDI-II) score on a scale ranging from 0 to 63. These scores are obtained from a self-report questionnaire completed by the participants, comprising 21 items scored from 0 to 3, whose sum provides the depression-severity label. The 150 pairs are evenly divided into official training, development, and test partitions, each containing 50 pairs. The partitioning aims to maintain similar distributions of age, gender, and depression severity across the three subsets, while recordings from the same original session are assigned to a single partition. While AVEC 2014 is a subset of AVEC 2013, we did not use AVEC 2013 for developing our FDA function as its rest tasks can not be conducted without a slide-guided guidance compared to the FreeForm and reading tasks provided by AVEC 2014, i.e., they are difficult to be directly conducted when using our OpenAHBU toolkit.

\subsection{Facial Apparent Personality Recognition (FAPR)}

OpenAHBU employs the ChaLearn First Impressions dataset for developing the FAPR function, which contains in-the-wild speaking videos annotated with observer-perceived Big Five personality traits. The dataset was constructed from YouTube videos of people facing and speaking in English to a camera under unconstrained conditions. The collection combined manual screening of source videos, automatic clip extraction based on face and eye detection, and further manual screening of the extracted clips. It comprises 10,000 15-second audiovisual clips extracted from 3,060 source videos across 2,764 YouTube channels. The videos are provided at 30 frames per second. Each clip is provided with continuous apparent-personality scores ranging from 0 to 1 for Openness, Conscientiousness, Extraversion, Agreeableness, and non-Neuroticism (Emotional Stability). External human annotators recruited through Amazon Mechanical Turk \cite{escalante2020modeling} compared pairs of clips and indicated which person appeared to express each trait more strongly. The pairwise judgements were converted into continuous scores by fitting a Bradley–Terry–Luce model using maximum likelihood estimation. OpenAHBU follows the official training, validation, and test partitions of 6,000, 2,000, and 2,000 clips, respectively. These annotations describe observers’ first impressions of the people in the clips rather than self-reported personality traits.

\section{Description of the Compared AFBU Toolkits and SOTA methods}

\subsection{Compared AFBU toolkits}

This subsection provides further details for the five existing academically available AFBU toolkits compared in Table 3 of the main paper. For each toolkit, we describe the evaluated functions, supported output labels, model design, and datasets used for model development. For all compared toolkits, we use the latest versions available in their official GitHub repositories at the time of evaluation. The released pre-trained models are evaluated using their official inference pipelines on the same test partitions as OpenAHBU, without retraining or adapting them to the corresponding datasets. For FAUD, each toolkit is evaluated only on the intersection between the AUs supported by the toolkit and those annotated by the target dataset. Continuous AU occurrence scores are converted into binary predictions using a threshold of 0.5, consistent with OpenAHBU. For SFER, the shared expression categories are matched directly. When a toolkit supports only seven expression categories, samples annotated as Contempt are excluded from its evaluation.

\textbf{DeepFace:} DeepFace \cite{serengil2020lightface} \footnote{\url{https://github.com/serengil/deepface}} is evaluated for SFER. Its facial-attribute analysis module recognises seven facial expression categories, including anger, disgust, fear, happiness, sadness, surprise, and neutral. The official inference pipeline supports face detection and alignment before converting the aligned face into a $48\times48$ greyscale image for facial expression recognition. A CNN-based model then produces the scores for the seven categories, and the category with the highest score is used as the predicted facial expression. The corresponding facial-attribute analysis study reports the evaluation of this model on FER2013. However, the exact composition of the data used to train the released checkpoint is not separately documented in the publication or official repository.

\textbf{OpenFace 2.0:} OpenFace 2.0 \cite{zadeh2018openface} \footnote{\url{https://github.com/TadasBaltrusaitis/OpenFace}} is evaluated for both FAUD and FAUE. For FAUD, it predicts the occurrences of 18 AUs: AU01, AU02, AU04, AU05, AU06, AU07, AU09, AU10, AU12, AU14, AU15, AU17, AU20, AU23, AU25, AU26, AU28, and AU45. For FAUE, it estimates the intensities of the same AUs except AU28, resulting in 17 intensity outputs. OpenFace 2.0 first performs facial landmark detection and alignment and then extracts HOG and facial geometric features. Separate linear SVM and SVR models are used for AU occurrence detection and intensity estimation, respectively. The occurrence and intensity models were developed using multiple AU datasets, including BP4D, DISFA, CK+, SEMAINE, Bosphorus, UNBC-McMaster, and the FERA 2011 and 2015 datasets \cite{zadeh2018openface}.

\textbf{Py-Feat:} Py-Feat \cite{cheong2023py} \footnote{\url{https://github.com/cosanlab/py-feat}} is evaluated for FAUD and SFER.  Its AU detector supports 20 AUs: AU01, AU02, AU04, AU05, AU06, AU07, AU09, AU10, AU12, AU14, AU15, AU17, AU18, AU20, AU23, AU24, AU25, AU26, AU28, and AU43. After face detection and alignment, Py-Feat extracts a 5,408-dimensional HOG representation from each face and applies PCA to retain 95\% of its variance. The resulting features are processed by separately trained SVM or XGBoost classifiers for AU occurrence detection. The XGBoost configuration is used as the default AU model because it directly provides the predicted probability of each AU. The AU models were developed using BP4D, BP4D+, DISFA, CK+, UNBC-McMaster, and EmotioNet \cite{cheong2023py}. For SFER, Py-Feat recognises Anger, Disgust, Fear, Happiness, Sadness, Surprise, and Neutral. It provides both a ResMaskNet-based model evaluated on FER2013 and an SVM-based model developed using ExpW, CK+, and JAFFE, with ResMaskNet used as the default SFER model \cite{cheong2023py}.

\textbf{LibreFace:} LibreFace \cite{chang2024libreface}\footnote{\url{https://github.com/ihp-lab/LibreFace}} is evaluated for FAUD, FAUE, and SFER. It employs MediaPipe to detect facial landmarks and align the input faces. Its facial representation model is developed by pre-training an MAE model with a ViT-B backbone on large-scale facial images from EmotioNet, AffectNet, and FFHQ. The resulting teacher model is further adapted to individual AFBU functions, and its feature representations are distilled into a ResNet-18 student model used for inference. For FAUE, LibreFace estimates the intensities of 12 AUs, including AU01, AU02, AU04, AU05, AU06, AU09, AU12, AU15, AU17, AU20, AU25, and AU26. The released toolkit additionally provides occurrence predictions for AU07, AU10, AU14, AU23, and AU24. DISFA and BP4D are used for the development of its AU intensity and occurrence models, respectively. Its SFER model recognises seven categories: anger, disgust, fear, happiness, sadness, surprise, and neutral, and is developed using AffectNet, FFHQ, and RAF-DB \cite{chang2024libreface}.

\textbf{OpenFace 3.0:} OpenFace 3.0 \cite{hu2025openface30lightweightmultitask}\footnote{\url{https://github.com/CMU-MultiComp-Lab/OpenFace-3.0}} is evaluated for FAUD and SFER. Its pipeline first employs RetinaFace for face detection and alignment, followed by a stacked Hourglass model for facial landmark detection. An EfficientNet backbone pre-trained on VGGFace2 extracts the facial representations, which are combined with the landmark features for downstream AFBU functions. For FAUD, a graph-based module models the relationships between different facial regions to predict eight AUs' occurrences: AU01, AU02, AU04, AU06, AU09, AU12, AU25, and AU26. The AU models are developed using DISFA and BP4D. For SFER, a fully connected prediction head recognises eight facial expression categories, including neutral, happiness, sadness, surprise, fear, disgust, anger, and contempt, with AffectNet used for model development. OpenFace 3.0 therefore provides a directly comparable output space for the eight-category SFER setting adopted in OpenAHBU.

\textbf{LibreFace 2.0:} LibreFace 2.0 \cite{guan2026libreface}\footnote{\url{https://github.com/ihp-lab/LibreFace}} is the latest release of LibreFace and is evaluated for FAUD, FAUE, and SFER. It employs MediaPipe to detect facial landmarks and align the input faces. Its model development follows the feature distillation strategy introduced in LibreFace, where face encoders are first pre-trained on large-scale facial images and then adapted to individual AFBU functions. The resulting facial representations are distilled into lightweight ResNet-18 and RepVGG models for inference. LibreFace 2.0 further combines the real AU samples from DISFA and BP4D with more than 240,000 synthetic frames from 708 identities. Specifically, Stable Diffusion 3.5 generates facial identities with different demographic attributes, while LivePortrait transfers the AU movements from DISFA and BP4D videos to the generated faces \cite{guan2026libreface}. For FAUE, LibreFace 2.0 estimates the intensities of 12 AUs, including AU01, AU02, AU04, AU05, AU06, AU09, AU12, AU15, AU17, AU20, AU25, and AU26. Its FAUD model additionally provides occurrence predictions for AU07, AU10, AU14, AU23, and AU24. DISFA and BP4D are used for developing the AU intensity and occurrence models, respectively. For SFER, it recognises seven categories: anger, disgust, fear, happiness, sadness, surprise, and neutral, with AffectNet, FFHQ, and RAF-DB used for model development.

\subsection{Compared SOTA methods}

To provide representative dataset-specific references for comparing the eight re-developed OpenAHBU functions, we select three reproducible SOTA methods published between 2021 and 2026 for each function. They cover established peer-reviewed venues in computer vision, artificial intelligence, affective computing, and biomedical analysis, such as CVPR, ECCV, ACM MM, FG, AAAI, IEEE Transactions on Affective Computing, Pattern Recognition, Nature Machine Intelligence, IEEE EMBC, and Frontiers in Pain Research. A SOTA method is included when its original paper reports results on at least one benchmark dataset used by OpenAHBU with the same primary evaluation metric and a comparable evaluation protocol. Under these conditions, we prioritise recent methods that report best performances and employed representative model designs.


\subsubsection{Facial AU Occurrence Detection (FAUD)}

Three open-source/reproducible SOTA FAUD methods are compared with FAUD models of the proposed OpenAHBU:
\begin{itemize}
    \item \textbf{MDHR} \cite{wang2024multi} first captures facial dynamics through feature differences between neighbouring frames at multiple spatial scales, and combines the adaptively weighted dynamic cues with static facial features. It then models AU dependencies within each facial region through regional feature extraction and auxiliary prediction of AU combinations, followed by graph attention across regions and temporal convolution over AU feature sequences. By considering the different spatial ranges and magnitudes of facial movements, this design exploits complementary appearance, motion and AU relationship cues to improve frame-level AU occurrence detection.

    \item \textbf{MAE-Face} \cite{ma2024facial} learns facial representations by pre-training a ViT-based masked autoencoder (MAE) \cite{he2022masked} on 2.17 million facial images without using their annotations, where randomly masked patches are reconstructed from the visible ones. The reconstruction decoder is then discarded, and the encoder is fine-tuned with a linear prediction head to detect AU occurrences. Such pre-training reduces the dependence on AU annotations, with the original study reporting competitive detection performance even when fine-tuning uses only 1\% of the available training frames.

    \item \textbf{FMAE} \cite{ning2024representation} is a facial masked autoencoder pre-trained on Face9M, which contains nine million facial images collected from multiple public datasets. During AU fine-tuning, separate linear heads predict AU occurrences and subject identities, while a gradient reversal layer encourages the shared encoder to retain AU-related features and suppress identity-related features. This combines large-scale facial representation learning with identity regularisation, reducing reliance on subject identity for AU prediction and improving detection on unseen subjects.
\end{itemize}

\subsubsection{Facial AU Intensity Estimation (FAUE)}

Three open-source/reproducible SOTA FAUE methods are compared with FAUE models of the proposed OpenAHBU:
\begin{itemize}
    \item \textbf{APs} \cite{sanchez2021affective} builds on Neural Processes \cite{garnelo2018neural} to encode facial features and initial AU intensity predictions from a pre-trained frame-level network as a distribution over a global latent variable. At inference time, context frames with lower estimated latent uncertainty are selected, and the resulting shared context is used to refine frame-level intensity predictions. This allows the model to account for uncertainty in temporal context and improves intensity estimation over the recurrent and self-attention baselines evaluated on BP4D and DISFA.

    \item \textbf{FRL} \cite{bulat2022pre} pre-trains a ResNet-50 on facial images using SwAV \cite{caron2020unsupervised}, which encourages consistent cluster assignments across different augmented views of the same image without requiring annotations. It then replaces the pre-training head with a linear regression head and fine-tunes the entire network using an $\ell_2$ loss to estimate AU intensities. This strategy provides transferable facial representations and improves intensity estimation over the supervised pre-training baselines on BP4D and DISFA, including settings with limited labelled data.

    \item \textbf{MAE-Face} \cite{ma2024facial} transfers facial representations learnt through masked image reconstruction \cite{he2022masked} on unlabelled facial images to AU intensity estimation. After removing the reconstruction decoder, it jointly fine-tunes the ViT encoder and a linear regression head using an $\ell_2$ loss, with AU intensity labels normalised to $[0,1]$. Such pre-training reduces the dependence on dense AU annotations, with the original study reporting competitive intensity estimation performance when fine-tuning uses only 1\% of the available training frames.
\end{itemize}

\subsubsection{Static Facial Expression Recognition (SFER)}

Three open-source/reproducible SOTA SFER methods are compared with SFER models of the proposed OpenAHBU:
\begin{itemize}
    \item \textbf{DDAMFN} \cite{zhang2023dual} employs a lightweight mixed feature network to extract facial features through mixed-size depthwise convolutions and coordinate attention \cite{hou2021coordinate}. Multiple dual-direction attention heads generate attention maps along the horizontal and vertical directions, while an attention loss encourages different heads to focus on distinct facial regions. Compared with its backbone alone, this design improves recognition accuracy on RAF-DB and AffectNet-7 with only a small increase in parameters and FLOPs.

    \item \textbf{POSTER++} \cite{mao2025poster++} simplifies POSTER \cite{zheng2023poster} by removing the image-to-landmark branch and using window-based cross-attention, where facial landmark features provide shared queries for different image windows. It replaces the pyramid network with a lightweight multi-scale interaction module that combines facial features across scales through attention. These changes reduce model size and computational cost while improving recognition accuracy over POSTER on RAF-DB and AffectNet.

    \item \textbf{MHAN} \cite{wang2026mhan} combines a mixed feature network (MFN) \cite{zhang2023dual} with efficient local attention (ELA) \cite{xu2024ela} to extract multi-scale facial features and emphasise expression-relevant regions. Its multiple attention heads combine horizontal and vertical attention with squeeze-and-excitation (SE) \cite{hu2018squeeze} to capture spatial information and adaptively adjust the importance of feature channels. The model further uses label smoothing to reduce overconfident predictions, and achieves competitive expression recognition performance with a low computational cost.
\end{itemize}

\subsubsection{Dynamic Facial Expression Recognition (DFER)}

Three open-source/reproducible SOTA DFER methods are compared with DFER models of the proposed OpenAHBU:
\begin{itemize}
    \item \textbf{MAE-DFER} \cite{sun2023mae} extends VideoMAE \cite{tong2022videomae} by jointly reconstructing facial appearance and differences between adjacent frames, allowing self-supervised pre-training to learn both static and dynamic facial information. Its Local-Global Interaction Transformer (LGI-Former) performs self-attention within local spatiotemporal regions and exchanges information between regions through a small set of representative tokens. After fine-tuning for expression classification, the resulting model achieves recognition performance comparable to the VideoMAE baseline pre-trained on the same facial video dataset, with substantially fewer FLOPs.

    \item \textbf{SVFAP} \cite{sun2024svfap} employs masked video reconstruction to pre-train a facial video encoder, which is subsequently fine-tuned for expression classification. Its Temporal Pyramid and Spatial Bottleneck Transformer (TPSBT) progressively downsamples temporal features using strided convolutions and aggregates spatial features into a small number of bottleneck tokens through spatial attention. In the DFEW ablation study, this design reduces the FLOPs of both pre-training and fine-tuning while improving recognition accuracy over the vanilla ViT baseline.

    \item \textbf{S4D} \cite{chen2025static} first pre-trains a shared ViT encoder on facial images and videos through masked reconstruction, and then jointly fine-tunes it for SFER and DFER using separate classification heads. During joint fine-tuning, Mixture of Adapter Experts (MoAE) modules are added in parallel with the feed-forward networks in the deeper Transformer layers, with a learned gate selecting lightweight adapter experts~\cite{houlsby2019parameter} for each input token. This design learns both shared and task-specific facial features, alleviating the negative transfer observed in direct multi-task fine-tuning and improving DFER performance.
\end{itemize}

\subsubsection{Valence/Arousal Intensity Estimation (VAIE)}

Three open-source/reproducible SOTA VAIE methods are compared with VAIE models of the proposed OpenAHBU:
\begin{itemize}
    \item \textbf{EMOFAN} \cite{toisoul2021estimation} builds on a pre-trained face alignment network (FAN) \cite{bulat2017far} to jointly predict facial landmarks, categorical expressions, and continuous valence and arousal values from a single face image. Features from multiple FAN layers are weighted by the predicted landmark heatmaps to focus the network on facial regions relevant to affect estimation. Knowledge distillation is further used to reduce the influence of noisy emotion labels, while the integrated architecture enables joint affect and landmark prediction in real time.

    \item \textbf{MAVENC} \cite{dresvyanskiy2024multi} follows a staged training strategy, with its visual encoders first pre-trained on multiple affective datasets and then fine-tuned on Aff-Wild2. For the facial video model used in our comparison, the selected EfficientNet-B1 \cite{tan2019efficientnet} encoder is frozen, and its frame-level features are processed by three Transformer encoder layers followed by a regression head. This design captures temporal facial patterns and achieves higher CCC than the statistical feature aggregation baseline on the Aff-Wild2 development set.

    \item \textbf{TGCT} \cite{li2025temporal} builds a spatial encoder on Swin Transformer \cite{liu2021swin}, using deformable landmark attention to adjust landmark-based sampling positions for each input frame and focus on emotion-related facial regions. Its temporal encoder summarises neighbouring frames into groups at multiple temporal scales, models dependencies between these groups, and passes the resulting contextual information back to frame-level features. Together with auxiliary learning objectives that predict affective differences between frames and between clips, this design improves prediction consistency and reduces inaccurate fluctuations in frame-level valence and arousal estimates.
\end{itemize}

\subsubsection{Facial Pain Level Estimation (FPLE)}

Three open-source/reproducible SOTA FPLE methods are compared with FPLE models of the proposed OpenAHBU:
\begin{itemize}
    \item \textbf{FTF} \cite{gkikas2023full} employs a Transformer in Transformer (TNT) \cite{han2021transformer} encoder, pre-trained on VGGFace2, to extract local and global facial features from each video frame. These features are concatenated and processed by a temporal Transformer comprising one cross-attention block and two self-attention blocks for pain classification. The cross-attention block uses a smaller set of learned queries, reducing the computational cost of modelling temporal information.

    \item \textbf{MAAP} \cite{gkikas2024multimodal} adopts a TNT-based spatial encoder, which is first pre-trained for face recognition and then jointly pre-trained on four facial expression datasets. In its video-only configuration, features from each full face frame and its four quadrants are combined through weighted addition, and the frame representations are concatenated for temporal modelling using cross-attention and self-attention. Its AugmNet module learns transformations of the video representations to regularise training and is inactive during inference, providing feature augmentation without additional inference cost.

    \item \textbf{PainFormer} \cite{gkikas2025painformer} combines spectral layers based on the fast Fourier transform with self-attention in a hierarchical encoder, which is pre-trained through multi-task learning on 14 dataset-specific tasks. For pain assessment on BioVid, the frame embeddings within each video are concatenated and processed by an Embedding-Mixer containing cross-attention and self-attention. The compared configuration combines predictions from RGB videos and their synthetic thermal and estimated depth representations through decision fusion, improving classification accuracy over the RGB-only configuration.
\end{itemize}

\subsubsection{Facial Depression Assessment (FDA)}

Three open-source/reproducible SOTA FDA methods are compared with FDA models of the proposed OpenAHBU:
\begin{itemize}
    \item \textbf{MTB-DFE} \cite{xu2024two} builds on a Temporal Pyramid Network (TPN) \cite{yang2020temporal} to extract facial behavioural features from short video segments at multiple spatio-temporal scales. Its Depression Feature Enhancement (DFE) module combines mutual temporal attention with feature disentanglement to enhance depression-related cues and suppress irrelevant information. The enhanced feature sequences are then encoded as a Spectral Graph Representation (SPG), whose nodes describe the temporal changes of individual feature dimensions, and processed by a graph neural network to combine short-term facial cues with video-level temporal patterns for depression assessment.

    \item \textbf{MGNN} \cite{wu2025depmgnn} represents each face video as a matrixial graph, whose nodes retain two-dimensional facial feature maps extracted from individual frames at multiple spatial scales. It learns matrix-valued spatial and temporal edge features to control information exchange among nodes, and adaptively combines the resulting representations across spatial scales to predict a video-level depression score. This design preserves spatial information during temporal modelling and accommodates videos of different lengths through graphs containing different numbers of nodes.

    \item \textbf{MTE-DFS} \cite{liao2026explainable} employs a Video Action Transformer \cite{girdhar2019video} to extract segment-level facial features and combines features from consecutive segments to represent facial behaviours over different temporal durations. Using video-level depression labels as weak supervision, its Discriminative Feature Selection (DFS) module estimates the relevance of these representations to the predicted depression category and selects the most relevant ones for score regression. The resulting scores are averaged to obtain the video-level depression estimate, while the relevance scores highlight the temporal intervals supporting the prediction and provide segment-level interpretability.
\end{itemize}

\subsubsection{Facial Apparent Personality Recognition (FAPR)}

Three open-source/reproducible SOTA FAPR methods are compared with FAPR models of the proposed OpenAHBU:
\begin{itemize}
    \item \textbf{VAT} \cite{liao2024open} adapts the Video Action Transformer \cite{girdhar2019video} to apparent personality recognition, using a CNN to extract frame-level facial features and Transformer attention to aggregate information across frames. Within each segment, the central frame feature serves as the initial query, and the resulting representation is mapped to five personality scores, with segment-level predictions averaged to obtain video-level estimates. This design combines facial appearance with its temporal context to capture facial behavioural cues for apparent personality recognition.

    \item \textbf{SVFAP} \cite{sun2024svfap} learns transferable facial representations through self-supervised reconstruction of masked spatio-temporal patches from unlabelled face videos. Its Temporal Pyramid and Spatial Bottleneck Transformer progressively reduces temporal resolution and summarises spatial information using a small set of bottleneck tokens, reducing the computational cost of spatio-temporal modelling. For FAPR, the reconstruction decoder is removed, and the pre-trained encoder is fine-tuned with global average pooling and a fully connected regression head to predict the five apparent personality scores.

    \item \textbf{MGNN} \cite{wu2025depmgnn} represents a face video as a matrixial graph, whose nodes retain two-dimensional facial feature maps extracted from individual frames at multiple spatial scales. It learns matrix-valued spatial and temporal edge features to control information exchange among nodes, and adaptively combines the updated representations across spatial scales to regress the five apparent personality traits. This design preserves the spatial structure of facial features during temporal modelling and accommodates videos of different lengths through graphs containing different numbers of nodes.
\end{itemize}

\noindent\textbf{Comparison settings.} As described in the main paper, the results of the compared SOTA method in Table 3 are taken directly from the corresponding papers, without re-training or re-evaluating their released models. When several versions of a SOTA method are available, we use the results and publication information from its latest peer-reviewed version. Code availability is not used as an exclusion criterion because the comparison is based on the published results. A result is included only when the original study reports the corresponding benchmark dataset and primary metric under a comparable evaluation protocol, while unavailable results are denoted by ``--''. It should be noted that these SOTA methods are generally trained and optimised for individual datasets, allowing them to adapt to dataset-specific annotations and data distributions. In contrast, our OpenAHBU uses a unified model trained on the heterogeneous datasets associated with each function, without further dataset-specific fine-tuning. Therefore, the comparison setting is favourable to the selected SOTA methods and is intended to examine whether the generic models implemented in OpenAHBU can remain competitive with recent dataset-specific methods.

\section{Speed and parameter scales of the re-developed function-specific models}

Table \ref{tab:model_efficiency} details the 49 task-specific models re-developed for the eight OpenAHBU functions, including eight FAUD models, five FAUE models, seven SFER models, five DFER models, ten VAIE models, four FPLE models, four FDA models, and six FAPR models. For each model, we report its: parameter count, floating-point operations (FLOPs), and inference speed measured on an NVIDIA L20 GPU, i.e., the inference speed is expressed as frames per second (FPS) for the frame-level functions and 
video clips per second (VPS) for the video-level functions. The number of frames sampled to process a single video varies across video-level methods according to their respective official implementations and will be specified for each method in the OpenAHBU's released code.

Following the function-specific method-selection criteria described in Sec. 3.2.1 of the main paper, the included methods were required to satisfy the openness, reproducibility, and predictive-performance criteria under the unified evaluation setting. Rather than selecting models solely according to predictive performance, OpenAHBU further retains models from different parameter-count tiers to accommodate various computational constraints. For example, the FAUD suite includes GraphAU(ViT-S), with 22.97M parameters, 15.86G FLOPs, and an inference speed of 58 FPS, as well as the large-parameter GraphAU(ViT-L), with 309.11M parameters, 60.74G FLOPs, and an inference speed of 48 FPS. Such configurations provide users with alternatives for both resource-constrained and large-parameter application settings.

Since each function may require different input resolutions, temporal window lengths, and prediction-head dimensions, the same underlying method can exhibit slightly different parameter counts and computational costs when adapted to different functions. Table \ref{tab:model_efficiency} should therefore be considered together with the detailed performance results reported in Sec. \ref{sec:experiments} of this Supplementary Material. By jointly considering predictive performance, model size, computational cost, and inference speed, users can select the function-specific model that best matches their application requirements and available computational resources.

\begin{table*}[htbp]
\large
\centering
\setlength{\doublerulesep}{0.3em}
\renewcommand{\arraystretch}{1.4}
\caption{Parameters (M), FLOPs (G) and Inference Speed (frames per second for frame-level models (FPS), and video clips per second for video-level models (VPS).) on a single Nvidia L20 GPU for all re-developed OpenAHBU models. The same model may exhibit differences in its parameter number and computational cost across different functions (e.g., MAE-Face-B in FAUD and SFER; MGNN (MFN) and MFNN (Res50) in FDA and FAPR) due to slight variations in the model configuration for each function (such as the dimensionality and number of layers for the MLP regression head, input image resolution, number of input frames, etc.). Results marked with $^{*}$ are obtained using methods that require video or video segment inputs, and thus their reported performance are calculated using only video datasets while excluding image datasets, e.g., AffectNet are excluded for MAE4DFER-B, FRU-Adapter-B, MAVENC(ViT-B) and MAVENC(Res50) VAIE models. For methods with inference speed reported in VPS, the average number of frames per video is 46 for DFER, 110 for FPLE, 1850 for FDA, and 450 for FAPR.}
\resizebox{1.0\linewidth}{!}{
\begin{tabular}{c|ccccc||c|ccccc}
\Xhline{2.5px}
\rowcolor{gray!15}
\textbf{Function} &
  \textbf{Model} &
  \begin{tabular}[c]{@{}c@{}}\textbf{\# Params}\\ (M)\end{tabular} &
  \begin{tabular}[c]{@{}c@{}}\textbf{FLOPs}\\ (G)\end{tabular} &
  \begin{tabular}[c]{@{}c@{}}\textbf{Inference Speed}\\ (FPS/VPS)\end{tabular} &
  \textbf{Results} &
\textbf{Task} &
  \textbf{Model} &
  \begin{tabular}[c]{@{}c@{}}\textbf{\# Params}\\ (M)\end{tabular} &
  \begin{tabular}[c]{@{}c@{}}\textbf{FLOPs}\\ (G)\end{tabular} &
  \begin{tabular}[c]{@{}c@{}}\textbf{Inference Speed}\\ (FPS/VPS)\end{tabular} &
  \textbf{Results}
\\ \hhline{-||-----||-||-----}

& GraphAU(ViT-S) & 22.97 & 15.86 & 58 & 60.69 & &
  DDAMFN & 4.14 & 0.55 & 68 & 75.67 \\
& GraphAU(Res50) & 33.25 & 18.22 & 94 & 60.46 & &
  MHAN & 4.20 & 0.55 & 67 & 75.91 \\
& MAE-Face-B & 86.10 & 16.87 & 131 & 63.42 & &
  EMOFAN & 13.96 & 16.98 & 123 & 74.55 \\
& FMAE(ViT-B) & 86.10 & 16.87 & 131 & 62.49 & &
  Poster++ & 43.85 & 8.61 & 90 & 74.70 \\
& GraphAU(ViT-B) & 87.18 & 28.16 & 70 & 62.78 & &
  MAE-Face-B & 87.00 & 16.87 & 131 & 75.17 \\
& MDHR(Res50) & 126.66 & 127.31 & 940 & $58.73^{*}$ & &
  FMAE-B & 87.00 & 16.87 & 131 & 74.84 \\
& FMAE(ViT-L) & 303.84 & 59.70 & 71 & 62.42 &
\multicolumn{1}{c|}{\multirow{-7}{*}{\begin{tabular}[c]{@{}c@{}}SFER\\ (ACC↑)\end{tabular}}} &
  FMAE-L & 305.42 & 59.70 & 70 & 76.49 \\
\multicolumn{1}{c|}{\multirow{-8}{*}{\begin{tabular}[c]{@{}c@{}}FAUD\\ (F1 score↑)\end{tabular}}}
& GraphAU(ViT-L) & 309.11 & 60.74 & 48 & 62.23 & & & & & & \\

\hhline{-||-----||-||-----}

& GraphAU(Res50) & 33.25 & 18.22 & 94 & 0.669 & &
  MGNN(MFN) & 7.56 & 37.11 & 17 & 57.91 \\
& MAE-Face-B & 86.10 & 16.87 & 131 & 0.728 & &
  SVFAP-B & 78.71 & 43.60 & 38 & 59.69 \\
& FMAE(ViT-B) & 86.10 & 16.87 & 131 & 0.687 & &
  MAE-DFER-B & 85.95 & 49.75 & 25 & 58.75 \\
& GraphAU(ViT-B) & 87.18 & 28.16 & 70 & 0.699 & &
  FRU-Adapter-B & 90.17 & 280.00 & 39 & 60.08 \\
\multicolumn{1}{c|}{\multirow{-5}{*}{\begin{tabular}[c]{@{}c@{}}FAUE\\ (ICC↑)\end{tabular}}}
& MDHR(Res50) & 126.66 & 127.31 & 940 & 0.660 &
\multicolumn{1}{c|}{\multirow{-5}{*}{\begin{tabular}[c]{@{}c@{}}DFER\\ (WAR↑)\end{tabular}}}
& TimeSFormer-B & 122.35 & 96.39 & 46 & 55.07 \\

\hhline{-||-----||-||-----}

& DDAMFN & 4.57 & 0.55 & 71 & 0.589 & &
  MGNN(MFN) & 7.53 & 10.49 & 43 & 62.32 \\
& MAVENC(MFN) & 23.47 & 45.70 & 3135 & $0.619^{*}$ & &
  SVFAP-B & 78.69 & 43.60 & 38 & 68.09 \\
& SVFAP-B & 78.16 & 43.61 & 602 & $0.605^{*}$ & &
  MAE-DFER-B & 85.93 & 49.75 & 25 & 69.53 \\
& MAE4DFER-B & 85.40 & 49.76 & 405 & $0.607^{*}$ &
\multicolumn{1}{c|}{\multirow{-4}{*}{\begin{tabular}[c]{@{}c@{}}FPLE\\ (ACC↑)\end{tabular}}} &
  TimeSFormer-B & 122.34 & 96.39 & 46 & 65.88 \\

\hhline{~~~~~~||-||-----}

& MAE-Face & 86.99 & 16.87 & 131 & 0.613 & &
  MGNN(MFN) & 7.55 & 37.11 & 17 & 0.728 \\
& FMAE-B & 86.99 & 16.87 & 131 & 0.595 & &
  MGNN(Res50) & 25.10 & 130.58 & 25 & 0.720 \\
& FRU-Adapter-B & 90.16 & 280.00 & 625 & $0.637^{*}$ & &
  SVFAP-B & 78.70 & 43.60 & 38 & 0.719 \\
& MAVENC(ViT-B) & 129.49 & 345.99 & 2568 & $0.611^{*}$ & &
  MAE-DFER-B & 85.94 & 49.75 & 25 & 0.718 \\
& FMAE-L & 305.41 & 59.70 & 71 & 0.597 & &
  FRU-Adapter-B & 90.16 & 280.00 & 39 & 0.722 \\
\multicolumn{1}{c|}{\multirow{-10}{*}{\begin{tabular}[c]{@{}c@{}}VAIE\\ (CCC↑)\end{tabular}}}
& MAVENC(Res50) & 334.01 & 221.06 & 1730 & $0.593^{*}$ &
\multicolumn{1}{c|}{\multirow{-6}{*}{\begin{tabular}[c]{@{}c@{}}FAPR\\ (CCC↑)\end{tabular}}} &
  TimeSFormer-B & 122.34 & 96.39 & 46 & 0.696 \\

\hhline{-||-----||-||-----}

& MGNN(MFN) & 5.00 & 43.73 & 11 & 6.936 & &
  SVFAP-B & 78.69 & 43.60 & 38 & 7.967 \\
\multicolumn{1}{c|}{\multirow{-2}{*}{\begin{tabular}[c]{@{}c@{}}FDA\\ (RMSE↓)\end{tabular}}}
& MGNN(Res50) & 25.10 & 296.00 & 3 & 6.623 &
\multicolumn{1}{c|}{\multirow{-2}{*}{\begin{tabular}[c]{@{}c@{}}FDA\\ (RMSE↓)\end{tabular}}}
& MAE-DFER-B & 85.93 & 49.75 & 25 & 7.533 \\

\Xhline{2.5px}
\end{tabular}
}
\label{tab:model_efficiency}
\end{table*}

\section{Additional implementation details}
\label{sec: implementation details}

\textbf{Dual-level resampling strategy.} As illustrated in Fig. \ref{fig:au_stas} and Fig. \ref{fig:fer_data}, the datasets used for function-specific model development exhibit substantial differences in both dataset sizes and label occurrence frequencies. To address these imbalances, the proposed OpenAHBU employs a \textbf{dual-level resampling strategy} that combines dataset-specific sampling weights with label frequency-based repeat factors to adjust how frequently each training example is sampled. Specifically, let $\mathcal{D}_d=\{x_{d,i}\}_{i=1}^{N_d}$ denote the training subset provided by the $d_\text{th}$ dataset with $N_d$ representing the number of its training examples before resampling.

At the dataset level, all examples in $\mathcal{D}_d$ are assigned a common sampling weight $w_d$. These weights are set in the model configuration files inversely proportional to the corresponding training subset sizes as:
\begin{equation}
    w_d \propto \frac{1}{N_d}
\end{equation}
Without this adjustment, sampling uniformly from the merged training set would expose the model to larger datasets more frequently. This can bias the model towards their data distributions, while providing fewer opportunities to learn the facial behaviours represented by smaller datasets, potentially limiting generalisation across datasets. Inverse-size weighting compensates for these size differences by assigning higher weights to examples from smaller datasets, providing more balanced exposure to different data sources during training. This weighting balances the aggregate contributions of the datasets before label-level repeat factors are applied, and the final sampling proportions also depend on the label distributions within each dataset.

At the label-level, label category frequencies are calculated separately within each dataset. Let $\mathcal{C}_{d,i}$ denote the set of category or AU labels associated with sample $x_{d,i}$, the occurrence frequency of each label category $c$ in $\mathcal{D}_d$ is defined as:
\begin{equation}
    f_{d,c} = \frac{1}{N_d}\sum_{i=1}^{N_d}\mathbb{I}\!\left(c\in\mathcal{C}_{d,i}\right)
\end{equation}
where $\mathbb{I}(\cdot)$ is the indicator function. For each observed label, its repeat factor is computed as:
\begin{equation}
    r_{d,c}=
    \begin{cases}
        1, & f_{d,c}\geq\tau, \\[2pt]
        \eta\left(\dfrac{\tau}{f_{d,c}}\right)^{\gamma},
        & 0<f_{d,c}<\tau
    \end{cases}
\end{equation}
where $\tau\in(0,1]$ is the frequency threshold for identifying under-represented labels, $\gamma\in{1/2,1}$ controls the strength of frequency-dependent reweighting, and $\eta\geq1$ scales the repeat factors of labels below the threshold. Labels with frequencies at or above $\tau$ retain a repeat factor of 1, whereas less frequent labels receive progressively larger repeat factors.
For an example associated with a single category, its label-level repeat factor is simply the repeat factor of that category. For an example associated with multiple labels, such as several AUs, we assign it the largest repeat factor among those labels:
\begin{equation}
    r_{d,i} = \max_{c\in\mathcal{C}_{d,i}} r_{d,c}
\end{equation}
where $r_{d,i}=1$ when $\mathcal{C}_{d,i}$ is empty. This allows an example containing an under-represented AU to be sampled more frequently even when the same example also contains commonly occurring AUs.

Consequently, the dataset-level weight and label-level repeat factor are then multiplied for each training example, where each product is divided by the smallest product across all training examples to obtain the final repeat factor as:
\begin{equation}
    R_{d,i} =
    \frac{w_d r_{d,i}}
    {\displaystyle\min_{d',j}
    \left(w_{d'}r_{d',j}\right)}
\end{equation}
where $d'$ ranges over all $D$ datasets, and $j$ ranges over the $N_{d'}$ training examples in dataset $d'$. The denominator is therefore the smallest combined factor across the entire merged training set. This normalisation sets the smallest repeat factor to 1.

Following the repeat factor sampling \cite{gupta2019lvis}, we construct a training index list in which $R_{d,i}$ specifies the expected occurrence number of each example $x_{d,i}$. To convert fractional repeat factors into integer occurrence counts without systematic rounding bias, we apply stochastic rounding \cite{connolly2021stochastic} as:
\begin{equation}
    \begin{aligned}
        K_{d,i} &= \left\lfloor R_{d,i}\right\rfloor+B_{d,i}, \\
        B_{d,i} &\sim \operatorname{Bernoulli}\!\left(
        R_{d,i}-\left\lfloor R_{d,i}\right\rfloor
        \right)
    \end{aligned}
\end{equation}
where $K_{d,i}$ denotes the number of occurrences of sample $x_{d,i}$ in each resampling round, $\lfloor R_{d,i}\rfloor$ is the integer part of its repeat factor, and $\operatorname{Bernoulli}(p)$ denotes a Bernoulli distribution. Thus, $B_{d,i}$ determines whether one additional occurrence is included, with probability $R_{d,i}-\lfloor R_{d,i}\rfloor$. This gives $\mathbb{E}[K_{d,i}]=R_{d,i}$, preserving the intended repetition count in expectation. The resulting indices are shuffled and distributed across training processes, with occurrence counts redrawn whenever a new index list is generated. The dataset weights and label-level resampling parameters are provided in the corresponding model configuration files.

\noindent \textbf{Software and hardware details.} All model training, evaluation, and computational-efficiency measurements performed for OpenAHBU were conducted on a computing platform equipped with an Intel Xeon 6986P-C CPU and NVIDIA L20 GPUs. The platform ran Ubuntu 22.04, with a software environment comprising Python 3.12, PyTorch 2.5.1, and CUDA 12.4. The complete environment specification, including all required third-party libraries and their versions, will be released together with the full source code to facilitate reproducibility.

\noindent \textbf{Hyper-parameters settings.} As described in Sec. 3.4.2 of the main paper, we employ sharpness-aware minimisation (SAM) \cite{foret2021sharpnessaware} to improve generalisation across the heterogeneous datasets used for function-specific model development. Specifically, all trainable models are optimised using its adaptive variant, ASAM \cite{kwon2021asam}, with the neighbourhood radius set to $\rho=2$ following its default configuration. A fixed random seed of 0 is used for all experiments. \textbf{The remaining model-specific hyper-parameters follow the official implementations of the corresponding methods.} \textbf{The model-specific configuration files is released together with the full source code to facilitate reproducibility.}

\section{Evaluation Metrics}
\label{metrics}

In this section, we detail the primary evaluation metrics employed for the eight re-developed OpenAHBU functions, together with their mathematical definitions.

\subsection{Facial AU Occurrence Detection (FAUD)}

We follow previous FAUD studies \cite{luo2022learning, wang2024multi, ma2024facial} using the \textbf{F1 score (Eq. \ref{eq:faud_f1})} and \textbf{the area under the receiver operating characteristic curve (ROC-AUC, Eq. \ref{eq:faud_auc})} to evaluate the performance of FAUD models in OpenAHBU, where F1 score is formulated as: 
\begin{equation}
\mathrm{F1}_k =
\frac{2\mathrm{TP}_k}
{2\mathrm{TP}_k+\mathrm{FP}_k+\mathrm{FN}k},
\qquad
\mathrm{F1}
=\frac{1}{K}\sum_{k=1}^{K}\mathrm{F1}_{k},
\label{eq:faud_f1}
\end{equation}
where $\mathrm{TP}_k$, $\mathrm{FP}_k$, and $\mathrm{FN}_k$ denote the numbers of true-positive, false-positive, and false-negative predictions for the $k_\text{th}$ AU, respectively, and $K$ is the number of evaluated AUs. A higher F1 score indicates better AU occurrence detection performance.

Meanwhile, the ROC-AUC evaluates the ability of FAUD models to distinguish each AU's occurrence from absence across different decision thresholds. The per-AU and macro-averaged ROC-AUC are defined as:
\begin{equation}
    \mathrm{AUC}_{k}
    =
    \int_{0}^{1} \mathrm{TPR}_{k}(u)\,\mathrm{d}u,
    \quad
    \mathrm{AUC}
    =
    \frac{1}{\lvert\mathcal{K}_{\mathrm{valid}}\rvert}
    \sum_{k \in \mathcal{K}_{\mathrm{valid}}}
    \mathrm{AUC}_{k},
    \label{eq:faud_auc}
\end{equation}
where $u$ denotes the false-positive rate, $\mathrm{FPR}_{k} = \mathrm{FP}_{k}/(\mathrm{FP}_{k}+\mathrm{TN}_{k})$, and $\mathrm{TPR}_{k}(u)$ denotes the corresponding true-positive rate, calculated as $\mathrm{TP}_{k}/(\mathrm{TP}_{k}+\mathrm{FN}_{k})$, along the ROC curve. Here, $\mathrm{TN}_{k}$ denotes the number of true-negative predictions for the $k_\text{th}$ AU. Samples with missing annotations are excluded, and $\mathcal{K}_{\mathrm{valid}}$ contains the AUs with both positive and negative annotated samples. ROC-AUC ranges from 0 to 1, with higher values indicating better AU discrimination. An ROC-AUC of 1 represents perfect discrimination, whereas an ROC-AUC of 0.5 corresponds to chance-level ranking.

\subsection{Facial AU Intensity Estimation (FAUE)}

We follow previous FAUE studies \cite{sanchez2021affective, bulat2022pre, ma2024facial} using \textbf{mean squared error (MSE, Eq. \ref{eq:mse})}, \textbf{Pearson correlation coefficient (PCC, Eq. \ref{eq:pcc})}, and \textbf{the intraclass correlation coefficient (ICC (3,1), Eq. \ref{eq:icc})} \cite{shrout1979intraclass} to evaluate the performance of FAUE models in OpenAHBU, where MSE and PCC are defined as: 
\begin{equation}
\mathrm{MSE} =
\frac{1}{n}\sum_{i=1}^{n}\left(p_i-g_i\right)^2,
\label{eq:mse}
\end{equation}
\begin{equation}
\mathrm{PCC} =
\rho_{p,g} =
\frac{
\sum_{i=1}^{n}\left(p_i-\mu_p\right)
\left(g_i-\mu_g\right)
}{
\sqrt{
\sum_{i=1}^{n}\left(p_i-\mu_p\right)^2
\sum_{i=1}^{n}\left(g_i-\mu_g\right)^2
}},
\label{eq:pcc}
\end{equation}
where $p_i$ and $g_i$ denote the predicted and ground-truth intensities of a given AU for the $i_\text{th}$ sample, respectively. $n$ is the number of samples, and $\mu_p$ and $\mu_g$ are the mean values of the predictions and ground-truth intensities.

Treating the predictions and ground-truth intensities as two fixed measurement sources, then the ICC (3,1) is defined as:
\begin{equation}
\mathrm{ICC}(3,1)
=
\frac{\mathrm{MS}_{R}-\mathrm{MS}_{E}}
{\mathrm{MS}_{R}+(q-1)\mathrm{MS}_{E}},
\label{eq:icc}
\end{equation}
where $\mathrm{MS}_{R}$ and $\mathrm{MS}_{E}$ denote the mean square for rows and the residual mean square obtained from a two-way mixed-effects analysis of variance, respectively, and $q=2$ is the number of measurement sources. All three metrics are computed separately for each AU and then averaged across the evaluated AUs. Lower MSE and higher PCC and ICC(3,1) indicate better AU intensity estimation performance.

\subsection{Static Facial Expression Recognition (SFER)}

We follow previous SFER studies \cite{zhang2023dual, mao2025poster++, wang2026mhan} using \textbf{Classification Accuracy (Eq. \ref{eq:acc})} and \textbf{F1-score (Eq. \ref{eq:faud_f1})} to evaluate the performance of SFER models in OpenAHBU, where Classification Accuracy is defined as: 
\begin{equation}
\mathrm{Accuracy} =
\frac{1}{n}\sum_{i=1}^{n}
\mathbb{I}\left(\hat{y}_i=y_i\right),
\label{eq:acc}
\end{equation}
where $\hat{y}_i$ and $y_i$ denote the predicted and ground-truth expression labels of the $i_\text{th}$ sample, respectively, and $\mathbb{I}(\cdot)$ is the indicator function, which equals one when its argument is true and zero otherwise. The F1 score is calculated for each expression class according to Eq. \ref{eq:faud_f1} ($K$ now denotes the number of evaluated expression classes) and then macro-averaged across all expression classes. Higher accuracy and F1 scores indicate better static facial expression recognition performance.

\subsection{Dynamic Facial Expression Recognition (DFER)}
We follow previous DFER studies \cite{sun2023mae, sun2024svfap, chen2025static} using \textbf{unweighted average recall (UAR)} and \textbf{weighted average recall (WAR)} to evaluate the performance of DFER models in OpenAHBU, which are defined as:

\begin{equation}
    \mathrm{UAR}
    =
    \frac{1}{C}
    \sum_{c=1}^{C}
    \frac{\mathrm{TP}_{c}}{N_{c}},
    \label{eq:uar}
\end{equation}

\begin{equation}
    \mathrm{WAR}
    =
    \sum_{c=1}^{C}
    \frac{N_{c}}{n}
    \frac{\mathrm{TP}_{c}}{N_{c}} 
    =
    \frac{1}{n}
    \sum_{c=1}^{C}
    \mathrm{TP}_{c}
    \label{eq:war}
\end{equation}
where $C$ is the number of dynamic expression classes, $N_{c}$ is the number of ground-truth samples belonging to the $c_\text{th}$ class, $\mathrm{TP}_{c}$ is the number of correctly classified samples from that class, and $n=\sum_{c=1}^{C}N_{c}$ is the total number of samples. UAR assigns equal importance to all expression classes, whereas WAR weights the recall of each class according to its sample proportion and is equivalent to overall classification accuracy in the single-label setting. Higher UAR and WAR indicate better dynamic facial expression recognition performance.

\subsection{Valence/Arousal Intensity Estimation (VAIE)}

We follow previous VAIE studies \cite{toisoul2021estimation, li2025temporal, dresvyanskiy2024multi} using \textbf{MSE (Eq. \ref{eq:mse})} and \textbf{the concordance correlation coefficient (CCC, Eq. \ref{eq:ccc})} to evaluate the performance of VAIE models in OpenAHBU, with CCC defined as:
\begin{equation}
\mathrm{CCC} =
\frac{
2\rho_{p,g}\sigma_p\sigma_g
}{
\sigma_p^2+\sigma_g^2+
\left(\mu_p-\mu_g\right)^2
},
\label{eq:ccc}
\end{equation}
where $\rho_{p,g}$ is the PCC between the predictions and ground-truth values defined in Eq. \ref{eq:pcc}, while $\mu_p$, $\mu_g$, $\sigma_p$, and $\sigma_g$ denote their respective means and standard deviations. 
Both MSE and CCC are calculated separately for valence and arousal, and their respective averages are reported as the overall VAIE performance as:
\begin{equation}
\mathrm{MSE}_{\mathrm{VA}} =
\frac{\mathrm{MSE}_{\mathrm{V}}+
\mathrm{MSE}_{\mathrm{A}}}{2},
\label{eq:mean_mse}
\end{equation}

\begin{equation}
\mathrm{CCC}_{\mathrm{VA}} =
\frac{\mathrm{CCC}_{\mathrm{V}}+
\mathrm{CCC}_{\mathrm{A}}}{2},
\label{eq:mean_ccc}
\end{equation}
where $\mathrm{MSE}_{\mathrm{V}}$ and $\mathrm{MSE}_{\mathrm{A}}$ denote the MSE values for
valence and arousal, respectively, and $\mathrm{CCC}_{\mathrm{V}}$ and $\mathrm{CCC}_{\mathrm{A}}$ denote the corresponding CCC values. A lower MSE indicates smaller prediction errors, while a higher CCC indicates better agreement between the predicted and ground-truth affect intensities.

\subsection{Facial Pain Level Estimation (FPLE)}

We follow previous FPLE studies \cite{gkikas2023full, gkikas2024multimodal, gkikas2025painformer} using \textbf{Accuracy}, and \textbf{F1-score} to evaluate the performance of \textbf{the binary classification FPLE models} in OpenAHBU. Accuracy is calculated using Eq. \ref{eq:uar} ($C$=2), while the class-wise F1 scores are calculated using Eq. \ref{eq:faud_f1} and macro-averaged across the pain and pain-free classes. Higher accuracy and F1 scores indicate better binary pain recognition performance.

Additionally, we report \textbf{the root mean square error (RMSE, Eq. \ref{eq:rmse})}, \textbf{PCC}, and \textbf{CCC} for \textbf{the five-level regression FPLE models} in OpenAHBU, where RMSE is defined as:
\begin{equation}
\mathrm{RMSE} =
\sqrt{
\frac{1}{n}
\sum_{i=1}^{n}
\left(p_i-g_i\right)^2
},
\label{eq:rmse}
\end{equation}
where $p_i$ and $g_i$ denote the predicted and ground-truth pain intensity levels of the $i$-th sample, respectively. PCC and CCC are calculated using Eqs. \ref{eq:pcc} and \ref{eq:ccc}, respectively. Lower RMSE and higher PCC and CCC indicate better five-level pain intensity estimation performance.

\subsection{Facial Depression Assessment (FDA)}
We follow previous FDA studies \cite{xu2024two, wu2025depmgnn, liao2026explainable} using \textbf{root mean square error (RMSE, Eq. \ref{eq:rmse})}, \textbf{mean absolute error (MAE, Eq. \ref{eq:mae})}, \textbf{PCC (Eq. \ref{eq:pcc})}, and \textbf{CCC (Eq. \ref{eq:ccc})} to evaluate the performance of FDA models in OpenAHBU. MAE is defined as:
\begin{equation}
\mbox{MAE} = \frac{1}{n} \sum_{i = 1}^n \vert p_i -g_i \vert,
\label{eq:mae}
\end{equation}
where $p_i$ and $g_i$ denote the predicted and ground-truth depression severity scores of the $i_\text{th}$ sample, respectively. Lower RMSE and MAE indicate smaller prediction errors, whereas higher PCC and CCC indicate stronger correlation and agreement between the predicted and ground-truth depression severity scores.

\subsection{Facial Apparent Personality Recognition (FAPR)}
We follow previous FAPR studies \cite{liao2024open, sun2024svfap, wu2025depmgnn} using \textbf{MSE (Eq. \ref{eq:mse})}, \textbf{PCC (Eq. \ref{eq:pcc})}, and \textbf{CCC (Eq. \ref{eq:ccc})} to evaluate the performance of FAPR models in OpenAHBU. Each metric is calculated separately for Openness, Conscientiousness, Extraversion, Agreeableness, and Neuroticism and then averaged across the five apparent personality traits. Lower MSE and higher PCC and CCC indicate better apparent personality recognition performance.

\section{Experiment Results}
\label{sec:experiments}

\subsection{Function-level OpenAHBU model performances}

\subsubsection{FAUD performances}

\begin{table*}[htbp]
\Huge
\centering
\setlength{\doublerulesep}{0.3em}
\renewcommand{\arraystretch}{1.5}
\caption{AU-wise F1 scores (\%) $\uparrow$ on the FAUD function-level test set formed from six benchmark datasets: BP4D, DISFA, RAF-AU, CK+, CASME2, and AFFW-AU. The best and second-best results among models with complete six-dataset coverage are highlighted in bold and underlined, respectively. Because MDHR requires temporal inputs, it is not evaluated on datasets processed as frame-level inputs under our protocol; unavailable results are denoted by ``-'', and its average is calculated over the available AUs only.}
\resizebox{1.0\linewidth}{!}{
\begin{tabular}{l|c|cccccccccccccccccc|c}
\Xhline{5px}
\rowcolor{gray!15} 
\cellcolor{gray!15} & \cellcolor{gray!15} &
  \multicolumn{18}{c|}{\cellcolor{gray!15} \textbf{AU}} &
  \cellcolor{gray!15} \\ \hhline{~~------------------}
\rowcolor{gray!15} 
\multirow{-2}{*}{\cellcolor{gray!15}\textbf{Methods}} &
    \multirow{-2}{*}{\cellcolor{gray!15}\textbf{Year}} &
  \textbf{01} &
  \textbf{02} &
  \cellcolor{gray!15}\textbf{04} &
  \cellcolor{gray!15}\textbf{05} &
  \cellcolor{gray!15}\textbf{06} &
  \cellcolor{gray!15}\textbf{07} &
  \textbf{09} &
  \textbf{10} &
  \textbf{12} &
  \textbf{14} &
  \textbf{15} &
  \textbf{16} &
  \textbf{17} &
  \textbf{23} &
  \textbf{24} &
  \textbf{25} &
  \textbf{26} &
  \textbf{27} &
  \multirow{-2}{*}{\cellcolor{gray!15}\textbf{Avg.}} \\ \hhline{=====================}
GraphAU(Res50) \cite{luo2022learning} &
  IJCAI 2022 &
  55.45 &
  47.77 &
  56.14 &
  70.07 &
  65.02 &
  74.34 &
  51.91 &
  75.14 &
  75.48 &
  68.54 &
  42.55 &
  65.48 &
  67.44 &
  36.75 &
  38.02 &
  82.99 &
  32.32 &
  82.93 &
  60.46 \\
GraphAU(ViT-S) \cite{luo2022learning} &
  IJCAI 2022 &
  56.40 &
  \cellcolor{gray!30}\textbf{49.00} &
  59.04 &
  66.32 &
  65.86 &
  71.18 &
  55.43 &
  75.45 &
  74.81 &
  67.21 &
  39.91 &
  61.36 &
  67.63 &
  37.03 &
  36.36 &
  82.11 &
  \cellcolor{gray!15}\underline{42.25} &
  85.02 &
  60.69 \\
GraphAU(ViT-B) \cite{luo2022learning} &
  IJCAI 2022 &
  \cellcolor{gray!30}\textbf{58.76} &
  47.30 &
  \cellcolor{gray!30}\textbf{61.35} &
  \cellcolor{gray!15}\underline{72.73} &
  \cellcolor{gray!15}\underline{67.11} &
  75.26 &
  62.71 &
  \cellcolor{gray!30}\textbf{77.74} &
  \cellcolor{gray!15}\underline{75.98} &
  \cellcolor{gray!15}\underline{68.98} &
  \cellcolor{gray!15}\underline{42.92} &
  65.82 &
  68.45 &
  37.98 &
  \cellcolor{gray!15}\underline{38.46} &
  \cellcolor{gray!30}\textbf{84.95} &
  37.31 &
  \cellcolor{gray!15}\underline{86.24} &
  \cellcolor{gray!15}\underline{62.78} \\
GraphAU(ViT-L) \cite{luo2022learning} &
  IJCAI 2022 &
  \cellcolor{gray!15}\underline{57.98} &
  \cellcolor{gray!15}\underline{47.80} &
  58.51 &
  71.78 &
  65.96 &
  73.50 &
  57.12 &
  76.69 &
  75.81 &
  \cellcolor{gray!30}\textbf{69.43} &
  42.72 &
  66.67 &
  \cellcolor{gray!15}\underline{68.56} &
  \cellcolor{gray!15}\underline{39.30} &
  38.25 &
  84.29 &
  41.90 &
  83.97 &
  62.23 \\ \hhline{=====================}
\color{gray!75}MDHR(Res50) \cite{wang2024multi} &
  CVPR 2024 &
  \color{gray!75}59.72 &
  \color{gray!75}50.78 &
  \color{gray!75}56.86 &
  \color{gray!75}- &
  \color{gray!75}64.88 &
  \color{gray!75}75.40 &
  \color{gray!75}49.03 &
  \color{gray!75}75.79 &
  \color{gray!75}76.45 &
  \color{gray!75}63.91 &
  \color{gray!75}40.87 &
  \color{gray!75}- &
  \color{gray!75}64.45 &
  \color{gray!75}39.79 &
  \color{gray!75}38.53 &
  \color{gray!75}84.65 &
  \color{gray!75}39.83 &
  \color{gray!75}- &
  \color{gray!75}58.73 \\ \hhline{=====================}
MAE-Face-B \cite{ma2024facial} &
  TAC 2024 &
  55.15 &
  46.02 &
  59.83 &
  \cellcolor{gray!30}\textbf{73.38} &
  66.96 &
  75.55 &
  \cellcolor{gray!30}\textbf{77.10} &
  76.83 &
  \cellcolor{gray!30}\textbf{76.15} &
  67.14 &
  41.99 &
  \cellcolor{gray!30}\textbf{67.81} &
  68.38 &
  38.23 &
  37.01 &
  84.14 &
  \cellcolor{gray!30}\textbf{43.79} &
  86.16 &
  \cellcolor{gray!30}\textbf{63.42} \\ 
  FMAE-B \cite{ning2024representation} &
  FG 2025 &
  57.51 &
  45.49 &
  58.29 &
  69.70 &
  \cellcolor{gray!30}\textbf{67.13} &
  \cellcolor{gray!30}\textbf{75.73} &
  64.16 &
  \cellcolor{gray!15}\underline{77.08} &
  75.95 &
  65.16 &
  \cellcolor{gray!30}\textbf{45.34} &
  \cellcolor{gray!15}\underline{66.91} &
  68.32 &
  39.21 &
  \cellcolor{gray!30}\textbf{39.22} &
  84.49 &
  40.00 &
  85.12 &
  62.49 \\ 
  FMAE-L \cite{ning2024representation} &
  FG 2025 &
  53.09 &
  44.18 &
  \cellcolor{gray!15}\underline{59.84} &
  70.46 &
  66.87 &
  \cellcolor{gray!15}\underline{75.69} &
  \cellcolor{gray!15}\underline{71.42} &
  76.69 &
  75.09 &
  68.84 &
  37.03 &
  66.89 &
  \cellcolor{gray!30}\textbf{68.69} &
  \cellcolor{gray!30}\textbf{39.34} &
  37.37 &
  \cellcolor{gray!15}\underline{84.72} &
  41.11 &
  \cellcolor{gray!30}\textbf{86.41} &
  62.42 \\ 
  \Xhline{5px}
\end{tabular}
}
\label{tab:aud_metric}
\end{table*}

\begin{table*}[htbp]
\Huge
\centering
\setlength{\doublerulesep}{0.3em}
\renewcommand{\arraystretch}{1.5}
\caption{AU-wise ROC-AUC scores (\%) $\uparrow$ on the FAUD function-level test set formed from six benchmark datasets: BP4D, DISFA, RAF-AU, CK+, CASME2, and AFFW-AU. The best and second-best results among models with complete six-dataset coverage are highlighted in bold and underlined, respectively. Because MDHR requires temporal inputs, it is not evaluated on datasets processed as frame-level inputs under our protocol; unavailable results are denoted by ``-'', and its average is calculated over the available AUs only.}
\resizebox{1.0\linewidth}{!}{
\begin{tabular}{l|c|cccccccccccccccccc|c}
\Xhline{5px}
\rowcolor{gray!15} 
\cellcolor{gray!15} & \cellcolor{gray!15} &
  \multicolumn{18}{c|}{\cellcolor{gray!15} \textbf{AU}} &
  \cellcolor{gray!15} \\ \hhline{~~------------------}
\rowcolor{gray!15} 
\multirow{-2}{*}{\cellcolor{gray!15}\textbf{Methods}} &
    \multirow{-2}{*}{\cellcolor{gray!15}\textbf{Year}} &
  \textbf{01} &
  \textbf{02} &
  \cellcolor{gray!15}\textbf{04} &
  \cellcolor{gray!15}\textbf{05} &
  \cellcolor{gray!15}\textbf{06} &
  \cellcolor{gray!15}\textbf{07} &
  \textbf{09} &
  \textbf{10} &
  \textbf{12} &
  \textbf{14} &
  \textbf{15} &
  \textbf{16} &
  \textbf{17} &
  \textbf{23} &
  \textbf{24} &
  \textbf{25} &
  \textbf{26} &
  \textbf{27} &
  \multirow{-2}{*}{\cellcolor{gray!15}\textbf{Avg.}} \\ \hhline{=====================}
GraphAU(Res50) \cite{luo2022learning} &
  IJCAI 2022 &
  85.17 &
  78.87 &
  84.16 &
  \cellcolor{gray!30}\textbf{90.89} &
  84.72 &
  87.71 &
  95.05 &
  90.24 &
  90.74 &
  71.77 &
  \cellcolor{gray!15}\underline{89.77} &
  89.40 &
  83.63 &
  \cellcolor{gray!15}\underline{84.99} &
  81.90 &
  88.82 &
  77.26 &
  96.83 &
  86.22 \\
GraphAU(ViT-S) \cite{luo2022learning} &
  IJCAI 2022 &
  83.19 &
  79.02 &
  85.35 &
  87.21 &
  84.68 &
  86.52 &
  97.08 &
  90.14 &
  90.02 &
  72.54 &
  88.86 &
  88.38 &
  82.34 &
  83.17 &
  82.30 &
  88.25 &
  77.70 &
  96.44 &
  85.73 \\
GraphAU(ViT-B) \cite{luo2022learning} &
  IJCAI 2022 &
  \cellcolor{gray!30}\textbf{85.60} &
  \cellcolor{gray!15}\underline{80.88} &
  \cellcolor{gray!30}\textbf{87.56} &
  90.02 &
  \cellcolor{gray!30}\textbf{85.95} &
  87.85 &
  \cellcolor{gray!30}\textbf{98.16} &
  \cellcolor{gray!30}\textbf{90.93} &
  91.50 &
  73.80 &
  88.71 &
  \cellcolor{gray!15}\underline{91.54} &
  \cellcolor{gray!30}\textbf{84.97} &
  83.93 &
  82.77 &
  \cellcolor{gray!15}\underline{89.69} &
  78.56 &
  96.46 &
  \cellcolor{gray!30}\textbf{87.16} \\
\end{tabular}
}
\label{tab:aud_auc}
\end{table*}

\begin{table*}[htbp]
\centering
\setlength{\doublerulesep}{0.3em}
\renewcommand{\arraystretch}{1.5}
\caption{AU-wise F1 scores (\%) $\uparrow$ on the BP4D dataset.}
\resizebox{1.0\linewidth}{!}{
\begin{tabular}{l|c|cccccccccccc|c}
\Xhline{1.5px}
\rowcolor{gray!15} 
\cellcolor{gray!15} & \cellcolor{gray!15} &
  \multicolumn{12}{c|}{\cellcolor{gray!15} \textbf{AU}} &
  \cellcolor{gray!15} \\ \hhline{~~------------}
\rowcolor{gray!15} 
\multirow{-2}{*}{\cellcolor{gray!15}\textbf{Methods}} &
    \multirow{-2}{*}{\cellcolor{gray!15}\textbf{Year}} &
  \textbf{01} &
  \textbf{02} &
  \cellcolor{gray!15}\textbf{04} &
  \cellcolor{gray!15}\textbf{06} &
  \cellcolor{gray!15}\textbf{07} &
  \textbf{10} &
  \textbf{12} &
  \textbf{14} &
  \textbf{15} &
  \textbf{17} &
  \textbf{23} &
  \textbf{24} &
  \multirow{-2}{*}{\cellcolor{gray!15}\textbf{Avg.}} \\ \hhline{===============}
GraphAU(Res50) \cite{luo2022learning} &
  IJCAI 2022 &
  55.47 &
  34.47 &
  54.11 &
  82.61 &
  82.68 &
  84.46 &
  86.30 &
  72.81 &
  60.76 &
  68.26 &
  53.04 &
  57.94 &
  66.08 \\
GraphAU(ViT-S) \cite{luo2022learning} &
  IJCAI 2022 &
  55.39 &
  40.31 &
  58.89 &
  80.81 &
  83.33 &
  83.78 &
  87.60 &
  70.56 &
  57.43 &
  67.98 &
  51.88 &
  59.57 &
  66.46 \\
GraphAU(ViT-B) \cite{luo2022learning} &
  IJCAI 2022 &
  64.10 &
  37.03 &
  61.01 &
  80.69 &
  82.76 &
  85.12 &
  89.40 &
  71.04 &
  56.98 &
  68.88 &
  52.30 &
  58.54 &
  67.32 \\
GraphAU(ViT-L) \cite{luo2022learning} &
  IJCAI 2022 &
  61.64 &
  32.49 &
  52.14 &
  80.68 &
  80.63 &
  85.80 &
  88.68 &
  70.35 &
  60.24 &
  68.04 &
  54.12 &
  61.81 &
  66.39 \\ \hhline{===============}
MDHR(Res50) \cite{wang2024multi} &
  CVPR 2024 &
  59.52 &
  41.40 &
  56.50 &
  81.47 &
  82.12 &
  83.36 &
  86.39 &
  65.80 &
  62.54 &
  65.13 &
  55.92 &
  59.43 &
  66.63\\ \hhline{===============}
MAE-Face \cite{ma2024facial} &
  TAC 2024 &
  60.64 &
  38.10 &
  59.14 &
  80.01 &
  83.27 &
  84.98 &
  88.61 &
  69.19 &
  59.88 &
  68.53 &
  55.50 &
  60.72 &
  67.38 \\
\end{tabular}
}
\label{tab:aud_metric_bp4d}
\end{table*}

\begin{table*}[htbp]
\centering
\setlength{\doublerulesep}{0.3em}
\renewcommand{\arraystretch}{1.5}
\caption{AU-wise ROC-AUC scores (\%) $\uparrow$ on the BP4D dataset.}
\resizebox{1.0\linewidth}{!}{
\begin{tabular}{l|c|cccccccccccc|c}
\Xhline{1.5px}
\rowcolor{gray!15} 
\cellcolor{gray!15} & \cellcolor{gray!15} &
  \multicolumn{12}{c|}{\cellcolor{gray!15} \textbf{AU}} &
  \cellcolor{gray!15} \\ \hhline{~~------------}
\rowcolor{gray!15} 
\multirow{-2}{*}{\cellcolor{gray!15}\textbf{Methods}} &
    \multirow{-2}{*}{\cellcolor{gray!15}\textbf{Year}} &
  \textbf{01} &
  \textbf{02} &
  \cellcolor{gray!15}\textbf{04} &
  \cellcolor{gray!15}\textbf{06} &
  \cellcolor{gray!15}\textbf{07} &
  \textbf{10} &
  \textbf{12} &
  \textbf{14} &
  \textbf{15} &
  \textbf{17} &
  \textbf{23} &
  \textbf{24} &
  \multirow{-2}{*}{\cellcolor{gray!15}\textbf{Avg.}} \\ \hhline{===============}
GraphAU(Res50) \cite{luo2022learning} &
  IJCAI 2022 &
  85.29 &
  76.80 &
  83.38 &
  91.44 &
  87.86 &
  89.65 &
  93.09 &
  74.52 &
  85.94 &
  82.37 &
  82.89 &
  88.86 &
  85.18 \\
GraphAU(ViT-S) \cite{luo2022learning} &
  IJCAI 2022 &
  84.31 &
  76.69 &
  86.37 &
  90.13 &
  87.20 &
  88.70 &
  92.82 &
  72.55 &
  82.84 &
  81.23 &
  80.54 &
  89.19 &
  84.38 \\
GraphAU(ViT-B) \cite{luo2022learning} &
  IJCAI 2022 &
  87.44 &
  80.16 &
  89.93 &
  89.99 &
  87.14 &
  89.32 &
  94.43 &
  71.80 &
  81.73 &
  83.95 &
  81.17 &
  89.20 &
  85.52 \\
GraphAU(ViT-L) \cite{luo2022learning} &
  IJCAI 2022 &
  86.05 &
  78.07 &
  88.87 &
  89.08 &
  83.53 &
  89.07 &
  94.38 &
  72.11 &
  83.96 &
  83.17 &
  82.10 &
  90.08 &
  85.04 \\ \hhline{===============}
MDHR(Res50) \cite{wang2024multi} &
  CVPR 2024 &
  84.64 &
  80.23 &
  84.28 &
  90.20 &
  86.06 &
  86.51 &
  93.18 &
  70.78 &
  86.77 &
  80.34 &
  84.44 &
  89.25 &
  84.72\\ \hhline{===============}
MAE-Face \cite{ma2024facial} &
  TAC 2024 &
  86.26 &
  78.42 &
  86.36 &
  88.96 &
  88.08 &
  88.65 &
  94.25 &
  71.68 &
  84.72 &
  82.73 &
  81.06 &
  88.89 &
  85.00 \\
\end{tabular}
}
\label{tab:aud_auc_bp4d}
\end{table*}

\begin{table*}[htbp]
\centering
\setlength{\doublerulesep}{0.3em}
\renewcommand{\arraystretch}{1.5}
\caption{AU-wise F1 scores (\%) $\uparrow$ on the DISFA dataset.}
\resizebox{0.8\linewidth}{!}{
\begin{tabular}{l|c|cccccccc|c}
\Xhline{1.3px}
\rowcolor{gray!15} 
\cellcolor{gray!15} & \cellcolor{gray!15} &
  \multicolumn{8}{c|}{\cellcolor{gray!15} \textbf{AU}} &
  \cellcolor{gray!15} \\ \hhline{~~--------}
\rowcolor{gray!15} 
\multirow{-2}{*}{\cellcolor{gray!15}\textbf{Methods}} &
    \multirow{-2}{*}{\cellcolor{gray!15}\textbf{Year}} &
  \textbf{01} &
  \textbf{02} &
  \cellcolor{gray!15}\textbf{04} &
  \cellcolor{gray!15}\textbf{06} &
  \textbf{09} &
  \textbf{12} &
  \textbf{25} &
  \textbf{26} &
  \multirow{-2}{*}{\cellcolor{gray!15}\textbf{Avg.}} \\ \hhline{===========}
GraphAU(Res50) \cite{luo2022learning} &
  IJCAI 2022 &
  76.72 &
  52.76 &
  64.87 &
  34.52 &
  49.89 &
  62.45 &
  86.96 &
  55.52 &
  60.46 \\
GraphAU(ViT-S) \cite{luo2022learning} &
  IJCAI 2022 &
  72.77 &
  59.28 &
  74.02 &
  44.23 &
  49.82 &
  70.42 &
  92.39 &
  62.60 &
  65.69 \\
GraphAU(ViT-B) \cite{luo2022learning} &
  IJCAI 2022 &
  76.91 &
  62.57 &
  74.78 &
  48.07 &
  62.27 &
  65.36 &
  93.61 &
  64.56 &
  68.52 \\
GraphAU(ViT-L) \cite{luo2022learning} &
  IJCAI 2022 &
  74.26 &
  67.47 &
  74.22 &
  54.59 &
  54.91 &
  70.70 &
  91.84 &
  66.67 &
  69.33 \\ \hhline{===========}
MDHR(Res50) \cite{wang2024multi} &
  CVPR 2024 &
  80.33 &
  41.26 &
  55.13 &
  34.62 &
  48.99 &
  65.67 &
  90.51 &
  54.23 &
  58.84 \\ \hhline{===========}
MAE-Face \cite{ma2024facial} &
  TAC 2024 &
  78.59 &
  64.01 &
  75.96 &
  59.39 &
  76.92 &
  70.01 &
  93.87 &
  69.48 &
  73.53 \\ 
  FMAE(ViT-B) \cite{ning2024representation} &
  FG 2025 &
  73.03 &
  63.05 &
  75.19 &
  52.35 &
  62.79 &
  73.36 &
  94.08 &
  65.24 &
  69.88 \\ 
  FMAE(ViT-L) \cite{ning2024representation} &
  FG 2025 &
  76.02 &
  72.05 &
  74.85 &
  54.97 &
  70.55 &
  73.88 &
  94.55 &
  67.49 &
  73.05\\ 
  \Xhline{1.3px}
\end{tabular}
}
\label{tab:aud_metric_disfa}
\end{table*}

\begin{table*}[htbp]
\centering
\setlength{\doublerulesep}{0.3em}
\renewcommand{\arraystretch}{1.5}
\caption{AU-wise ROC-AUC scores (\%) $\uparrow$ on the DISFA dataset.}
\resizebox{0.8\linewidth}{!}{
\begin{tabular}{l|c|cccccccc|c}
\Xhline{1.3px}
\rowcolor{gray!15} 
\cellcolor{gray!15} & \cellcolor{gray!15} &
  \multicolumn{8}{c|}{\cellcolor{gray!15} \textbf{AU}} &
  \cellcolor{gray!15} \\ \hhline{~~--------}
\rowcolor{gray!15} 
\multirow{-2}{*}{\cellcolor{gray!15}\textbf{Methods}} &
    \multirow{-2}{*}{\cellcolor{gray!15}\textbf{Year}} &
  \textbf{01} &
  \textbf{02} &
  \cellcolor{gray!15}\textbf{04} &
  \cellcolor{gray!15}\textbf{06} &
  \textbf{09} &
  \textbf{12} &
  \textbf{25} &
  \textbf{26} &
  \multirow{-2}{*}{\cellcolor{gray!15}\textbf{Avg.}} \\ \hhline{===========}
GraphAU(Res50) \cite{luo2022learning} &
  IJCAI 2022 &
  97.64 &
  93.86 &
  93.06 &
  85.64 &
  95.26 &
  92.48 &
  95.11 &
  89.44 &
  92.81 \\
GraphAU(ViT-S) \cite{luo2022learning} &
  IJCAI 2022 &
  79.71 &
  85.98 &
  93.79 &
  86.02 &
  97.25 &
  91.62 &
  98.02 &
  90.56 &
  90.37 \\
GraphAU(ViT-B) \cite{luo2022learning} &
  IJCAI 2022 &
  96.18 &
  97.83 &
  95.09 &
  92.94 &
  98.19 &
  95.09 &
  98.78 &
  93.02 &
  95.89 \\
GraphAU(ViT-L) \cite{luo2022learning} &
  IJCAI 2022 &
  96.14 &
  92.74 &
  96.73 &
  94.60 &
  96.48 &
  94.90 &
  99.19 &
  93.80 &
  95.57 \\ \hhline{===========}
MDHR(Res50) \cite{wang2024multi} &
  CVPR 2024 &
  98.64 &
  96.84 &
  89.98 &
  85.99 &
  94.27 &
  89.54 &
  96.99 &
  94.28 &
  93.32 \\ \hhline{===========}
MAE-Face \cite{ma2024facial} &
  TAC 2024 &
  98.03 &
  98.09 &
  96.84 &
  94.88 &
  98.39 &
  95.72 &
  99.09 &
  95.32 &
  97.04 \\ 
  FMAE(ViT-B) \cite{ning2024representation} &
  FG 2025 &
  95.43 &
  96.91 &
  95.18 &
  92.87 &
  97.49 &
  95.96 &
  99.08 &
  95.13 &
  96.01 \\ 
  FMAE(ViT-L) \cite{ning2024representation} &
  FG 2025 &
  93.75 &
  96.62 &
  96.54 &
  94.08 &
  98.08 &
  94.52 &
  99.38 &
  95.01 &
  96.00\\ 
  \Xhline{1.3px}
\end{tabular}
}
\label{tab:aud_auc_disfa}
\end{table*}

\begin{table*}[htbp]
\centering
\setlength{\doublerulesep}{0.3em}
\renewcommand{\arraystretch}{1.5}
\caption{AU-wise F1 scores (\%) $\uparrow$ on the RAF-AU dataset.}
\resizebox{1.0\linewidth}{!}{
\begin{tabular}{l|c|ccccccccccccc|c}
\Xhline{1.5px}
\rowcolor{gray!15} 
\cellcolor{gray!15} & \cellcolor{gray!15} &
  \multicolumn{13}{c|}{\cellcolor{gray!15} \textbf{AU}} &
  \cellcolor{gray!15} \\ \hhline{~~-------------}
\rowcolor{gray!15} 
\multirow{-2}{*}{\cellcolor{gray!15}\textbf{Methods}} &
    \multirow{-2}{*}{\cellcolor{gray!15}\textbf{Year}} &
  \textbf{01} &
  \textbf{02} &
  \cellcolor{gray!15}\textbf{04} &
  \cellcolor{gray!15}\textbf{05} &
  \cellcolor{gray!15}\textbf{06} &
  \cellcolor{gray!15}\textbf{09} &
  \textbf{10} &
  \textbf{12} &
  \textbf{16} &
  \textbf{17} &
  \textbf{25} &
  \textbf{26} &
  \textbf{27} &
  \multirow{-2}{*}{\cellcolor{gray!15}\textbf{Avg.}} \\ \hhline{================}
GraphAU(Res50) \cite{luo2022learning} &
  IJCAI 2022 &
  72.73 &
  71.92 &
  83.95 &
  71.36 &
  53.81 &
  75.07 &
  73.29 &
  76.47 &
  65.48 &
  61.75 &
  94.63 &
  70.23 &
  82.93 &
  73.36 \\
GraphAU(ViT-S) \cite{luo2022learning} &
  IJCAI 2022 &
  79.36 &
  74.92 &
  82.35 &
  69.43 &
  51.09 &
  69.21 &
  69.69 &
  74.53 &
  61.21 &
  62.63 &
  93.46 &
  69.30 &
  83.87 &
  72.39 \\
GraphAU(ViT-B) \cite{luo2022learning} &
  IJCAI 2022 &
  79.26 &
  77.54 &
  84.74 &
  73.52 &
  50.81 &
  77.01 &
  76.87 &
  79.84 &
  66.06 &
  63.85 &
  95.40 &
  76.99 &
  85.89 &
  75.98 \\
GraphAU(ViT-L) \cite{luo2022learning} &
  IJCAI 2022 &
  80.44 &
  77.81 &
  83.31 &
  73.09 &
  53.66 &
  75.28 &
  74.61 &
  77.94 &
  66.89 &
  66.36 &
  94.92 &
  77.90 &
  84.46 &
  75.90 \\ \hhline{================}
MAE-Face \cite{ma2024facial} &
  TAC 2024 &
  82.43 &
  79.66 &
  84.96 &
  74.02 &
  48.89 &
  78.61 &
  76.67 &
  79.03 &
  67.81 &
  68.63 &
  95.73 &
  75.05 &
  86.16 &
  76.74 \\ 
  FMAE(ViT-B) \cite{ning2024representation} &
  FG 2025 &
  77.80 &
  74.29 &
  84.36 &
  70.83 &
  49.44 &
  77.11 &
  72.91 &
  76.15 &
  66.91 &
  67.34 &
  94.19 &
  71.98 &
  85.12 &
  74.49 \\
\end{tabular}
}
\label{tab:aud_metric_raf_au}
\end{table*}

\begin{table*}[htbp]
\centering
\setlength{\doublerulesep}{0.3em}
\renewcommand{\arraystretch}{1.5}
\caption{AU-wise ROC-AUC scores (\%) $\uparrow$ on the RAF-AU dataset.}
\resizebox{1.0\linewidth}{!}{
\begin{tabular}{l|c|ccccccccccccc|c}
\Xhline{1.5px}
\rowcolor{gray!15} 
\cellcolor{gray!15} & \cellcolor{gray!15} &
  \multicolumn{13}{c|}{\cellcolor{gray!15} \textbf{AU}} &
  \cellcolor{gray!15} \\ \hhline{~~-------------}
\rowcolor{gray!15} 
\multirow{-2}{*}{\cellcolor{gray!15}\textbf{Methods}} &
    \multirow{-2}{*}{\cellcolor{gray!15}\textbf{Year}} &
  \textbf{01} &
  \textbf{02} &
  \cellcolor{gray!15}\textbf{04} &
  \cellcolor{gray!15}\textbf{05} &
  \cellcolor{gray!15}\textbf{06} &
  \cellcolor{gray!15}\textbf{09} &
  \textbf{10} &
  \textbf{12} &
  \textbf{16} &
  \textbf{17} &
  \textbf{25} &
  \textbf{26} &
  \textbf{27} &
  \multirow{-2}{*}{\cellcolor{gray!15}\textbf{Avg.}} \\ \hhline{================}
GraphAU(Res50) \cite{luo2022learning} &
  IJCAI 2022 &
  91.11 &
  92.82 &
  93.55 &
  90.53 &
  87.77 &
  94.66 &
  89.55 &
  92.91 &
  89.40 &
  94.79 &
  96.96 &
  89.93 &
  96.83 &
  92.37 \\
GraphAU(ViT-S) \cite{luo2022learning} &
  IJCAI 2022 &
  93.71 &
  94.16 &
  92.53 &
  86.26 &
  85.17 &
  93.27 &
  86.55 &
  91.74 &
  88.38 &
  94.33 &
  96.09 &
  89.45 &
  96.44 &
  91.39 \\
GraphAU(ViT-B) \cite{luo2022learning} &
  IJCAI 2022 &
  93.48 &
  94.37 &
  94.41 &
  89.81 &
  90.13 &
  95.94 &
  90.50 &
  94.00 &
  91.54 &
  96.00 &
  97.23 &
  91.47 &
  96.46 &
  93.49 \\
GraphAU(ViT-L) \cite{luo2022learning} &
  IJCAI 2022 &
  92.91 &
  94.04 &
  93.29 &
  88.88 &
  90.37 &
  94.93 &
  90.22 &
  93.32 &
  91.67 &
  96.63 &
  96.49 &
  91.83 &
  97.27 &
  93.22 \\ \hhline{================}
MAE-Face \cite{ma2024facial} &
  TAC 2024 &
  94.27 &
  95.26 &
  94.35 &
  89.49 &
  85.40 &
  95.72 &
  91.57 &
  91.74 &
  90.82 &
  96.70 &
  97.51 &
  92.09 &
  97.31 &
  93.25 \\ 
  FMAE(ViT-B) \cite{ning2024representation} &
  FG 2025 &
  93.41 &
  92.66 &
  93.69 &
  86.98 &
  81.69 &
  94.67 &
  89.50 &
  90.94 &
  88.95 &
  96.33 &
  96.93 &
  91.18 &
  97.78 &
  91.90 \\
\end{tabular}
}
\label{tab:aud_auc_raf_au}
\end{table*}

\begin{table*}[htbp]
\centering
\setlength{\doublerulesep}{0.3em}
\renewcommand{\arraystretch}{1.5}
\caption{AU-wise F1 scores (\%) $\uparrow$ on the CK+ dataset.}
\resizebox{0.85\linewidth}{!}{
\begin{tabular}{l|c|ccccccccc|c}
\Xhline{1.3px}
\rowcolor{gray!15} 
\cellcolor{gray!15} & \cellcolor{gray!15} &
  \multicolumn{9}{c|}{\cellcolor{gray!15} \textbf{AU}} &
  \cellcolor{gray!15} \\ \hhline{~~---------}
\rowcolor{gray!15} 
\multirow{-2}{*}{\cellcolor{gray!15}\textbf{Methods}} &
    \multirow{-2}{*}{\cellcolor{gray!15}\textbf{Year}} &
  \textbf{01} &
  \textbf{02} &
  \cellcolor{gray!15}\textbf{04} &
  \cellcolor{gray!15}\textbf{05} &
  \cellcolor{gray!15}\textbf{06} &
  \textbf{07} &
  \textbf{12} &
  \textbf{17} &
  \textbf{25} &
  \multirow{-2}{*}{\cellcolor{gray!15}\textbf{Avg.}} \\ \hhline{============}
GraphAU(Res50) \cite{luo2022learning} &
  IJCAI 2022 &
  33.33 &
  0.00 &
  35.29 &
  30.77 &
  0.00 &
  0.00 &
  50.00 &
  16.00 &
  28.57 &
  21.55 \\
GraphAU(ViT-S) \cite{luo2022learning} &
  IJCAI 2022 &
  18.18 &
  0.00 &
  35.29 &
  18.18 &
  0.00 &
  0.00 &
  58.82 &
  34.48 &
  25.00 &
  21.11 \\
GraphAU(ViT-B) \cite{luo2022learning} &
  IJCAI 2022 &
  30.77 &
  0.00 &
  35.29 &
  22.22 &
  0.00 &
  0.00 &
  63.16 &
  22.22 &
  28.57 &
  22.47 \\
GraphAU(ViT-L) \cite{luo2022learning} &
  IJCAI 2022 &
  33.33 &
  0.00 &
  33.33 &
  50.00 &
  0.00 &
  0.00 &
  60.00 &
  38.71 &
  28.57 &
  27.10 \\ \hhline{============}
MAE-Face \cite{ma2024facial} &
  TAC 2024 &
  28.57 &
  0.00 &
  33.33 &
  44.44 &
  0.00 &
  28.57 &
  60.00 &
  48.49 &
  28.57 &
  30.22 \\ 
  FMAE(ViT-B) \cite{ning2024representation} &
  FG 2025 &
  30.77 &
  0.00 &
  35.29 &
  33.33 &
  0.00 &
  0.00 &
  70.00 &
  44.44 &
  28.57 &
  26.93 \\ 
  FMAE(ViT-L) \cite{ning2024representation} &
  FG 2025 &
  18.18 &
  0.00 &
  35.29 &
  33.33 &
  0.00 &
  28.57 &
  73.68 &
  40.00 &
  28.57 &
  28.62 \\ 
  \Xhline{1.3px}
\end{tabular}
}
\label{tab:aud_metric_ck}
\end{table*}

\begin{table*}[htbp]
\centering
\setlength{\doublerulesep}{0.3em}
\renewcommand{\arraystretch}{1.5}
\caption{AU-wise ROC-AUC scores (\%) $\uparrow$ on the CK+ dataset.}
\resizebox{0.85\linewidth}{!}{
\begin{tabular}{l|c|ccccccccc|c}
\Xhline{1.3px}
\rowcolor{gray!15} 
\cellcolor{gray!15} & \cellcolor{gray!15} &
  \multicolumn{9}{c|}{\cellcolor{gray!15} \textbf{AU}} &
  \cellcolor{gray!15} \\ \hhline{~~---------}
\rowcolor{gray!15} 
\multirow{-2}{*}{\cellcolor{gray!15}\textbf{Methods}} &
    \multirow{-2}{*}{\cellcolor{gray!15}\textbf{Year}} &
  \textbf{01} &
  \textbf{02} &
  \cellcolor{gray!15}\textbf{04} &
  \cellcolor{gray!15}\textbf{05} &
  \cellcolor{gray!15}\textbf{06} &
  \textbf{07} &
  \textbf{12} &
  \textbf{17} &
  \textbf{25} &
  \multirow{-2}{*}{\cellcolor{gray!15}\textbf{Avg.}} \\ \hhline{============}
GraphAU(Res50) \cite{luo2022learning} &
  IJCAI 2022 &
  79.11 &
  69.54 &
  84.62 &
  90.27 &
  34.62 &
  68.14 &
  79.26 &
  80.82 &
  78.18 &
  73.84 \\
GraphAU(ViT-S) \cite{luo2022learning} &
  IJCAI 2022 &
  73.26 &
  80.22 &
  81.84 &
  93.06 &
  44.76 &
  58.57 &
  89.93 &
  88.32 &
  69.54 &
  75.50 \\
GraphAU(ViT-B) \cite{luo2022learning} &
  IJCAI 2022 &
  73.93 &
  53.00 &
  69.08 &
  88.51 &
  82.87 &
  74.43 &
  74.96 &
  76.35 &
  65.11 &
  73.14 \\
GraphAU(ViT-L) \cite{luo2022learning} &
  IJCAI 2022 &
  75.85 &
  65.95 &
  79.33 &
  88.61 &
  31.82 &
  62.71 &
  80.67 &
  84.59 &
  56.83 &
  69.60 \\ \hhline{============}
MAE-Face \cite{ma2024facial} &
  TAC 2024 &
  83.93 &
  71.46 &
  90.19 &
  76.60 &
  56.29 &
  89.57 &
  81.33 &
  82.04 &
  69.66 &
  77.90 \\ 
  FMAE(ViT-B) \cite{ning2024representation} &
  FG 2025 &
  74.44 &
  63.91 &
  88.44 &
  89.13 &
  50.00 &
  81.43 &
  86.52 &
  84.22 &
  62.11 &
  75.58 \\ 
  FMAE(ViT-L) \cite{ning2024representation} &
  FG 2025 &
  83.63 &
  83.45 &
  84.57 &
  95.45 &
  41.96 &
  79.71 &
  84.74 &
  87.40 &
  76.50 &
  79.71 \\ 
  \Xhline{1.3px}
\end{tabular}
}
\label{tab:aud_auc_ck}
\end{table*}

\begin{table*}[htbp]
\centering
\setlength{\doublerulesep}{0.3em}
\renewcommand{\arraystretch}{1.5}
\caption{AU-wise F1 scores (\%) $\uparrow$ on the CASME2 dataset.}
\resizebox{0.8\linewidth}{!}{
\begin{tabular}{l|c|ccccccc|c}
\Xhline{1.3px}
\rowcolor{gray!15} 
\cellcolor{gray!15} & \cellcolor{gray!15} &
  \multicolumn{7}{c|}{\cellcolor{gray!15} \textbf{AU}} &
  \cellcolor{gray!15} \\ \hhline{~~-------}
\rowcolor{gray!15} 
\multirow{-2}{*}{\cellcolor{gray!15}\textbf{Methods}} &
    \multirow{-2}{*}{\cellcolor{gray!15}\textbf{Year}} &
  \textbf{01} &
  \textbf{02} &
  \cellcolor{gray!15}\textbf{04} &
  \cellcolor{gray!15}\textbf{07} &
  \textbf{12} &
  \textbf{14} &
  \textbf{17} &
  \multirow{-2}{*}{\cellcolor{gray!15}\textbf{Avg.}} \\ \hhline{==========}
GraphAU(Res50) \cite{luo2022learning} &
  IJCAI 2022 &
  0.00 &
  0.00 &
  75.97 &
  25.76 &
  43.77 &
  21.26 &
  43.64 &
  30.06 \\
GraphAU(ViT-S) \cite{luo2022learning} &
  IJCAI 2022 &
  0.00 &
  0.00 &
  79.07 &
  13.33 &
  40.23 &
  25.47 &
  42.72 &
  28.69 \\
GraphAU(ViT-B) \cite{luo2022learning} &
  IJCAI 2022 &
  0.00 &
  0.00 &
  83.80 &
  9.11 &
  46.77 &
  25.45 &
  53.00 &
  31.16 \\
GraphAU(ViT-L) \cite{luo2022learning} &
  IJCAI 2022 &
  0.00 &
  0.00 &
  83.47 &
  1.17 &
  47.04 &
  33.21 &
  66.67 &
  33.08 \\ \hhline{==========}
MDHR(Res50) \cite{wang2024multi} &
  CVPR 2024 &
  0.00 &
  0.00 &
  78.31 &
  0.62 &
  41.99 &
  32.18 &
  41.16 &
  27.75 \\ \hhline{==========}
MAE-Face \cite{ma2024facial} &
  TAC 2024 &
  0.00 &
  0.00 &
  81.54 &
  0.00 &
  45.06 &
  32.94 &
  64.88 &
  32.06 \\ 
  FMAE(ViT-B) \cite{ning2024representation} &
  FG 2025 &
  0.00 &
  0.00 &
  80.46 &
  6.19 &
  26.61 &
  20.70 &
  51.71 &
  26.52 \\ 
  FMAE(ViT-L) \cite{ning2024representation} &
  FG 2025 &
  0.00 &
  0.00 &
  79.04 &
  4.71 &
  41.40 &
  34.22 &
  69.32 &
  32.67 \\ 
  \Xhline{1.3px}
\end{tabular}
}
\label{tab:aud_metric_casme2}
\end{table*}

\begin{table*}[htbp]
\centering
\setlength{\doublerulesep}{0.3em}
\renewcommand{\arraystretch}{1.5}
\caption{AU-wise ROC-AUC scores (\%) $\uparrow$ on the CASME2 dataset.}
\resizebox{0.8\linewidth}{!}{
\begin{tabular}{l|c|ccccccc|c}
\Xhline{1.3px}
\rowcolor{gray!15} 
\cellcolor{gray!15} & \cellcolor{gray!15} &
  \multicolumn{7}{c|}{\cellcolor{gray!15} \textbf{AU}} &
  \cellcolor{gray!15} \\ \hhline{~~-------}
\rowcolor{gray!15} 
\multirow{-2}{*}{\cellcolor{gray!15}\textbf{Methods}} &
    \multirow{-2}{*}{\cellcolor{gray!15}\textbf{Year}} &
  \textbf{01} &
  \textbf{02} &
  \cellcolor{gray!15}\textbf{04} &
  \cellcolor{gray!15}\textbf{07} &
  \textbf{12} &
  \textbf{14} &
  \textbf{17} &
  \multirow{-2}{*}{\cellcolor{gray!15}\textbf{Avg.}} \\ \hhline{==========}
GraphAU(Res50) \cite{luo2022learning} &
  IJCAI 2022 &
  54.87 &
  26.87 &
  86.05 &
  46.77 &
  82.91 &
  61.90 &
  84.76 &
  63.45 \\
GraphAU(ViT-S) \cite{luo2022learning} &
  IJCAI 2022 &
  79.11 &
  87.49 &
  85.74 &
  54.15 &
  79.68 &
  64.49 &
  75.28 &
  75.13 \\
GraphAU(ViT-B) \cite{luo2022learning} &
  IJCAI 2022 &
  55.28 &
  95.94 &
  92.22 &
  52.41 &
  87.36 &
  68.99 &
  84.33 &
  76.65 \\
GraphAU(ViT-L) \cite{luo2022learning} &
  IJCAI 2022 &
  92.05 &
  91.95 &
  90.70 &
  55.21 &
  82.55 &
  64.38 &
  88.62 &
  80.78 \\ \hhline{==========}
MDHR(Res50) \cite{wang2024multi} &
  CVPR 2024 &
  70.19 &
  46.13 &
  85.88 &
  36.98 &
  88.88 &
  75.60 &
  84.08 &
  69.68 \\ \hhline{==========}
MAE-Face \cite{ma2024facial} &
  TAC 2024 &
  65.25 &
  78.19 &
  87.21 &
  41.04 &
  90.40 &
  74.08 &
  92.36 &
  75.50 \\ 
  FMAE(ViT-B) \cite{ning2024representation} &
  FG 2025 &
  80.71 &
  45.97 &
  85.33 &
  54.15 &
  82.43 &
  68.46 &
  89.83 &
  72.41 \\ 
  FMAE(ViT-L) \cite{ning2024representation} &
  FG 2025 &
  89.68 &
  74.89 &
  82.29 &
  35.40 &
  90.59 &
  81.89 &
  91.82 &
  78.08 \\ 
  \Xhline{1.3px}
\end{tabular}
}
\label{tab:aud_auc_casme2}
\end{table*}

\begin{table*}[htbp]
\centering
\setlength{\doublerulesep}{0.3em}
\renewcommand{\arraystretch}{1.5}
\caption{AU-wise F1 scores (\%) $\uparrow$ on the AFFW-AU dataset.}
\resizebox{1.0\linewidth}{!}{
\begin{tabular}{l|c|cccccccccccc|c}
\Xhline{1.5px}
\rowcolor{gray!15} 
\cellcolor{gray!15} & \cellcolor{gray!15} &
  \multicolumn{12}{c|}{\cellcolor{gray!15} \textbf{AU}} &
  \cellcolor{gray!15} \\ \hhline{~~------------}
\rowcolor{gray!15} 
\multirow{-2}{*}{\cellcolor{gray!15}\textbf{Methods}} &
    \multirow{-2}{*}{\cellcolor{gray!15}\textbf{Year}} &
  \textbf{01} &
  \textbf{02} &
  \cellcolor{gray!15}\textbf{04} &
  \cellcolor{gray!15}\textbf{06} &
  \cellcolor{gray!15}\textbf{07} &
  \textbf{10} &
  \textbf{12} &
  \textbf{15} &
  \textbf{23} &
  \textbf{24} &
  \textbf{25} &
  \textbf{26} &
  \multirow{-2}{*}{\cellcolor{gray!15}\textbf{Avg.}} \\ \hhline{===============}
GraphAU(Res50) \cite{luo2022learning} &
  IJCAI 2022 &
  54.97 &
  49.85 &
  54.86 &
  62.26 &
  73.32 &
  73.58 &
  73.73 &
  29.93 &
  21.45 &
  18.47 &
  82.84 &
  29.34 &
  52.05 \\
GraphAU(ViT-S) \cite{luo2022learning} &
  IJCAI 2022 &
  55.79 &
  50.77 &
  55.58 &
  62.48 &
  70.62 &
  74.38 &
  72.18 &
  17.98 &
  9.12 &
  11.89 &
  82.36 &
  39.64 &
  50.23 \\
GraphAU(ViT-B) \cite{luo2022learning} &
  IJCAI 2022 &
  57.40 &
  49.56 &
  59.33 &
  64.98 &
  74.47 &
  76.41 &
  73.64 &
  33.40 &
  21.46 &
  16.47 &
  84.73 &
  33.15 &
  53.77 \\
GraphAU(ViT-L) \cite{luo2022learning} &
  IJCAI 2022 &
  56.97 &
  50.15 &
  57.19 &
  62.63 &
  72.62 &
  74.88 &
  73.42 &
  24.41 &
  19.27 &
  11.62 &
  84.05 &
  39.05 &
  52.19 \\ \hhline{===============}
MDHR(Res50) \cite{wang2024multi} &
  CVPR 2024 &
  59.39 &
  53.22 &
  56.01 &
  62.22 &
  74.53 &
  74.54 &
  74.70 &
  9.41 &
  11.28 &
  9.91 &
  84.47 &
  38.48 &
  50.68 \\ \hhline{===============}
MAE-Face \cite{ma2024facial} &
  TAC 2024 &
  53.64 &
  46.43 &
  57.17 &
  64.43 &
  74.61 &
  75.48 &
  73.65 &
  22.70 &
  20.33 &
  18.29 &
  83.80 &
  40.92 &
  52.62 \\
\end{tabular}
}
\label{tab:aud_metric_affw}
\end{table*}

\begin{table*}[htbp]
\centering
\setlength{\doublerulesep}{0.3em}
\renewcommand{\arraystretch}{1.5}
\caption{AU-wise ROC-AUC scores (\%) $\uparrow$ on the AFFW-AU dataset.}
\resizebox{1.0\linewidth}{!}{
\begin{tabular}{l|c|cccccccccccc|c}
\Xhline{1.5px}
\rowcolor{gray!15} 
\cellcolor{gray!15} & \cellcolor{gray!15} &
  \multicolumn{12}{c|}{\cellcolor{gray!15} \textbf{AU}} &
  \cellcolor{gray!15} \\ \hhline{~~------------}
\rowcolor{gray!15} 
\multirow{-2}{*}{\cellcolor{gray!15}\textbf{Methods}} &
    \multirow{-2}{*}{\cellcolor{gray!15}\textbf{Year}} &
  \textbf{01} &
  \textbf{02} &
  \cellcolor{gray!15}\textbf{04} &
  \cellcolor{gray!15}\textbf{06} &
  \cellcolor{gray!15}\textbf{07} &
  \textbf{10} &
  \textbf{12} &
  \textbf{15} &
  \textbf{23} &
  \textbf{24} &
  \textbf{25} &
  \textbf{26} &
  \multirow{-2}{*}{\cellcolor{gray!15}\textbf{Avg.}} \\ \hhline{===============}
GraphAU(Res50) \cite{luo2022learning} &
  IJCAI 2022 &
  83.79 &
  77.76 &
  83.49 &
  82.59 &
  87.76 &
  90.17 &
  90.01 &
  87.86 &
  81.44 &
  74.94 &
  86.76 &
  75.66 &
  83.52 \\
GraphAU(ViT-S) \cite{luo2022learning} &
  IJCAI 2022 &
  81.98 &
  78.10 &
  83.93 &
  82.27 &
  86.47 &
  90.21 &
  88.72 &
  86.62 &
  78.35 &
  75.79 &
  85.70 &
  76.25 &
  82.87 \\
GraphAU(ViT-B) \cite{luo2022learning} &
  IJCAI 2022 &
  83.80 &
  78.27 &
  86.08 &
  83.63 &
  87.71 &
  90.82 &
  90.10 &
  86.42 &
  79.89 &
  76.45 &
  87.47 &
  76.41 &
  83.92 \\
GraphAU(ViT-L) \cite{luo2022learning} &
  IJCAI 2022 &
  83.15 &
  79.04 &
  85.10 &
  82.88 &
  87.47 &
  90.44 &
  90.48 &
  88.41 &
  80.80 &
  76.87 &
  86.68 &
  76.89 &
  84.02 \\ \hhline{===============}
MDHR(Res50) \cite{wang2024multi} &
  CVPR 2024 &
  86.73 &
  82.56 &
  83.65 &
  82.72 &
  88.51 &
  90.90 &
  90.85 &
  88.05 &
  82.80 &
  75.67 &
  88.70 &
  80.73 &
  85.16 \\ \hhline{===============}
MAE-Face \cite{ma2024facial} &
  TAC 2024 &
  82.66 &
  75.73 &
  83.93 &
  83.27 &
  88.27 &
  90.57 &
  90.32 &
  88.33 &
  82.42 &
  78.69 &
  87.86 &
  76.24 &
  84.02 \\
\end{tabular}
}
\label{tab:aud_auc_affw}
\end{table*}

\begin{figure*}[htbp]
    \centering
    \includegraphics[width=1.0\linewidth]{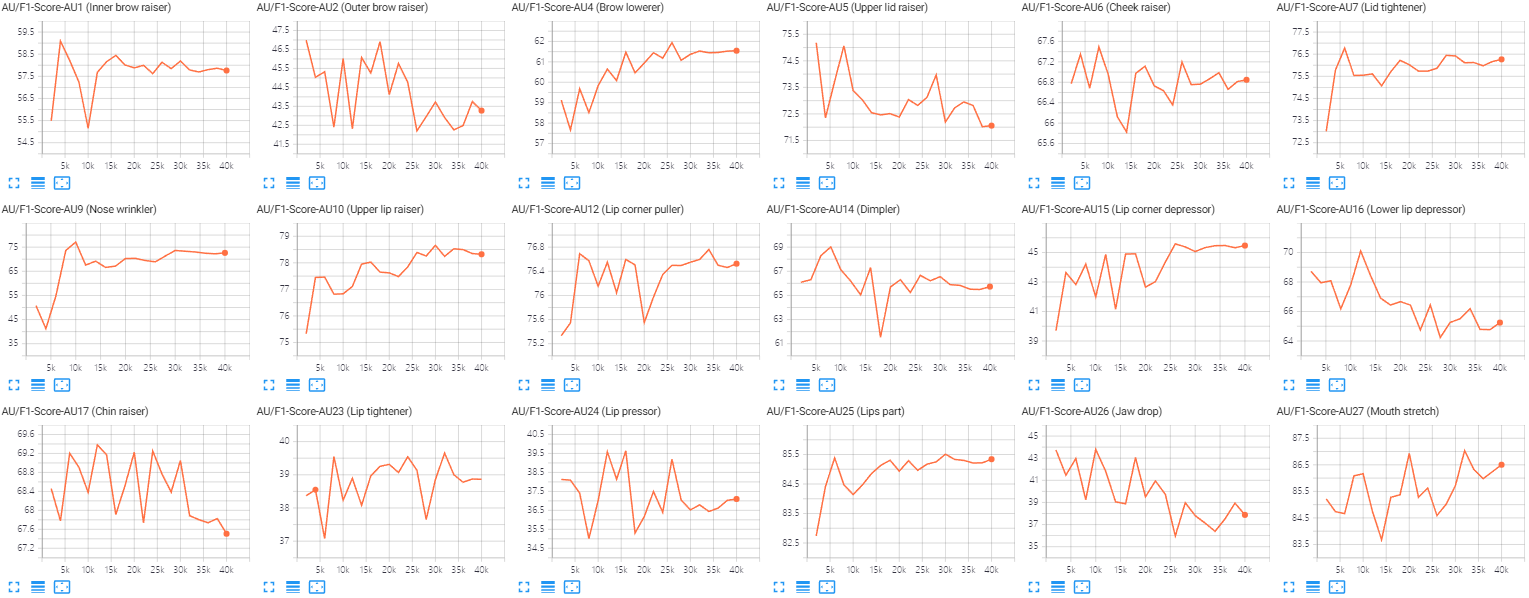}
    \caption{AU-wise validation F1-score trajectories of MAE-Face-B on function-level FAUD validation set during training. Each subplot corresponds to one of the 18 evaluated AUs. The horizontal axis denotes the training iteration, and the vertical axis denotes the F1 score on the function-level validation set.}
    \label{fig:faud_maeface_curve}
\end{figure*}

\begin{figure*}[htbp]
    \centering
    \includegraphics[width=1.0\linewidth]{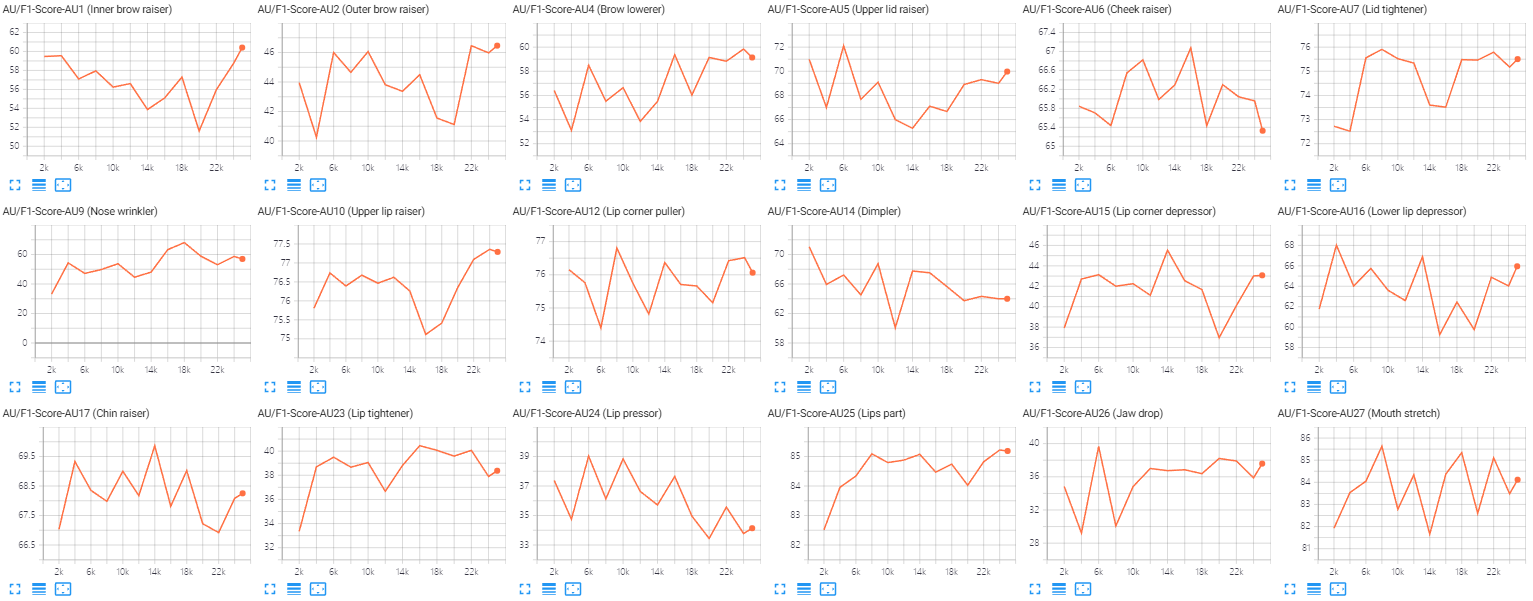}
    \caption{AU-wise validation F1-score trajectories of GraphAU (ViT-L) on function-level FAUD validation set during training. Each subplot corresponds to one of the 18 evaluated AUs. The horizontal axis denotes the training iteration, and the vertical axis denotes the F1 score on the function-level validation set.}
    \label{fig:faud_graphau_curve}
\end{figure*}

\noindent \textbf{Overall and dataset-specific performance.}
We evaluate eight FAUD models re-developed from four open-source methods, including GraphAU \cite{luo2022learning}, MDHR \cite{wang2024multi}, MAE-Face \cite{ma2024facial}, and FMAE \cite{ning2024representation}, under the unified multi-dataset protocol described in the main paper. Table \ref{tab:aud_metric} and \ref{tab:aud_auc} report the AU-wise F1 scores and ROC-AUC scores on the function-level test set formed from BP4D, DISFA, RAF-AU, CK+, CASME2, and AFFW-AU, while Tables \ref{tab:aud_metric_bp4d}-\ref{tab:aud_auc_affw} provide the corresponding dataset-specific results.

The results in Table \ref{tab:aud_metric} and \ref{tab:aud_auc} further support the parameter-performance relationship discussed in Sec. 4.3 of the main paper. MAE-Face-B achieves the highest average F1 score of 63.42\%, representing a relative performance gain of 1.02\% over the second-best GraphAU (ViT-B). Notably, MAE-Face-B achieves a relative performance gain of 1.91\% over the substantially larger GraphAU (ViT-L), with only 86.10M parameters. Increasing model parameters within the same method similarly provides no consistent benefit: GraphAU (ViT-B) achieves a relative performance gain of 0.88\% over GraphAU (ViT-L), while FMAE-B achieves a relative performance gain of 0.11\% over FMAE-L despite containing approximately 3.5 times fewer parameters. The AU-wise results further show that no single configuration dominates all AUs. MAE-Face-B performs best on AU05, AU09, AU12, AU16, and AU26; GraphAU (ViT-B) leads on AU01, AU04, AU10, and AU25; and FMAE-B performs best on AU06, AU07, AU15, and AU24. The remaining AUs favour GraphAU (ViT-S), GraphAU (ViT-L), or FMAE-L, demonstrating that the preferred model can also depend on the AUs required by a particular application.

The dataset-specific results in Tables \ref{tab:aud_metric_bp4d}-\ref{tab:aud_auc_affw} present a similar pattern. FMAE-B obtains the highest average F1 score on BP4D, MAE-Face-B leads on DISFA, RAF-AU, and CK+, GraphAU (ViT-L) performs best on CASME2, and GraphAU (ViT-B) leads on AFFW-AU. Despite these dataset-dependent changes, MAE-Face-B remains within a relative performance difference of 3.08\% from the best result on every dataset and ranks among the top three configurations throughout all six datasets, indicating the most stable cross-dataset performance among the evaluated models.

\begin{table*}[htbp]
\large
\centering
\setlength{\doublerulesep}{0.3em}
\renewcommand{\arraystretch}{1.3}
\caption{AU-wise FAUE results on the function-level test set formed from BP4D and DISFA. We report the intraclass correlation coefficient (ICC(3,1)) $\uparrow$, Pearson correlation coefficient (PCC) $\uparrow$, and mean squared error (MSE) $\downarrow$. The best and second-best results are highlighted in bold and underlined, respectively.}
\resizebox{1.0\linewidth}{!}{
\begin{tabular}{cl|c|cccccccccccccc|c}
\Xhline{2.5px}
\rowcolor{gray!15} 
\cellcolor{gray!15} &
  \cellcolor{gray!15} & \cellcolor{gray!15} &
  \multicolumn{14}{c|}{\cellcolor{gray!15} \textbf{AU}} &
  \cellcolor{gray!15} \\ \hhline{~~~--------------}
\rowcolor{gray!15} 
\multirow{-2}{*}{\cellcolor{gray!15}} & 
  \multirow{-2}{*}{\cellcolor{gray!15} \textbf{Methods}} &
      \multirow{-2}{*}{\cellcolor{gray!15}\textbf{Year}} &
  \textbf{01} &
  \textbf{02} &
  \cellcolor{gray!15}\textbf{04} &
  \cellcolor{gray!15}\textbf{05} &
  \cellcolor{gray!15}\textbf{06} &
  \textbf{09} &
  \textbf{10} &
  \textbf{12} &
  \textbf{14} &
  \textbf{15} &
  \textbf{17} &
  \textbf{20} &
  \textbf{25} &
  \textbf{26} &
  \multirow{-2}{*}{\cellcolor{gray!15}\textbf{Avg.}} \\ \hhline{==================}
 &
  GraphAU(Res50) \cite{luo2022learning} &
  IJCAI 2022 &
  0.805 &
  \cellcolor{gray!15}\underline{0.757} &
  0.728 &
  0.689 &
  \cellcolor{gray!15}\underline{0.777} &
  0.594 &
  0.723 &
  0.819 &
  0.406 &
  \cellcolor{gray!30}\textbf{0.736} &
  0.608 &
  0.116 &
  0.900 &
  0.702 &
  0.669 \\
 &
  GraphAU(ViT-B) \cite{luo2022learning} &
  IJCAI 2022 &
  \cellcolor{gray!15}\underline{0.833} &
  0.736 &
  \cellcolor{gray!30}\textbf{0.809} &
  0.753 &
  0.735 &
  0.768 &
  0.748 &
  0.834 &
  0.404 &
  0.552 &
  0.650 &
  \cellcolor{gray!15}\underline{0.260} &
  \cellcolor{gray!30}\textbf{0.955} &
  \cellcolor{gray!15}\underline{0.744} &
  \cellcolor{gray!15}\underline{0.699} \\
\multirow{-2}{*}{\rotatebox[origin=c]{90}{ICC(3, 1)↑}} &
  MDHR(Res50) \cite{wang2024multi} &
  CVPR 2024 &
  0.811 &
  0.665 &
  0.625 &
  0.665 &
  0.745 &
  0.699 &
  0.725 &
  0.730 &
  0.410 &
  \cellcolor{gray!15}\underline{0.639} &
  \cellcolor{gray!15}\underline{0.677} &
  0.247 &
  0.934 &
  0.665 &
  0.660 \\ &
MAE-Face \cite{ma2024facial} &
  TAC 2024 &
  \cellcolor{gray!30}\textbf{0.893} &
  \cellcolor{gray!30}\textbf{0.795} &
  \cellcolor{gray!15}\underline{0.804} &
  \cellcolor{gray!15}\underline{0.758} &
  \cellcolor{gray!30}\textbf{0.785} &
  \cellcolor{gray!15}\underline{0.783} &
  \cellcolor{gray!30}\textbf{0.770} &
  \cellcolor{gray!30}\textbf{0.848} &
  \cellcolor{gray!30}\textbf{0.475} &
  0.520 &
  \cellcolor{gray!30}\textbf{0.727} &
  \cellcolor{gray!30}\textbf{0.328} &
  \cellcolor{gray!30}\textbf{0.955} &
  \cellcolor{gray!30}\textbf{0.752} &
  \cellcolor{gray!30}\textbf{0.728} \\ &
  FMAE(ViT-B) \cite{sun2023mae} &
  FG 2025 &
  0.815 &
  0.701 &
  0.674 &
  \cellcolor{gray!30}\textbf{0.773} &
  0.770 &
  \cellcolor{gray!30}\textbf{0.801} &
  \cellcolor{gray!15}\underline{0.756} &
  \cellcolor{gray!15}\underline{0.842} &
  \cellcolor{gray!15}\underline{0.421} &
  0.568 &
  0.630 &
  0.249 &
  0.953 &
  0.668 &
  0.687 \\
  \hhline{==================}
 &
  GraphAU(Res50) \cite{luo2022learning} &
  IJCAI 2022 &
  0.874 &
  \cellcolor{gray!30}\textbf{0.817} &
  0.750 &
  0.791 &
  \cellcolor{gray!30}\textbf{0.787} &
  0.595 &
  0.738 &
  0.826 &
  0.443 &
  \cellcolor{gray!30}\textbf{0.787} &
  0.620 &
  0.117 &
  0.908 &
  0.735 &
  0.699 \\
 &
  GraphAU(ViT-B) \cite{luo2022learning} &
  IJCAI 2022 &
  \cellcolor{gray!15}\underline{0.875} &
  0.754 &
  \cellcolor{gray!30}\textbf{0.821} &
  \cellcolor{gray!15}\underline{0.816} &
  0.776 &
  \cellcolor{gray!15}\underline{0.789} &
  \cellcolor{gray!30}\textbf{0.782} &
  0.840 &
  0.493 &
  0.557 &
  0.656 &
  0.277 &
  0.956 &
  \cellcolor{gray!30}\textbf{0.765} &
  \cellcolor{gray!15}\underline{0.726} \\
\multirow{-1.5}{*}{\rotatebox[origin=c]{90}{PCC↑}} &
  MDHR(Res50) \cite{wang2024multi} &
  CVPR 2024 &
  0.863 &
  0.670 &
  0.658 &
  0.682 &
  0.752 &
  0.701 &
  0.741 &
  0.834 &
  0.425 &
  \cellcolor{gray!15}\underline{0.713} &
  \cellcolor{gray!15}\underline{0.707} &
  \cellcolor{gray!30}\textbf{0.345} &
  0.936 &
  0.752 &
  0.698 \\ &
  MAE-Face \cite{ma2024facial} &
  TAC 2024 &
  \cellcolor{gray!30}\textbf{0.902} &
  \cellcolor{gray!15}\underline{0.796} &
  \cellcolor{gray!15}\underline{0.815} &
  \cellcolor{gray!30}\textbf{0.820} &
  \cellcolor{gray!30}\textbf{0.787} &
  0.783 &
  \cellcolor{gray!15}\underline{0.773} &
  \cellcolor{gray!30}\textbf{0.849} &
  \cellcolor{gray!30}\textbf{0.523} &
  0.574 &
  \cellcolor{gray!30}\textbf{0.727} &
  \cellcolor{gray!15}\underline{0.339} &
  \cellcolor{gray!30}\textbf{0.957} &
  \cellcolor{gray!15}\underline{0.761} &
  \cellcolor{gray!30}\textbf{0.743} \\ &
  FMAE(ViT-B) \cite{sun2023mae} &
  FG 2025 &
  0.841 &
  0.722 &
  0.700 &
  0.778 &
  0.785 &
  \cellcolor{gray!30}\textbf{0.802} &
  \cellcolor{gray!15}\underline{0.773} &
  \cellcolor{gray!15}\underline{0.843} &
  \cellcolor{gray!15}\underline{0.494} &
  0.587 &
  0.632 &
  0.256 &
  \cellcolor{gray!30}\textbf{0.957} &
  0.672 &
  0.703 \\
  \hhline{==================}
 &
  GraphAU(Res50) \cite{luo2022learning} &
  IJCAI 2022 &
  0.148 &
  \cellcolor{gray!15}\underline{0.127} &
  0.539 &
  0.050 &
  \cellcolor{gray!30}\textbf{0.576} &
  0.250 &
  1.148 &
  0.541 &
  1.673 &
  \cellcolor{gray!30}\textbf{0.053} &
  0.568 &
  0.155 &
  0.255 &
  0.291 &
  0.455 \\
 &
  GraphAU(ViT-B) \cite{luo2022learning} &
  IJCAI 2022 &
  0.137 &
  0.184 &
  \cellcolor{gray!15}\underline{0.377} &
  \cellcolor{gray!15}\underline{0.043} &
  0.605 &
  \cellcolor{gray!30}\textbf{0.099} &
  \cellcolor{gray!30}\textbf{0.886} &
  \cellcolor{gray!30}\textbf{0.503} &
  \cellcolor{gray!15}\underline{1.564} &
  0.155 &
  0.506 &
  \cellcolor{gray!15}\underline{0.076} &
  \cellcolor{gray!15}\underline{0.127} &
  \cellcolor{gray!30}\textbf{0.265} &
  \cellcolor{gray!30}\textbf{0.395} \\
\multirow{-1.5}{*}{\rotatebox[origin=c]{90}{MSE↓}} &
  MDHR(Res50) \cite{wang2024multi} &
  CVPR 2024 &
  \cellcolor{gray!15}\underline{0.102} &
  \cellcolor{gray!30}\textbf{0.089} &
  \cellcolor{gray!30}\textbf{0.345} &
  \cellcolor{gray!30}\textbf{0.042} &
  0.697 &
  0.200 &
  1.049 &
  0.631 &
  1.810 &
  \cellcolor{gray!15}\underline{0.062} &
  \cellcolor{gray!30}\textbf{0.378} &
  \cellcolor{gray!30}\textbf{0.056} &
  0.195 &
  0.315 &
  0.426 \\ &
  MAE-Face \cite{ma2024facial} &
  TAC 2024 &
  \cellcolor{gray!30}\textbf{0.101} &
  0.140 &
  \cellcolor{gray!15}\underline{0.377} &
  0.099 &
  0.596 &
  0.116 &
  0.951 &
  0.539 &
  \cellcolor{gray!30}\textbf{1.507} &
  0.228 &
  \cellcolor{gray!15}\underline{0.421} &
  0.119 &
  \cellcolor{gray!30}\textbf{0.122} &
  \cellcolor{gray!15}\underline{0.272} &
  \cellcolor{gray!15}\underline{0.399} \\ &
  FMAE(ViT-B) \cite{sun2023mae} &
  FG 2025 &
  0.159 &
  0.186 &
  0.576 &
  0.049 &
  \cellcolor{gray!30}\textbf{0.576} &
  \cellcolor{gray!15}\underline{0.102} &
  \cellcolor{gray!15}\underline{0.906} &
  \cellcolor{gray!15}\underline{0.517} &
  1.711 &
  0.170 &
  0.565 &
  0.084 &
  0.128 &
  0.381 &
  0.437 \\ 
  \Xhline{2.5px}
\end{tabular}
}
\label{tab:auie_metric}
\end{table*}

\begin{table*}[htbp]
\centering
\setlength{\doublerulesep}{0.3em}
\renewcommand{\arraystretch}{1.3}
\caption{AU-wise FAUE results on the BP4D dataset. We report ICC(3,1) $\uparrow$, PCC $\uparrow$, and MSE $\downarrow$.}
\resizebox{0.8\linewidth}{!}{
\begin{tabular}{cl|c|ccccc|c}
\Xhline{1.3px}
\rowcolor{gray!15} 
\cellcolor{gray!15} &
  \cellcolor{gray!15} & \cellcolor{gray!15} &
  \multicolumn{5}{c|}{\cellcolor{gray!15} \textbf{AU}} &
  \cellcolor{gray!15} \\ \hhline{~~~-----}
\rowcolor{gray!15} 
\multirow{-2}{*}{\cellcolor{gray!15}} & 
  \multirow{-2}{*}{\cellcolor{gray!15} \textbf{Methods}} &
      \multirow{-2}{*}{\cellcolor{gray!15}\textbf{Year}} &
  \cellcolor{gray!15}\textbf{06} &
  \textbf{10} &
  \textbf{12} &
  \textbf{14} &
  \textbf{17} &
  \multirow{-2}{*}{\cellcolor{gray!15}\textbf{Avg.}} \\ \hhline{=========}
 &
  GraphAU(Res50) \cite{luo2022learning} &
  IJCAI 2022 &
  0.779 &
  0.723 &
  0.799 &
  0.406 &
  0.619 &
  0.665 \\
 &
  GraphAU(ViT-B) \cite{luo2022learning} &
  IJCAI 2022 &
  0.719 &
  0.748 &
  0.801 &
  0.404 &
  0.702 &
  0.675 \\
\multirow{-2}{*}{\rotatebox[origin=c]{90}{ICC(3, 1)↑}} &
  MDHR(Res50) \cite{wang2024multi} &
  CVPR 2024 &
  0.757 &
  0.725 &
  0.723 &
  0.410 &
  0.637 &
  0.650 \\ &
MAE-Face \cite{ma2024facial} &
  TAC 2024 &
  0.751 &
  0.770 &
  0.821 &
  0.475 &
  0.720 &
  0.707 \\ &
  FMAE(ViT-B) \cite{sun2023mae} &
  FG 2025 &
  0.745 &
  0.756 &
  0.818 &
  0.421 &
  0.699 &
  0.688 \\
  \hhline{=========}
 &
  GraphAU(Res50) \cite{luo2022learning} &
  IJCAI 2022 &
  0.788 &
  0.738 &
  0.805 &
  0.443 &
  0.646 &
  0.684 \\
 &
  GraphAU(ViT-B) \cite{luo2022learning} &
  IJCAI 2022 &
  0.756 &
  0.782 &
  0.807 &
  0.493 &
  0.714 &
  0.711 \\
\multirow{-1.5}{*}{\rotatebox[origin=c]{90}{PCC↑}} &
  MDHR(Res50) \cite{wang2024multi} &
  CVPR 2024 &
  0.761 &
  0.741 &
  0.821 &
  0.425 &
  0.673 &
  0.684 \\ &
  MAE-Face \cite{ma2024facial} &
  TAC 2024 &
  0.753 &
  0.773 &
  0.822 &
  0.523 &
  0.721 &
  0.718 \\ &
  FMAE(ViT-B) \cite{sun2023mae} &
  FG 2025 &
  0.758 &
  0.773 &
  0.819 &
  0.494 &
  0.705 &
  0.710 \\
  \hhline{=========}
 &
  GraphAU(Res50) \cite{luo2022learning} &
  IJCAI 2022 &
  0.776 &
  1.148 &
  0.773 &
  1.673 &
  0.673 &
  1.009 \\
 &
  GraphAU(ViT-B) \cite{luo2022learning} &
  IJCAI 2022 &
  0.925 &
  0.886 &
  0.776 &
  1.564 &
  0.557 &
  0.941 \\
\multirow{-1.5}{*}{\rotatebox[origin=c]{90}{MSE↓}} &
  MDHR(Res50) \cite{wang2024multi} &
  CVPR 2024 &
  0.895 &
  1.049 &
  0.866 &
  1.810 &
  0.604 &
  1.045 \\ &
  MAE-Face \cite{ma2024facial} &
  TAC 2024 &
  0.940 &
  0.951 &
  0.823 &
  1.507 &
  0.592 &
  0.963 \\ &
  FMAE(ViT-B) \cite{sun2023mae} &
  FG 2025 &
  0.876 &
  0.906 &
  0.748 &
  1.711 &
  0.588 &
  0.966 \\ 
  \Xhline{1.3px}
\end{tabular}
}
\label{tab:auie_metric_bp4d}
\end{table*}

\begin{table*}[htbp]
\large
\centering
\setlength{\doublerulesep}{0.3em}
\renewcommand{\arraystretch}{1.3}
\caption{AU-wise FAUE results on the DISFA dataset. We report ICC(3,1) $\uparrow$, PCC $\uparrow$, and MSE $\downarrow$.}
\resizebox{1.0\linewidth}{!}{
\begin{tabular}{cl|c|cccccccccccc|c}
\Xhline{2.5px}
\rowcolor{gray!15} 
\cellcolor{gray!15} &
  \cellcolor{gray!15} & \cellcolor{gray!15} &
  \multicolumn{12}{c|}{\cellcolor{gray!15} \textbf{AU}} &
  \cellcolor{gray!15} \\ \hhline{~~~------------}
\rowcolor{gray!15} 
\multirow{-2}{*}{\cellcolor{gray!15}} & 
  \multirow{-2}{*}{\cellcolor{gray!15} \textbf{Methods}} &
      \multirow{-2}{*}{\cellcolor{gray!15}\textbf{Year}} &
  \textbf{01} &
  \textbf{02} &
  \cellcolor{gray!15}\textbf{04} &
  \cellcolor{gray!15}\textbf{05} &
  \cellcolor{gray!15}\textbf{06} &
  \textbf{09} &
  \textbf{12} &
  \textbf{15} &
  \textbf{17} &
  \textbf{20} &
  \textbf{25} &
  \textbf{26} &
  \multirow{-2}{*}{\cellcolor{gray!15}\textbf{Avg.}} \\ \hhline{================}
 &
  GraphAU(Res50) \cite{luo2022learning} &
  IJCAI 2022 &
  0.805 &
  0.757 &
  0.728 &
  0.689 &
  0.562 &
  0.594 &
  0.771 &
  0.736 &
  0.482 &
  0.116 &
  0.900 &
  0.702 &
  0.653 \\
 &
  GraphAU(ViT-B) \cite{luo2022learning} &
  IJCAI 2022 &
  0.833 &
  0.736 &
  0.809 &
  0.753 &
  0.629 &
  0.768 &
  0.853 &
  0.552 &
  0.417 &
  0.260 &
  0.955 &
  0.744 &
  0.693 \\
\multirow{-2}{*}{\rotatebox[origin=c]{90}{ICC(3, 1)↑}} &
  MDHR(Res50) \cite{wang2024multi} &
  CVPR 2024 &
  0.811 &
  0.665 &
  0.625 &
  0.665 &
  0.532 &
  0.699 &
  0.734 &
  0.639 &
  0.526 &
  0.247 &
  0.934 &
  0.665 &
  0.645 \\ &
MAE-Face \cite{ma2024facial} &
  TAC 2024 &
  0.892 &
  0.795 &
  0.804 &
  0.758 &
  0.692 &
  0.783 &
  0.853 &
  0.520 &
  0.612 &
  0.328 &
  0.955 &
  0.752 &
  0.729 \\ &
  FMAE(ViT-B) \cite{sun2023mae} &
  FG 2025 &
  0.815 &
  0.701 &
  0.674 &
  0.773 &
  0.606 &
  0.801 &
  0.816 &
  0.568 &
  0.382 &
  0.249 &
  0.953 &
  0.668 &
  0.667 \\
  \hhline{================}
 &
  GraphAU(Res50) \cite{luo2022learning} &
  IJCAI 2022 &
  0.874 &
  0.817 &
  0.750 &
  0.791 &
  0.563 &
  0.595 &
  0.790 &
  0.787 &
  0.495 &
  0.117 &
  0.908 &
  0.735 &
  0.685 \\
 &
  GraphAU(ViT-B) \cite{luo2022learning} &
  IJCAI 2022 &
  0.875 &
  0.754 &
  0.821 &
  0.816 &
  0.655 &
  0.789 &
  0.860 &
  0.557 &
  0.429 &
  0.277 &
  0.956 &
  0.765 &
  0.713 \\
\multirow{-1.5}{*}{\rotatebox[origin=c]{90}{PCC↑}} &
  MDHR(Res50) \cite{wang2024multi} &
  CVPR 2024 &
  0.863 &
  0.670 &
  0.658 &
  0.682 &
  0.533 &
  0.701 &
  0.820 &
  0.713 &
  0.526 &
  0.345 &
  0.936 &
  0.752 &
  0.683 \\ &
  MAE-Face \cite{ma2024facial} &
  TAC 2024 &
  0.902 &
  0.796 &
  0.815 &
  0.820 &
  0.708 &
  0.783 &
  0.858 &
  0.574 &
  0.625 &
  0.339 &
  0.957 &
  0.761 &
  0.745 \\ &
  FMAE(ViT-B) \cite{sun2023mae} &
  FG 2025 &
  0.841 &
  0.722 &
  0.700 &
  0.778 &
  0.643 &
  0.802 &
  0.833 &
  0.587 &
  0.403 &
  0.256 &
  0.957 &
  0.672 &
  0.683 \\
  \hhline{================}
 &
  GraphAU(Res50) \cite{luo2022learning} &
  IJCAI 2022 &
  0.148 &
  0.127 &
  0.539 &
  0.050 &
  0.345 &
  0.250 &
  0.273 &
  0.053 &
  0.446 &
  0.155 &
  0.255 &
  0.291 &
  0.244 \\
 &
  GraphAU(ViT-B) \cite{luo2022learning} &
  IJCAI 2022 &
  0.137 &
  0.184 &
  0.377 &
  0.043 &
  0.238 &
  0.099 &
  0.190 &
  0.155 &
  0.447 &
  0.076 &
  0.127 &
  0.265 &
  0.195 \\
\multirow{-1.5}{*}{\rotatebox[origin=c]{90}{MSE↓}} &
  MDHR(Res50) \cite{wang2024multi} &
  CVPR 2024 &
  0.102 &
  0.089 &
  0.345 &
  0.042 &
  0.442 &
  0.200 &
  0.328 &
  0.062 &
  0.085 &
  0.056 &
  0.195 &
  0.315 &
  0.188 \\ &
  MAE-Face \cite{ma2024facial} &
  TAC 2024 &
  0.101 &
  0.140 &
  0.377 &
  0.099 &
  0.200 &
  0.116 &
  0.213 &
  0.228 &
  0.224 &
  0.119 &
  0.122 &
  0.272 &
  0.184 \\ &
  FMAE(ViT-B) \cite{sun2023mae} &
  FG 2025 &
  0.159 &
  0.186 &
  0.576 &
  0.049 &
  0.232 &
  0.102 &
  0.251 &
  0.170 &
  0.540 &
  0.084 &
  0.128 &
  0.382 &
  0.238 \\ 
  \Xhline{2.5px}
\end{tabular}
}
\label{tab:auie_metric_disfa}
\end{table*}

\textbf{Application-oriented model selection.} 
The computational characteristics reported in Table \ref{tab:model_efficiency} provide complementary guidance for selecting a FAUD model. MAE-Face-B offers the most favourable overall balance, combining the highest average F1 score with 86.10M parameters, 16.87G FLOPs, and an inference speed of 131 FPS on L20 GPU. FMAE-B has the same computational profile and is a suitable alternative when performance on BP4D or on AU06, AU07, AU15, and AU24 is prioritised. For model-size-constrained applications, GraphAU (ViT-S) provides the smallest configuration, requiring 22.97M parameters and 15.86G FLOPs, although its inference speed of 58 FPS shows that parameter count alone does not determine runtime efficiency. When temporal inputs are available and maximum measured throughput is prioritised, MDHR (Res50) achieves 940 FPS under the adopted evaluation protocol, but its 126.66M parameters and 127.31G FLOPs require substantially greater computational resources. By comparison, GraphAU (ViT-L) and FMAE-L contain more than 300M parameters but provide no improvement in average F1 score over their medium-sized counterparts. These high-capacity configurations are therefore more appropriate when their gains on particular datasets or AUs justify the additional computational cost, rather than as general-purpose FAUD models.

\textbf{AU-wise convergence analysis.} 
The results above show that increasing model parameters does not consistently improve the final FAUD performance. Figs. \ref{fig:faud_maeface_curve} and \ref{fig:faud_graphau_curve} further examine this behaviour by presenting the AU-wise validation F1 trajectories of MAE-Face-B and GraphAU (ViT-L) during training. The curves reveal that different AUs reach their highest validation performance at different stages. For MAE-Face-B, several AUs, including AU01, AU04, AU07, AU10, AU12, AU15, and AU25, become comparatively stable after their initial optimisation stages. In contrast, the substantially larger GraphAU (ViT-L) exhibits more pronounced peak-to-final performance reversals on several AUs. For example, the validation F1 scores for AU05, AU09, AU14, and AU24 reach relatively high values during intermediate training stages but subsequently decline, even though other AUs, such as AU01, AU02, AU04, AU10, AU23, and AU25, continue to improve or recover.

This asynchronous convergence means that continued optimisation can improve some AUs while simultaneously degrading others, preventing a single checkpoint from preserving the best performance for every AU. The effect is more pronounced for GraphAU (ViT-L) on several AUs, providing training-process evidence consistent with the AU-dependent overfitting discussed in the main paper. Together with the non-monotonic parameter-performance relationship in Table \ref{tab:aud_metric}, these observations explain why increasing model parameters does not necessarily improve the overall FAUD performance under the heterogeneous datasets used in OpenAHBU.

\subsubsection{FAUE performances}

\noindent \textbf{Overall and dataset-specific performance.}
Table \ref{tab:auie_metric} reports the AU-wise FAUE results of the five model configurations on the function-level test set formed from BP4D and DISFA, while Tables \ref{tab:auie_metric_bp4d} and \ref{tab:auie_metric_disfa} provide the corresponding dataset-specific results. MAE-Face-B achieves the highest average ICC(3,1) of 0.728 and PCC of 0.743, representing relative performance gains of 4.15\% and 2.34\%, respectively, over the second-best GraphAU (ViT-B). GraphAU (ViT-B), however, obtains the lowest MSE of 0.395, corresponding to a relative error reduction of 1.00\% over the 0.399 obtained by MAE-Face-B. These results further support the observation in Sec.4.3 of the main paper that increasing the number of parameters does not consistently improve FAUE performance. In particular, MDHR (Res50), which contains the most parameters among the evaluated FAUE models, achieves the lowest average ICC(3,1) of 0.660, whereas MAE-Face-B achieves the best correlation-based performance with fewer parameters. The dataset-specific results show a consistent advantage for MAE-Face-B: it achieves the highest ICC(3,1) and PCC on BP4D, reaching 0.707 and 0.718, respectively, while GraphAU (ViT-B) obtains the lowest MSE of 0.941. On DISFA, MAE-Face-B performs best across all three metrics, achieving an ICC(3,1) of 0.729, a PCC of 0.745, and an MSE of 0.184. This consistent performance across both datasets explains its superior function-level results in Table \ref{tab:auie_metric}.

\noindent \textbf{Application-oriented model selection.}
The computational measures in Table \ref{tab:model_efficiency} provide further guidance for selecting FAUE models under different application requirements. MAE-Face-B provides the most favourable balance between predictive performance and computation, combining the highest average ICC(3,1) and PCC with 86.10M parameters, 16.87G FLOPs, and an inference speed of 131 FPS on L20 GPU. It is therefore the preferred general-purpose configuration when both correlation-based performance and real-time inference are required. GraphAU (Res50) contains the fewest parameters at 33.25M and reaches 94 FPS, making it more suitable when model size is the primary constraint. GraphAU (ViT-B) is preferable when minimising intensity-estimation error is prioritised, as it achieves the lowest function-level and BP4D MSE, although its inference speed decreases to 70 FPS with 28.16G FLOPs. FMAE-B has the same number of parameters, FLOPs, and inference speed as MAE-Face-B and can be considered when performance on AU05 or AU09 is prioritised, for which it obtains the highest ICC(3,1). MDHR (Res50) provides the highest measured throughput of 940 FPS when temporal inputs are available, but its 126.66M parameters and 127.31G FLOPs require substantially greater computational resources. It is therefore more appropriate for high-throughput video processing when sufficient computational resources are available. These results also demonstrate that neither parameter count nor FLOPs alone determines predictive performance or practical inference speed, and model selection should jointly consider the target AUs' performance and available computational resources.

\subsubsection{SFER performances}

\begin{table*}[htbp]
\centering
\renewcommand{\arraystretch}{1.3}
\caption{Dataset-specific accuracy (\%) $\uparrow$ and macro-averaged F1 score (\%) $\uparrow$ of the SFER models on AffectNet-8, RAF-DB, FER+, SFEW, and AFFW-FER. The best and second-best results for each dataset and metric are highlighted in bold and underlined, respectively.}
\resizebox{1.0\linewidth}{!}{
\begin{tabular}{l|c|cccccccccc}
\Xhline{1.2px}
\rowcolor{gray!15}
\cellcolor{gray!15} &
  \cellcolor{gray!15} &
  \multicolumn{2}{c}{\cellcolor{gray!15}\textbf{AffectNet-8} \cite{mollahosseini2017affectnet}} &
  \multicolumn{2}{c}{\cellcolor{gray!15}\textbf{RAF-DB} \cite{shan2018reliable}} &
  \multicolumn{2}{c}{\cellcolor{gray!15}\textbf{FER+} \cite{BarsoumICMI2016}} &
  \multicolumn{2}{c}{\cellcolor{gray!15}\textbf{SFEW} \cite{dhall2015video}} &
  \multicolumn{2}{c}{\cellcolor{gray!15}\textbf{AFFW-FER} \cite{kollias2018aff}} \\ \hhline{~~||--||--||--||--||--}
\rowcolor{gray!15}
\multirow{-2}{*}{\cellcolor{gray!15}\textbf{Methods}} &
  \multirow{-2}{*}{\cellcolor{gray!15}\textbf{Year}} &
  Acc. (\%) &
  F1 (\%) &
  Acc. (\%) &
  F1 (\%) &
  Acc. (\%) &
  F1 (\%) &
  Acc. (\%) &
  F1 (\%) &
  Acc. (\%) &
  F1 (\%) \\ \hhline{============}
EMOFAN \cite{toisoul2021estimation}   & NMI 2021  
& 61.75 
& 61.73 
& 90.32 
& 84.98 
& 89.57 
& 80.51 
& \cellcolor{gray!30}\textbf{63.37} 
& \cellcolor{gray!15}\underline{56.05} 
& 67.72 
& 41.78 \\
DDAMFN \cite{zhang2023dual}  & Ele. 2023 
& 63.58 
& 63.51 
& 92.37 
& 87.89 
& 90.61 
& 82.10 
& 61.39 
& 53.04 
& 70.40 
& 44.21 \\
Poster++ \cite{mao2025poster++} & PR 2025  
& 62.65 
& 62.41 
& 90.48 
& 85.08 
& 90.10 
& 80.27 
& 62.13 
& 54.02 
& 68.15 
& 41.84 \\
MHAN \cite{wang2026mhan}   & PR 2026   
& \cellcolor{gray!30}\textbf{63.75} 
& \cellcolor{gray!30}\textbf{63.72} 
& 92.60 
& 87.98 
& 90.83 
& \cellcolor{gray!15}\underline{82.53} 
& 61.88 
& 53.87 
& 70.51 
& \cellcolor{gray!15}\underline{44.84} \\ 
\hhline{============}
MAE-Face-B \cite{ma2024facial} & TAC 2024  
& 63.63 
& 63.57 
& \cellcolor{gray!15}\underline{92.89} 
& \cellcolor{gray!15}\underline{88.85} 
& \cellcolor{gray!15}\underline{91.37} 
& 81.93 
& 55.20 
& 50.61 
& \cellcolor{gray!30}\textbf{72.78} 
& \cellcolor{gray!30}\textbf{50.70} \\
FMAE-B \cite{ning2024representation}  & FG 2025   
& 63.08 
& 63.01
& 92.18 
& 87.13
& 90.61 
& 81.99 
& 61.63 
& 54.12 
& 66.69 
& 41.73 \\
FMAE-L \cite{ning2024representation}  & FG 2025   
& \cellcolor{gray!15}\underline{63.72} 
& \cellcolor{gray!15}\underline{63.60} 
& \cellcolor{gray!30}\textbf{93.22} 
& \cellcolor{gray!30}\textbf{88.99} 
& \cellcolor{gray!30}\textbf{92.23} 
& \cellcolor{gray!30}\textbf{84.35} 
& \cellcolor{gray!15}\underline{62.38} 
& \cellcolor{gray!30}\textbf{56.96}
& \cellcolor{gray!15}\underline{70.92} 
& 43.28 \\ \Xhline{1.2px}
\end{tabular}
}
\label{tab:fer}
\end{table*}

\begin{table*}[htbp]
\centering
\renewcommand{\arraystretch}{1.3}
\caption{Class-wise F1 scores (\%) $\uparrow$ on AffectNet-8. The Average column reports the macro-averaged F1 score across the eight expression categories.}
\resizebox{1.0\linewidth}{!}{
\begin{tabular}{l|c|ccccccccc}
\Xhline{1.2px}
\rowcolor{gray!15}
{\cellcolor{gray!15}\textbf{Methods}} &
{\cellcolor{gray!15}\textbf{Year}} &
{\cellcolor{gray!15}\textbf{Neutral}} &
{\cellcolor{gray!15}\textbf{Happiness}} &
{\cellcolor{gray!15}\textbf{Sadness}} &
{\cellcolor{gray!15}\textbf{Surprise}} &
{\cellcolor{gray!15}\textbf{Fear}} &
{\cellcolor{gray!15}\textbf{Anger}} &
{\cellcolor{gray!15}\textbf{Disgust}} &
{\cellcolor{gray!15}\textbf{Contempt}} &
{\cellcolor{gray!15}\textbf{Average}} \\ \hhline{===========}
EMOFAN \cite{toisoul2021estimation}   & NMI 2021  & 51.53 & 73.46 & 62.76 & 59.39 & 66.59 & 61.00 & 57.14 & 61.97 & 61.73\\
DDAMFN \cite{zhang2023dual}  & Ele. 2023 & 53.98 & 75.46 & 64.82 & 61.60 & 66.38 & 60.84 & 60.13 & 64.89 & 63.51\\
Poster++ \cite{mao2025poster++} & PR 2025   & 54.29 & 74.15 & 64.75 & 62.05 & 65.80 & 58.37 & 61.07 & 58.78 & 62.41 \\
MHAN \cite{wang2026mhan}   & PR 2026   & 53.51 & 74.93 & 65.32 & 62.46 & 67.89 & 61.55 & 59.33 & 64.72 & 63.72 \\ \hhline{===========}
MAE-Face \cite{ma2024facial} & TAC 2024  & 51.35 & 76.52 & 66.31 & 62.62 & 65.45 & 62.63 & 59.34 & 64.34 & 63.57 \\
FMAE-B \cite{ning2024representation}  & FG 2025   & 53.18 & 75.11 & 65.24 & 61.48 & 66.11 & 61.56 & 58.61 & 62.82 & 63.01 \\
FMAE-L \cite{ning2024representation}  & FG 2025   & 55.66 & 74.46 & 66.47 & 60.31 & 68.42 & 59.80 & 61.13 & 62.54 & 63.60 \\ \Xhline{1.2px}
\end{tabular}
}
\label{tab:sfer_metric_affectnet}
\end{table*}

\begin{table*}[htbp]
\centering
\renewcommand{\arraystretch}{1.3}
\caption{Class-wise F1 scores (\%) $\uparrow$ on RAF-DB. The Average column reports the macro-averaged F1 score across the seven evaluated expression categories. Contempt is not included in the RAF-DB evaluation and is denoted by ``-''.}
\resizebox{1.0\linewidth}{!}{
\begin{tabular}{l|c|ccccccccc}
\Xhline{1.2px}
\rowcolor{gray!15}
{\cellcolor{gray!15}\textbf{Methods}} &
{\cellcolor{gray!15}\textbf{Year}} &
{\cellcolor{gray!15}\textbf{Neutral}} &
{\cellcolor{gray!15}\textbf{Happiness}} &
{\cellcolor{gray!15}\textbf{Sadness}} &
{\cellcolor{gray!15}\textbf{Surprise}} &
{\cellcolor{gray!15}\textbf{Fear}} &
{\cellcolor{gray!15}\textbf{Anger}} &
{\cellcolor{gray!15}\textbf{Disgust}} &
{\cellcolor{gray!15}\textbf{Contempt}} &
{\cellcolor{gray!15}\textbf{Average}} \\ \hhline{===========}
EMOFAN \cite{toisoul2021estimation}   & NMI 2021  & 89.44 & 95.53 & 89.86 & 88.76 & 69.33 & 86.08 & 75.88 & - & 84.98\\
DDAMFN \cite{zhang2023dual}  & Ele. 2023 & 92.09 & 96.95 & 91.24 & 91.16 & 77.85 & 86.57 & 79.39 & - & 87.89\\
Poster++ \cite{mao2025poster++} & PR 2025   & 90.07 & 95.96 & 89.51 & 88.86 & 71.05 & 85.80 & 74.34 & - & 85.08 \\
MHAN \cite{wang2026mhan}   & PR 2026   & 92.34 & 97.14 & 91.34 & 91.98 & 77.33 & 86.98 & 78.75 & - & 87.98 \\ \hhline{===========}
MAE-Face \cite{ma2024facial} & TAC 2024  & 91.74 & 96.92 & 92.05 & 92.15 & 75.52 & 89.84 & 83.73 & - & 88.85 \\
FMAE-B \cite{ning2024representation}  & FG 2025   & 91.32 & 96.91 & 91.11 & 91.24 & 71.62 & 87.69 & 80.00 & - & 87.13 \\
FMAE-L \cite{ning2024representation}  & FG 2025   & 92.01 & 97.61 & 91.88 & 91.19 & 73.83 & 91.41 & 84.97 & - & 88.99 \\ \Xhline{1.2px}
\end{tabular}
}
\label{tab:sfer_metric_raf_db}
\end{table*}

\begin{table*}[htbp]
\centering
\renewcommand{\arraystretch}{1.3}
\caption{Class-wise F1 scores (\%) $\uparrow$ on FER+. The Average column reports the macro-averaged F1 score across the eight expression categories.}
\resizebox{1.0\linewidth}{!}{
\begin{tabular}{l|c|ccccccccc}
\Xhline{1.2px}
\rowcolor{gray!15}
{\cellcolor{gray!15}\textbf{Methods}} &
{\cellcolor{gray!15}\textbf{Year}} &
{\cellcolor{gray!15}\textbf{Neutral}} &
{\cellcolor{gray!15}\textbf{Happiness}} &
{\cellcolor{gray!15}\textbf{Sadness}} &
{\cellcolor{gray!15}\textbf{Surprise}} &
{\cellcolor{gray!15}\textbf{Fear}} &
{\cellcolor{gray!15}\textbf{Anger}} &
{\cellcolor{gray!15}\textbf{Disgust}} &
{\cellcolor{gray!15}\textbf{Contempt}} &
{\cellcolor{gray!15}\textbf{Average}} \\ \hhline{===========}
EMOFAN \cite{toisoul2021estimation}   & NMI 2021  & 90.06 & 95.22 & 80.40 & 89.94 & 72.22 & 88.97 & 68.97 & 58.33 & 80.51\\
DDAMFN \cite{zhang2023dual}  & Ele. 2023 & 90.53 & 95.80 & 81.84 & 92.46 & 74.83 & 91.04 & 68.75 & 61.54 & 82.10\\
Poster++ \cite{mao2025poster++} & PR 2025   & 90.63 & 95.11 & 82.31 & 91.44 & 72.48 & 88.97 & 62.86 & 58.33 & 80.27 \\
MHAN \cite{wang2026mhan}   & PR 2026   & 90.84 & 96.06 & 82.13 & 92.36 & 75.34 & 91.01 & 70.97 & 61.54 & 82.53 \\ \hhline{===========}
MAE-Face \cite{ma2024facial} & TAC 2024  & 91.64 & 96.60 & 83.78 & 92.00 & 72.34 & 91.94 & 63.16 & 64.00 & 81.93 \\
FMAE-B \cite{ning2024representation}  & FG 2025   & 91.33 & 95.85 & 82.32 & 90.89 & 72.60 & 89.48 & 70.97 & 61.54 & 81.87 \\
FMAE-L \cite{ning2024representation}  & FG 2025   & 93.08 & 97.08 & 85.23 & 92.06 & 71.72 & 90.58 & 70.97 & 74.07 & 84.35 \\ \Xhline{1.2px}
\end{tabular}
}
\label{tab:sfer_metric_fer_plus}
\end{table*}

\begin{table*}[htbp]
\centering
\renewcommand{\arraystretch}{1.3}
\caption{Class-wise F1 scores (\%) $\uparrow$ on SFEW. The Average column reports the macro-averaged F1 score across the seven evaluated expression categories. Contempt is not included in the SFEW evaluation and is denoted by ``-''.}
\resizebox{1.0\linewidth}{!}{
\begin{tabular}{l|c|ccccccccc}
\Xhline{1.2px}
\rowcolor{gray!15}
{\cellcolor{gray!15}\textbf{Methods}} &
{\cellcolor{gray!15}\textbf{Year}} &
{\cellcolor{gray!15}\textbf{Neutral}} &
{\cellcolor{gray!15}\textbf{Happiness}} &
{\cellcolor{gray!15}\textbf{Sadness}} &
{\cellcolor{gray!15}\textbf{Surprise}} &
{\cellcolor{gray!15}\textbf{Fear}} &
{\cellcolor{gray!15}\textbf{Anger}} &
{\cellcolor{gray!15}\textbf{Disgust}} &
{\cellcolor{gray!15}\textbf{Contempt}} &
{\cellcolor{gray!15}\textbf{Average}} \\ \hhline{===========}
EMOFAN \cite{toisoul2021estimation}   & NMI 2021  & 68.60 & 82.76 & 65.04 & 56.36 & 44.44 & 70.15 & 5.00 & - & 56.05\\
DDAMFN \cite{zhang2023dual}  & Ele. 2023 & 61.62 & 82.67 & 67.20 & 42.35 & 44.74 & 72.73 & 0.00 & - & 53.04\\
Poster++ \cite{mao2025poster++} & PR 2025   & 65.24 & 81.05 & 65.12 & 55.81 & 37.50 & 69.01 & 4.44 & - & 54.02 \\
MHAN \cite{wang2026mhan}   & PR 2026   & 63.68 & 82.67 & 64.41 & 47.73 & 42.11 & 71.11 & 5.41 & - & 53.87 \\ \hhline{===========}
MAE-Face \cite{ma2024facial} & TAC 2024  & 60.26 & 84.29 & 57.89 & 47.17 & 38.46 & 60.43 & 5.80 & - & 50.61 \\
FMAE-B \cite{ning2024representation}  & FG 2025   & 63.41 & 81.88 & 66.13 & 50.45 & 38.46 & 73.38 & 5.13 & - & 54.12 \\
FMAE-L \cite{ning2024representation}  & FG 2025   & 60.47 & 81.33 & 62.90 & 58.70 & 46.34 & 69.93 & 19.05 & - & 56.96 \\ \Xhline{1.2px}
\end{tabular}
}
\label{tab:sfer_metric_sfew}
\end{table*}

\begin{table*}[htbp]
\centering
\renewcommand{\arraystretch}{1.3}
\caption{Class-wise F1 scores (\%) $\uparrow$ on AFFW-FER. The Average column reports the macro-averaged F1 score across the seven evaluated expression categories. Contempt is not included in the AFFW-FER evaluation and is denoted by ``-''.}
\resizebox{1.0\linewidth}{!}{
\begin{tabular}{l|c|ccccccccc}
\Xhline{1.2px}
\rowcolor{gray!15}
{\cellcolor{gray!15}\textbf{Methods}} &
{\cellcolor{gray!15}\textbf{Year}} &
{\cellcolor{gray!15}\textbf{Neutral}} &
{\cellcolor{gray!15}\textbf{Happiness}} &
{\cellcolor{gray!15}\textbf{Sadness}} &
{\cellcolor{gray!15}\textbf{Surprise}} &
{\cellcolor{gray!15}\textbf{Fear}} &
{\cellcolor{gray!15}\textbf{Anger}} &
{\cellcolor{gray!15}\textbf{Disgust}} &
{\cellcolor{gray!15}\textbf{Contempt}} &
{\cellcolor{gray!15}\textbf{Average}} \\ \hhline{===========}
EMOFAN \cite{toisoul2021estimation}   & NMI 2021  & 81.84 & 68.73 & 55.57 & 44.68 & 9.86 & 27.70 & 4.06 & - & 41.78\\
DDAMFN \cite{zhang2023dual}  & Ele. 2023 & 83.24 & 70.93 & 62.84 & 43.81 & 12.48 & 33.99 & 2.20 & - & 44.21\\
Poster++ \cite{mao2025poster++} & PR 2025   & 82.31 & 69.60 & 61.49 & 39.02 & 12.82 & 23.60 & 4.05 & - & 41.84 \\
MHAN \cite{wang2026mhan}   & PR 2026   & 83.22 & 71.14 & 61.76 & 44.14 & 16.72 & 33.93 & 2.99 & - & 44.84 \\ \hhline{===========}
MAE-Face \cite{ma2024facial} & TAC 2024  & 83.46 & 71.27 & 67.04 & 52.55 & 18.85 & 24.62 & 37.12 & - & 50.70 \\
FMAE-B \cite{ning2024representation}  & FG 2025   & 78.82 & 67.89 & 58.74 & 46.94 & 7.48 & 28.44 & 3.83 & - & 41.73 \\
FMAE-L \cite{ning2024representation}  & FG 2025   & 83.98 & 69.95 & 61.35 & 46.95 & 6.45 & 27.52 & 6.75 & - & 43.28 \\ \Xhline{1.2px}
\end{tabular}
}
\label{tab:sfer_metric_affw}
\end{table*}

\noindent \textbf{Overall and dataset-specific performance.} 
Table \ref{tab:fer} reports the accuracy and F1 scores of the seven SFER models on AffectNet-8, RAF-DB, FER+, SFEW, and AFFW-FER. Averaged across the five datasets, FMAE-L achieves the highest accuracy of 76.49\% and the highest F1 score of 67.44\%. Compared with MHAN, which obtains the second-highest average accuracy of 75.91\%, FMAE-L provides a relative performance gain of 0.76\% but increases the number of parameters from 4.20M to 305.42M. Similarly, its average F1 score provides only a 0.45\% relative performance gain over the 67.13\% obtained by MAE-Face-B. The dataset-level results nevertheless show that the additional parameters of FMAE-L provide clear benefits on several datasets. It achieves the highest accuracy and F1 score on both RAF-DB and FER+, as well as the highest F1 score on SFEW. However, this advantage is not consistent across different data distributions (e.g., AffectNet and AFFW-FER). MHAN performs best on AffectNet-8, EMOFAN achieves the highest accuracy on SFEW, and MAE-Face-B leads on AFFW-FER. In particular, MAE-Face-B achieves an F1 score of 50.70\% on AFFW-FER, representing a relative performance gain of 13.07\% over the second-best MHAN. These results extend the analysis in Sec.~4.3 of the main paper: increasing the number of parameters generally benefits SFER, but the improvement can be modest and remains dependent on the evaluated dataset.

\noindent \textbf{Class-wise performance.} 
Tables \ref{tab:sfer_metric_affectnet}-\ref{tab:sfer_metric_affw} further identify the expression categories responsible for these dataset-level differences. On AffectNet-8, FMAE-L performs best on Neutral, Sadness, Fear, and Disgust, while MAE-Face-B leads on Happiness, Surprise, and Anger, and DDAMFN performs best on Contempt. On RAF-DB, FMAE-L provides the highest F1 scores for Happiness, Anger, and Disgust, whereas MAE-Face-B performs best on Sadness and Surprise, and DDAMFN remains preferable for Fear. The advantages of FMAE-L become more evident on FER+, where it performs best on Neutral, Happiness, Sadness, and Contempt and obtains the highest average F1 score of 84.35\%. On the more challenging SFEW dataset, FMAE-L achieves the best results for Surprise, Fear, and Disgust, while EMOFAN, MAE-Face-B, DDAMFN, and FMAE-B retain advantages on other categories. AFFW-FER presents a different pattern: MAE-Face-B performs best on Happiness, Sadness, Surprise, Fear, and Disgust, whereas FMAE-L leads only on Neutral. This class-wise difference explains why FMAE-L obtains a relatively high accuracy but a substantially lower F1 score than MAE-Face-B on AFFW-FER.

\begin{figure*}[htbp]
    \centering
    \begin{subfigure}[t]{0.8\textwidth}
        \centering
        \includegraphics[width=\linewidth]{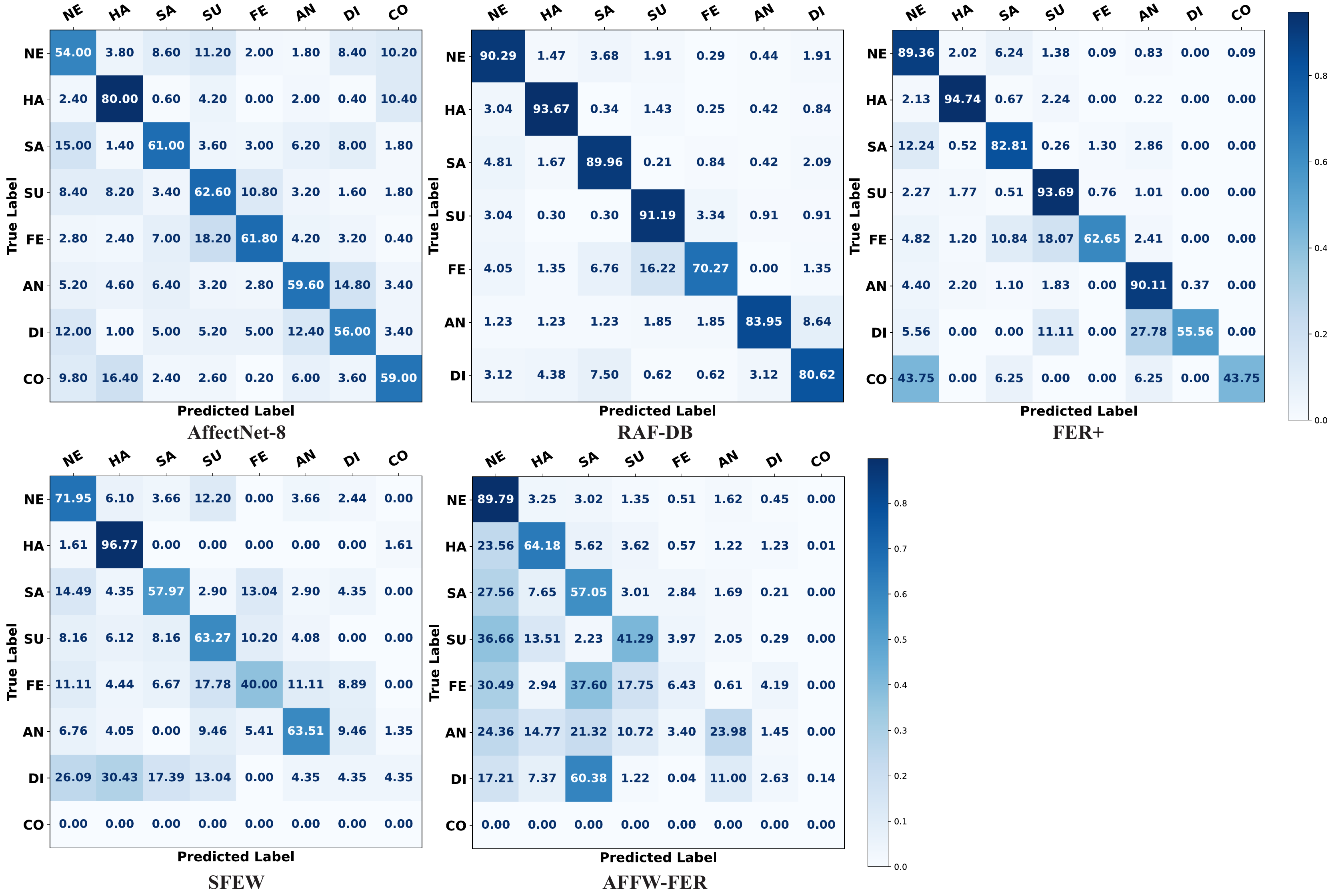}
        \caption{Row-normalised confusion matrices (\%) of EMOFAN.}
        \label{fig:cm_emofan}
    \end{subfigure}
    
    \vspace{0.5cm}
    
    \begin{subfigure}[t]{0.8\textwidth}
        \centering
        \includegraphics[width=\linewidth]{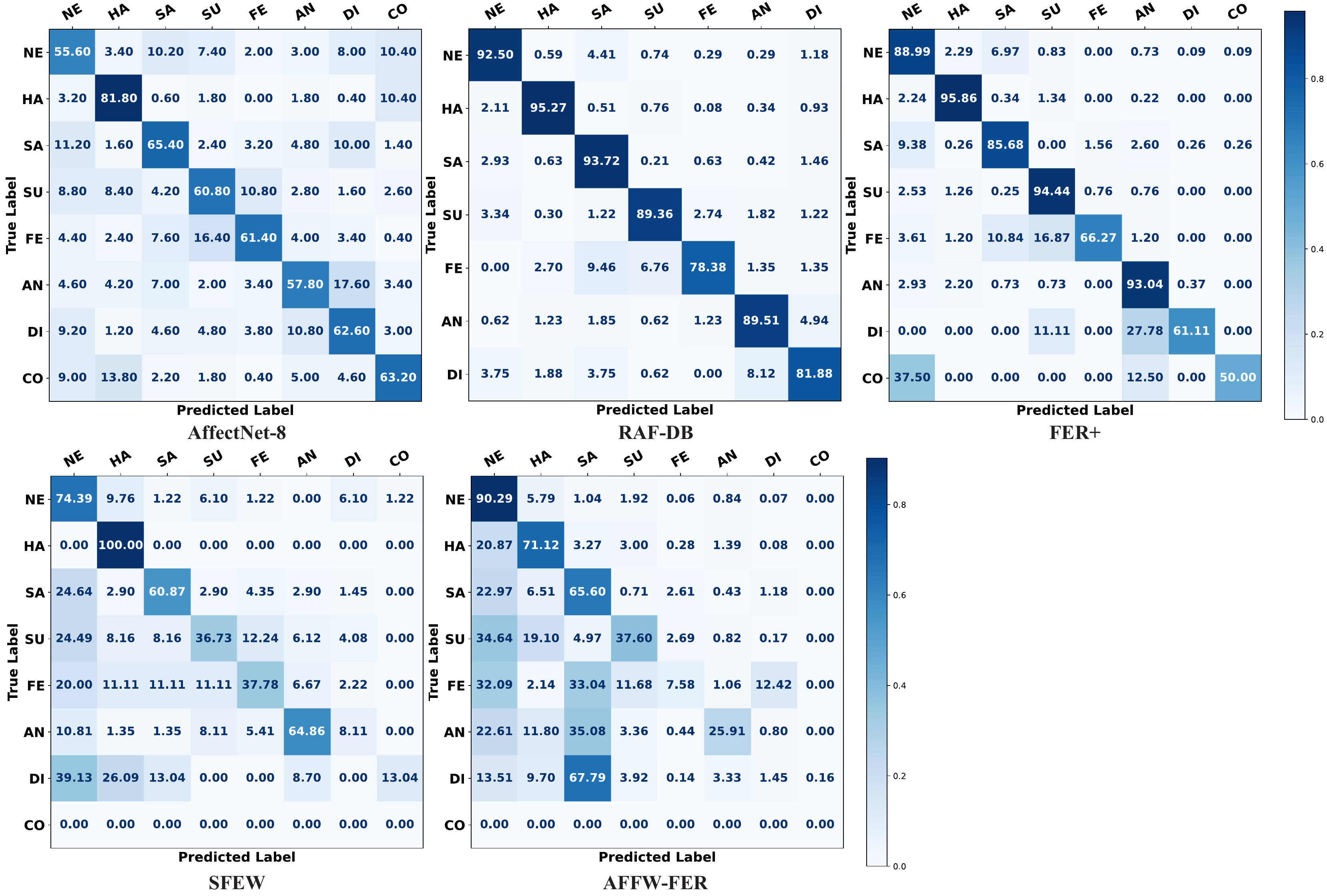}
        \caption{Row-normalised confusion matrices (\%) of DDAMFN.}
        \label{fig:cm_ddamfn}
    \end{subfigure}
\caption{Row-normalised confusion matrices (\%) of EMOFAN and EMOFAN on AffectNet-8, RAF-DB, FER+, SFEW, and AFFW-FER. NE: Neutral; HA: Happiness; SA: Sadness; SU: Surprise; FE: Fear; AN: Anger; DI: Disgust; CO: Contempt.}
\label{fig:sfer_confu1}
\end{figure*}

\begin{figure*}[htbp]
    \centering

    \begin{subfigure}[t]{0.8\textwidth}
        \centering
        \includegraphics[width=\linewidth]{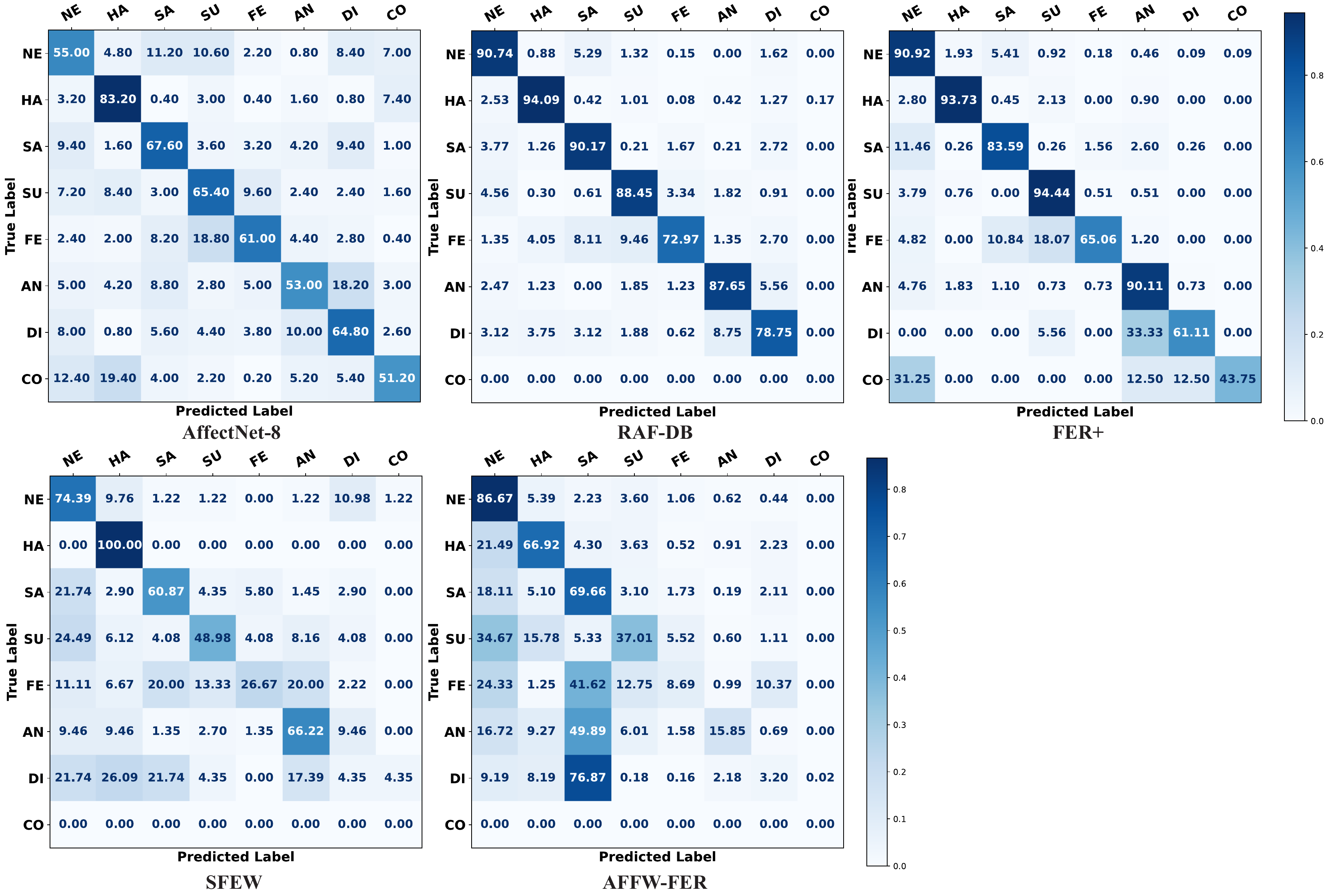}
        \caption{Row-normalised confusion matrices (\%) of Poster++.}
        \label{fig:cm_posterv2}
    \end{subfigure}

    \vspace{0.5cm}

    \begin{subfigure}[t]{0.8\textwidth}
        \centering
        \includegraphics[width=\linewidth]{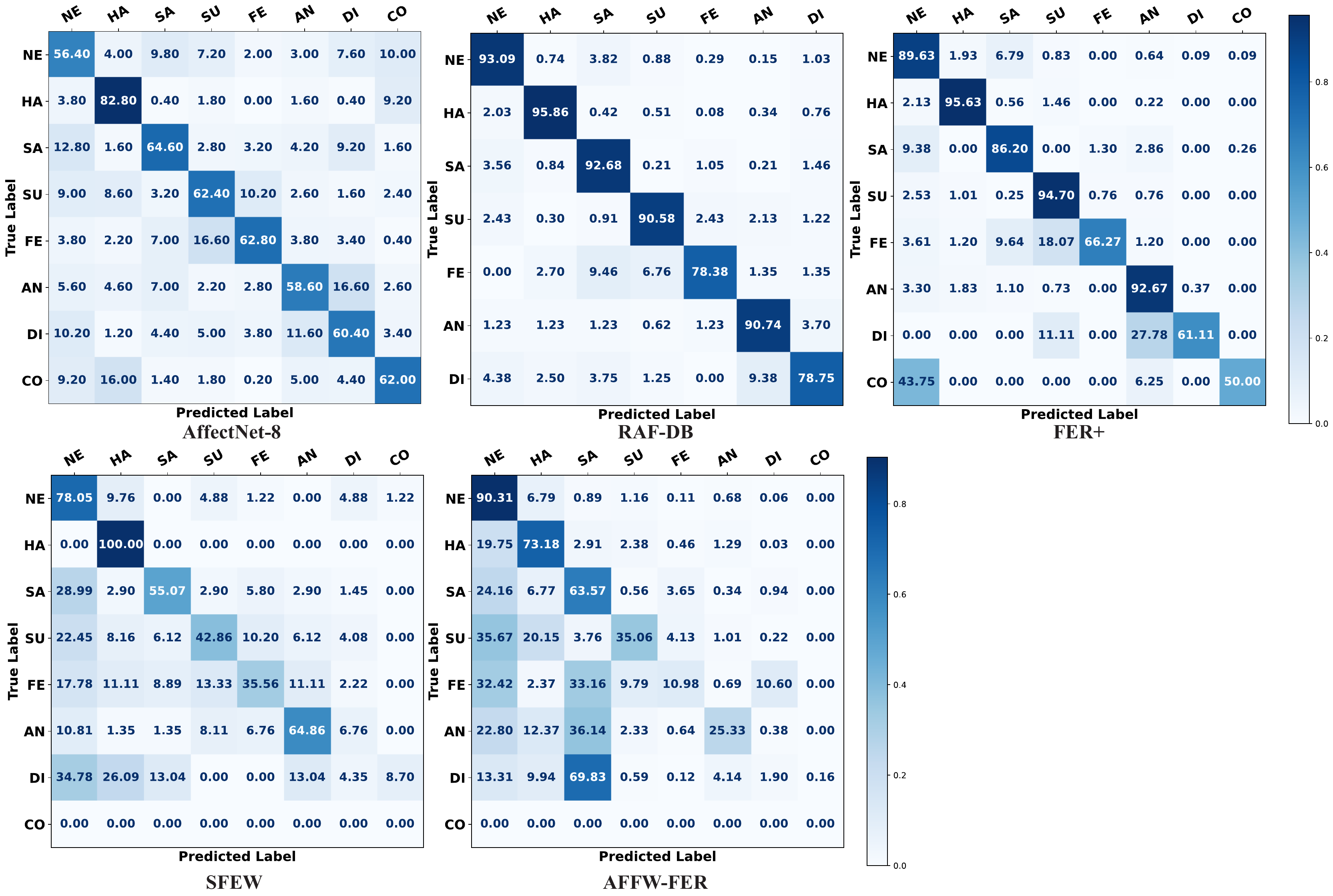}
        \caption{Row-normalised confusion matrices (\%) of MHAN.}
        \label{fig:cm_mhan}
    \end{subfigure}

\caption{Row-normalised confusion matrices (\%) of Poster++ and MHAN on AffectNet-8, RAF-DB, FER+, SFEW, and AFFW-FER. NE: Neutral; HA: Happiness; SA: Sadness; SU: Surprise; FE: Fear; AN: Anger; DI: Disgust; CO: Contempt.}
\label{fig:sfer_confu2}
\end{figure*}

\begin{figure*}[htbp]
    \centering

    \begin{subfigure}[t]{0.8\textwidth}
        \centering
        \includegraphics[width=\linewidth]{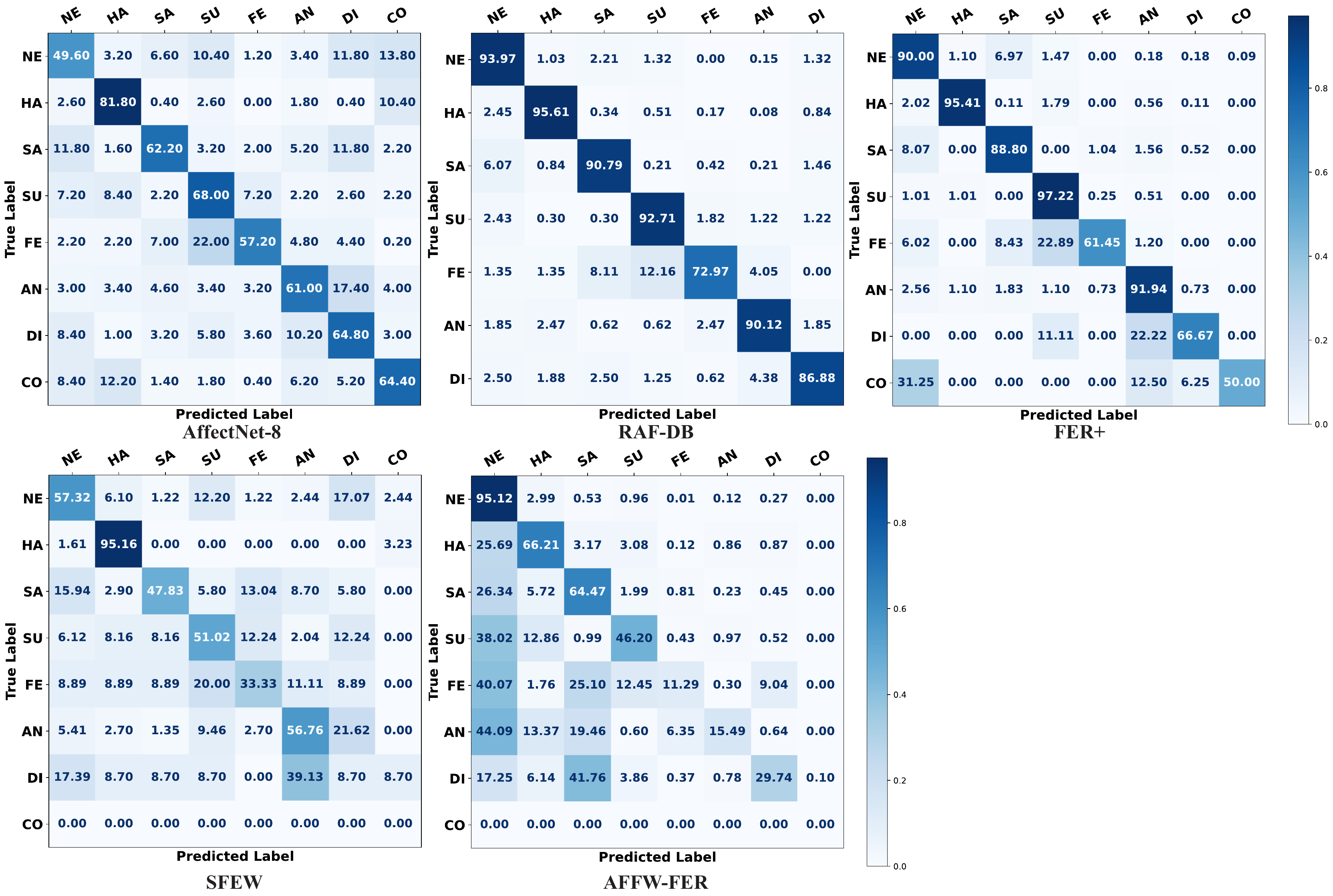}
        \caption{Row-normalised confusion matrices (\%) of MAE-Face.}
        \label{fig:cm_mae_face}
    \end{subfigure}

    \vspace{0.5cm}

    \begin{subfigure}[t]{0.8\textwidth}
        \centering
        \includegraphics[width=\linewidth]{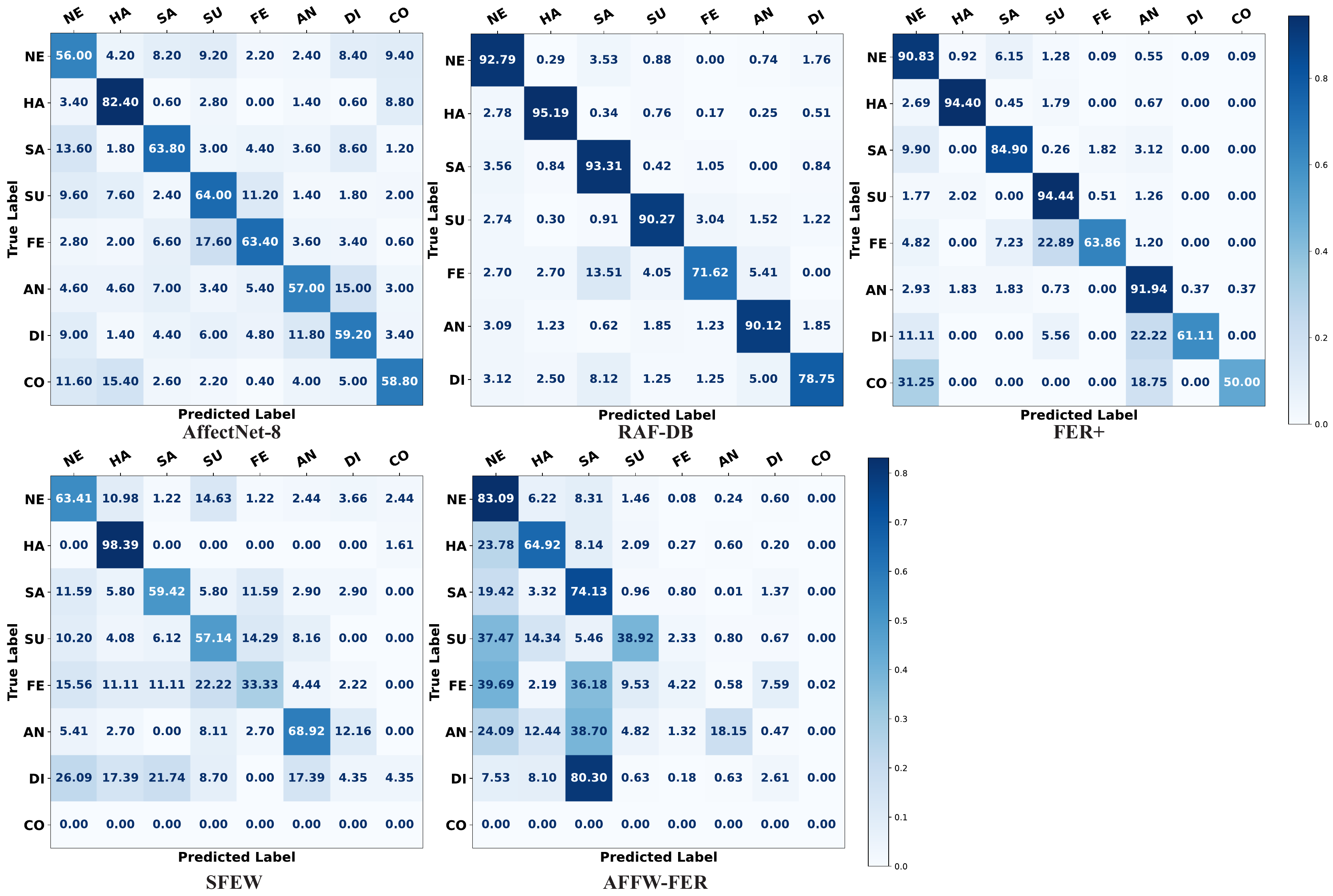}
        \caption{Row-normalised confusion matrices (\%) of FMAE-B.}
        \label{fig:cm_fmae_b}
    \end{subfigure}

\caption{Row-normalised confusion matrices (\%) of MAE-Face and FMAE-B on AffectNet-8, RAF-DB, FER+, SFEW, and AFFW-FER. NE: Neutral; HA: Happiness; SA: Sadness; SU: Surprise; FE: Fear; AN: Anger; DI: Disgust; CO: Contempt.}
\label{fig:sfer_confu3}
\end{figure*}

\begin{figure*}[h]
    \centering
    \includegraphics[width=0.8\linewidth]{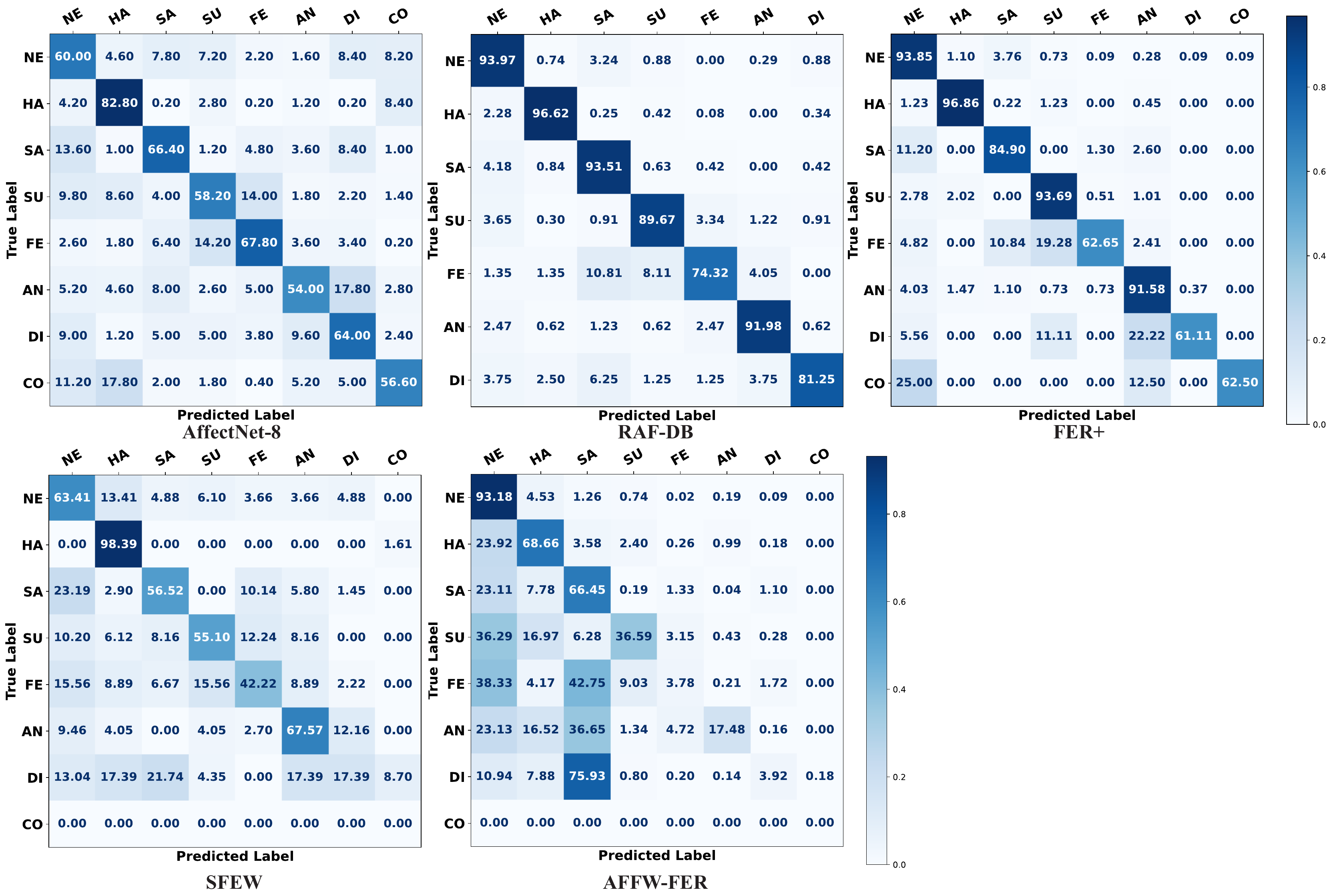}
    \caption{Row-normalised confusion matrices (\%) of FMAE-L on AffectNet-8, RAF-DB, FER+, SFEW, and AFFW-FER. NE: Neutral; HA: Happiness; SA: Sadness; SU: Surprise; FE: Fear; AN: Anger; DI: Disgust; CO: Contempt.}
    \label{fig:cm_fmae_l}
\end{figure*}

\noindent \textbf{Confusion patterns.} 
The row-normalised confusion matrices in Figs. \ref{fig:sfer_confu1}, \ref{fig:sfer_confu2}, \ref{fig:sfer_confu3}, and  \ref{fig:cm_fmae_l} provide further evidence for the class-wise results. Across models and datasets, Happiness and Neutral generally produce the strongest diagonal responses, whereas Fear and Disgust exhibit more dispersed predictions. Fear is frequently confused with Surprise and Sadness on AffectNet-8, RAF-DB, and FER+, while the same category is often predicted as Neutral or Sadness on SFEW and AFFW-FER. Disgust is commonly confused with Anger on AffectNet-8 and with Sadness on AFFW-FER. These errors become more pronounced on SFEW and AFFW-FER, where the diagonal responses for Fear and Disgust are substantially weaker than those for Neutral and Happiness. The matrices also clarify the different advantages of the best-performing models. FMAE-L produces stronger diagonal responses for several categories on RAF-DB and FER+ and improves the recognition of Surprise, Fear, and Disgust on SFEW. In contrast, MAE-Face-B substantially improves Disgust recognition on AFFW-FER and reduces the tendency to assign minority expression samples to Neutral or Sadness, leading to its considerably higher macro-averaged F1 score. Therefore, the confusion matrices confirm that the performance differences in Table \ref{tab:fer} arise not only from overall classification accuracy but also from how each model handles difficult and under-represented expression categories.

\noindent \textbf{Application-oriented model selection.}
The computational measures in Table \ref{tab:model_efficiency} enable users to select SFER models according to their performance and computational requirements. FMAE-L is preferable when the highest average performance is prioritised and sufficient computational resources are available, but it requires 305.42M parameters and 59.70G FLOPs and reaches 70 FPS. MHAN provides the most favourable parameter-performance balance, achieving an average accuracy only 0.76\% lower than FMAE-L with approximately 73 times fewer parameters. Its 4.20M parameters and 0.55G FLOPs make it suitable for applications with strict model-size and computation constraints, although its measured inference speed is 67 FPS. DDAMFN provides the smallest configuration at 4.14M parameters and can be considered when minimising model size is the primary requirement. MAE-Face-B is more appropriate when inference speed or performance on challenging in-the-wild data is prioritised. It reaches 131 FPS with 87.00M parameters and 16.87G FLOPs and provides the best results on AFFW-FER. EMOFAN offers a smaller 13.96M-parameter alternative with a high inference speed of 123 FPS, whereas FMAE-B shares the 131 FPS inference speed of MAE-Face-B and can be selected when its advantages on specific categories, such as Anger on SFEW, are required. These results show that the model with the most parameters is not necessarily preferable for every dataset or expression category, and users should jointly consider the target data distribution, required categories, model size, and inference speed.

\subsubsection{DFER performances}

\noindent\textbf{Overall and dataset-specific performance.} Table \ref{tab:dfer_metric} reports the UAR and WAR of the five DFER models on DFEW, FERV39k, and MAFW. Averaged across the three datasets, SVFAP-B achieves the highest UAR of 49.86\%, representing a relative performance gain of 0.18\% over the second-best MAE-DFER-B. FRU-Adapter-B obtains the highest average WAR of 60.08\%, providing a relative performance gain of 0.65\% over SVFAP-B. These results further extend the analysis in Sec. 4.3 of the main paper: increasing the number of parameters generally benefits DFER performance, although the relationship is not strictly monotonic. In particular, TimeSFormer-B contains the most parameters among the evaluated models but obtains the lowest average UAR and WAR. In contrast, MGNN (MFN), with only 7.56M parameters, provides relative performance gains of 2.17\% and 5.16\% over TimeSFormer-B in average UAR and WAR, respectively. The dataset-specific results further show that the preferred configuration depends on the evaluated data distribution. FRU-Adapter-B achieves the highest WAR of 77.45\% on DFEW and performs best in both UAR and WAR on MAFW, reaching 38.71\% and 50.57\%, respectively. SVFAP-B leads on FERV39k with a UAR of 46.05\% and a WAR of 54.10\%, whereas MAE-DFER-B obtains the highest DFEW UAR of 67.08\%. Therefore, no single model consistently performs best across both evaluation metrics and all three datasets.

\noindent\textbf{Class-wise performance.}
Tables \ref{tab:dfer_metric_dfew}-\ref{tab:dfer_metric_mafw} further identify the expression categories responsible for these dataset-level differences. On DFEW, FRU-Adapter-B performs best on happiness, sadness, and fear, while MAE-DFER-B leads on surprise and disgust. SVFAP-B and MGNN (MFN) achieve the highest accuracies for neutral and anger, respectively. Happiness remains the most accurately recognised category across all models, with accuracies ranging from 92.23\% to 96.00\%, whereas disgust remains substantially more difficult, ranging from 18.18\% to 34.48\%. On FERV39k, SVFAP-B achieves the best results for neutral and surprise and consequently obtains the highest UAR and WAR. FRU-Adapter-B performs best on sadness, MAE-DFER-B leads on happiness, MGNN (MFN) performs best on fear and anger, and TimeSFormer-B obtains the highest disgust accuracy. The MAFW results present a different pattern. FRU-Adapter-B leads on sadness, fear, and anxiety and shares the best results for disgust and contempt, explaining its superior overall performance on this dataset. MAE-DFER-B performs best on happiness and disappointment, SVFAP-B leads on surprise and helplessness, and MGNN (MFN) remains preferable for neutral and anger. Nevertheless, contempt, helplessness, and disappointment remain difficult for all models, and MGNN (MFN) records zero correct predictions for these three categories. These results demonstrate that DFER model selection should also consider the expression categories required by a particular application.

\begin{figure*}[h]
\centering
\begin{subfigure}[t]{1.0\textwidth}
    \centering
    \includegraphics[width=1.0\linewidth]{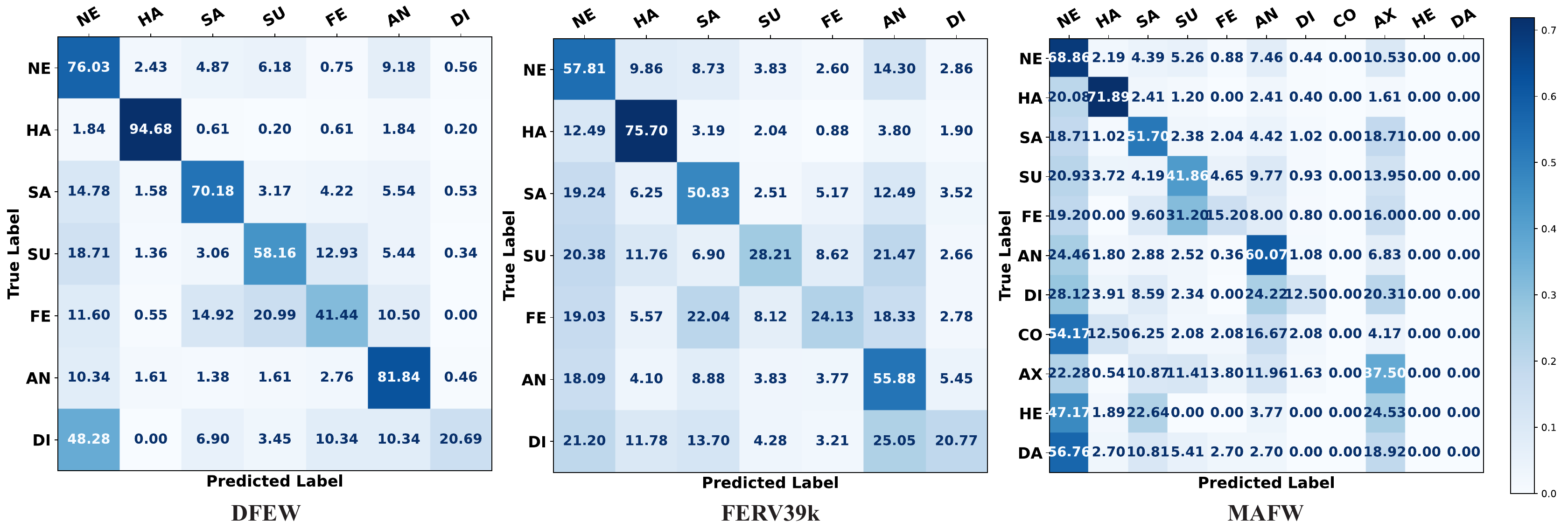}
    \caption{Row-normalised confusion matrices (\%) for MGNN (MFN) \cite{wu2025depmgnn}.}
    \label{fig:cm_mgnn}
\end{subfigure}

\vspace{0.5cm}

\begin{subfigure}[t]{1.0\textwidth}
    \centering
    \includegraphics[width=1.0\linewidth]{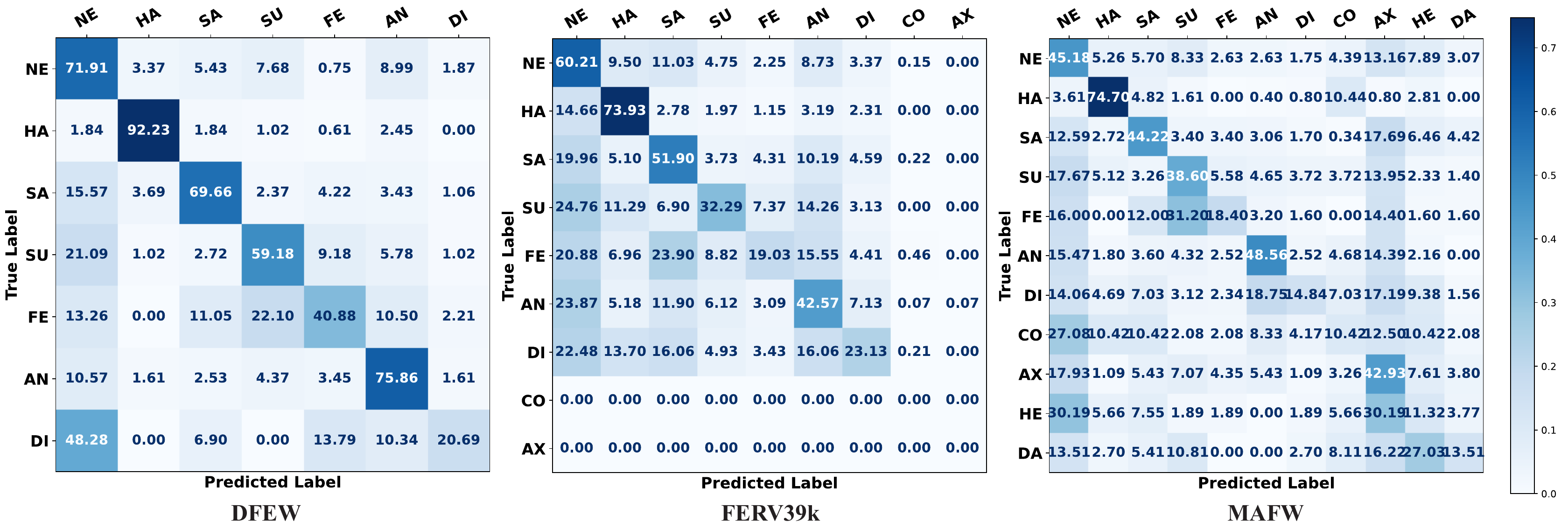}
    \caption{Row-normalised confusion matrices (\%) for TimeSFormer \cite{bertasius2021spacetime}.}
    \label{fig:cm_timesformer}
\end{subfigure}

\vspace{0.5cm}

\begin{subfigure}[t]{1.0\textwidth}
    \centering
    \includegraphics[width=1.0\linewidth]{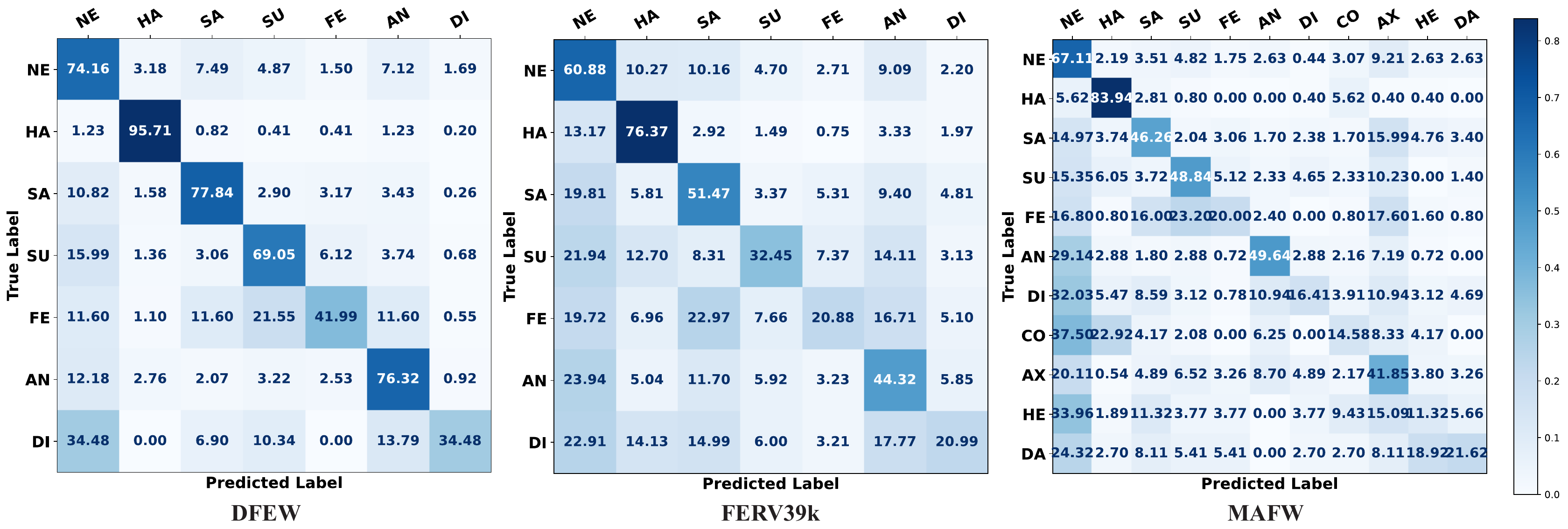}
    \caption{Row-normalised confusion matrices (\%) for MAE-DFER-B \cite{sun2023mae}.}
    \label{fig:cm_mae4dfer}
\end{subfigure}

\caption{Row-normalised confusion matrices (\%) for MGNN (MFN), TimeSFormer and MAE-DFER-B on DFEW, FERV39k, MAFW, respectively. NE: Neutral, HA: Happiness, SA: Sadness, SU: Surprise, FE: Fear, AN: Anger, DI: Disgust, CO: Contempt, AX: Anxiety, HE: Helplessness, DA: Disappointment.}
\label{fig:dfer_confu1}
\end{figure*}

\begin{figure*}[h]
\centering
\begin{subfigure}[t]{1.0\textwidth}
    \centering
    \includegraphics[width=1.0\linewidth]{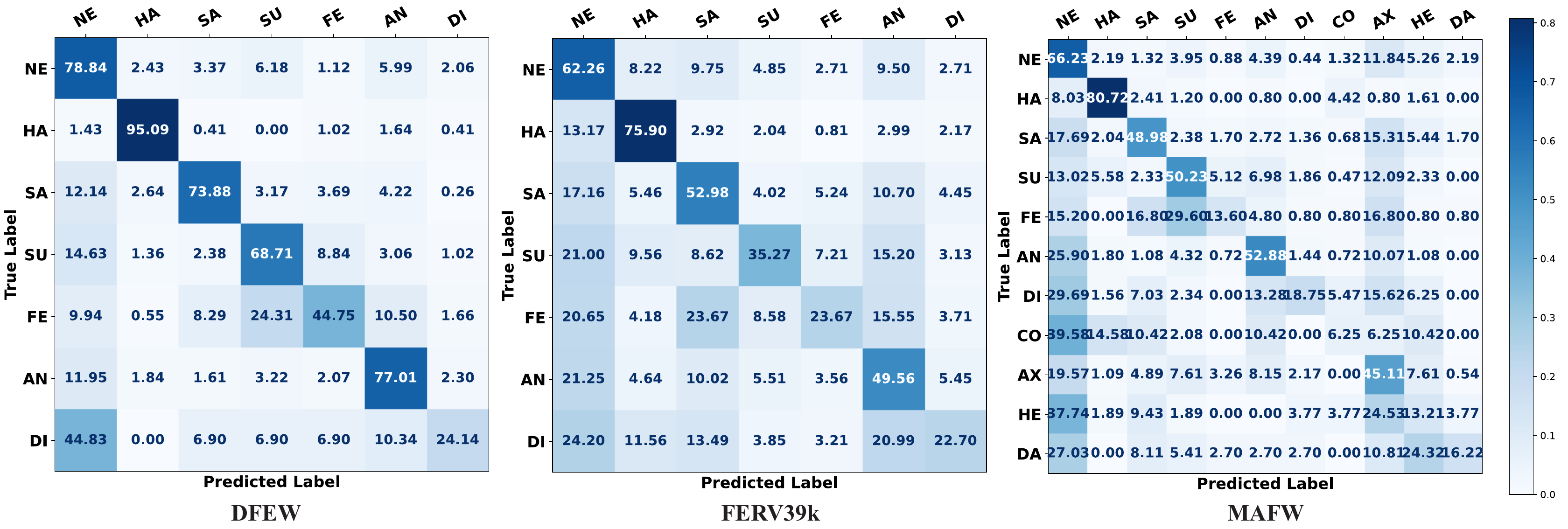}
    \caption{Row-normalised confusion matrices (\%) for SVFAP-B \cite{sun2024svfap}.}
    \label{fig:cm_svfap}
\end{subfigure}

\vspace{0.5cm}

\begin{subfigure}[t]{1.0\textwidth}
    \centering
    \includegraphics[width=1.0\linewidth]{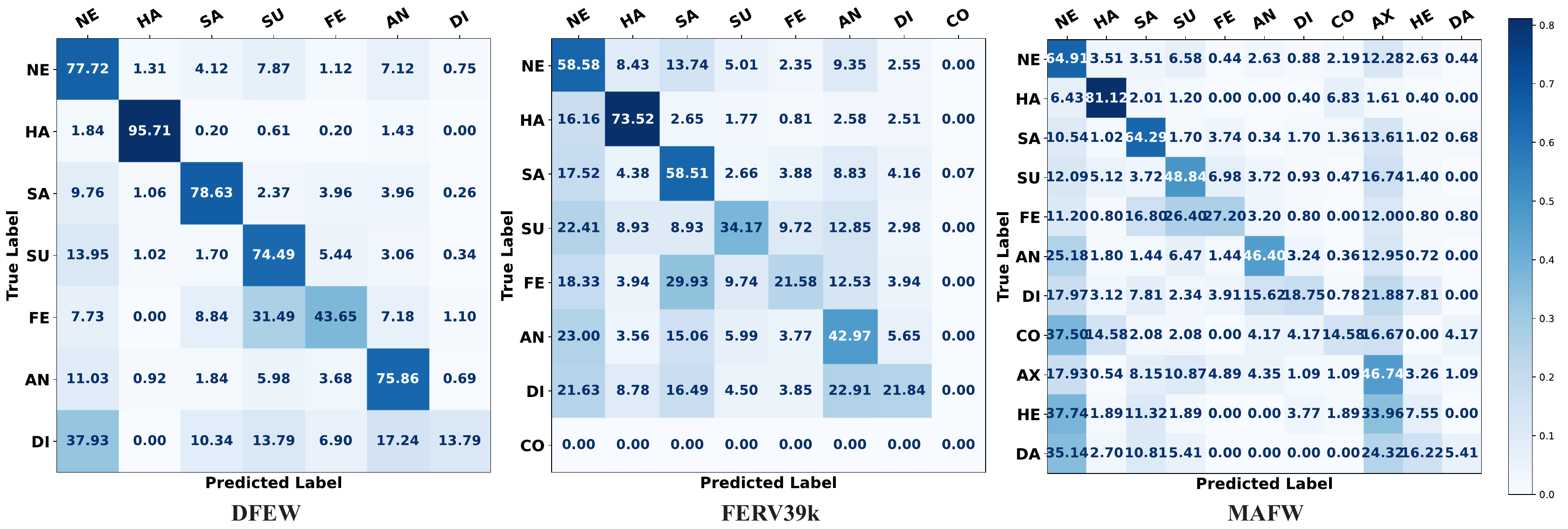}
    \caption{Row-normalised confusion matrices (\%) for FRU-Adpater-B \cite{her2025fru}.}
    \label{fig:cm_fru_adapter}
\end{subfigure}

\caption{Row-normalised confusion matrices (\%) for SVFAP-B and FRU-Adpater-B on DFEW, FERV39k, MAFW, respectively. NE: Neutral, HA: Happiness, SA: Sadness, SU: Surprise, FE: Fear, AN: Anger, DI: Disgust, CO: Contempt, AX: Anxiety, HE: Helplessness, DA: Disappointment.}
\label{fig:dfer_confu2}
\end{figure*}

\noindent\textbf{Confusion patterns.}
The row-normalised confusion matrices in Figs. \ref{fig:dfer_confu1}-\ref{fig:dfer_confu2} provide further evidence for the class-wise results. Across models and datasets, happiness generally produces the strongest diagonal responses, whereas fear, disgust, and the additional MAFW categories exhibit more dispersed predictions. On DFEW, between 34.48\% and 48.28\% of disgust samples are predicted as neutral, while between 20.99\% and 31.49\% of fear samples are assigned to surprise. On FERV39k, fear is frequently confused with sadness or neutral, while disgust is commonly predicted as neutral or anger. These errors explain the consistently lower accuracies of fear and disgust in Table \ref{tab:dfer_metric_ferv39k}. The recognition problem becomes more challenging on MAFW, where contempt, helplessness, and disappointment exhibit weak diagonal responses and are frequently assigned to neutral or anxiety. The matrices also clarify the different advantages of the evaluated models. MAE-DFER-B produces stronger diagonal responses for surprise and disgust on DFEW, SVFAP-B improves neutral and surprise recognition on FERV39k, and FRU-Adapter-B provides stronger responses for sadness, fear, and anxiety on MAFW. Therefore, the confusion matrices confirm that the overall performance differences in Table \ref{tab:dfer_metric} arise from improvements in particular expression categories rather than uniform improvements across all classes.

\noindent\textbf{Application-oriented model selection.} The computational measures in Table \ref{tab:model_efficiency} provide complementary guidance for selecting DFER models under different application requirements. FRU-Adapter-B is preferable when the highest average WAR is prioritised and sufficient computational resources are available. It achieves an average WAR of 60.08\% with 90.17M parameters and an inference speed of 39 VPS on L20 GPU, but requires 280.00G FLOPs. SVFAP-B provides the most favourable overall balance between predictive performance and computation, achieving the highest average UAR and the second-highest average WAR with 78.71M parameters, 43.60G FLOPs, and an inference speed of 38 VPS on L20 GPU. FRU-Adapter-B provides only a relative performance gain of 0.65\% over SVFAP-B in average WAR while requiring 11.46M additional parameters and 236.40G additional FLOPs. MGNN (MFN) provides the smallest configuration at 7.56M parameters and requires only 37.11G FLOPs, making it the most economical DFER configuration when model size is the primary constraint. However, its inference speed is limited to 17 VPS. TimeSFormer-B achieves the highest measured inference speed of 46 VPS, but contains 122.35M parameters and obtains the lowest average UAR and WAR. It is therefore more suitable when processing throughput is prioritised over predictive performance and model size. MAE-DFER-B provides an intermediate configuration with 85.95M parameters, 49.75G FLOPs, and 25 VPS and is more appropriate when class-balanced performance on DFEW, particularly the recognition of surprise and disgust, is required. These results show that users should jointly consider the required expression categories, model size, flops, and inference speed when selecting a DFER model.

\begin{table}[htbp]
\Huge
\centering
\setlength{\doublerulesep}{0.3em}
\renewcommand{\arraystretch}{1.6}
\caption{Dataset-specific and average UAR (\%) $\uparrow$ and WAR (\%) $\uparrow$ of the DFER models on DFEW, FERV39k, and MAFW. The average results are computed across the three datasets. The best and second-best results are highlighted in bold and underlined, respectively.}
\resizebox{1.0\linewidth}{!}{
\begin{tabular}{l|c|cccccc}
\Xhline{5px}
\rowcolor{gray!15}
\cellcolor{gray!15} &
  \cellcolor{gray!15} &
  \multicolumn{2}{c}{\cellcolor{gray!15} \textbf{DFEW} \cite{jiang2020dfew}} &
  \multicolumn{2}{c}{\cellcolor{gray!15} \textbf{FERV39k} \cite{wang2022ferv39k}} &
  \multicolumn{2}{c}{\cellcolor{gray!15} \textbf{MAFW} \cite{liu2022mafw}} \\ \hhline{~~||--||--||--}
\rowcolor{gray!15}
\multirow{-2}{*}{\cellcolor{gray!15} \textbf{Methods}} &
    \multirow{-2}{*}{\cellcolor{gray!15}\textbf{Year}} &
  \cellcolor{gray!15}UAR &
  \cellcolor{gray!15}WAR &
  UAR &
  WAR &
  UAR &
  WAR \\ 
\hhline{========}
MGNN(MFN) \cite{wu2025depmgnn} & AAAI 2025 
& 63.29    
& 74.46 
& \cellcolor{gray!15}\underline{44.76} 
& \cellcolor{gray!15}\underline{53.10} 
& 32.69 
& 46.17 \\
\hhline{========}
TimeSFormer-B \cite{bertasius2021spacetime} & ICML 2021
& 61.49 
& 71.89 
& 43.29 
& 51.23
& 32.97 
& 42.09 \\
MAE-DFER-B \cite{sun2023mae} & MM 2023 
& \cellcolor{gray!30}\textbf{67.08} 
& 76.04 
& 43.91
& 52.10 
& \cellcolor{gray!15}\underline{38.32} 
& 48.12 \\
SVFAP-B \cite{sun2024svfap} & TAC 2024 
& \cellcolor{gray!15}\underline{66.06} 
& \cellcolor{gray!15}\underline{76.51}
& \cellcolor{gray!30}\textbf{46.05} 
& \cellcolor{gray!30}\textbf{54.10} 
& 37.47 
& \cellcolor{gray!15}\underline{48.45} \\ 
FRU-Adapter-B \cite{her2025fru} & Ele. 2025 
& 65.69 
& \cellcolor{gray!30}\textbf{77.45} 
& 44.45 
& 52.21 
& \cellcolor{gray!30}\textbf{38.71} 
& \cellcolor{gray!30}\textbf{50.57} \\ 
\Xhline{5px}
\end{tabular}
}
\label{tab:dfer_metric}
\end{table}

\begin{table*}[htbp]
\centering
\setlength{\doublerulesep}{0.3em}
\renewcommand{\arraystretch}{1.3}
\caption{Class-wise accuracy (\%) and overall UAR (\%) $\uparrow$ and WAR (\%) $\uparrow$ of the DFER models on DFEW.}
\resizebox{1.0\linewidth}{!}{
\begin{tabular}{l|c|ccccccccc}
\Xhline{1.2px}
\rowcolor{gray!15}
\cellcolor{gray!15} &
  \cellcolor{gray!15} &
  \multicolumn{7}{c}{\cellcolor{gray!15} \textbf{Accuracy of Each Emotion (\%)}} &
  \multicolumn{2}{c}{\cellcolor{gray!15} \textbf{Metric (\%)}} \\ \hhline{~~-------||--}
\rowcolor{gray!15} 
\multirow{-2}{*}{\cellcolor{gray!15}Methods} &
  \multirow{-2}{*}{\cellcolor{gray!15}Year} &
  Neutral &
  Happiness &
  Sadness &
  Surprise &
  Fear &
  Anger &
  Disgust &
  UAR &
  WAR \\ \hhline{===========}
MGNN(MFN) \cite{wu2025depmgnn}    & AAAI 2025 & 76.03 & 94.68 & 70.18 & 58.16 & 41.44 & 81.84 & 20.69 & 63.29 & 74.46 \\
TimeSFormer-B \cite{bertasius2021spacetime} & ICML 2021 & 71.91 & 92.23 & 69.66 & 59.18 & 40.88 & 75.86 & 20.69 & 61.49 & 71.89 \\
MAE-DFER-B \cite{sun2023mae}    & MM 2023   & 74.16 & 95.71 & 77.84 & 69.05 & 41.99 & 76.32 & 34.48 & 67.08 & 76.04 \\
SVFAP-B \cite{sun2024svfap}      & TAC 2024  & 78.84 & 95.09 & 73.88 & 68.71 & 44.75 & 77.01 & 24.14 & 66.06 & 76.51 \\
FRU-Adapter-B \cite{her2025fru} & Elc. 2025 & 74.84 & 96.00 & 81.42 & 66.97 & 50.00 & 77.46 & 18.18 & 65.69 & 77.45 \\ 
\Xhline{1.2px}
\end{tabular}
}
\label{tab:dfer_metric_dfew}
\end{table*}

\begin{table*}[htbp]
\centering
\setlength{\doublerulesep}{0.3em}
\renewcommand{\arraystretch}{1.3}
\caption{Class-wise accuracy (\%) and overall UAR (\%) $\uparrow$ and WAR (\%) $\uparrow$ of the DFER models on FERV39k.}
\resizebox{1.0\linewidth}{!}{
\begin{tabular}{l|c|ccccccccc}
\Xhline{1.2px}
\rowcolor{gray!15}
\cellcolor{gray!15} &
  \cellcolor{gray!15} &
  \multicolumn{7}{c}{\cellcolor{gray!15} \textbf{Accuracy of Each Emotion (\%)}} &
  \multicolumn{2}{c}{\cellcolor{gray!15} \textbf{Metric (\%)}} \\ \hhline{~~-------||--}
\rowcolor{gray!15} 
\multirow{-2}{*}{\cellcolor{gray!15}Methods} &
  \multirow{-2}{*}{\cellcolor{gray!15}Year} &
  Neutral &
  Happiness &
  Sadness &
  Surprise &
  Fear &
  Anger &
  Disgust &
  UAR &
  WAR \\ \hhline{===========}
MGNN(MFN) \cite{wu2025depmgnn}    & AAAI 2025 & 57.81 & 75.70 & 50.83 & 28.21 & 24.13 & 55.88 & 20.77 & 44.76 & 53.10 \\
TimeSFormer-B \cite{bertasius2021spacetime} & ICML 2021 & 60.21 & 73.93 & 51.90 & 32.29 & 19.03 & 42.57 & 23.13 & 43.29 & 51.23 \\
MAE-DFER-B \cite{sun2023mae}    & MM 2023   & 60.88 & 76.37 & 51.47 & 32.45 & 20.88 & 44.32 & 20.99 & 43.91 & 52.10 \\
SVFAP-B \cite{sun2024svfap}      & TAC 2024  & 62.26 & 75.90 & 52.98 & 35.27 & 23.67 & 49.56 & 22.70 & 46.05 & 54.10 \\
FRU-Adapter-B \cite{her2025fru} & Elc. 2025 & 58.58 & 73.52 & 58.51 & 34.17 & 21.58 & 42.97 & 21.84 & 44.45 & 52.21 \\ 
\Xhline{1.2px}
\end{tabular}
}
\label{tab:dfer_metric_ferv39k}
\end{table*}

\begin{table*}[htbp]
\centering
\large
\setlength{\doublerulesep}{0.3em}
\renewcommand{\arraystretch}{1.3}
\caption{Class-wise accuracy (\%) and overall UAR (\%) $\uparrow$ and WAR (\%) $\uparrow$ of the DFER models on MAFW. NE: Neutral, HA: Happiness, SA: Sadness, SU: Surprise, FE: Fear, AN: Anger, DI: Disgust, CO: Contempt, AX: Anxiety, HE: Helplessness, DA: Disappointment.}
\resizebox{1.0\linewidth}{!}{
\begin{tabular}{l|c|ccccccccccccc}
\Xhline{2.0px}
\rowcolor{gray!15}
\cellcolor{gray!15} &
  \cellcolor{gray!15} &
  \multicolumn{11}{c}{\cellcolor{gray!15} \textbf{Accuracy of Each Emotion (\%)}} &
  \multicolumn{2}{c}{\cellcolor{gray!15} \textbf{Metric (\%)}} \\ \hhline{~~-----------||--}
\rowcolor{gray!15} 
\multirow{-2}{*}{\cellcolor{gray!15}Methods} &
  \multirow{-2}{*}{\cellcolor{gray!15}Year} &
  NE &
  HA &
  SA &
  SU &
  FE &
  AN &
  DI &
  CO &
  AX &
  HL &
  DS &
  UAR &
  WAR \\ \hhline{===============}
MGNN(MFN) \cite{wu2025depmgnn}    & AAAI 2025 & 68.86 & 71.89 & 51.70 & 41.86 & 15.20 & 60.07 & 12.50 & 0.00 & 37.50 & 0.00 & 0.00 & 32.69 & 46.17 \\
TimeSFormer-B \cite{bertasius2021spacetime} & ICML 2021 & 45.18 & 74.70 & 44.22 & 38.60 & 18.40 & 48.56 & 14.84 & 10.42 & 42.93 & 11.32 & 13.51 & 32.97 & 42.09 \\
MAE-DFER-B \cite{sun2023mae}    & MM 2023   & 67.11 & 83.94 & 46.26 & 48.84 & 20.00 & 49.64 & 16.41 & 14.58 & 41.85 & 11.32 & 21.62 & 38.32 & 48.12 \\
SVFAP-B \cite{sun2024svfap}      & TAC 2024  & 66.23 & 80.72 & 48.98 & 50.23 & 13.60 & 52.88 & 18.75 & 6.25 & 45.11 & 13.21 & 16.22 & 37.47 & 48.45 \\
FRU-Adapter-B \cite{her2025fru} & Elc. 2025 & 64.91 & 81.12 & 64.29 & 48.84 & 27.20 & 46.40 & 18.75 & 14.58 & 46.74 & 7.55 & 5.41 & 38.71 & 50.57 \\ 
\Xhline{2.0px}
\end{tabular}
}
\label{tab:dfer_metric_mafw}
\end{table*}

\subsubsection{VAIE performances}

\begin{table*}[htbp]
\large
\centering
\setlength{\doublerulesep}{0.3em}
\renewcommand{\arraystretch}{1.5}
\caption{Dataset-specific concordance correlation coefficient (CCC) $\uparrow$ results of the VAIE models on AFFW-VA, AFEW-VA, SEWA, and AffectNet. Mean denotes the average of the valence and arousal CCC values. Static models process individual frames without temporal modelling, whereas temporal models exploit sequential facial cues and are therefore not evaluated on the image-based AffectNet dataset. The best and second-best results within each dataset are highlighted in bold and underlined, respectively.}
\resizebox{1.0\linewidth}{!}{
\begin{tabular}{cl|c|ccccccccc|ccc}
\Xhline{2.5px}
\rowcolor{gray!15}  
\cellcolor{gray!15}  &
  \cellcolor{gray!15}  &  \cellcolor{gray!15}  &
  \multicolumn{3}{c}{\cellcolor{gray!15} \textbf{AFFW-VA} \cite{kollias2018aff}} &
  \multicolumn{3}{c}{\cellcolor{gray!15} \textbf{AFEW-VA} \cite{kossaifi2017afew}} &
  \multicolumn{3}{c|}{\cellcolor{gray!15} \textbf{SEWA} \cite{kossaifi2019sewa}} &
  \multicolumn{3}{c}{\cellcolor{gray!15} \textbf{AffectNet} \cite{mollahosseini2017affectnet}} \\  \hhline{~~~---||---||---||---} 
\rowcolor{gray!15}  
\multirow{-2}{*}{\cellcolor{gray!15} } &
  \multirow{-2}{*}{\cellcolor{gray!15} \textbf{Methods}} &
  \multirow{-2}{*}{\cellcolor{gray!15}\textbf{Year}} &
  Valence &
  Arousal &
  Mean &
  \cellcolor{gray!15} Valence &
  \cellcolor{gray!15} Arousal &
  Mean &
  \cellcolor{gray!15} Valence &
  \cellcolor{gray!15} Arousal &
  Mean &
  \cellcolor{gray!15} Valence &
  \cellcolor{gray!15} Arousal &
  Mean \\  \hhline{===============} 
  & DDAMFN \cite{zhang2023dual} & Ele. 2023 
  & 0.401 
  & 0.583 
  & 0.492 
  & 0.574 
  & 0.523 
  & 0.548 
  & \cellcolor{gray!15}\underline{0.814} 
  & 0.681 
  & \cellcolor{gray!15}\underline{0.747} 
  & \cellcolor{gray!15}\underline{0.615} 
  & 0.524 
  & \cellcolor{gray!15}\underline{0.569} \\
  & MAE-Face \cite{ma2024facial} & TAC 2024 
  & 0.392 
  & 0.546 
  & 0.469 
  & 0.665 
  & \cellcolor{gray!30}\textbf{0.626} 
  & \cellcolor{gray!15}\underline{0.646} 
  & 0.805 
  & 0.665 
  & 0.735 
  & \cellcolor{gray!30}\textbf{0.636} 
  & \cellcolor{gray!30}\textbf{0.566} 
  & \cellcolor{gray!30}\textbf{0.601} \\
\multirow{-2.0}{*}{\rotatebox[origin=c]{90}{Static}}   
  & FMAE-B \cite{ning2024representation} & FG 2025 
  & 0.399 
  & 0.578 
  & 0.488 
  & 0.579 
  & \cellcolor{gray!15}\underline{0.617} 
  & 0.598 
  & 0.800 
  & 0.681 
  & 0.741 
  & 0.584 
  & 0.522 
  & 0.553 \\
  & FMAE-L \cite{ning2024representation} & FG 2025 
  & 0.382 
  & 0.569 
  & 0.475 
  & 0.591 
  & 0.590 
  & 0.591 
  & \cellcolor{gray!30}\textbf{0.824} 
  & \cellcolor{gray!30}\textbf{0.691} 
  & \cellcolor{gray!30}\textbf{0.757} 
  & 0.580 
  & \cellcolor{gray!15}\underline{0.550} 
  & 0.565 \\
  \hhline{===============} 
  & MAVENC(MFN) \cite{dresvyanskiy2024multi}    & CVPRW 2024 
  & \cellcolor{gray!30}\textbf{0.498} 
  & \cellcolor{gray!15}\underline{0.620} 
  & \cellcolor{gray!30}\textbf{0.559} 
  & 0.663 
  & 0.493 
  & 0.578 
  & 0.782 
  & 0.660 
  & 0.721 
  & - & - & - \\
\multirow{-0.2}{*}{\rotatebox[origin=c]{90}{Temporal}} 
  & MAVENC(Res50) \cite{dresvyanskiy2024multi}    & CVPRW 2024 
  & 0.372 
  & 0.547 
  & 0.459 
  & 0.708 
  & 0.543 
  & 0.626 
  & 0.741 
  & 0.646 
  & 0.694 
  & - & - & - \\
  & MAVENC(ViT-B) \cite{dresvyanskiy2024multi}    & CVPRW 2024 
  & \cellcolor{gray!15}\underline{0.470} 
  & \cellcolor{gray!30}\textbf{0.630} 
  & \cellcolor{gray!15}\underline{0.550} 
  & 0.603 
  & 0.560 
  & 0.581 
  & 0.756 
  & 0.648 
  & 0.702 
  & - & - & - \\
  & MAE-DFER-B \cite{sun2023mae}  & MM 2023 
  & 0.406 
  & 0.566 
  & 0.486 
  & \cellcolor{gray!15}\underline{0.729} 
  & 0.496 
  & 0.612 
  & 0.794 
  & 0.651 
  & 0.722 
  & - & - & - \\
  & SVFAP-B \cite{sun2024svfap}   & TAC 2024 
  & 0.382 
  & 0.550 
  & 0.466 
  & 0.694
  & 0.520 
  & 0.607 
  & 0.802 
  & \cellcolor{gray!15}\underline{0.684} 
  & 0.743 
  & - & - & - \\
  & FRU-Adapter-B \cite{her2025fru}   & Ele. 2025 
  & 0.421 
  & 0.561 
  & 0.491 
  & \cellcolor{gray!30}\textbf{0.759} 
  & 0.615 
  & \cellcolor{gray!30}\textbf{0.687} 
  & 0.802
  & 0.665 
  & 0.733 
  & - & - & - \\
\Xhline{2.5px}
\end{tabular}
}
\label{tab:va_metric}
\end{table*}

\begin{table*}[htbp]
\large
\centering
\setlength{\doublerulesep}{0.3em}
\renewcommand{\arraystretch}{1.5}
\caption{Dataset-specific mean squared error (MSE) $\downarrow$ results of the VAIE models on AFFW-VA, AFEW-VA, SEWA, and AffectNet. Mean denotes the average of the valence and arousal MSE values. Static models process individual frames without temporal modelling, whereas temporal models exploit sequential facial cues and are therefore not evaluated on the image-based AffectNet dataset. The best and second-best results within each dataset are highlighted in bold and underlined, respectively.}
\resizebox{1.0\linewidth}{!}{
\begin{tabular}{cl|c|ccccccccc|ccc}
\Xhline{2.5px}
\rowcolor{gray!15}  
\cellcolor{gray!15}  &
  \cellcolor{gray!15}  &  \cellcolor{gray!15}  &
  \multicolumn{3}{c}{\cellcolor{gray!15} \textbf{AFFW-VA} \cite{kollias2018aff}} &
  \multicolumn{3}{c}{\cellcolor{gray!15} \textbf{AFEW-VA} \cite{kossaifi2017afew}} &
  \multicolumn{3}{c|}{\cellcolor{gray!15} \textbf{SEWA} \cite{kossaifi2019sewa}} &
  \multicolumn{3}{c}{\cellcolor{gray!15} \textbf{AffectNet} \cite{mollahosseini2017affectnet}} \\  \hhline{~~~---||---||---||---} 
\rowcolor{gray!15}  
\multirow{-2}{*}{\cellcolor{gray!15} } &
  \multirow{-2}{*}{\cellcolor{gray!15} \textbf{Methods}} &
  \multirow{-2}{*}{\cellcolor{gray!15}\textbf{Year}} &
  Valence &
  Arousal &
  Mean &
  \cellcolor{gray!15} Valence &
  \cellcolor{gray!15} Arousal &
  Mean &
  \cellcolor{gray!15} Valence &
  \cellcolor{gray!15} Arousal &
  Mean &
  \cellcolor{gray!15} Valence &
  \cellcolor{gray!15} Arousal &
  Mean \\  \hhline{===============} 
  & DDAMFN \cite{zhang2023dual} & Ele. 2023 
  & \cellcolor{gray!30}\textbf{0.143} 
  & \cellcolor{gray!15}\underline{0.054} 
  & \cellcolor{gray!30}\textbf{0.099} 
  & 0.070 
  & 0.071 
  & 0.071 
  & 0.075 
  & 0.074 
  & 0.075 
  & \cellcolor{gray!15}\underline{0.175} 
  & 0.132 
  & \cellcolor{gray!15}\underline{0.154} \\
  & MAE-Face \cite{ma2024facial} & TAC 2024 
  & 0.149 
  & 0.057 
  & \cellcolor{gray!15}\underline{0.103} 
  & 0.051 
  & \cellcolor{gray!30}\textbf{0.062} 
  & \cellcolor{gray!15}\underline{0.056} 
  & 0.077 
  & 0.078 
  & 0.078 
  & \cellcolor{gray!30}\textbf{0.166} 
  & \cellcolor{gray!30}\textbf{0.125} 
  & \cellcolor{gray!30}\textbf{0.145} \\
\multirow{-2.0}{*}{\rotatebox[origin=c]{90}{Static}}   
  & FMAE-B \cite{ning2024representation} & FG 2025 
  & 0.169 
  & 0.061 
  & 0.115 
  & 0.082 
  & 0.066 
  & 0.074 
  & 0.073 
  & 0.073 
  & 0.073 
  & 0.193 
  & 0.140 
  & 0.167 \\
  & FMAE-L \cite{ning2024representation} & FG 2025 
  & 0.161 
  & 0.059 
  & 0.110 
  & 0.071 
  & 0.067 
  & 0.069 
  & \cellcolor{gray!15}\underline{0.071} 
  & \cellcolor{gray!30}\textbf{0.071} 
  & \cellcolor{gray!30}\textbf{0.071} 
  & 0.198 
  & \cellcolor{gray!15}\underline{0.129} 
  & 0.163 \\
  \hhline{===============} 
  & MAVENC(MFN) \cite{dresvyanskiy2024multi}    & CVPRW 2024 
  & \cellcolor{gray!15}\underline{0.147} 
  & 0.063 
  & 0.105 
  & 0.056 
  & 0.082 
  & 0.069 
  & 0.093 
  & 0.086 
  & 0.090 
  & - & - & - \\
\multirow{-0.2}{*}{\rotatebox[origin=c]{90}{Temporal}} 
  & MAVENC(Res50) \cite{dresvyanskiy2024multi}    & CVPRW 2024 
  & 0.196 
  & 0.080 
  & 0.138 
  & \cellcolor{gray!15}\underline{0.045} 
  & 0.075 
  & 0.060 
  & 0.120 
  & 0.094 
  & 0.107 
  & - & - & - \\
  & MAVENC(ViT-B) \cite{dresvyanskiy2024multi}    & CVPRW 2024 
  & 0.164 
  & 0.065 
  & 0.114 
  & 0.064 
  & 0.064 
  & 0.064 
  & 0.109 
  & 0.083 
  & 0.096 
  & - & - & - \\
  & MAE-DFER-B \cite{sun2023mae}  & MM 2023 
  & 0.156 
  & 0.064 
  & 0.110 
  & 0.046 
  & 0.087 
  & 0.067 
  & \cellcolor{gray!30}\textbf{0.068} 
  & 0.081 
  & 0.075 
  & - & - & - \\
  & SVFAP-B \cite{sun2024svfap}   & TAC 2024 
  & 0.161 
  & 0.070 
  & 0.115 
  & 0.049
  & 0.079 
  & 0.064 
  & 0.073 
  & \cellcolor{gray!15}\underline{0.071} 
  & \cellcolor{gray!15}\underline{0.072} 
  & - & - & - \\
  & FRU-Adapter-B \cite{her2025fru}   & Ele. 2025 
  & 0.152 
  & \cellcolor{gray!30}\textbf{0.054} 
  & 0.103 
  & \cellcolor{gray!30}\textbf{0.044} 
  & \cellcolor{gray!15}\underline{0.063} 
  & \cellcolor{gray!30}\textbf{0.053} 
  & 0.079
  & 0.074 
  & 0.077 
  & - & - & - \\
\Xhline{2.5px}
\end{tabular}
}
\label{tab:va_mse}
\end{table*}

\begin{figure*}[!h]
    \centering
    \begin{subfigure}[b]{\linewidth}
        \centering
        \includegraphics[width=\linewidth]{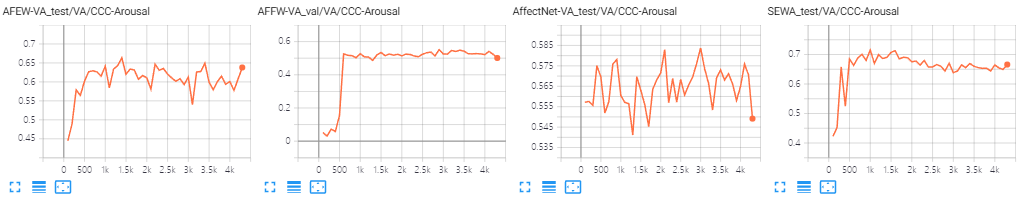}
        \caption{Arousal}
        \label{fig:vaie_maeface_sub1}
    \end{subfigure}
    
    \vspace{1em}  
    
    \begin{subfigure}[b]{\linewidth}
        \centering
        \includegraphics[width=\linewidth]{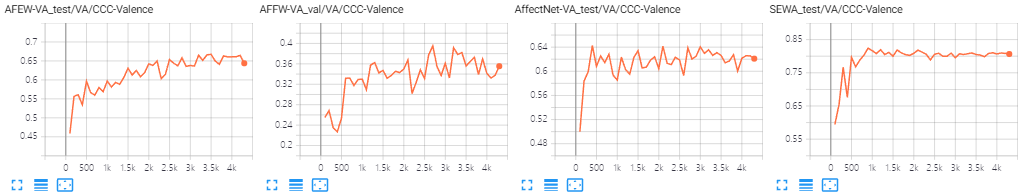}
        \caption{Valecne}
        \label{fig:vaie_maeface_sub2}
    \end{subfigure}
    
    \caption{Validation CCC trajectories of MAE-Face during training on AFEW-VA, AFFW-VA, AffectNet, and SEWA. The top and bottom rows present the arousal and valence results, respectively, with the datasets arranged from left to right in the same order.}
    \label{fig:vaie_maeface}
\end{figure*}

\begin{figure*}[!h]
    \centering
    \begin{subfigure}[b]{\linewidth}
        \centering
        \includegraphics[width=\linewidth]{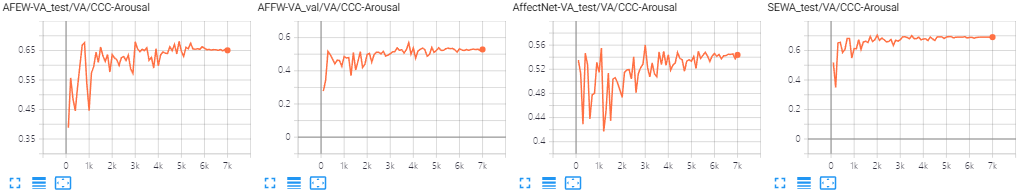}
        \caption{Arousal}
        \label{fig:vaie_fmae_sub1}
    \end{subfigure}
    
    \vspace{1em}  
    
    \begin{subfigure}[b]{\linewidth}
        \centering
        \includegraphics[width=\linewidth]{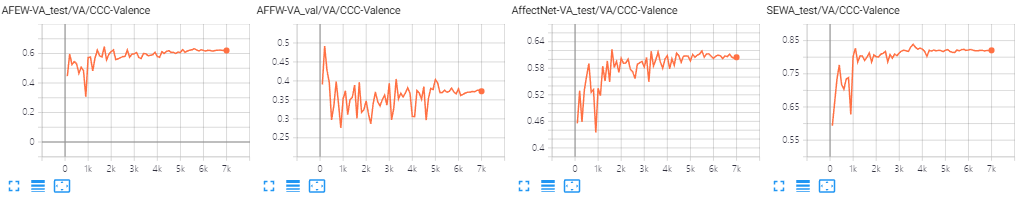}
        \caption{Valence}
        \label{fig:vaie_fmae_sub2}
    \end{subfigure}
    
    \caption{Validation CCC trajectories of FMAE-L during training on AFEW-VA, AFFW-VA, AffectNet, and SEWA. The top and bottom rows present the arousal and valence results, respectively, with the datasets arranged from left to right in the same order.}
    \label{fig:vaie_fmae-vit-l}
\end{figure*}

\noindent\textbf{Overall and dataset-specific performance.} Table \ref{tab:va_metric} and \ref{tab:va_mse} report the valence, arousal, and mean CCC and MSE results of the ten VAIE models on AFFW-VA, AFEW-VA, SEWA, and AffectNet. Since the temporal models are not evaluated on the image-based AffectNet dataset, we separately analyse the static models with complete four-dataset coverage and the temporal models evaluated on the three video datasets.

Among the static models, MAE-Face achieves the highest four-dataset average CCC of 0.613, representing a relative performance gain of 2.64\% over the second-best FMAE-L. DDAMFN achieves a competitive average CCC of 0.589 using only 4.57M parameters, approximately 1/20 of the parameters required by MAE-Face. Moreover, increasing the number of parameters from FMAE-B to FMAE-L provides only a relative performance gain of 0.34\% in average CCC, despite increasing the parameter count from 86.99M to 305.41M. These results further support the observation in Sec. 4.3 of the main paper that VAIE models exhibit only modest performance gains from increased parameters.

Among the temporal models, FRU-Adapter-B obtains the highest three-video-dataset average CCC of 0.637, representing a relative performance gain of 2.85\% over MAVENC (MFN). However, the dataset-specific results show that no single model consistently performs best across datasets or affective dimensions. MAVENC (MFN) achieves the highest mean CCC of 0.559 on AFFW-VA, with MAVENC (MFN) and MAVENC (ViT-B) leading on valence and arousal, respectively. FRU-Adapter-B performs best on AFEW-VA with a mean CCC of 0.687 and the highest valence CCC of 0.759, whereas MAE-Face obtains the highest arousal CCC of 0.626. On SEWA, FMAE-L performs best for both valence and arousal, reaching a mean CCC of 0.757. MAE-Face similarly leads on both affective dimensions of AffectNet and achieves a mean CCC of 0.601. These dataset- and dimension-dependent results indicate that the benefit of increasing model parameters is determined by both the target data distribution and the affective dimension being estimated.

\noindent\textbf{Valence-arousal convergence analysis.} Figs. \ref{fig:vaie_maeface} and \ref{fig:vaie_fmae-vit-l} further examine this behaviour by presenting the validation CCC trajectories of MAE-Face and FMAE-L during training. The curves first demonstrate that valence and arousal are not learned at the same rate. For MAE-Face, arousal on AFEW-VA and AFFW-VA reaches a comparatively stable level during the early training stages, while valence continues to improve or fluctuate over a longer period. On SEWA, arousal reaches its highest validation performance early and subsequently decreases, whereas valence remains comparatively stable after its initial optimisation. This asynchronous convergence means that continued optimisation may improve one affective dimension while simultaneously degrading the other, preventing a single checkpoint from preserving the best valence and arousal performance. 

The effect becomes more pronounced for the substantially larger FMAE-L. On AFEW-VA, the valence curve exhibits a considerable early peak-to-later performance reversal before gradually stabilising. On AFFW-VA, valence continues to fluctuate after arousal has reached a comparatively stable range, while both AffectNet curves show more pronounced early-stage fluctuations than those of MAE-Face. These behaviours are consistent with the lower average performance of FMAE-L on AFEW-VA and AffectNet. In contrast, FMAE-L produces comparatively stable later-stage trajectories on SEWA and achieves the best final results on this dataset, showing that the influence of increased parameters remains dataset-dependent. Together with the non-monotonic parameter-performance relationship in Table \ref{tab:va_metric}, these curves provide training-process evidence that models with more parameters are more susceptible to dimension-dependent overfitting: one affective dimension may begin to degrade before the other is adequately fit, and increasing model parameters therefore does not necessarily improve the overall VAIE performance.

\noindent\textbf{Application-oriented model selection.} 
The dataset-specific results and computational measures in Table \ref{tab:model_efficiency} provide complementary guidance for selecting VAIE models. MAE-Face offers the most favourable general-purpose configuration in image/frame-level VAIE models, combining the highest four-dataset average CCC with 86.99M parameters, 16.87G FLOPs, and an inference speed of 131 FPS on L20 GPU. DDAMFN is preferable when model size and computation are the primary constraints. It requires only 4.57M parameters and 0.55G FLOPs while retaining an average CCC of 0.589. FMAE-L is more appropriate when performance on SEWA is prioritised and sufficient computational resources are available, but it requires 305.41M parameters and 59.70G FLOPs and reaches 71 FPS. FMAE-B shares the 131 FPS inference speed and 16.87G FLOPs of MAE-Face and can be considered when its stronger SEWA arousal performance is required.

For temporal VAIE models, FRU-Adapter-B is preferable when the highest AFEW-VA performance is prioritised, although its 280.00G FLOPs impose a substantially higher computational cost. MAE-DFER-B provides a less computationally demanding alternative for AFEW-VA valence estimation, reaching a valence CCC of 0.729 with 49.76G FLOPs. MAVENC (MFN) provides the most favourable configuration for high-throughput AFFW-VA processing, achieving the best mean CCC on this dataset with 23.47M parameters and an inference speed of 3,135 VPS. MAVENC (ViT-B) can instead be selected when AFFW-VA arousal estimation is prioritised, although it requires 345.99G FLOPs. SVFAP-B provides a more balanced temporal configuration for SEWA, achieving a mean CCC of 0.743 with 78.16M parameters, 43.61G FLOPs, and 602 VPS. Therefore, VAIE model selection should jointly consider the performance of valence and arousal, model size, flops, and inference speed.

\subsubsection{FPLE performances}

\begin{table}[htbp]
\Huge
\centering
\setlength{\doublerulesep}{0.3em}
\renewcommand{\arraystretch}{1.6}
\caption{Class-wise and mean accuracy and F1 scores (\%) for binary pain classification on BioVid. Pain levels 0 and 4 are treated as the No Pain and Pain classes, respectively. Mean denotes the macro-average across the two classes. The best and second-best results are highlighted in bold and underlined, respectively.}
\resizebox{1.0\linewidth}{!}{
\begin{tabular}{l|c|cccc|cc}
\Xhline{4.5px}
\rowcolor{gray!15} 
\cellcolor{gray!15} & \cellcolor{gray!15} &
  \multicolumn{2}{c}{\cellcolor{gray!15}\textbf{No Pain}} &
  \multicolumn{2}{c|}{\cellcolor{gray!15}\textbf{Pain}} &
  \multicolumn{2}{c}{\cellcolor{gray!15}\textbf{Mean}} \\ \hhline{~~--||--||--} 
\rowcolor{gray!15} 
\multirow{-2}{*}{\cellcolor{gray!15}\textbf{Methods}} &
\multirow{-2}{*}{\cellcolor{gray!15}\textbf{Year}} &
  Acc.↑ &
  F1↑ &
  Acc.↑ &
  F1↑ &
  Acc.↑ &
  F1↑ \\ \hhline{========}
MGNN(MFN) \cite{wu2025depmgnn} &  AAAI 2025
& \cellcolor{gray!30}\textbf{89.26} 
& 77.28 
& 58.24 
& 68.93 
& 73.75 
& 73.10 \\
TimeSFormer-B \cite{bertasius2021spacetime} & ICML 2021 
& 84.26 
& \cellcolor{gray!30}\textbf{77.48} 
& \cellcolor{gray!15}\underline{66.76} 
& \cellcolor{gray!15}\underline{73.17} 
& \cellcolor{gray!15}\underline{75.51} 
& \cellcolor{gray!15}\underline{75.33} \\
MAE-DFER-B \cite{sun2023mae} & MM 2023  
& \cellcolor{gray!15}\underline{86.03} 
& 77.23 
& 63.24 
& 71.37 
& 74.63 
& 74.30 \\
SVFAP-B \cite{sun2024svfap} &   TAC 2024   
& 81.18 
& \cellcolor{gray!15}\underline{77.47} 
& \cellcolor{gray!30}\textbf{71.62} 
& \cellcolor{gray!30}\textbf{75.21} 
& \cellcolor{gray!30}\textbf{76.40} 
& \cellcolor{gray!30}\textbf{76.34} \\
\Xhline{4.5px}
\end{tabular}
}
\label{tab:pain_biovid}
\end{table}

\begin{table}[htbp]
\Huge
\centering
\setlength{\doublerulesep}{0.3em}
\renewcommand{\arraystretch}{1.6}
\caption{Class-wise and mean accuracy and F1 scores (\%) for binary pain classification on MIntPAIN. Pain levels 0 and 4 are treated as the No Pain and Pain classes, respectively. Mean denotes the macro-average across the two classes. The best and second-best results are highlighted in bold and underlined, respectively.}
\resizebox{1.0\linewidth}{!}{
\begin{tabular}{l|c|cccc|cc}
\Xhline{4.5px}
\rowcolor{gray!15} 
\cellcolor{gray!15} & \cellcolor{gray!15} &
  \multicolumn{2}{c}{\cellcolor{gray!15}\textbf{No Pain}} &
  \multicolumn{2}{c|}{\cellcolor{gray!15}\textbf{Pain}} &
  \multicolumn{2}{c}{\cellcolor{gray!15}\textbf{Mean}} \\ \hhline{~~--||--||--} 
\rowcolor{gray!15} 
\multirow{-2}{*}{\cellcolor{gray!15}\textbf{Methods}} &
\multirow{-2}{*}{\cellcolor{gray!15}\textbf{Year}} &
  Acc.↑ &
  F1↑ &
  Acc.↑ &
  F1↑ &
  Acc.↑ &
  F1↑ \\ \hhline{========}
MGNN(MFN) \cite{wu2025depmgnn} & AAAI 2025  
& \cellcolor{gray!30}\textbf{100.00}   
& 89.24    
& 2.56     
& 5.00     
& 51.28    
& 47.12    \\
TimeSFormer-B \cite{bertasius2021spacetime} & ICML 2021 
& \cellcolor{gray!15}\underline{99.68}    
& 90.10    
& 12.82    
& 22.47    
& 56.25   
& 56.29    \\
MAE-DFER-B \cite{sun2023mae} & MM 2023  
& 99.37    
& \cellcolor{gray!30}\textbf{91.65}    
& \cellcolor{gray!30}\textbf{29.49}    
& \cellcolor{gray!30}\textbf{44.66}    
& \cellcolor{gray!30}\textbf{64.43}    
& \cellcolor{gray!30}\textbf{68.16}    \\
SVFAP-B \cite{sun2024svfap} &   TAC 2024   
& 99.05    
& \cellcolor{gray!15}\underline{90.57}    
& \cellcolor{gray!15}\underline{20.51}    
& \cellcolor{gray!15}\underline{32.99}    
& \cellcolor{gray!15}\underline{59.78}    
& \cellcolor{gray!15}\underline{61.78}    \\
\Xhline{4.5px}
\end{tabular}
}
\label{tab:pain_mintpain}
\end{table}

\begin{table}[htbp]
\Huge
\centering
\setlength{\doublerulesep}{0.3em}
\renewcommand{\arraystretch}{1.6}
\caption{Five-level pain intensity regression results on BioVid and MIntPAIN using pain levels 0-4. We report the Pearson Correlation Coefficient (PCC) $\uparrow$, Concordance Correlation Coefficient (CCC) $\uparrow$, and Root Mean Square Error (RMSE) $\downarrow$. The best and second-best results are highlighted in bold and underlined, respectively.}
\resizebox{1.0\linewidth}{!}{
\begin{tabular}{l|c|cccccc}
\Xhline{4.5px}
\rowcolor{gray!15}
\cellcolor{gray!15} & \cellcolor{gray!15} &
  \multicolumn{3}{c}{\cellcolor{gray!15}\textbf{BioVid} \cite{walter2013biovid}} &
  \multicolumn{3}{c}{\cellcolor{gray!15}\textbf{MIntPAIN} \cite{haque2018deep}} \\ \hhline{~~---||---} 
\rowcolor{gray!15}
\multirow{-2}{*}{\cellcolor{gray!15}\textbf{Methods}} &
\multirow{-2}{*}{\cellcolor{gray!15}\textbf{Year}} &
  PCC↑ &
  CCC↑ &
  RMSE↓ &
  PCC↑ &
  CCC↑ &
  RMSE↓ \\ \hhline{========}
MGNN(MFN) \cite{wu2025depmgnn} & AAAI 2025    
& 0.459 
& 0.452 
& \cellcolor{gray!30}\textbf{1.367} 
& 0.153 
& 0.034 
& 1.537 \\
TimeSFormer-B \cite{bertasius2021spacetime} & ICML 2021 
& 0.422 
& 0.420 
& 1.466 
& 0.247 
& 0.104 
& 1.551 \\
MAE-DFER-B \cite{sun2023mae} &  MM 2023  
& \cellcolor{gray!30}\textbf{0.464} 
& \cellcolor{gray!15}\underline{0.458} 
& 1.412 
& \cellcolor{gray!15}\underline{0.365} 
& \cellcolor{gray!15}\underline{0.259} 
& \cellcolor{gray!15}\underline{1.499} \\
SVFAP-B \cite{sun2024svfap} &  TAC 2024   
& \cellcolor{gray!15}\underline{0.463} 
& \cellcolor{gray!30}\textbf{0.461} 
& \cellcolor{gray!15}\underline{1.408} 
& \cellcolor{gray!30}\textbf{0.393} 
& \cellcolor{gray!30}\textbf{0.372} 
& \cellcolor{gray!30}\textbf{1.448} \\ 
\Xhline{4.5px}
\end{tabular}
}
\label{tab:pain_reg}
\end{table}

\noindent\textbf{Binary pain classification.} 
Tables \ref{tab:pain_biovid} and \ref{tab:pain_mintpain} report the class-wise and mean accuracy and F1 scores of the four FPLE models for binary pain classification, where pain levels 0 and 4 are treated as the No Pain and Pain classes, respectively. For the average performance of BioVid and MIntPAIN, MAE-DFER-B achieves the highest accuracy of 69.53\% and the highest mean F1 score of 71.23\%, representing relative performance gains of 2.12\% and 3.14\% over the second-best SVFAP-B, respectively. These results further support the broadly positive parameter-performance relationship for FPLE discussed in Sec. 4.3 of the main paper.

On BioVid, SVFAP-B achieves the highest mean accuracy of 76.40\% and F1 score of 76.34\%. Its advantage mainly arises from the Pain class, for which it achieves an accuracy of 71.62\% and an F1 score of 75.21\%. However, on MIntPAIN dataset, MAE-DFER-B performs the best result, reaching a mean accuracy of 64.43\% and an F1 score of 68.16\%. All four models achieve No Pain accuracies above 99\% on MIntPAIN, whereas their Pain accuracies range from only 2.56\% to 29.49\%. MAE-DFER-B provides the strongest recognition of the Pain class, achieving an accuracy of 29.49\% and an F1 score of 44.66\%, which explains its clear advantage in the mean results. These results show that increasing model parameters does not uniformly improve binary pain recognition: TimeSFormer-B contains the most parameters but does not lead on either dataset, while SVFAP-B outperforms MAE-DFER-B on BioVid despite containing fewer parameters.

\noindent\textbf{Five-level pain intensity regression.}
Table \ref{tab:pain_reg} further evaluates the four models on the additional five-level pain intensity regression setting using all pain levels from 0 to 4. On BioVid, no single model performs best across all three metrics. MAE-DFER-B achieves the highest PCC of 0.464, SVFAP-B obtains the highest CCC of 0.461, and MGNN (MFN) provides the lowest RMSE of 1.367. The small differences between MAE-DFER-B and SVFAP-B in PCC and CCC indicate comparable correlation-based performance, whereas the lower RMSE of MGNN (MFN) shows that the preferred model also depends on the evaluation metric.

MIntPAIN is more challenging for five-level regression, as all models produce lower PCC and CCC values and higher RMSE values than on BioVid. SVFAP-B nevertheless performs best across all three metrics, achieving a PCC of 0.393, a CCC of 0.372, and an RMSE of 1.448. Compared with MAE-DFER-B, SVFAP-B provides relative performance gains of 7.67\% and 43.63\% in PCC and CCC, respectively. Averaged across the two datasets, SVFAP-B also obtains the highest PCC of 0.428 and CCC of 0.417 and the lowest RMSE of 1.428. Therefore, while MAE-DFER-B is preferable for binary pain classification, SVFAP-B provides the most consistent overall performance for five-level pain intensity regression.

\noindent\textbf{Application-oriented model selection.}
The classification and regression results above, together with the computational measures in Table \ref{tab:model_efficiency}, provide complementary guidance for selecting FPLE models. MAE-DFER-B is preferable when binary pain recognition is prioritised, particularly for MIntPAIN, where recognising the Pain class remains difficult. It achieves the highest two-dataset average accuracy and F1 score with 85.93M parameters, 49.75G FLOPs, and an inference speed of 25 VPS on L20 GPU. SVFAP-B provides the most favourable balance when both binary classification and five-level regression are required. It performs best for binary classification on BioVid and achieves the strongest overall regression results while requiring 78.69M parameters and 43.60G FLOPs and reaching 38 VPS. MGNN (MFN) provides the smallest and least computationally demanding configuration, requiring only 7.53M parameters and 10.49G FLOPs while reaching 43 VPS. It also achieves the lowest regression RMSE on BioVid, making it suitable when model size or computation is the primary constraint. These results show that FPLE model selection should jointly consider the binary or five-level prediction setting, evaluation metric, model size, flops, and inference speed.

\subsubsection{FDA performances}

\begin{table}[htbp]
\centering
\Huge
\setlength{\doublerulesep}{0.3em}
\renewcommand{\arraystretch}{1.3}
\caption{FDA Results on AVEC 2014 Dataset.}
\resizebox{1.0\linewidth}{!}{
\begin{tabular}{l|c|cccc}
\Xhline{1.0px}
\rowcolor{gray!15} 
\cellcolor{gray!15} & \cellcolor{gray!15} & 
\multicolumn{4}{c}{\cellcolor{gray!15} \textbf{AVEC 2014}} \\ \hhline{~~----} 
\rowcolor{gray!15} 
\multirow{-2}{*}{\cellcolor{gray!15}\textbf{Methods}} & 
\multirow{-2}{*}{\cellcolor{gray!15}\textbf{Year}} &
RMSE↓        & MAE↓         & PCC↑         & CCC↑        \\ \hhline{======}
MGNN(MFN) \cite{wu2025depmgnn} & AAAI 2025 
& \cellcolor{gray!15}\underline{6.936} 
& \cellcolor{gray!30}\textbf{5.279} 
& \cellcolor{gray!15}\underline{0.797} 
& \cellcolor{gray!15}\underline{0.778}       
\\
MGNN(Res50) \cite{wu2025depmgnn} & AAAI 2025 
& \cellcolor{gray!30}\textbf{6.623} 
& \cellcolor{gray!15}\underline{5.448} 
& \cellcolor{gray!30}\textbf{0.817}
& \cellcolor{gray!30}\textbf{0.799}       
\\
MAE-DFER-B \cite{sun2023mae} & MM 2023 
& 7.533 
& 5.678 
& 0.760 
& 0.739       
\\
SVFAP-B \cite{sun2024svfap} & TAC 2024
& 7.967 
& 5.933 
& 0.722 
& 0.695       
\\ 
\Xhline{4.5px}
\end{tabular}
}
\label{tab:dep_avec2014}
\end{table}

\begin{table*}[htbp]
\Huge
\centering
\setlength{\doublerulesep}{0.3em}
\renewcommand{\arraystretch}{1.5}
\caption{Comprehensive Results of FAPR on the ChaLearn First Impression Dataset \cite{ponce2016chalearn}. We report the Pearson Correlation Coefficient (PCC), the  Concordance Correlation Coefficient (CCC), and the Mean Squared Error (MSE). Results for methods with $*$ are taken directly from their original publications. Top-performing results are highlighted in bold, while second-best outcomes are indicated with underlining.}
\resizebox{1.0\linewidth}{!}{
\begin{tabular}{l|c|ccccccccccccccc|ccc}
\Xhline{4.5px}
\rowcolor{gray!15} 
\cellcolor{gray!15} & \cellcolor{gray!15} &
  \multicolumn{3}{c}{\cellcolor{gray!15}\textbf{Openness}} &
  \multicolumn{3}{c}{\cellcolor{gray!15}\textbf{Conscientiousness}} &
  \multicolumn{3}{c}{\cellcolor{gray!15}\textbf{Extraversion}} &
  \multicolumn{3}{c}{\cellcolor{gray!15}\textbf{Agreeableness}} &
  \multicolumn{3}{c|}{\cellcolor{gray!15}\textbf{Neuroticism}} &
  \multicolumn{3}{c}{\cellcolor{gray!15}\textbf{Mean}} \\ \hhline{~~---||---||---||---||---||---} 
\rowcolor{gray!15} 
\multirow{-2}{*}{\cellcolor{gray!15}\textbf{Methods}} &
\multirow{-2}{*}{\cellcolor{gray!15}\textbf{Year}} &
  PCC↑ &
  CCC↑ &
  MSE↓ &
  PCC↑ &
  CCC↑ &
  MSE↓ &
  PCC↑ &
  CCC↑ &
  MSE↓ &
  PCC↑ &
  CCC↑ &
  MSE↓ &
  PCC↑ &
  CCC↑ &
  MSE↓ &
  PCC↑ &
  CCC↑ &
  MSE↓ \\ \hhline{====================}
MGNN(MFN) \cite{wu2025depmgnn} &
  AAAI 2025 &
  \cellcolor{gray!15}\underline{0.705} &
  \cellcolor{gray!15}\underline{0.705} &
  \cellcolor{gray!30}\textbf{0.012} &
  \cellcolor{gray!30}\textbf{0.780} &
  \cellcolor{gray!30}\textbf{0.779} &
  \cellcolor{gray!30}\textbf{0.010} &
  \cellcolor{gray!30}\textbf{0.782} &
  \cellcolor{gray!30}\textbf{0.782} &
  \cellcolor{gray!30}\textbf{0.010} &
  \cellcolor{gray!30}\textbf{0.635} &
  \cellcolor{gray!30}\textbf{0.634} &
  \cellcolor{gray!30}\textbf{0.013} &
  \cellcolor{gray!30}\textbf{0.738} &
  \cellcolor{gray!30}\textbf{0.737} &
  \cellcolor{gray!30}\textbf{0.012} &
  \cellcolor{gray!30}\textbf{0.728} &
  \cellcolor{gray!30}\textbf{0.728} &
  \cellcolor{gray!30}\textbf{0.011} \\
\color{gray!75}MGNN(Res50)* \cite{wu2025depmgnn}&
  AAAI 2025 &
  \color{gray!75}- &
  \color{gray!75}0.683 &
  \color{gray!75}- &
  \color{gray!75}- &
  \color{gray!75}0.758 &
  \color{gray!75}- &
  \color{gray!75}- &
  \color{gray!75}0.759 &
  \color{gray!75}- &
  \color{gray!75}- &
  \color{gray!75}0.610 &
  \color{gray!75}- &
  \color{gray!75}- &
  \color{gray!75}0.718 &
  \color{gray!75}- &
  \color{gray!75}- &
  \color{gray!75}0.705 &
  \color{gray!75}- \\
MGNN(Res50) \cite{wu2025depmgnn} &
  AAAI 2025 &
  0.700 &
  0.700 &
  \cellcolor{gray!30}\textbf{0.012} &
  0.768 &
  0.768 &
  \cellcolor{gray!15}\underline{0.011} &
  0.778 &
  0.778 &
  \cellcolor{gray!30}\textbf{0.010} &
  0.626 &
  0.625 &
  \cellcolor{gray!30}\textbf{0.013} &
  0.732 &
  0.731 &
  \cellcolor{gray!30}\textbf{0.012} &
  0.721 &
  0.720 &
  \cellcolor{gray!30}\textbf{0.011} \\ \hhline{====================}
TimeSFormer-B \cite{bertasius2021spacetime} &
  ICML 2021 &
  0.685 &
  0.670 &
  0.015 &
  0.762 &
  0.755 &
  0.012 &
  0.756 &
  0.749 &
  0.012 &
  0.612 &
  0.597 &
  0.015 &
  0.717 &
  0.711 &
  0.014 &
  0.706 &
  0.696 &
  0.013 \\
MAE-DFER-B \cite{sun2023mae} &
  MM 2023 &
  0.702 &
  0.697 &
  0.014 &
  0.761 &
  0.760 &
  0.012 &
  0.776 &
  0.774 &
  0.011 &
  \cellcolor{gray!15}\underline{0.631} &
  \cellcolor{gray!15}\underline{0.628} &
  0.014 &
  \cellcolor{gray!15}\underline{0.734} &
  \cellcolor{gray!15}\underline{0.733} &
  0.013 &
  0.721 &
  0.718 &
  0.013 \\
\color{gray!75}SVFAP-B* \cite{sun2024svfap} &
  TAC 2024 &
  \color{gray!75}- &
  \color{gray!75}0.651 &
  \color{gray!75}- &
  \color{gray!75}- &
  \color{gray!75}0.714 &
  \color{gray!75}- &
  \color{gray!75}- &
  \color{gray!75}0.735 &
  \color{gray!75}- &
  \color{gray!75}- &
  \color{gray!75}0.550 &
  \color{gray!75}- &
  \color{gray!75}- &
  \color{gray!75}0.672 &
  \color{gray!75}- &
  \color{gray!75}- &
  \color{gray!75}0.665 &
  \color{gray!75}- \\
SVFAP-B \cite{sun2024svfap} &
  TAC 2024 &
  0.703 &
  0.697 &
  0.014 &
  \cellcolor{gray!15}\underline{0.773} &
  0.768 &
  0.012 &
  \cellcolor{gray!15}\underline{0.780} &
  \cellcolor{gray!15}\underline{0.779} &
  \cellcolor{gray!30}\textbf{0.010} &
  0.624 &
  0.619 &
  0.014 &
  0.731 &
  0.730 &
  0.013 &
  \cellcolor{gray!15}\underline{0.722} &
  0.719 &
  0.013 \\ 
  FRU-Adapter-B \cite{her2025fru} &
  Ele. 2025 &
  \cellcolor{gray!30}\textbf{0.707} &
  \cellcolor{gray!30}\textbf{0.707} &
  0.013 &
  0.771 &
  \cellcolor{gray!15}\underline{0.769} &
  \cellcolor{gray!15}\underline{0.011} &
  0.779 &
  0.778 &
  \cellcolor{gray!30}\textbf{0.010} &
  0.622 &
  0.622 &
  \cellcolor{gray!30}\textbf{0.013} &
  0.734 &
  0.732 &
  \cellcolor{gray!30}\textbf{0.012} &
  0.723 &
  \cellcolor{gray!15}\underline{0.722} &
  0.012 \\
  \Xhline{4.5px}
\end{tabular}
}
\label{tab:pr_metric}
\end{table*}

\noindent \textbf{Performance analysis.}
Table \ref{tab:dep_avec2014} provides a more complete view of the non-monotonic parameter-performance relationship observed for FDA in Fig. 4(g) of the main paper. The two MGNN configurations occupy the leading positions, but their advantages differ across evaluation metrics. MGNN (ResNet-50) reports the lowest RMSE of 6.623 together with the highest PCC of 0.817 and CCC of 0.799. Relative to MGNN (MFN), MGNN (ResNet-50) reduces RMSE by 4.51\% and provides relative performance gains of 2.51\% and 2.70\% in PCC and CCC, respectively. MGNN (MFN), however, achieves the lowest MAE of 5.279, representing a relative error reduction of 3.10\% over MGNN (ResNet-50). More importantly, MGNN (MFN) outperforms MAE-DFER-B and SVFAP-B with substantially larger parameters across all four metrics. MAE-DFER-B and SVFAP-B yield RMSE values of 7.533 and 7.967 and CCC values of 0.739 and 0.695, respectively. The results therefore reinforce the conclusion in the main paper that increasing model parameters does not consistently improve FDA performance. For AVEC2014, the temporal modelling strategy contributes more to the final performance than simply increasing the number of model parameters.

\noindent\textbf{Application-oriented model selection.} MGNN (ResNet-50) is preferable when the lowest RMSE and the strongest correlation-based performance are required and sufficient computation is available. According to Table \ref{tab:model_efficiency}, it requires 25.10M parameters and 296.00G FLOPs and processes 3 VPS on L20 GPU, making it more appropriate for offline analysis than time-sensitive processing. MGNN (MFN) provides a more favourable balance between predictive performance and computation. It retains strong PCC and CCC results, achieves the lowest MAE, and reduces the parameter count and FLOPs to 5.00M and 43.73G, respectively. It is therefore the preferred general-purpose FDA configuration when model size and predictive performance must be considered together.

When inference speed is prioritised, MAE-DFER-B and SVFAP-B provide faster alternatives. MAE-DFER-B reaches 25 VPS and consistently outperforms SVFAP-B across all four FDA metrics, whereas SVFAP-B increases the inference speed to 38 VPS with 78.69M parameters and 43.60G FLOPs. Thus, MAE-DFER-B offers a more favourable intermediate choice between FDA performance and throughput, while SVFAP-B is more suitable when the highest measured processing speed is required and a larger performance reduction is acceptable.

\subsubsection{FAPR performances}

\noindent\textbf{Performance analysis.} The results in Table \ref{tab:pr_metric} make the parameter-performance relationship for FAPR particularly clear. MGNN (MFN), the smallest evaluated configuration, ranks first in the aggregate results with a mean PCC and CCC of 0.728 and a mean MSE of 0.011. Its advantage is not confined to the average score. MGNN (MFN) produces the highest PCC and CCC for Conscientiousness, Extraversion, Agreeableness, and Neuroticism, while FRU-Adapter-B performs best for Openness. Across all models, Agreeableness produces the lowest correlation values, indicating that this personality trait remains more difficult to estimate than Conscientiousness and Extraversion under the adopted evaluation protocol.

The comparison with the results reported in the original publications further supports the OpenAHBU re-development strategy. The re-developed MGNN (ResNet-50) improves the mean CCC from 0.705 to 0.720, corresponding to a relative performance gain of 2.13\%. The improvement is more pronounced for SVFAP-B, whose mean CCC increases from 0.665 to 0.719, providing a relative performance gain of 8.12\%. Nevertheless, neither configuration surpasses MGNN (MFN). TimeSFormer-B contains the most parameters but only reaches a mean CCC of only 0.696. These results extend the observation in Fig. 4(h) of the main paper: increasing model parameters provides no consistent benefit for FAPR, and an appropriately designed compact temporal model can perform better across multiple apparent personality traits.

\noindent\textbf{Application-oriented model selection.} MGNN (MFN) is the preferred general-purpose FAPR model because it combines the best overall performance with the smallest model size. It requires 7.55M parameters and 37.11G FLOPs, although its inference speed is limited to 17 VPS. MGNN (ResNet-50) increases the inference speed to 25 VPS and retains a competitive mean CCC of 0.720, making it a suitable alternative when higher processing throughput is needed without a substantial reduction in predictive performance.

FRU-Adapter-B should be considered when Openness estimation is particularly important. It achieves the highest Openness PCC and CCC and the second-highest mean CCC of 0.722, but its 280.00G FLOPs impose a considerably higher computational cost. SVFAP-B offers a more favourable performance-computation balance for faster processing, reaching 38 VPS with 43.60G FLOPs while retaining a mean CCC of 0.719. MAE-DFER-B provides comparable overall performance with a mean CCC of 0.718 and can be selected when its stronger results on Agreeableness or Neuroticism are relevant to the target application. TimeSFormer-B achieves the highest measured inference speed of 46 VPS, but its 122.34M parameters and lower mean CCC make it suitable primarily when throughput is prioritised over model size and predictive performance. Consequently, FAPR model selection should consider both the required personality traits and the available computational resources rather than relying on model parameters alone.

\subsection{Synergistic multi-task enhancement results}

This section supplements the SMTE experiments provided in Sec. 4.4 of the main paper with a function-wise analysis of the eight re-developed OpenAHBU functions. 
Specifically, for each function, we first report and analyse the detailed SMTE results obtained using the best-performing model. We then repeat the cross-function interactions using the second best-performing model for each function, selected by jointly considering its overall performance and number of parameters, to examine whether the observed interactions remain consistent across different model configurations.

\subsubsection{Relationship among facial behaviours and long-term psychological states}

\textbf{Relationships among frame-level facial behaviours:} 

\begin{itemize}

    \item \textbf{AUs and facial expressions.}
    Mehu et al. \cite{mehu2012reliable} analysed coded AU intensities alongside observers' judgements of actors' expressions. Stronger inner brow raising (AU1) was associated with more accurate recognition of sadness facial expression, while higher intensities of the AU1+AU4 combination, AU16 and AU23 were associated with more accurate recognition of anger facial expression. Using FACS to code posed expressions from 230 participants, Du et al. \cite{du2014compound} identified AU combinations associated with different facial expressions, including AU4+AU7+AU24 for anger, AU9+AU10+AU17 for disgust, and AU1+AU2+AU25+AU26 for surprise. Scherer et al. \cite{scherer2018appraisal} further examined animated faces with controlled AU activations, linking brow lowering (AU4) to perceived anger, and nose wrinkling (AU9) or upper lip raising (AU10) to perceived disgust.

    \item \textbf{AUs and valence/arousal.}
    Liu et al. \cite{liu2022facial} collected emotion labels and valence/arousal ratings for randomly generated facial animations, linking nose wrinkling (AU9) to perceived negative valence and high arousal. Their analyses further showed that several AUs, including AU4, AU9, AU10, AU12 and AU15, provided information about both perceived emotion categories and valence/arousal. Namba et al. \cite{namba2023development} analysed AU intensities in participants' facial expressions of recalled personal events, reporting positive associations between self-reported valence and AUs 6, 10, 12 and 14, and a negative association with AU4. Similarly, Zhang et al. \cite{zhang2024sensing} correlated 20 automatically estimated AUs during emotional films with participants' valence/arousal ratings collected by asking them to recall their viewing experience. Their exploratory analyses showed negative correlations between valence and AUs 1, 4, 15 and 18, and positive correlations with AUs 6, 7, 9, 12, 20 and 25. AUs 2, 6, 7, 9, 12 and 17 were also positively associated with arousal.

    \item \textbf{Facial expressions and valence/arousal.}
    Russell and Bullock \cite{russell1985multidimensional} applied multidimensional scaling to judgements of facial expression similarity, revealing a two-dimensional structure corresponding to valence and arousal. Using morphed facial images and the Affect Grid, Fujimura et al. \cite{fujimura2012categorical} observed category boundaries in valence ratings along the happiness--fear continuum and in arousal ratings along the surprise--sadness continuum. More recently, Baudouin et al. \cite{baudouin2025arousal} collected emotion labels and valence/arousal ratings for facial photographs. Happy and sad expressions received positive and negative valence ratings, respectively, while fearful expressions received higher arousal ratings than sad expressions.
\end{itemize}

\textbf{Relationships between frame-level facial behaviours and video-level facial behaviours/long-term psychological states:}

\begin{itemize}
    \item \textbf{AUs and dynamic facial expressions.}
    Using FACS to analyse actors' emotional portrayals, Krumhuber and Scherer \cite{krumhuber2011affect} reported that the relative peak timing of AU4, AU7, AU10 and AU25 differed across emotions, with AU4 peaking earlier in sadness than in anger and fear. By relating observers' emotion judgements to randomly generated facial animations, Jack et al. \cite{jack2014dynamic} identified early upper-lid raising (AU5) shared by fear and surprise, and early nose wrinkling (AU9) shared by anger and disgust. Delis et al. \cite{delis2016space} further varied the temporal ordering of AU components in facial animations, showing that the original ordering produced higher emotion-matching hit rates than the inverted ordering for happiness, surprise, fear and disgust.

    \item \textbf{AUs and pain.}
    By coding facial responses during shoulder movement tests, Prkachin and Solomon \cite{prkachin2008structure} found that an index combining brow lowering (AU4), eye tightening (AU6/7), nose wrinkling or upper lip raising (AU9/10), and eye closure (AU43) was associated with patients' self-reported pain. Using FACS to compare responses to painful heat and disgust-inducing pictures in the same participants, Kunz et al. \cite{kunz2013pain} observed that pain and disgust shared AU4, AU6/7 and AU9/10, while eye tightening was displayed more strongly during pain. Through cluster analysis of FACS-coded responses to painful heat in two independent samples, Kunz and Lautenbacher \cite{kunz2014faces} identified distinct patterns involving multiple AUs, including brow lowering (AU4) with eye tightening (AU6/7), and mouth opening (AU25/26/27) with eye tightening. Here, slash-separated AU numbers indicate facial actions analysed as a group.

    \item \textbf{AUs and depression.}
    By coding changes within smiles elicited by a comedy clip, Reed et al. \cite{reed2007impact} linked the occurrence of AU14 or AU15 after the onset and before the offset of an AU12 smile to higher Beck Depression Inventory scores among participants who smiled. Falkenberg et al. \cite{falkenberg2012mood} used FACS to assess facial responses during mood induction with facial pictures and cartoons, reporting lower mean smile intensity, measured by AU6 and AU12 combined, in patients with major depressive disorder than in healthy controls. In a three-year study of adolescent girls with no previous depressive disorder, Ozturk et al. \cite{ozturk2024digital} analysed automatically estimated AUs during baseline clinical interviews and found that higher mean AU4 and AU26 intensities were associated with subsequent onset of depressive disorders.

    \item \textbf{AUs and apparent personality.}
    By relating automatically estimated AUs in posed smiling photographs to observers' personality judgements, Witkower et al. \cite{witkower2024smile} found that higher AU12 and AU14 intensities were associated with higher perceived conscientiousness. They also reported that smiles combining AU6, AU12 and AU25 were associated with higher perceived conscientiousness and trustworthiness.

    \item \textbf{Facial expressions and depression.}
    Through EMFACS coding of responses to emotional films, Renneberg et al. \cite{renneberg2005facial} observed that women with major depression displayed happiness expressions less frequently, including fewer high-intensity displays, than non-clinical controls during a positive film.

    \item \textbf{Facial expressions and apparent personality.}
    Kenny et al. \cite{kenny1992consensus} collected separate behavioural ratings and personality judgements from 20-second videos, linking more smiling to higher perceived extraversion and agreeableness.
\end{itemize}

\textbf{Relationships among video-level facial behaviours and long-term psychological states:} 

\begin{itemize}
    \item \textbf{Dynamic facial expressions and pain.}
    By asking observers to evaluate one-second facial videos, Simon et al. \cite{simon2008recognition} showed that pain expressions were distinguishable from six basic emotional expressions, although they were occasionally confused with fear and disgust. In a within-participant experiment, Kunz et al. \cite{kunz2013pain} recorded facial responses to painful heat and disgust-inducing pictures, finding that the two states shared facial actions but differed in their strength and combinations.

    \item \textbf{Dynamic facial expressions and depression.}
    Rottenberg et al. \cite{rottenberg2002sadness} coded facial responses to emotional films and followed patients with major depression for six months. Patients who did not recover had shown lower facial reactivity to an amusing film at baseline. By counting smiles in recordings of European American and Asian American women watching an amusing film, Chentsova-Dutton et al. \cite{chentsova2010further} also observed smaller increases in enjoyment smiles in depressed participants than in controls.

    \item \textbf{Dynamic facial expressions and apparent personality.}
    Kenny et al. \cite{kenny1992consensus} used separate raters to assess smiling behaviour and personality impressions from 20-second videos. Greater smiling was associated with higher perceived extraversion and agreeableness.

    \item \textbf{Pain and depression.}
    Lautenbacher et al. \cite{lautenbacher2017understanding} compared video-recorded responses to heat stimulation in patients with major depression and matched controls. Patients with depression showed increased brow contraction and a stronger association between facial responses and pain ratings, despite no group differences in the psychophysical pain measures.

    \item \textbf{Apparent personality and depression.}
    From brief intake recordings, Sasso and Strunk \cite{sasso2013thin} obtained observers' ratings of clients' Big Five personality traits and related them to symptom changes during 16 weeks of cognitive therapy. Higher perceived extraversion was associated with greater reductions in depressive symptoms.
\end{itemize}

These findings provide psychological evidence for the relationships among facial behaviours and long-term psychological states discussed in the main paper and motivate the SMTE design, which aims to enhance each target function using complementary cues contained in its most-related support function's predictions.

\subsubsection{FAUD enhancement results}

\begin{table*}[htbp]
\centering
\Huge
\setlength{\doublerulesep}{0.3em}
\renewcommand{\arraystretch}{1.4}
\caption{AU-wise F1 scores (\%) of the baseline and SMTE-enhanced FAUD models on the function-level test set. Results are reported for the best-performing MAE-Face-B model and the representative second configuration, FMAE-B, selected by jointly considering overall performance and model parameters. FAUE, SFER, and VAIE denote the support functions whose predictions are incorporated by SMTE, while All denotes the joint use of predictions from all three support functions. Values in parentheses indicate the relative metric changes compared with the corresponding FAUD baseline. Positive changes are highlighted in bold.}
\resizebox{1.0\linewidth}{!}{
\begin{tabular}{c|cccccccccccccccccc|c}
\Xhline{4.5px}
\rowcolor{gray!15}
\cellcolor{gray!15} &
  \multicolumn{18}{c|}{\cellcolor{gray!15}AU} &
  \cellcolor{gray!15} \\ \hhline{~------------------}
\rowcolor{gray!15}
\multirow{-2}{*}{\cellcolor{gray!15}Methods} &
  01 &
  02 &
  04 &
  05 &
  06 &
  07 &
  09 &
  10 &
  12 &
  14 &
  15 &
  16 &
  17 &
  23 &
  24 &
  25 &
  26 &
  27 &
  \multirow{-2}{*}{\cellcolor{gray!15}Avg.} \\ \hhline{====================}
MAE-Face-B &
  55.15 &
  46.02 &
  59.83 &
  73.38 &
  66.96 &
  75.55 &
  77.10 &
  76.83 &
  76.15 &
  67.14 &
  41.99 &
  67.81 &
  68.38 &
  38.23 &
  37.01 &
  84.14 &
  43.79 &
  86.16 &
  63.42 \\ \hline
\begin{tabular}[c]{@{}c@{}}MAE-Face-B\\ (+FAUE)\end{tabular} &
  \begin{tabular}[c]{@{}c@{}}\cellcolor{gray!30}57.44\\ \cellcolor{gray!30}\textbf{(+4.15\%)}\end{tabular} &
  \begin{tabular}[c]{@{}c@{}}\cellcolor{gray!30}47.21\\ \cellcolor{gray!30}\textbf{(+2.59\%)}\end{tabular} &
  \begin{tabular}[c]{@{}c@{}}\cellcolor{gray!30}60.69\\ \cellcolor{gray!30}\textbf{(+1.44\%)}\end{tabular} &
  \begin{tabular}[c]{@{}c@{}}72.16\\ (-1.66\%)\end{tabular} &
  \begin{tabular}[c]{@{}c@{}}66.54\\ (-0.63\%)\end{tabular} &
  \begin{tabular}[c]{@{}c@{}}\cellcolor{gray!30}76.26\\ \cellcolor{gray!30}\textbf{(+0.94\%)}\end{tabular} &
  \begin{tabular}[c]{@{}c@{}}75.78\\ (-1.71\%)\end{tabular} &
  \begin{tabular}[c]{@{}c@{}}\cellcolor{gray!30}77.15\\ \cellcolor{gray!30}\textbf{(+0.42\%)}\end{tabular} &
  \begin{tabular}[c]{@{}c@{}}\cellcolor{gray!30}76.24\\ \cellcolor{gray!30}\textbf{(+0.12\%)}\end{tabular} &
  \begin{tabular}[c]{@{}c@{}}\cellcolor{gray!30}67.28\\ \cellcolor{gray!30}\textbf{(+0.21\%)}\end{tabular} &
  \begin{tabular}[c]{@{}c@{}}\cellcolor{gray!30}43.58\\ \cellcolor{gray!30}\textbf{(+3.79\%)}\end{tabular} &
  \begin{tabular}[c]{@{}c@{}}\cellcolor{gray!30}68.29\\ \cellcolor{gray!30}\textbf{(+0.71\%)}\end{tabular} &
  \begin{tabular}[c]{@{}c@{}}\cellcolor{gray!30}68.46\\ \cellcolor{gray!30}\textbf{(+0.12\%)}\end{tabular} &
  \begin{tabular}[c]{@{}c@{}}\cellcolor{gray!30}39.10\\ \cellcolor{gray!30}\textbf{(+2.28\%)}\end{tabular} &
  \begin{tabular}[c]{@{}c@{}}\cellcolor{gray!30}37.44\\ \cellcolor{gray!30}\textbf{(+1.16\%)}\end{tabular} &
  \begin{tabular}[c]{@{}c@{}}\cellcolor{gray!30}84.63\\ \cellcolor{gray!30}\textbf{(+0.58\%)}\end{tabular} &
  \begin{tabular}[c]{@{}c@{}}42.38\\ (-3.22\%)\end{tabular} &
  \begin{tabular}[c]{@{}c@{}}\cellcolor{gray!30}87.58\\ \cellcolor{gray!30}\textbf{(+1.65\%)}\end{tabular} &
  \begin{tabular}[c]{@{}c@{}}\cellcolor{gray!30}63.79\\ \cellcolor{gray!30}\textbf{(+0.58\%)}\end{tabular} \\
  \hline
\begin{tabular}[c]{@{}c@{}}MAE-Face-B\\ (+SFER)\end{tabular} &
  \begin{tabular}[c]{@{}c@{}}\cellcolor{gray!30}59.14\\ \cellcolor{gray!30}\textbf{(+7.23\%)}\end{tabular} &
  \begin{tabular}[c]{@{}c@{}}\cellcolor{gray!30}49.44\\ \cellcolor{gray!30}\textbf{(+7.43\%)}\end{tabular} &
  \begin{tabular}[c]{@{}c@{}}\cellcolor{gray!30}62.22\\ \cellcolor{gray!30}\textbf{(+3.99\%)}\end{tabular} &
  \begin{tabular}[c]{@{}c@{}}71.67\\ (-2.33\%)\end{tabular} &
  \begin{tabular}[c]{@{}c@{}}\cellcolor{gray!30}68.18\\ \cellcolor{gray!30}\textbf{(+1.82\%)}\end{tabular} &
  \begin{tabular}[c]{@{}c@{}}\cellcolor{gray!30}76.45\\ \cellcolor{gray!30}\textbf{(+1.19\%)}\end{tabular} &
  \begin{tabular}[c]{@{}c@{}}74.89\\ (-2.87\%)\end{tabular} &
  \begin{tabular}[c]{@{}c@{}}\cellcolor{gray!30}77.64\\ \cellcolor{gray!30}\textbf{(+1.05\%)}\end{tabular} &
  \begin{tabular}[c]{@{}c@{}}\cellcolor{gray!30}77.03\\ \cellcolor{gray!30}\textbf{(+1.16\%)}\end{tabular} &
  \begin{tabular}[c]{@{}c@{}}\cellcolor{gray!30}67.79\\ \cellcolor{gray!30}\textbf{(+0.97\%)}\end{tabular} &
  \begin{tabular}[c]{@{}c@{}}\cellcolor{gray!30}43.67\\ \cellcolor{gray!30}\textbf{(+4.00\%)}\end{tabular} &
  \begin{tabular}[c]{@{}c@{}}\cellcolor{gray!30}68.29\\ \cellcolor{gray!30}\textbf{(+0.71\%)}\end{tabular} &
  \begin{tabular}[c]{@{}c@{}}68.33\\ (-0.07\%)\end{tabular} &
  \begin{tabular}[c]{@{}c@{}}\cellcolor{gray!30}39.47\\ \cellcolor{gray!30}\textbf{(+3.24\%)}\end{tabular} &
  \begin{tabular}[c]{@{}c@{}}\cellcolor{gray!30}38.72\\ \cellcolor{gray!30}\textbf{(+4.62\%)}\end{tabular} &
  \begin{tabular}[c]{@{}c@{}}\cellcolor{gray!30}84.63\\ \cellcolor{gray!30}\textbf{(+0.58\%)}\end{tabular} &
  \begin{tabular}[c]{@{}c@{}}\cellcolor{gray!30}44.05\\ \cellcolor{gray!30}\textbf{(+0.59\%)}\end{tabular} &
  \begin{tabular}[c]{@{}c@{}}\cellcolor{gray!30}87.77\\ \cellcolor{gray!30}\textbf{(+1.87\%)}\end{tabular} &
  \begin{tabular}[c]{@{}c@{}}\cellcolor{gray!30}64.41\\ \cellcolor{gray!30}\textbf{(+1.56\%)}\end{tabular} \\
  \hline
\begin{tabular}[c]{@{}c@{}}MAE-Face-B\\ (+VAIE)\end{tabular} &
  \begin{tabular}[c]{@{}c@{}}\cellcolor{gray!30}58.38\\ \cellcolor{gray!30}\textbf{(+5.86\%)}\end{tabular} &
  \begin{tabular}[c]{@{}c@{}}\cellcolor{gray!30}46.66\\ \cellcolor{gray!30}\textbf{(+1.39\%)}\end{tabular} &
  \begin{tabular}[c]{@{}c@{}}\cellcolor{gray!30}61.26\\ \cellcolor{gray!30}\textbf{(+2.39\%)}\end{tabular} &
  \begin{tabular}[c]{@{}c@{}}71.36\\ (-2.75\%)\end{tabular} &
  \begin{tabular}[c]{@{}c@{}}\cellcolor{gray!30}67.18\\ \cellcolor{gray!30}\textbf{(+0.33\%)}\end{tabular} &
  \begin{tabular}[c]{@{}c@{}}\cellcolor{gray!30}76.25\\ \cellcolor{gray!30}\textbf{(+0.93\%)}\end{tabular} &
  \begin{tabular}[c]{@{}c@{}}76.22\\ (-1.14\%)\end{tabular} &
  \begin{tabular}[c]{@{}c@{}}\cellcolor{gray!30}77.66\\ \cellcolor{gray!30}\textbf{(+1.08\%)}\end{tabular} &
  \begin{tabular}[c]{@{}c@{}}\cellcolor{gray!30}77.00\\ \cellcolor{gray!30}\textbf{(+1.12\%)}\end{tabular} &
  \begin{tabular}[c]{@{}c@{}}\cellcolor{gray!30}67.24\\ \cellcolor{gray!30}\textbf{(+0.15\%)}\end{tabular} &
  \begin{tabular}[c]{@{}c@{}}\cellcolor{gray!30}43.37\\ \cellcolor{gray!30}\textbf{(+3.29\%)}\end{tabular} &
  \begin{tabular}[c]{@{}c@{}}\cellcolor{gray!30}68.07\\ \cellcolor{gray!30}\textbf{(+0.38\%)}\end{tabular} &
  \begin{tabular}[c]{@{}c@{}}68.02\\ (-0.53\%)\end{tabular} &
  \begin{tabular}[c]{@{}c@{}}\cellcolor{gray!30}39.23\\ \cellcolor{gray!30}\textbf{(+2.88\%)}\end{tabular} &
  \begin{tabular}[c]{@{}c@{}}36.76\\ (-0.68\%)\end{tabular} &
  \begin{tabular}[c]{@{}c@{}}84.77\\ (-0.75\%)\end{tabular} &
  \begin{tabular}[c]{@{}c@{}}41.59\\ (-5.02\%)\end{tabular} &
  \begin{tabular}[c]{@{}c@{}}\cellcolor{gray!30}87.50\\ \cellcolor{gray!30}\textbf{(+1.56\%)}\end{tabular} &
  \begin{tabular}[c]{@{}c@{}}\cellcolor{gray!30}63.81\\ \cellcolor{gray!30}\textbf{(+0.61\%)}\end{tabular} \\
  \hline
\begin{tabular}[c]{@{}c@{}}MAE-Face-B\\ (+All)\end{tabular} &
  \begin{tabular}[c]{@{}c@{}}\cellcolor{gray!30}57.84\\ \cellcolor{gray!30}\textbf{(+4.88\%)}\end{tabular} &
  \begin{tabular}[c]{@{}c@{}}\cellcolor{gray!30}49.50\\ \cellcolor{gray!30}\textbf{(+7.56\%)}\end{tabular} &
  \begin{tabular}[c]{@{}c@{}}\cellcolor{gray!30}62.37\\ \cellcolor{gray!30}\textbf{(+4.25\%)}\end{tabular} &
  \begin{tabular}[c]{@{}c@{}}72.68\\ (-0.95\%)\end{tabular} &
  \begin{tabular}[c]{@{}c@{}}\cellcolor{gray!30}68.58\\ \cellcolor{gray!30}\textbf{(+2.42\%)}\end{tabular} &
  \begin{tabular}[c]{@{}c@{}}\cellcolor{gray!30}76.71\\ \cellcolor{gray!30}\textbf{(+1.54\%)}\end{tabular} &
  \begin{tabular}[c]{@{}c@{}}75.97\\ (-1.47\%)\end{tabular} &
  \begin{tabular}[c]{@{}c@{}}\cellcolor{gray!30}77.65\\ \cellcolor{gray!30}\textbf{(+1.07\%)}\end{tabular} &
  \begin{tabular}[c]{@{}c@{}}\cellcolor{gray!30}77.18\\ \cellcolor{gray!30}\textbf{(+1.35\%)}\end{tabular} &
  \begin{tabular}[c]{@{}c@{}}\cellcolor{gray!30}67.64\\ \cellcolor{gray!30}\textbf{(+0.74\%)}\end{tabular} &
  \begin{tabular}[c]{@{}c@{}}\cellcolor{gray!30}43.20\\ \cellcolor{gray!30}\textbf{(+2.88\%)}\end{tabular} &
  \begin{tabular}[c]{@{}c@{}}\cellcolor{gray!30}68.57\\ \cellcolor{gray!30}\textbf{(+1.12\%)}\end{tabular} &
  \begin{tabular}[c]{@{}c@{}}68.36\\ (-0.03\%)\end{tabular} &
  \begin{tabular}[c]{@{}c@{}}\cellcolor{gray!30}39.09\\ \cellcolor{gray!30}\textbf{(+2.25\%)}\end{tabular} &
  \begin{tabular}[c]{@{}c@{}}\cellcolor{gray!30}38.19\\ \cellcolor{gray!30}\textbf{(+3.19\%)}\end{tabular} &
  \begin{tabular}[c]{@{}c@{}}\cellcolor{gray!30}84.81\\ \cellcolor{gray!30}\textbf{(+0.80\%)}\end{tabular} &
  \begin{tabular}[c]{@{}c@{}}\cellcolor{gray!30}44.43\\ \cellcolor{gray!30}\textbf{(+1.46\%)}\end{tabular} &
  \begin{tabular}[c]{@{}c@{}}\cellcolor{gray!30}88.13\\ \cellcolor{gray!30}\textbf{(+2.29\%)}\end{tabular} &
  \begin{tabular}[c]{@{}c@{}}\cellcolor{gray!30}64.49\\ \cellcolor{gray!30}\textbf{(+1.69\%)}\end{tabular} \\ \hhline{====================}
FMAE-B &
  57.51 &
  45.49 &
  58.29 &
  69.70 &
  67.13 &
  75.73 &
  64.16 &
  77.08 &
  75.95 &
  65.16 &
  45.34 &
  66.91 &
  68.32 &
  39.21 &
  39.22 &
  84.49 &
  40.00 &
  85.12 &
  62.49 \\ \hline
\begin{tabular}[c]{@{}c@{}}FMAE-B\\ (+FAUE)\end{tabular} &
  \begin{tabular}[c]{@{}c@{}}\cellcolor{gray!30}58.79\\ \cellcolor{gray!30}\textbf{(+2.23\%)}\end{tabular} &
  \begin{tabular}[c]{@{}c@{}}\cellcolor{gray!30}45.65\\ \cellcolor{gray!30}\textbf{(+0.35\%)}\end{tabular} &
  \begin{tabular}[c]{@{}c@{}}58.13\\ (-0.27\%)\end{tabular} &
  \begin{tabular}[c]{@{}c@{}}69.25\\ (-0.65\%)\end{tabular} &
  \begin{tabular}[c]{@{}c@{}}\cellcolor{gray!30}67.46\\ \cellcolor{gray!30}\textbf{(+0.49\%)}\end{tabular} &
  \begin{tabular}[c]{@{}c@{}}\cellcolor{gray!30}76.06\\ \cellcolor{gray!30}\textbf{(+0.44\%)}\end{tabular} &
  \begin{tabular}[c]{@{}c@{}}61.06\\ (-4.83\%)\end{tabular} &
  \begin{tabular}[c]{@{}c@{}}\cellcolor{gray!30}77.17\\ \cellcolor{gray!30}\textbf{(+0.12\%)}\end{tabular} &
  \begin{tabular}[c]{@{}c@{}}75.87\\ (-0.11\%)\end{tabular} &
  \begin{tabular}[c]{@{}c@{}}64.77\\ (-0.60\%)\end{tabular} &
  \begin{tabular}[c]{@{}c@{}}43.74\\ (-3.53\%)\end{tabular} &
  \begin{tabular}[c]{@{}c@{}}\cellcolor{gray!30}67.62\\ \cellcolor{gray!30}\textbf{(+1.06\%)}\end{tabular} &
  \begin{tabular}[c]{@{}c@{}}67.98\\ (-0.50\%)\end{tabular} &
  \begin{tabular}[c]{@{}c@{}}38.38\\ (-2.12\%)\end{tabular} &
  \begin{tabular}[c]{@{}c@{}}36.83\\ (-6.09\%)\end{tabular} &
  \begin{tabular}[c]{@{}c@{}}\cellcolor{gray!30}84.64\\ \cellcolor{gray!30}\textbf{(+0.18\%)}\end{tabular} &
  \begin{tabular}[c]{@{}c@{}}39.77\\ (-0.58\%)\end{tabular} &
  \begin{tabular}[c]{@{}c@{}}\cellcolor{gray!30}85.20\\ \cellcolor{gray!30}\textbf{(+0.09\%)}\end{tabular} &
  \begin{tabular}[c]{@{}c@{}}62.13\\ (-0.58\%)\end{tabular} \\
  \hline
\begin{tabular}[c]{@{}c@{}}FMAE-B\\ (+SFER)\end{tabular} &
  \begin{tabular}[c]{@{}c@{}}\cellcolor{gray!30}60.84\\ \cellcolor{gray!30}\textbf{(+5.79\%)}\end{tabular} &
  \begin{tabular}[c]{@{}c@{}}\cellcolor{gray!30}48.44\\ \cellcolor{gray!30}\textbf{(+6.48\%)}\end{tabular} &
  \begin{tabular}[c]{@{}c@{}}\cellcolor{gray!30}61.20\\ \cellcolor{gray!30}\textbf{(+4.99\%)}\end{tabular} &
  \begin{tabular}[c]{@{}c@{}}68.59\\ (-1.59\%)\end{tabular} &
  \begin{tabular}[c]{@{}c@{}}\cellcolor{gray!30}68.92\\ \cellcolor{gray!30}\textbf{(+2.67\%)}\end{tabular} &
  \begin{tabular}[c]{@{}c@{}}\cellcolor{gray!30}76.24\\ \cellcolor{gray!30}\textbf{(+0.67\%)}\end{tabular} &
  \begin{tabular}[c]{@{}c@{}}62.34\\ (-2.84\%)\end{tabular} &
  \begin{tabular}[c]{@{}c@{}}\cellcolor{gray!30}77.48\\ \cellcolor{gray!30}\textbf{(+0.52\%)}\end{tabular} &
  \begin{tabular}[c]{@{}c@{}}\cellcolor{gray!30}76.79\\ \cellcolor{gray!30}\textbf{(+1.11\%)}\end{tabular} &
  \begin{tabular}[c]{@{}c@{}}\cellcolor{gray!30}65.85\\ \cellcolor{gray!30}\textbf{(+1.06\%)}\end{tabular} &
  \begin{tabular}[c]{@{}c@{}}44.69\\ (-1.43\%)\end{tabular} &
  \begin{tabular}[c]{@{}c@{}}\cellcolor{gray!30}68.06\\ \cellcolor{gray!30}\textbf{(+1.72\%)}\end{tabular} &
  \begin{tabular}[c]{@{}c@{}}68.16\\ (-0.09\%)\end{tabular} &
  \begin{tabular}[c]{@{}c@{}}37.70\\ (-3.85\%)\end{tabular} &
  \begin{tabular}[c]{@{}c@{}}\cellcolor{gray!30}40.14\\ \cellcolor{gray!30}\textbf{(+2.35\%)}\end{tabular} &
  \begin{tabular}[c]{@{}c@{}}\cellcolor{gray!30}84.72\\ \cellcolor{gray!30}\textbf{(+0.27\%)}\end{tabular} &
  \begin{tabular}[c]{@{}c@{}}\cellcolor{gray!30}41.40\\ \cellcolor{gray!30}\textbf{(+3.50\%)}\end{tabular} &
  \begin{tabular}[c]{@{}c@{}}85.03\\ (-0.11\%)\end{tabular} &
  \begin{tabular}[c]{@{}c@{}}\cellcolor{gray!30}63.14\\ \cellcolor{gray!30}\textbf{(+1.04\%)}\end{tabular} \\
  \hline
\begin{tabular}[c]{@{}c@{}}FMAE-B\\ (+VAIE)\end{tabular} &
  \begin{tabular}[c]{@{}c@{}}\cellcolor{gray!30}57.69\\ \cellcolor{gray!30}\textbf{(+0.31\%)}\end{tabular} &
  \begin{tabular}[c]{@{}c@{}}45.12\\ (-0.81\%)\end{tabular} &
  \begin{tabular}[c]{@{}c@{}}\cellcolor{gray!30}59.17\\ \cellcolor{gray!30}\textbf{(+1.51\%)}\end{tabular} &
  \begin{tabular}[c]{@{}c@{}}\cellcolor{gray!30}70.17\\ \cellcolor{gray!30}\textbf{(+0.67\%)}\end{tabular} &
  \begin{tabular}[c]{@{}c@{}}\cellcolor{gray!30}67.70\\ \cellcolor{gray!30}\textbf{(+0.85\%)}\end{tabular} &
  \begin{tabular}[c]{@{}c@{}}\cellcolor{gray!30}76.17\\ \cellcolor{gray!30}\textbf{(+0.58\%)}\end{tabular} &
  \begin{tabular}[c]{@{}c@{}}62.34\\ (-2.84\%)\end{tabular} &
  \begin{tabular}[c]{@{}c@{}}\cellcolor{gray!30}77.23\\ \cellcolor{gray!30}\textbf{(+0.19\%)}\end{tabular} &
  \begin{tabular}[c]{@{}c@{}}\cellcolor{gray!30}76.34\\ \cellcolor{gray!30}\textbf{(+0.51\%)}\end{tabular} &
  \begin{tabular}[c]{@{}c@{}}65.15\\ (-0.02\%)\end{tabular} &
  \begin{tabular}[c]{@{}c@{}}43.55\\ (-3.95\%)\end{tabular} &
  \begin{tabular}[c]{@{}c@{}}\cellcolor{gray!30}68.09\\ \cellcolor{gray!30}\textbf{(+1.76\%)}\end{tabular} &
  \begin{tabular}[c]{@{}c@{}}68.02\\ (-0.44\%)\end{tabular} &
  \begin{tabular}[c]{@{}c@{}}38.09\\ (-2.86\%)\end{tabular} &
  \begin{tabular}[c]{@{}c@{}}37.38\\ (-4.69\%)\end{tabular} &
  \begin{tabular}[c]{@{}c@{}}\cellcolor{gray!30}84.55\\ \cellcolor{gray!30}\textbf{(+0.07\%)}\end{tabular} &
  \begin{tabular}[c]{@{}c@{}}39.12\\ (-2.20\%)\end{tabular} &
  \begin{tabular}[c]{@{}c@{}}85.03\\ (-0.11\%)\end{tabular} &
  \begin{tabular}[c]{@{}c@{}}62.27\\ (-0.35\%)\end{tabular} \\
  \hline
\begin{tabular}[c]{@{}c@{}}FMAE-B\\ (+All)\end{tabular} &
  \begin{tabular}[c]{@{}c@{}}\cellcolor{gray!30}59.99\\ \cellcolor{gray!30}\textbf{(+4.31\%)}\end{tabular} &
  \begin{tabular}[c]{@{}c@{}}\cellcolor{gray!30}47.89\\ \cellcolor{gray!30}\textbf{(+5.27\%)}\end{tabular} &
  \begin{tabular}[c]{@{}c@{}}\cellcolor{gray!30}60.84\\ \cellcolor{gray!30}\textbf{(+4.37\%)}\end{tabular} &
  \begin{tabular}[c]{@{}c@{}}69.07\\ (-0.90\%)\end{tabular} &
  \begin{tabular}[c]{@{}c@{}}\cellcolor{gray!30}69.12\\ \cellcolor{gray!30}\textbf{(+2.96\%)}\end{tabular} &
  \begin{tabular}[c]{@{}c@{}}\cellcolor{gray!30}75.98\\ \cellcolor{gray!30}\textbf{(+0.33\%)}\end{tabular} &
  \begin{tabular}[c]{@{}c@{}}62.12\\ (-3.18\%)\end{tabular} &
  \begin{tabular}[c]{@{}c@{}}\cellcolor{gray!30}77.18\\ \cellcolor{gray!30}\textbf{(+0.13\%)}\end{tabular} &
  \begin{tabular}[c]{@{}c@{}}\cellcolor{gray!30}76.98\\ \cellcolor{gray!30}\textbf{(+1.36\%)}\end{tabular} &
  \begin{tabular}[c]{@{}c@{}}\cellcolor{gray!30}66.08\\ \cellcolor{gray!30}\textbf{(+1.41\%)}\end{tabular} &
  \begin{tabular}[c]{@{}c@{}}45.06\\ (-0.62\%)\end{tabular} &
  \begin{tabular}[c]{@{}c@{}}\cellcolor{gray!30}67.12\\ \cellcolor{gray!30}\textbf{(+0.31\%)}\end{tabular} &
  \begin{tabular}[c]{@{}c@{}}68.15\\ (-0.25\%)\end{tabular} &
  \begin{tabular}[c]{@{}c@{}}37.29\\ (-4.90\%)\end{tabular} &
  \begin{tabular}[c]{@{}c@{}}\cellcolor{gray!30}39.42\\ \cellcolor{gray!30}\textbf{(+0.51\%)}\end{tabular} &
  \begin{tabular}[c]{@{}c@{}}84.46\\ (-0.04\%)\end{tabular} &
  \begin{tabular}[c]{@{}c@{}}\cellcolor{gray!30}40.56\\ \cellcolor{gray!30}\textbf{(+1.40\%)}\end{tabular} &
  \begin{tabular}[c]{@{}c@{}}\cellcolor{gray!30}85.88\\ \cellcolor{gray!30}\textbf{(+0.89\%)}\end{tabular} &
  \begin{tabular}[c]{@{}c@{}}\cellcolor{gray!30}62.95\\ \cellcolor{gray!30}\textbf{(+0.74\%)}\end{tabular} \\
  \Xhline{4.5px}
\end{tabular}
}
\label{tab:faud_sec}
\end{table*}

\noindent\textbf{Enhancement of the best-performing FAUD model.} Table \ref{tab:faud_sec} expands the SMTE-based FAUD results reported in the main paper by presenting the AU-wise effects of FAUE, SFER, and VAIE predictions on MAE-Face-B. Consistent with Fig. 5 of the main paper, SFER provides the strongest individual support, improving the average F1 score by 1.56\%. Its benefit is distributed across 15 of the 18 AUs and is particularly evident for AU01, AU02, AU04, AU15, AU23, and AU24. This broader improvement supports the explanation given in the main paper: categorical facial expression predictions provide global facial contexts that complement the local facial changes used for AU occurrence detection. Nevertheless, SFER predictions do not benefit every AU, reducing the F1 scores for AU05 and AU09 and producing a marginal decrease for AU17. Therefore, even the most effective support function provides AU-dependent rather than uniform improvements.

FAUE and VAIE predictions produce smaller overall gains of 0.58\% and 0.61\%, respectively. FAUE supplies explicit information about the intensities of activated AUs, whereas VAIE describes continuous global affective states. These two types of predictions improve several common AUs, including AU01, AU02, AU04, AU07, AU10, AU12, AU15, AU16, AU23, and AU27, indicating that both local AU intensities and global affective states contain informative complementary cues for FAUD.

Jointly incorporating predictions from all three support functions increases the average F1 score by 1.69\%, only marginally exceeding the 1.56\% gain provided by SFER alone. This limited additional benefit comes with a substantially higher computational cost because the FAUE, SFER, and VAIE models must all perform inference before their predictions can be incorporated by SMTE. Thus, although multiple support functions provide more facial information, the small additional gain does not justify the increased end-to-end computation in the default OpenAHBU setting. This result explains why the main paper focuses on the single most effective support function for each target function.

\noindent\textbf{Enhancement of the second best FAUD model.} 

Repeating the same interactions with the second-best model reveals both consistent and model-dependent cross-function relationships. SFER remains the most effective individual support for FMAE-B, improving its average F1 score by 1.04\%. The improvements on AU01, AU02, and AU04 are consistent with those obtained by MAE-Face-B. This common pattern indicates that the global facial contexts provided by SFER are informative for several AUs across both FAUD models.

The remaining interactions are less consistent. FAUE and VAIE reduce the average F1 score of FMAE-B by 0.58\% and 0.35\%, respectively. For both support functions, the FMAE-B results are evenly divided between nine improved and nine degraded AUs. The largest decreases occur on AU09, AU15, AU23, and AU24, showing that FMAE-B cannot exploit these support predictions as effectively as MAE-Face-B. Thus, the effect of a support function is determined not only by the relationship between the two AFBU functions but also by how the target model represents the associated facial information.

When all three support functions are used together, FMAE-B improves by 0.74\%, which is lower than the 1.04\% gain obtained using SFER alone. Therefore, jointly introducing FAUE and VAIE predictions provides no additional performance benefit but requires inference from two additional support models. This result further demonstrates that multi-function interaction does not consistently improve SMTE across different model configurations, while substantially increasing the required computation. Selecting only the most effective support function therefore provides a more favourable balance between enhancement performance and computational cost.

\subsubsection{FAUE enhancement results}

\begin{table*}[htbp]
\centering
\Huge
\setlength{\doublerulesep}{0.3em}
\renewcommand{\arraystretch}{1.4}
\caption{AU-wise ICC(3,1) results of the baseline and SMTE-enhanced FAUE models on the function-level test set. Results are reported for the best-performing MAE-Face-B model and the representative second configuration, FMAE-B, selected by jointly considering overall performance and model parameters. FAUD, SFER, and VAIE denote the support functions whose predictions are incorporated by SMTE, while All denotes their joint use. Values in parentheses indicate the relative metric changes compared with the corresponding FAUE baseline. Positive changes are highlighted in bold.}
\resizebox{1.0\linewidth}{!}{
\begin{tabular}{c|cccccccccccccc|c}
\Xhline{4.5px}
\rowcolor{gray!15}
\cellcolor{gray!15} &
  \multicolumn{14}{c|}{\cellcolor{gray!15}\textbf{AU}} &
  \cellcolor{gray!15}\\ \hhline{~--------------}
\rowcolor{gray!15} 
\multirow{-2}{*}{\cellcolor{gray!15}\textbf{Methods}} &
  01 &
  02 &
  04 &
  05 &
  06 &
  09 &
  10 &
  12 &
  14 &
  15 &
  17 &
  20 &
  25 &
  26 &
  \multirow{-2}{*}{\cellcolor{gray!15}\textbf{Avg.}} \\ 
  \hhline{================}
MAE-Face-B &
  0.893 &
  0.795 &
  0.804 &
  0.758 &
  0.785 &
  0.783 &
  0.770 &
  0.848 &
  0.475 &
  0.520 &
  0.727 &
  0.328 &
  0.955 &
  0.752 &
  0.728 \\ \hline
\begin{tabular}[c]{@{}c@{}}MAE-Face-B\\ (+FAUD)\end{tabular} &
  \begin{tabular}[c]{@{}c@{}}0.892\\ (-0.11\%)\end{tabular} &
  \begin{tabular}[c]{@{}c@{}}\cellcolor{gray!30}0.813\\ \cellcolor{gray!30}\textbf{(+2.26\%)}\end{tabular} &
  \begin{tabular}[c]{@{}c@{}}\cellcolor{gray!30}0.812\\ \cellcolor{gray!30}\textbf{(+1.00\%)}\end{tabular} &
  \begin{tabular}[c]{@{}c@{}}\cellcolor{gray!30}0.836\\ \cellcolor{gray!30}\textbf{(+10.29\%)}\end{tabular} &
  \begin{tabular}[c]{@{}c@{}}0.779\\ (-0.76\%)\end{tabular} &
  \begin{tabular}[c]{@{}c@{}}\cellcolor{gray!30}0.837\\ \cellcolor{gray!30}\textbf{(+6.90\%)}\end{tabular} &
  \begin{tabular}[c]{@{}c@{}}0.765\\ (-0.65\%)\end{tabular} &
  \begin{tabular}[c]{@{}c@{}}\cellcolor{gray!30}0.857\\ \cellcolor{gray!30}\textbf{(+1.06\%)}\end{tabular} &
  \begin{tabular}[c]{@{}c@{}}0.387\\ (-18.53\%)\end{tabular} &
  \begin{tabular}[c]{@{}c@{}}\cellcolor{gray!30}0.537\\ \cellcolor{gray!30}\textbf{(+3.27\%)}\end{tabular} &
  \begin{tabular}[c]{@{}c@{}}0.721\\ (-0.83\%)\end{tabular} &
  \begin{tabular}[c]{@{}c@{}}\cellcolor{gray!30}0.359\\ \cellcolor{gray!30}\textbf{(+9.45\%)}\end{tabular} &
  \begin{tabular}[c]{@{}c@{}}0.953\\ (-0.21\%)\end{tabular} &
  \begin{tabular}[c]{@{}c@{}}\cellcolor{gray!30}0.767\\ \cellcolor{gray!30}\textbf{(+1.99\%)}\end{tabular} &
  \begin{tabular}[c]{@{}c@{}}\cellcolor{gray!30}0.737\\ \cellcolor{gray!30}\textbf{(+1.24\%)}\end{tabular} \\
  \hline
\begin{tabular}[c]{@{}c@{}}MAE-Face-B\\ (+SFER)\end{tabular} &
  \begin{tabular}[c]{@{}c@{}}\cellcolor{gray!30}0.896\\ \cellcolor{gray!30}\textbf{(+0.34\%)}\end{tabular} &
  \begin{tabular}[c]{@{}c@{}}\cellcolor{gray!30}0.806\\ \cellcolor{gray!30}\textbf{(+1.38\%)}\end{tabular} &
  \begin{tabular}[c]{@{}c@{}}0.795\\ (-1.12\%)\end{tabular} &
  \begin{tabular}[c]{@{}c@{}}\cellcolor{gray!30}0.821\\ \cellcolor{gray!30}\textbf{(+8.31\%)}\end{tabular} &
  \begin{tabular}[c]{@{}c@{}}0.781\\ (-0.51\%)\end{tabular} &
  \begin{tabular}[c]{@{}c@{}}0.763\\ (-2.55\%)\end{tabular} &
  \begin{tabular}[c]{@{}c@{}}0.770\\ (+0.00\%)\end{tabular} &
  \begin{tabular}[c]{@{}c@{}}\cellcolor{gray!30}0.852\\ \cellcolor{gray!30}\textbf{(+0.47\%)}\end{tabular} &
  \begin{tabular}[c]{@{}c@{}}0.441\\ (-7.16\%)\end{tabular} &
  \begin{tabular}[c]{@{}c@{}}\cellcolor{gray!30}0.534\\ \cellcolor{gray!30}\textbf{(+2.69\%)}\end{tabular} &
  \begin{tabular}[c]{@{}c@{}}\cellcolor{gray!30}0.734\\ \cellcolor{gray!30}\textbf{(+0.96\%)}\end{tabular} &
  \begin{tabular}[c]{@{}c@{}}\cellcolor{gray!30}0.345\\ \cellcolor{gray!30}\textbf{(+5.18\%)}\end{tabular} &
  \begin{tabular}[c]{@{}c@{}}0.950\\ (-0.52\%)\end{tabular} &
  \begin{tabular}[c]{@{}c@{}}\cellcolor{gray!30}0.764\\ \cellcolor{gray!30}\textbf{(+1.60\%)}\end{tabular} &
  \begin{tabular}[c]{@{}c@{}}\cellcolor{gray!30}0.732\\ \cellcolor{gray!30}\textbf{(+0.55\%)}\end{tabular} \\
  \hline
\begin{tabular}[c]{@{}c@{}}MAE-Face-B\\ (+VAIE)\end{tabular} &
  \begin{tabular}[c]{@{}c@{}}0.891\\ (-0.22\%)\end{tabular} &
  \begin{tabular}[c]{@{}c@{}}\cellcolor{gray!30}0.813\\ \cellcolor{gray!30}\textbf{(+2.26\%)}\end{tabular} &
  \begin{tabular}[c]{@{}c@{}}\cellcolor{gray!30}0.807\\ \cellcolor{gray!30}\textbf{(+0.37\%)}\end{tabular} &
  \begin{tabular}[c]{@{}c@{}}\cellcolor{gray!30}0.831\\ \cellcolor{gray!30}\textbf{(+9.63\%)}\end{tabular} &
  \begin{tabular}[c]{@{}c@{}}0.782\\ (-0.38\%)\end{tabular} &
  \begin{tabular}[c]{@{}c@{}}0.761\\ (-2.81\%)\end{tabular} &
  \begin{tabular}[c]{@{}c@{}}0.755\\ (-1.95\%)\end{tabular} &
  \begin{tabular}[c]{@{}c@{}}\cellcolor{gray!30}0.856\\ \cellcolor{gray!30}\textbf{(+0.94\%)}\end{tabular} &
  \begin{tabular}[c]{@{}c@{}}0.475\\ (+0.00\%)\end{tabular} &
  \begin{tabular}[c]{@{}c@{}}\cellcolor{gray!30}0.536\\ \cellcolor{gray!30}\textbf{(+3.08\%)}\end{tabular} &
  \begin{tabular}[c]{@{}c@{}}0.723\\ (-0.55\%)\end{tabular} &
  \begin{tabular}[c]{@{}c@{}}\cellcolor{gray!30}0.365\\ \cellcolor{gray!30}\textbf{(+11.28\%)}\end{tabular} &
  \begin{tabular}[c]{@{}c@{}}0.953\\ (-0.21\%)\end{tabular} &
  \begin{tabular}[c]{@{}c@{}}\cellcolor{gray!30}0.757\\ \cellcolor{gray!30}\textbf{(+0.66\%)}\end{tabular} &
  \begin{tabular}[c]{@{}c@{}}\cellcolor{gray!30}0.736\\ \cellcolor{gray!30}\textbf{(+1.10\%)}\end{tabular} \\
  \hline
\begin{tabular}[c]{@{}c@{}}MAE-Face-B\\ (+All)\end{tabular} &
  \begin{tabular}[c]{@{}c@{}}0.893\\ (+0.00\%)\end{tabular} &
  \begin{tabular}[c]{@{}c@{}}\cellcolor{gray!30}0.817\\ \cellcolor{gray!30}\textbf{(+2.77\%)}\end{tabular} &
  \begin{tabular}[c]{@{}c@{}}\cellcolor{gray!30}0.815\\ \cellcolor{gray!30}\textbf{(+1.37\%)}\end{tabular} &
  \begin{tabular}[c]{@{}c@{}}\cellcolor{gray!30}0.827\\ \cellcolor{gray!30}\textbf{(+9.10\%)}\end{tabular} &
  \begin{tabular}[c]{@{}c@{}}0.784\\ (-0.13\%)\end{tabular} &
  \begin{tabular}[c]{@{}c@{}}\cellcolor{gray!30}0.834\\ \cellcolor{gray!30}\textbf{(+6.51\%)}\end{tabular} &
  \begin{tabular}[c]{@{}c@{}}0.761\\ (-1.17\%)\end{tabular} &
  \begin{tabular}[c]{@{}c@{}}\cellcolor{gray!30}0.857\\ \cellcolor{gray!30}\textbf{(+1.06\%)}\end{tabular} &
  \begin{tabular}[c]{@{}c@{}}0.390\\ (-17.89\%)\end{tabular} &
  \begin{tabular}[c]{@{}c@{}}\cellcolor{gray!30}0.534\\ \cellcolor{gray!30}\textbf{(+2.69\%)}\end{tabular} &
  \begin{tabular}[c]{@{}c@{}}0.725\\ (-0.28\%)\end{tabular} &
  \begin{tabular}[c]{@{}c@{}}\cellcolor{gray!30}0.357\\ \cellcolor{gray!30}\textbf{(+8.84\%)}\end{tabular} &
  \begin{tabular}[c]{@{}c@{}}0.952\\ (-0.31\%)\end{tabular} &
  \begin{tabular}[c]{@{}c@{}}\cellcolor{gray!30}0.767\\ \cellcolor{gray!30}\textbf{(+1.99\%)}\end{tabular} &
  \begin{tabular}[c]{@{}c@{}}\cellcolor{gray!30}0.737\\ \cellcolor{gray!30}\textbf{(+1.24\%)}\end{tabular} \\ 
  \hhline{================}
FMAE-B &
  0.815 &
  0.701 &
  0.674 &
  0.773 &
  0.770 &
  0.801 &
  0.756 &
  0.842 &
  0.421 &
  0.568 &
  0.630 &
  0.249 &
  0.953 &
  0.668 &
  0.687 \\ \hline
\begin{tabular}[c]{@{}c@{}}FMAE-B\\ (+FAUD)\end{tabular} &
  \begin{tabular}[c]{@{}c@{}}\cellcolor{gray!30}0.863\\ \cellcolor{gray!30}\textbf{(+5.89\%)}\end{tabular} &
  \begin{tabular}[c]{@{}c@{}}\cellcolor{gray!30}0.781\\ \cellcolor{gray!30}\textbf{(+11.41\%)}\end{tabular} &
  \begin{tabular}[c]{@{}c@{}}\cellcolor{gray!30}0.765\\ \cellcolor{gray!30}\textbf{(+13.50\%)}\end{tabular} &
  \begin{tabular}[c]{@{}c@{}}\cellcolor{gray!30}0.785\\ \cellcolor{gray!30}\textbf{(+1.55\%)}\end{tabular} &
  \begin{tabular}[c]{@{}c@{}}\cellcolor{gray!30}0.789\\ \cellcolor{gray!30}\textbf{(+2.47\%)}\end{tabular} &
  \begin{tabular}[c]{@{}c@{}}\cellcolor{gray!30}0.822\\ \cellcolor{gray!30}\textbf{(+2.62\%)}\end{tabular} &
  \begin{tabular}[c]{@{}c@{}}\cellcolor{gray!30}0.763\\ \cellcolor{gray!30}\textbf{(+0.93\%)}\end{tabular} &
  \begin{tabular}[c]{@{}c@{}}\cellcolor{gray!30}0.863\\ \cellcolor{gray!30}\textbf{(+2.49\%)}\end{tabular} &
  \begin{tabular}[c]{@{}c@{}}0.409\\ (-2.85\%)\end{tabular} &
  \begin{tabular}[c]{@{}c@{}}0.521\\ (-8.27\%)\end{tabular} &
  \begin{tabular}[c]{@{}c@{}}\cellcolor{gray!30}0.663\\ \cellcolor{gray!30}\textbf{(+5.24\%)}\end{tabular} &
  \begin{tabular}[c]{@{}c@{}}0.240\\ (-3.61\%)\end{tabular} &
  \begin{tabular}[c]{@{}c@{}}0.949\\ (-0.42\%)\end{tabular} &
  \begin{tabular}[c]{@{}c@{}}\cellcolor{gray!30}0.711\\ \cellcolor{gray!30}\textbf{(+6.44\%)}\end{tabular} &
  \begin{tabular}[c]{@{}c@{}}\cellcolor{gray!30}0.709\\ \cellcolor{gray!30}\textbf{(+3.20\%)}\end{tabular} \\
  \hline
\begin{tabular}[c]{@{}c@{}}FMAE-B\\ (+SFER)\end{tabular} &
  \begin{tabular}[c]{@{}c@{}}\cellcolor{gray!30}0.824\\ \cellcolor{gray!30}\textbf{(+1.10\%)}\end{tabular} &
  \begin{tabular}[c]{@{}c@{}}\cellcolor{gray!30}0.716\\ \cellcolor{gray!30}\textbf{(+2.14\%)}\end{tabular} &
  \begin{tabular}[c]{@{}c@{}}\cellcolor{gray!30}0.714\\ \cellcolor{gray!30}\textbf{(+5.93\%)}\end{tabular} &
  \begin{tabular}[c]{@{}c@{}}\cellcolor{gray!30}0.796\\ \cellcolor{gray!30}\textbf{(+2.98\%)}\end{tabular} &
  \begin{tabular}[c]{@{}c@{}}\cellcolor{gray!30}0.781\\ \cellcolor{gray!30}\textbf{(+1.43\%)}\end{tabular} &
  \begin{tabular}[c]{@{}c@{}}0.775\\ (-3.25\%)\end{tabular} &
  \begin{tabular}[c]{@{}c@{}}0.750\\ (-0.79\%)\end{tabular} &
  \begin{tabular}[c]{@{}c@{}}\cellcolor{gray!30}0.847\\ \cellcolor{gray!30}\textbf{(+0.59\%)}\end{tabular} &
  \begin{tabular}[c]{@{}c@{}}\cellcolor{gray!30}0.463\\ \cellcolor{gray!30}\textbf{(+9.98\%)}\end{tabular} &
  \begin{tabular}[c]{@{}c@{}}0.524\\ (-7.75\%)\end{tabular} &
  \begin{tabular}[c]{@{}c@{}}\cellcolor{gray!30}0.647\\ \cellcolor{gray!30}\textbf{(+2.70\%)}\end{tabular} &
  \begin{tabular}[c]{@{}c@{}}0.241\\ (-3.21\%)\end{tabular} &
  \begin{tabular}[c]{@{}c@{}}0.933\\ (-2.10\%)\end{tabular} &
  \begin{tabular}[c]{@{}c@{}}0.654\\ (-2.10\%)\end{tabular} &
  \begin{tabular}[c]{@{}c@{}}\cellcolor{gray!30}0.690\\ \cellcolor{gray!30}\textbf{(+0.44\%)}\end{tabular} \\
  \hline
\begin{tabular}[c]{@{}c@{}}FMAE-B\\ (+VAIE)\end{tabular} &
  \begin{tabular}[c]{@{}c@{}}\cellcolor{gray!30}0.827\\ \cellcolor{gray!30}\textbf{(+1.47\%)}\end{tabular} &
  \begin{tabular}[c]{@{}c@{}}0.697\\ (-0.57\%)\end{tabular} &
  \begin{tabular}[c]{@{}c@{}}\cellcolor{gray!30}0.722\\ \cellcolor{gray!30}\textbf{(+7.12\%)}\end{tabular} &
  \begin{tabular}[c]{@{}c@{}}\cellcolor{gray!30}0.787\\ \cellcolor{gray!30}\textbf{(+1.81\%)}\end{tabular} &
  \begin{tabular}[c]{@{}c@{}}\cellcolor{gray!30}0.782\\ \cellcolor{gray!30}\textbf{(+1.56\%)}\end{tabular} &
  \begin{tabular}[c]{@{}c@{}}0.767\\ (-4.24\%)\end{tabular} &
  \begin{tabular}[c]{@{}c@{}}0.747\\ (-1.19\%)\end{tabular} &
  \begin{tabular}[c]{@{}c@{}}\cellcolor{gray!30}0.847\\ \cellcolor{gray!30}\textbf{(+0.59\%)}\end{tabular} &
  \begin{tabular}[c]{@{}c@{}}\cellcolor{gray!30}0.466\\ \cellcolor{gray!30}\textbf{(+10.69\%)}\end{tabular} &
  \begin{tabular}[c]{@{}c@{}}0.535\\ (-5.81\%)\end{tabular} &
  \begin{tabular}[c]{@{}c@{}}\cellcolor{gray!30}0.655\\ \cellcolor{gray!30}\textbf{(+3.97\%)}\end{tabular} &
  \begin{tabular}[c]{@{}c@{}}\cellcolor{gray!30}0.254\\ \cellcolor{gray!30}\textbf{(+2.01\%)}\end{tabular} &
  \begin{tabular}[c]{@{}c@{}}0.937\\ (-1.68\%)\end{tabular} &
  \begin{tabular}[c]{@{}c@{}}0.629\\ (-5.84\%)\end{tabular} &
  \begin{tabular}[c]{@{}c@{}}\cellcolor{gray!30}0.690\\ \cellcolor{gray!30}\textbf{(+0.44\%)}\end{tabular} \\
  \hline
\begin{tabular}[c]{@{}c@{}}FMAE-B\\ (+All)\end{tabular} &
  \begin{tabular}[c]{@{}c@{}}\cellcolor{gray!30}0.869\\ \cellcolor{gray!30}\textbf{(+6.63\%)}\end{tabular} &
  \begin{tabular}[c]{@{}c@{}}\cellcolor{gray!30}0.765\\ \cellcolor{gray!30}\textbf{(+9.13\%)}\end{tabular} &
  \begin{tabular}[c]{@{}c@{}}\cellcolor{gray!30}0.778\\ \cellcolor{gray!30}\textbf{(+15.43\%)}\end{tabular} &
  \begin{tabular}[c]{@{}c@{}}\cellcolor{gray!30}0.789\\ \cellcolor{gray!30}\textbf{(+2.07\%)}\end{tabular} &
  \begin{tabular}[c]{@{}c@{}}\cellcolor{gray!30}0.794\\ \cellcolor{gray!30}\textbf{(+3.12\%)}\end{tabular} &
  \begin{tabular}[c]{@{}c@{}}\cellcolor{gray!30}0.842\\ \cellcolor{gray!30}\textbf{(+5.12\%)}\end{tabular} &
  \begin{tabular}[c]{@{}c@{}}0.751\\ (-0.66\%)\end{tabular} &
  \begin{tabular}[c]{@{}c@{}}\cellcolor{gray!30}0.850\\ \cellcolor{gray!30}\textbf{(+0.95\%)}\end{tabular} &
  \begin{tabular}[c]{@{}c@{}}0.393\\ (-6.65\%)\end{tabular} &
  \begin{tabular}[c]{@{}c@{}}0.546\\ (-3.87\%)\end{tabular} &
  \begin{tabular}[c]{@{}c@{}}\cellcolor{gray!30}0.678\\ \cellcolor{gray!30}\textbf{(+7.62\%)}\end{tabular} &
  \begin{tabular}[c]{@{}c@{}}0.233\\ (-6.43\%)\end{tabular} &
  \begin{tabular}[c]{@{}c@{}}0.953\\ (+0.00\%)\end{tabular} &
  \begin{tabular}[c]{@{}c@{}}\cellcolor{gray!30}0.692\\ \cellcolor{gray!30}\textbf{(+3.59\%)}\end{tabular} &
  \begin{tabular}[c]{@{}c@{}}\cellcolor{gray!30}0.710\\ \cellcolor{gray!30}\textbf{(+3.35\%)}\end{tabular} \\
  \Xhline{4.5px}
\end{tabular}
}
\label{tab:faue_sec}
\end{table*}

\noindent\textbf{Enhancement of the best-performing FAUE model.} Table \ref{tab:faue_sec} provides the AU-wise evidence underlying the FAUE results reported in the main paper. For MAE-Face-B, predictions from all three frame-level support functions improve the average ICC(3,1). FAUD provides the largest gain of 1.24\%, followed closely by VAIE with 1.10\%, while SFER produces a smaller improvement of 0.55\%. These results are consistent with the relationships identified in Fig. 5 of the main paper.

The improvement from FAUD is particularly pronounced for AU05, AU09, and AU20. Since FAUD explicitly indicates which AUs are activated, its predictions provide direct guidance for estimating their intensities. VAIE produces similarly large improvements for AU05 and AU20, indicating that continuous global facial affect intensities also contain complementary information for AU intensity estimation. Both support functions additionally improve AU02, AU04, AU12, AU15, and AU26. These shared improvements explain why FAUD and VAIE provide similar overall gains despite describing facial behaviour at different levels. SFER predictions also benefit AU05, AU15, and AU20, but their smaller average contribution suggests that categorical facial expression information provides less direct guidance for FAUE than AU occurrences or continuous affective states.

Using FAUD, SFER, and VAIE predictions together produces the same 1.24\% average improvement as FAUD alone. Since the joint setting additionally requires inference from the SFER and VAIE models before applying SMTE, it provides no further performance benefit while substantially increasing the overall computation. This result is consistent with the multi-function interaction observed for FAUD and further supports selecting only the most effective support function.

\noindent \textbf{Enhancement of the second-best FAUE model.} 
The results obtained with FMAE-B preserve the main relationship observed for MAE-Face-B. FAUD remains the most effective support function and improves the average ICC(3,1) by 3.20\%, whereas SFER and VAIE each provide a gain of 0.44\%. The consistent advantage of FAUD across both target models confirms that AU occurrence predictions provide informative guidance for AU intensity estimation and that this relationship is not limited to the best-performing FAUE model.

The magnitude and AU-wise distribution of the improvements, however, change considerably. FMAE-B benefits more strongly from FAUD than MAE-Face-B, particularly on AU01, AU02, AU04, AU17, and AU26. In contrast, FAUD improves AU15 and AU20 for MAE-Face-B but reduces their performance for FMAE-B. Similar model-dependent changes appear under SFER and VAIE support. Both predictions substantially improve AU14 for FMAE-B, whereas they provide no benefit to the same AU for MAE-Face-B. VAIE also produces a smaller average gain for FMAE-B, despite remaining beneficial for several individual AUs. These differences show that FAUD-to-FAUE is the most consistent cross-function relationship, while the specific AUs enhanced by each support function depend on the representations learned by the target model.

Jointly incorporating all three support functions improves FMAE-B by 3.35\%, only slightly exceeding the 3.20\% gain provided by FAUD alone and changing the average ICC(3,1) from 0.709 to 0.710. This marginal difference does not compensate for the additional computation required to perform inference with all three support models, leading to the same model-selection conclusion as for MAE-Face-B.

\subsubsection{SFER enhancement results}

\begin{table*}[htbp]
\centering
\setlength{\doublerulesep}{0.3em}
\renewcommand{\arraystretch}{1.4}
\caption{Dataset-specific accuracy and F1 scores (\%) of the baseline and SMTE-enhanced SFER models. Results are reported for the best-performing FMAE-L model and a representative intermediate-performing MAE-Face model used to examine the influence of baseline performance on cross-function enhancement. FAUD, FAUE, and VAIE denote the support functions whose predictions are incorporated by SMTE, while All denotes their joint use. Values in parentheses indicate the relative metric changes compared with the corresponding SFER baseline. Positive changes are highlighted in bold.}
\resizebox{1.0\linewidth}{!}{
\begin{tabular}{c|cccccccccc}
\Xhline{1.2px}
\rowcolor{gray!15}  
\cellcolor{gray!15}  &
  \multicolumn{2}{c}{\cellcolor{gray!15} \textbf{AffectNet-8}} &
  \multicolumn{2}{c}{\cellcolor{gray!15} \textbf{RAF-DB}} &
  \multicolumn{2}{c}{\cellcolor{gray!15} \textbf{FER+}} &
  \multicolumn{2}{c}{\cellcolor{gray!15} \textbf{SFEW}} &
  \multicolumn{2}{c}{\cellcolor{gray!15} \textbf{AFFW-FER}} \\ \hhline{~--||--||--||--||--}
\rowcolor{gray!15} 
\multirow{-2}{*}{\cellcolor{gray!15}\textbf{Method}} &
  Acc &
  F1 &
  Acc &
  F1 &
  Acc &
  F1 &
  Acc &
  F1 &
  Acc &
  F1 \\ \hhline{===========}
FMAE-L &
  63.72 &
  63.60 &
  93.22 &
  88.99 &
  92.23 &
  84.35 &
  62.38 &
  56.96 &
  70.92 &
  43.28 \\ \hline
\begin{tabular}[c]{@{}c@{}}FMAE-L\\ (+FAUD)\end{tabular} &
  \begin{tabular}[c]{@{}c@{}}\cellcolor{gray!30}64.38\\ \cellcolor{gray!30}\textbf{(+1.04\%)}\end{tabular} &
  \begin{tabular}[c]{@{}c@{}}\cellcolor{gray!30}64.30\\ \cellcolor{gray!30}\textbf{(+1.10\%)}\end{tabular} &
  \begin{tabular}[c]{@{}c@{}}92.83\\ (-0.42\%)\end{tabular} &
  \begin{tabular}[c]{@{}c@{}}88.27\\ (-0.81\%)\end{tabular} &
  \begin{tabular}[c]{@{}c@{}}91.94\\ (-0.31\%)\end{tabular} &
  \begin{tabular}[c]{@{}c@{}}83.67\\ (-0.81\%)\end{tabular} &
  \begin{tabular}[c]{@{}c@{}}61.39\\ (-1.59\%)\end{tabular} &
  \begin{tabular}[c]{@{}c@{}}55.64\\ (-2.32\%)\end{tabular} &
  \begin{tabular}[c]{@{}c@{}}70.51\\ (-0.58\%)\end{tabular} &
  \begin{tabular}[c]{@{}c@{}}42.47\\ (-1.87\%)\end{tabular} \\ \hline
\begin{tabular}[c]{@{}c@{}}FMAE-L\\ (+FAUE)\end{tabular} &
  \begin{tabular}[c]{@{}c@{}}\cellcolor{gray!30}64.13\\ \cellcolor{gray!30}\textbf{(+0.64\%)}\end{tabular} &
  \begin{tabular}[c]{@{}c@{}}\cellcolor{gray!30}64.05\\ \cellcolor{gray!30}\textbf{(+0.71\%)}\end{tabular} &
  \begin{tabular}[c]{@{}c@{}}92.89\\ (-0.35\%)\end{tabular} &
  \begin{tabular}[c]{@{}c@{}}88.50\\ (-0.55\%)\end{tabular} &
  \begin{tabular}[c]{@{}c@{}}91.98\\ (-0.27\%)\end{tabular} &
  \begin{tabular}[c]{@{}c@{}}83.66\\ (-0.82\%)\end{tabular} &
  \begin{tabular}[c]{@{}c@{}}62.13\\ (-0.40\%)\end{tabular} &
  \begin{tabular}[c]{@{}c@{}}56.22\\ (-1.30\%)\end{tabular} &
  \begin{tabular}[c]{@{}c@{}}70.49\\ (-0.61\%)\end{tabular} &
  \begin{tabular}[c]{@{}c@{}}42.70\\ (-1.34\%)\end{tabular} \\ \hline
\begin{tabular}[c]{@{}c@{}}FMAE-L\\ (+VAIE)\end{tabular} &
  \begin{tabular}[c]{@{}c@{}}\cellcolor{gray!30}64.35\\ \cellcolor{gray!30}\textbf{(+0.99\%)}\end{tabular} &
  \begin{tabular}[c]{@{}c@{}}\cellcolor{gray!30}64.27\\ \cellcolor{gray!30}\textbf{(+1.05\%)}\end{tabular} &
  \begin{tabular}[c]{@{}c@{}}92.73\\ (-0.53\%)\end{tabular} &
  \begin{tabular}[c]{@{}c@{}}88.13\\ (-0.97\%)\end{tabular} &
  \begin{tabular}[c]{@{}c@{}}91.98\\ (-0.27\%)\end{tabular} &
  \begin{tabular}[c]{@{}c@{}}84.06\\ (-0.34\%)\end{tabular} &
  \begin{tabular}[c]{@{}c@{}}\cellcolor{gray!30}63.61\\ \cellcolor{gray!30}\textbf{(+1.93\%)}\end{tabular} &
  \begin{tabular}[c]{@{}c@{}}\cellcolor{gray!30}58.10\\ \cellcolor{gray!30}\textbf{(+2.00\%)}\end{tabular} &
  \begin{tabular}[c]{@{}c@{}}70.70\\ (-0.31\%)\end{tabular} &
  \begin{tabular}[c]{@{}c@{}}42.88\\ (-0.92\%)\end{tabular} \\ \hline
\begin{tabular}[c]{@{}c@{}}FMAE-L\\ (+All)\end{tabular} &
  \begin{tabular}[c]{@{}c@{}}\cellcolor{gray!30}64.40\\ \cellcolor{gray!30}\textbf{(+1.07\%)}\end{tabular} &
  \begin{tabular}[c]{@{}c@{}}\cellcolor{gray!30}64.30\\ \cellcolor{gray!30}\textbf{(+1.10\%)}\end{tabular} &
  \begin{tabular}[c]{@{}c@{}}92.93\\ (-0.31\%)\end{tabular} &
  \begin{tabular}[c]{@{}c@{}}88.47\\ (-0.58\%)\end{tabular} &
  \begin{tabular}[c]{@{}c@{}}91.85\\ (-0.41\%)\end{tabular} &
  \begin{tabular}[c]{@{}c@{}}83.48\\ (-1.03\%)\end{tabular} &
  \begin{tabular}[c]{@{}c@{}}61.88\\ (-0.80\%)\end{tabular} &
  \begin{tabular}[c]{@{}c@{}}56.01\\ (-1.67\%)\end{tabular} &
  \begin{tabular}[c]{@{}c@{}}70.73\\ (-0.27\%)\end{tabular} &
  \begin{tabular}[c]{@{}c@{}}42.92\\ (-0.83\%)\end{tabular} \\ \hhline{===========}
MAE-Face &
  63.63 &
  63.57 &
  92.89 &
  88.85 &
  91.37 &
  81.93 &
  55.20 &
  50.61 &
  72.78 &
  50.70 \\ \hline
\begin{tabular}[c]{@{}c@{}}MAE-Face\\ (+FAUD)\end{tabular} &
  \begin{tabular}[c]{@{}c@{}}\cellcolor{gray!30}64.73\\ \cellcolor{gray!30}\textbf{(+1.73\%)}\end{tabular} &
  \begin{tabular}[c]{@{}c@{}}\cellcolor{gray!30}64.71\\ \cellcolor{gray!30}\textbf{(+1.79\%)}\end{tabular} &
  \begin{tabular}[c]{@{}c@{}}92.86\\ (-0.03\%)\end{tabular} &
  \begin{tabular}[c]{@{}c@{}}88.70\\ (-0.17\%)\end{tabular} &
  \begin{tabular}[c]{@{}c@{}}\cellcolor{gray!30}91.50\\ \cellcolor{gray!30}\textbf{(+0.14\%)}\end{tabular} &
  \begin{tabular}[c]{@{}c@{}}\cellcolor{gray!30}82.42\\ \cellcolor{gray!30}\textbf{(+0.60\%)}\end{tabular} &
  \begin{tabular}[c]{@{}c@{}}\cellcolor{gray!30}58.91\\ \cellcolor{gray!30}\textbf{(+6.72\%)}\end{tabular} &
  \begin{tabular}[c]{@{}c@{}}\cellcolor{gray!30}52.76\\ \cellcolor{gray!30}\textbf{(+4.25\%)}\end{tabular} &
  \begin{tabular}[c]{@{}c@{}}\cellcolor{gray!30}72.90\\ \cellcolor{gray!30}\textbf{(+0.16\%)}\end{tabular} &
  \begin{tabular}[c]{@{}c@{}}49.77\\ (-1.83\%)\end{tabular} \\ \hline
\begin{tabular}[c]{@{}c@{}}MAE-Face\\ (+FAUE)\end{tabular} &
  \begin{tabular}[c]{@{}c@{}}\cellcolor{gray!30}64.62\\ \cellcolor{gray!30}\textbf{(+1.56\%)}\end{tabular} &
  \begin{tabular}[c]{@{}c@{}}\cellcolor{gray!30}64.59\\ \cellcolor{gray!30}\textbf{(+1.60\%)}\end{tabular} &
  \begin{tabular}[c]{@{}c@{}}92.83\\ (-0.06\%)\end{tabular} &
  \begin{tabular}[c]{@{}c@{}}88.72\\ (-0.15\%)\end{tabular} &
  \begin{tabular}[c]{@{}c@{}}\cellcolor{gray!30}91.47\\ \cellcolor{gray!30}\textbf{(+0.11\%)}\end{tabular} &
  \begin{tabular}[c]{@{}c@{}}\cellcolor{gray!30}82.62\\ \cellcolor{gray!30}\textbf{(+0.84\%)}\end{tabular} &
  \begin{tabular}[c]{@{}c@{}}\cellcolor{gray!30}57.43\\ \cellcolor{gray!30}\textbf{(+4.04\%)}\end{tabular} &
  \begin{tabular}[c]{@{}c@{}}\cellcolor{gray!30}51.91\\ \cellcolor{gray!30}\textbf{(+2.57\%)}\end{tabular} &
  \begin{tabular}[c]{@{}c@{}}72.39\\ (-0.54\%)\end{tabular} &
  \begin{tabular}[c]{@{}c@{}}48.87\\ (-3.61\%)\end{tabular} \\ \hline
\begin{tabular}[c]{@{}c@{}}MAE-Face\\ (+VAIE)\end{tabular} &
  \begin{tabular}[c]{@{}c@{}}\cellcolor{gray!30}64.65\\ \cellcolor{gray!30}\textbf{(+1.60\%)}\end{tabular} &
  \begin{tabular}[c]{@{}c@{}}\cellcolor{gray!30}64.59\\ \cellcolor{gray!30}\textbf{(+1.60\%)}\end{tabular} &
  \begin{tabular}[c]{@{}c@{}}92.73\\ (-0.17\%)\end{tabular} &
  \begin{tabular}[c]{@{}c@{}}88.48\\ (-0.42\%)\end{tabular} &
  \begin{tabular}[c]{@{}c@{}}\cellcolor{gray!30}91.50\\ \cellcolor{gray!30}\textbf{(+0.14\%)}\end{tabular} &
  \begin{tabular}[c]{@{}c@{}}\cellcolor{gray!30}82.04\\ \cellcolor{gray!30}\textbf{(+0.13\%)}\end{tabular} &
  \begin{tabular}[c]{@{}c@{}}\cellcolor{gray!30}58.66\\ \cellcolor{gray!30}\textbf{(+6.27\%)}\end{tabular} &
  \begin{tabular}[c]{@{}c@{}}\cellcolor{gray!30}52.68\\ \cellcolor{gray!30}\textbf{(+4.09\%)}\end{tabular} &
  \begin{tabular}[c]{@{}c@{}}\cellcolor{gray!30}73.61\\ \cellcolor{gray!30}\textbf{(+1.14\%)}\end{tabular} &
  \begin{tabular}[c]{@{}c@{}}50.09\\ (-1.20\%)\end{tabular} \\ \hline
\begin{tabular}[c]{@{}c@{}}MAE-Face\\ (+All)\end{tabular} &
  \begin{tabular}[c]{@{}c@{}}\cellcolor{gray!30}64.70\\ \cellcolor{gray!30}\textbf{(+1.68\%)}\end{tabular} &
  \begin{tabular}[c]{@{}c@{}}\cellcolor{gray!30}64.63\\ \cellcolor{gray!30}\textbf{(+1.67\%)}\end{tabular} &
  \begin{tabular}[c]{@{}c@{}}\cellcolor{gray!30}92.99\\ \cellcolor{gray!30}\textbf{(+0.11\%)}\end{tabular} &
  \begin{tabular}[c]{@{}c@{}}\cellcolor{gray!30}88.91\\ \cellcolor{gray!30}\textbf{(+0.07\%)}\end{tabular} &
  \begin{tabular}[c]{@{}c@{}}91.31\\ (-0.07\%)\end{tabular} &
  \begin{tabular}[c]{@{}c@{}}\cellcolor{gray!30}83.04\\ \cellcolor{gray!30}\textbf{(+1.35\%)}\end{tabular} &
  \begin{tabular}[c]{@{}c@{}}\cellcolor{gray!30}60.40\\ \cellcolor{gray!30}\textbf{(+9.42\%)}\end{tabular} &
  \begin{tabular}[c]{@{}c@{}}\cellcolor{gray!30}54.68\\ \cellcolor{gray!30}\textbf{(+8.04\%)}\end{tabular} &
  \begin{tabular}[c]{@{}c@{}}\cellcolor{gray!30}73.49\\ \cellcolor{gray!30}\textbf{(+0.98\%)}\end{tabular} &
  \begin{tabular}[c]{@{}c@{}}50.07\\ (-1.24\%)\end{tabular} 
  \\ \Xhline{1.2px}
\end{tabular}
}
\label{tab:sfer_sec}
\end{table*}

\noindent\textbf{Enhancement of the best-performing SFER model.}
Table \ref{tab:sfer_sec} provides the dataset-specific evidence underlying the SFER results reported in the main paper. For the best-performing FMAE-L model, FAUD and FAUE predictions reduce the average accuracy by 0.37\% and 0.22\%, respectively. Both support functions improve the results on AffectNet-8 but provide no benefit on the remaining four datasets. This result does not necessarily indicate that AU occurrences and intensities are unrelated to facial expression recognition. Instead, the strong FMAE-L baseline may already capture most of the AU-related facial patterns contained in FAUD and FAUE predictions, leaving limited complementary information that can be further exploited by the current SMTE mechanism.

VAIE is the only individual support function that improves the average performance of FMAE-L, providing an accuracy gain of 0.24\%. Its predictions improve both accuracy and F1 score on AffectNet-8 and SFEW, with the largest gains appearing on the more challenging SFEW dataset. Unlike the local AU information provided by FAUD and FAUE, VAIE describes continuous global affective states that are strongly associated with categorical facial expressions. The performance gain from VAIE therefore indicates that it contains the most informative complementary cues for the best-performing SFER model among the evaluated frame-level support functions.

Jointly incorporating FAUD, FAUE, and VAIE predictions does not improve the average performance of FMAE-L and additionally requires inference from all three support models. This result is consistent with the observations for FAUD and FAUE: using multiple support functions introduces substantially greater computation but does not consistently provide additional performance gains.

\noindent\textbf{Enhancement of another representative SFER model.} 
To further examine whether the limited improvements above arise from the strong FMAE-L baseline, we repeat the same interactions using MAE-Face, which provides more moderate SFER performance. In contrast to FMAE-L, MAE-Face benefits from each of the three support functions. Averaged across the five dataset-level relative accuracy changes, FAUD and FAUE improve performance by 1.74\% and 1.02\%, respectively, while VAIE produces the largest individual gain of 1.80\%. The improvements are mainly observed on AffectNet-8 and SFEW, where all three support functions increase both accuracy and F1 score.

These results clarify the model-dependent effects observed for FMAE-L. AU occurrence and intensity predictions can provide useful information for SFER when the target model has not already captured the associated local facial patterns. Their contribution becomes limited for the stronger FMAE-L model, whereas the global affective information supplied by VAIE remains complementary to both SFER models. The consistent advantage of VAIE therefore supports the close relationship between continuous valence-arousal states and categorical facial expressions identified in the main paper.

Using all three support functions increases the mean dataset-level relative accuracy of MAE-Face by 2.42\%, with the largest improvement appearing on SFEW. Although this result shows that multiple support predictions can be complementary when the target baseline leaves greater room for improvement, it requires all three support models to perform inference before SMTE. The resulting gain remains modest relative to the additional computational cost.

\subsubsection{VAIE enhancement results}

\begin{table*}[htbp]
\centering
\large
\setlength{\doublerulesep}{0.3em}
\renewcommand{\arraystretch}{1.4}
\caption{Dataset-specific valence, arousal, and mean CCC results of the baseline and SMTE-enhanced VAIE models. Results are reported for the best-performing MAE-Face model and the second-best FMAE-L model. FAUD, FAUE, and SFER denote the support functions whose predictions are incorporated by SMTE, while All denotes their joint use. Values in parentheses indicate the relative metric changes compared with the corresponding VAIE baseline. Positive changes are highlighted in bold.}
\resizebox{1.0\linewidth}{!}{
\begin{tabular}{c|cccccccccccc}
\Xhline{2px}
\rowcolor{gray!15} 
\cellcolor{gray!15} &
  \multicolumn{3}{c}{\cellcolor{gray!15}\textbf{AFFW-VA}} &
  \multicolumn{3}{c}{\cellcolor{gray!15}\textbf{AFEW-VA}} &
  \multicolumn{3}{c}{\cellcolor{gray!15}\textbf{SEWA}} &
  \multicolumn{3}{c}{\cellcolor{gray!15}\textbf{AffectNet}} \\ \hhline{~---||---||---||---}
\rowcolor{gray!15}
\multirow{-2}{*}{\cellcolor{gray!15}\textbf{Method}} &
  Valence &
  Arousal &
  Mean &
  Valence &
  Arousal &
  Mean &
  Valence &
  Arousal &
  Mean &
  Valence &
  Arousal &
  Mean \\ \hhline{=============}
MAE-Face &
  0.392 &
  0.546 &
  0.469 &
  0.665 &
  0.626 &
  0.646 &
  0.805 &
  0.665 &
  0.735 &
  0.636 &
  0.566 &
  0.601 \\ \hline
\begin{tabular}[c]{@{}c@{}}MAE-Face\\ (+FAUD)\end{tabular} &
  \begin{tabular}[c]{@{}c@{}}\cellcolor{gray!30}0.438\\ \cellcolor{gray!30}\textbf{(+11.73\%)}\end{tabular} &
  \begin{tabular}[c]{@{}c@{}}\cellcolor{gray!30}0.552\\ \cellcolor{gray!30}\textbf{(+1.10\%)}\end{tabular} &
  \begin{tabular}[c]{@{}c@{}}\cellcolor{gray!30}0.495\\ \cellcolor{gray!30}\textbf{(+5.54\%)}\end{tabular} &
  \begin{tabular}[c]{@{}c@{}}0.641\\ (-3.61\%)\end{tabular} &
  \begin{tabular}[c]{@{}c@{}}0.586\\ (-6.39\%)\end{tabular} &
  \begin{tabular}[c]{@{}c@{}}0.613\\ (-5.11\%)\end{tabular} &
  \begin{tabular}[c]{@{}c@{}}0.793\\ (-1.49\%)\end{tabular} &
  \begin{tabular}[c]{@{}c@{}}0.651\\ (-2.11\%)\end{tabular} &
  \begin{tabular}[c]{@{}c@{}}0.722\\ (-1.77\%)\end{tabular} &
  \begin{tabular}[c]{@{}c@{}}0.595\\ (-6.45\%)\end{tabular} &
  \begin{tabular}[c]{@{}c@{}}0.553\\ (-2.30\%)\end{tabular} &
  \begin{tabular}[c]{@{}c@{}}0.574\\ (-4.49\%)\end{tabular} \\ \hline
\begin{tabular}[c]{@{}c@{}}MAE-Face\\ (+FAUE)\end{tabular} &
  \begin{tabular}[c]{@{}c@{}}\cellcolor{gray!30}0.412\\ \cellcolor{gray!30}\textbf{(+5.10\%)}\end{tabular} &
  \begin{tabular}[c]{@{}c@{}}0.539\\ (-1.28\%)\end{tabular} &
  \begin{tabular}[c]{@{}c@{}}\cellcolor{gray!30}0.475\\ \cellcolor{gray!30}\textbf{(+1.28\%)}\end{tabular} &
  \begin{tabular}[c]{@{}c@{}}0.630\\ (-5.26\%)\end{tabular} &
  \begin{tabular}[c]{@{}c@{}}\cellcolor{gray!30}0.628\\ \cellcolor{gray!30}\textbf{(+0.32\%)}\end{tabular} &
  \begin{tabular}[c]{@{}c@{}}0.629\\ (-2.63\%)\end{tabular} &
  \begin{tabular}[c]{@{}c@{}}0.780\\ (-3.11\%)\end{tabular} &
  \begin{tabular}[c]{@{}c@{}}0.652\\ (-1.95\%)\end{tabular} &
  \begin{tabular}[c]{@{}c@{}}0.716\\ (-2.59\%)\end{tabular} &
  \begin{tabular}[c]{@{}c@{}}0.597\\ (-6.13\%)\end{tabular} &
  \begin{tabular}[c]{@{}c@{}}0.544\\ (-3.89\%)\end{tabular} &
  \begin{tabular}[c]{@{}c@{}}0.571\\ (-4.99\%)\end{tabular} \\ \hline
\begin{tabular}[c]{@{}c@{}}MAE-Face\\ (+SFER)\end{tabular} &
  \begin{tabular}[c]{@{}c@{}}\cellcolor{gray!30}0.481\\ \cellcolor{gray!30}\textbf{(+22.70\%)}\end{tabular} &
  \begin{tabular}[c]{@{}c@{}}0.543\\ (-0.55\%)\end{tabular} &
  \begin{tabular}[c]{@{}c@{}}\cellcolor{gray!30}0.512\\ \cellcolor{gray!30}\textbf{(+9.17\%)}\end{tabular} &
  \begin{tabular}[c]{@{}c@{}}\cellcolor{gray!30}0.673\\ \cellcolor{gray!30}\textbf{(+1.20\%)}\end{tabular} &
  \begin{tabular}[c]{@{}c@{}}\cellcolor{gray!30}0.630\\ \cellcolor{gray!30}\textbf{(+0.64\%)}\end{tabular} &
  \begin{tabular}[c]{@{}c@{}}\cellcolor{gray!30}0.652\\ \cellcolor{gray!30}\textbf{(+0.93\%)}\end{tabular} &
  \begin{tabular}[c]{@{}c@{}}0.780\\ (-3.11\%)\end{tabular} &
  \begin{tabular}[c]{@{}c@{}}0.663\\ (-0.30\%)\end{tabular} &
  \begin{tabular}[c]{@{}c@{}}0.721\\ (-1.90\%)\end{tabular} &
  \begin{tabular}[c]{@{}c@{}}0.617\\ (-2.99\%)\end{tabular} &
  \begin{tabular}[c]{@{}c@{}}\cellcolor{gray!30}0.581\\ \cellcolor{gray!30}\textbf{(+2.65\%)}\end{tabular} &
  \begin{tabular}[c]{@{}c@{}}0.599\\ (-0.33\%)\end{tabular} \\ \hline
\begin{tabular}[c]{@{}c@{}}MAE-Face\\ (+All)\end{tabular} &
  \begin{tabular}[c]{@{}c@{}}\cellcolor{gray!30}0.451\\ \cellcolor{gray!30}\textbf{(+15.05\%)}\end{tabular} &
  \begin{tabular}[c]{@{}c@{}}0.537\\ (-1.65\%)\end{tabular} &
  \begin{tabular}[c]{@{}c@{}}\cellcolor{gray!30}0.494\\ \cellcolor{gray!30}\textbf{(+5.33\%)}\end{tabular} &
  \begin{tabular}[c]{@{}c@{}}0.654\\ (-1.65\%)\end{tabular} &
  \begin{tabular}[c]{@{}c@{}}\cellcolor{gray!30}0.629\\ \cellcolor{gray!30}\textbf{(+0.48\%)}\end{tabular} &
  \begin{tabular}[c]{@{}c@{}}0.642\\ (-0.62\%)\end{tabular} &
  \begin{tabular}[c]{@{}c@{}}0.786\\ (-2.36\%)\end{tabular} &
  \begin{tabular}[c]{@{}c@{}}0.664\\ (-0.15\%)\end{tabular} &
  \begin{tabular}[c]{@{}c@{}}0.725\\ (-1.36\%)\end{tabular} &
  \begin{tabular}[c]{@{}c@{}}0.607\\ (-4.56\%)\end{tabular} &
  \begin{tabular}[c]{@{}c@{}}0.565\\ (-0.18\%)\end{tabular} &
  \begin{tabular}[c]{@{}c@{}}0.586\\ (-2.50\%)\end{tabular} \\ \hhline{=============}
FMAE-L &
  0.382 &
  0.569 &
  0.475 &
  0.591 &
  0.590 &
  0.591 &
  0.824 &
  0.691 &
  0.757 &
  0.580 &
  0.550 &
  0.565 \\ \hline
\begin{tabular}[c]{@{}c@{}}FMAE-L\\ (+FAUD)\end{tabular} &
  \begin{tabular}[c]{@{}c@{}}\cellcolor{gray!30}0.410\\ \cellcolor{gray!30}\textbf{(+7.33\%)}\end{tabular} &
  \begin{tabular}[c]{@{}c@{}}\cellcolor{gray!30}0.571\\ \cellcolor{gray!30}\textbf{(+0.35\%)}\end{tabular} &
  \begin{tabular}[c]{@{}c@{}}\cellcolor{gray!30}0.491\\ \cellcolor{gray!30}\textbf{(+3.37\%)}\end{tabular} &
  \begin{tabular}[c]{@{}c@{}}\cellcolor{gray!30}0.614\\ \cellcolor{gray!30}\textbf{(+3.89\%)}\end{tabular} &
  \begin{tabular}[c]{@{}c@{}}0.579\\ (-1.86\%)\end{tabular} &
  \begin{tabular}[c]{@{}c@{}}\cellcolor{gray!30}0.597\\ \cellcolor{gray!30}\textbf{(+1.02\%)}\end{tabular} &
  \begin{tabular}[c]{@{}c@{}}0.819\\ (-0.61\%)\end{tabular} &
  \begin{tabular}[c]{@{}c@{}}0.666\\ (-3.62\%)\end{tabular} &
  \begin{tabular}[c]{@{}c@{}}0.743\\ (-1.85\%)\end{tabular} &
  \begin{tabular}[c]{@{}c@{}}0.574\\ (-1.03\%)\end{tabular} &
  \begin{tabular}[c]{@{}c@{}}\cellcolor{gray!30}0.553\\ \cellcolor{gray!30}\textbf{(+0.55\%)}\end{tabular} &
  \begin{tabular}[c]{@{}c@{}}0.563\\ (-0.35\%)\end{tabular} \\ \hline
\begin{tabular}[c]{@{}c@{}}FMAE-L\\ (+FAUE)\end{tabular} &
  \begin{tabular}[c]{@{}c@{}}\cellcolor{gray!30}0.406\\ \cellcolor{gray!30}\textbf{(+6.28\%)}\end{tabular} &
  \begin{tabular}[c]{@{}c@{}}\cellcolor{gray!30}0.571\\ \cellcolor{gray!30}\textbf{(+0.35\%)}\end{tabular} &
  \begin{tabular}[c]{@{}c@{}}\cellcolor{gray!30}0.488\\ \cellcolor{gray!30}\textbf{(+2.74\%)}\end{tabular} &
  \begin{tabular}[c]{@{}c@{}}0.591\\ (+0.00\%)\end{tabular} &
  \begin{tabular}[c]{@{}c@{}}\cellcolor{gray!30}0.631\\ \cellcolor{gray!30}\textbf{(+6.95\%)}\end{tabular} &
  \begin{tabular}[c]{@{}c@{}}\cellcolor{gray!30}0.611\\ \cellcolor{gray!30}\textbf{(+3.38\%)}\end{tabular} &
  \begin{tabular}[c]{@{}c@{}}0.814\\ (-1.21\%)\end{tabular} &
  \begin{tabular}[c]{@{}c@{}}0.678\\ (-1.88\%)\end{tabular} &
  \begin{tabular}[c]{@{}c@{}}0.746\\ (-1.45\%)\end{tabular} &
  \begin{tabular}[c]{@{}c@{}}\cellcolor{gray!30}0.589\\ \cellcolor{gray!30}\textbf{(+1.55\%)}\end{tabular} &
  \begin{tabular}[c]{@{}c@{}}0.545\\ (-0.91\%)\end{tabular} &
  \begin{tabular}[c]{@{}c@{}}\cellcolor{gray!30}0.567\\ \cellcolor{gray!30}\textbf{(+0.35\%)}\end{tabular} \\ \hline
\begin{tabular}[c]{@{}c@{}}FMAE-L\\ (+SFER)\end{tabular} &
  \begin{tabular}[c]{@{}c@{}}\cellcolor{gray!30}0.470\\ \cellcolor{gray!30}\textbf{(+23.04\%)}\end{tabular} &
  \begin{tabular}[c]{@{}c@{}}0.549\\ (-3.51\%)\end{tabular} &
  \begin{tabular}[c]{@{}c@{}}\cellcolor{gray!30}0.509\\ \cellcolor{gray!30}\textbf{(+7.16\%)}\end{tabular} &
  \begin{tabular}[c]{@{}c@{}}\cellcolor{gray!30}0.633\\ \cellcolor{gray!30}\textbf{(+7.11\%)}\end{tabular} &
  \begin{tabular}[c]{@{}c@{}}\cellcolor{gray!30}0.639\\ \cellcolor{gray!30}\textbf{(+8.31\%)}\end{tabular} &
  \begin{tabular}[c]{@{}c@{}}\cellcolor{gray!30}0.636\\ \cellcolor{gray!30}\textbf{(+7.61\%)}\end{tabular} &
  \begin{tabular}[c]{@{}c@{}}0.805\\ (-2.31\%)\end{tabular} &
  \begin{tabular}[c]{@{}c@{}}0.678\\ (-1.88\%)\end{tabular} &
  \begin{tabular}[c]{@{}c@{}}0.741\\ (-2.11\%)\end{tabular} &
  \begin{tabular}[c]{@{}c@{}}\cellcolor{gray!30}0.606\\ \cellcolor{gray!30}\textbf{(+4.48\%)}\end{tabular} &
  \begin{tabular}[c]{@{}c@{}}\cellcolor{gray!30}0.578\\ \cellcolor{gray!30}\textbf{(+5.09\%)}\end{tabular} &
  \begin{tabular}[c]{@{}c@{}}\cellcolor{gray!30}0.592\\ \cellcolor{gray!30}\textbf{(+4.78\%)}\end{tabular} \\ \hline
\begin{tabular}[c]{@{}c@{}}FMAE-L\\ (+All)\end{tabular} &
  \begin{tabular}[c]{@{}c@{}}\cellcolor{gray!30}0.451\\ \cellcolor{gray!30}\textbf{(+18.06\%)}\end{tabular} &
  \begin{tabular}[c]{@{}c@{}}0.556\\ (-2.28\%)\end{tabular} &
  \begin{tabular}[c]{@{}c@{}}\cellcolor{gray!30}0.503\\ \cellcolor{gray!30}\textbf{(+5.89\%)}\end{tabular} &
  \begin{tabular}[c]{@{}c@{}}\cellcolor{gray!30}0.655\\ \cellcolor{gray!30}\textbf{(+10.83\%)}\end{tabular} &
  \begin{tabular}[c]{@{}c@{}}\cellcolor{gray!30}0.629\\ \cellcolor{gray!30}\textbf{(+6.61\%)}\end{tabular} &
  \begin{tabular}[c]{@{}c@{}}\cellcolor{gray!30}0.642\\ \cellcolor{gray!30}\textbf{(+8.63\%)}\end{tabular} &
  \begin{tabular}[c]{@{}c@{}}0.812\\ (-1.46\%)\end{tabular} &
  \begin{tabular}[c]{@{}c@{}}0.686\\ (-0.72\%)\end{tabular} &
  \begin{tabular}[c]{@{}c@{}}0.749\\ (-1.06\%)\end{tabular} &
  \begin{tabular}[c]{@{}c@{}}\cellcolor{gray!30}0.592\\ \cellcolor{gray!30}\textbf{\cellcolor{gray!30}(+2.07\%)}\end{tabular} &
  \begin{tabular}[c]{@{}c@{}}\cellcolor{gray!30}0.573\\ \cellcolor{gray!30}\textbf{(+4.18\%)}\end{tabular} &
  \begin{tabular}[c]{@{}c@{}}\cellcolor{gray!30}0.583\\ \cellcolor{gray!30}\textbf{(+3.19\%)}\end{tabular} \\
\Xhline{2px}
\end{tabular}
}
\label{tab:vaie_sec}
\end{table*}

\noindent\textbf{Enhancement of the best-performing VAIE model.}
Table \ref{tab:vaie_sec} expands the VAIE results reported in the main paper by presenting the effects of FAUD, FAUE, and SFER predictions on valence and arousal estimation across the four datasets. For the best-performing MAE-Face model, FAUD and FAUE reduce the average CCC by 1.96\% and 2.45\%, respectively. Although both support functions improve the results on AFFW-VA, their predictions provide no benefit on the remaining datasets. This limited enhancement is consistent with the strong MAE-Face baseline, which may already capture most of the AU-related facial information supplied by FAUD and FAUE, leaving little additional information for SMTE to exploit.

SFER is the only individual support function that improves the overall performance of MAE-Face, increasing its average CCC by 1.31\%. The improvement is most evident on AFFW-VA, where SFER substantially increases the valence CCC and produces a 9.17\% gain in the mean result. It also improves both valence and arousal on AFEW-VA. These results support the close relationship identified in the main paper between categorical facial expressions and continuous valence-arousal states. Among the evaluated frame-level functions, SFER therefore provides the most informative complementary cues for the best-performing VAIE model.

Jointly using FAUD, FAUE, and SFER does not improve the overall MAE-Face result and requires inference from all three support models. As observed for the preceding frame-level functions, the additional computation introduced by multi-function interaction does not lead to a corresponding performance benefit.

\noindent\textbf{Enhancement of the second-best VAIE model.}
The FMAE-L results show that the limited contribution of FAUD and FAUE to MAE-Face does not mean that AU-related predictions contain no complementary information for VAIE. FMAE-L has a lower average baseline performance and benefits from all three individual support functions. Averaged across the four dataset-level relative changes in mean CCC, FAUD improves performance by 0.55\%, while FAUE provides a larger gain of 1.26\%. Both functions improve FMAE-L on AFFW-VA, AFEW-VA, and AffectNet, indicating that AU occurrence and intensity predictions can support valence-arousal estimation when the target model has not already captured the corresponding local facial patterns.

SFER remains the most effective support function and provides an average improvement of 4.36\%. Its predictions improve the mean CCC by 7.16\% on AFFW-VA, 7.61\% on AFEW-VA, and 4.78\% on AffectNet. The consistent advantage of SFER across both VAIE models confirms that categorical facial expressions contain the most informative complementary cues for estimating continuous affective states. The results on SEWA nevertheless decrease under every support function, showing that these cross-function relationships remain dependent on the evaluated data distribution.

Combining the predictions of FAUD, FAUE, and SFER produces an average gain of 4.16\%, which is lower than the 4.36\% improvement obtained using SFER predictions alone. This finding provides further evidence that selecting the single most effective support function offers a more favourable balance between SMTE enhancement and computational cost.

\subsubsection{DFER enhancement results}

\begin{table*}[htbp]
\centering
\Huge
\setlength{\doublerulesep}{0.3em}
\renewcommand{\arraystretch}{1.4}
\caption{Dataset-specific UAR and WAR results of the baseline and SMTE-enhanced DFER models on DFEW, FERV39k, and MAFW. Results are reported for the best-performing FRU-Adapter-B model and the second-best SVFAP-B model. FAUD, FAUE, SFER, VAIE, FPLE, FDA, and FAPR denote the support functions whose prediction sequences are incorporated by SMTE, while All denotes their joint use. Values in parentheses indicate the relative metric changes compared with the corresponding DFER baseline. Positive changes are highlighted in bold.}
\resizebox{1.0\linewidth}{!}{
\begin{tabular}{c|cccccc|c|cccccc}
\Xhline{4.5px}
\rowcolor{gray!15} 
\cellcolor{gray!15} &
  \multicolumn{2}{c}{\cellcolor{gray!15}\textbf{DFEW}} &
  \multicolumn{2}{c}{\cellcolor{gray!15}\textbf{FERV39k}} &
  \multicolumn{2}{c|}{\cellcolor{gray!15}\textbf{MAFW}} &
  \cellcolor{gray!15} &
  \multicolumn{2}{c}{\cellcolor{gray!15}\textbf{DFEW}} &
  \multicolumn{2}{c}{\cellcolor{gray!15}\textbf{FERV39k}} &
  \multicolumn{2}{c}{\cellcolor{gray!15}\textbf{MAFW}} \\ \hhline{~--||--||--~--||--||--}
\rowcolor{gray!15} 
\multirow{-2}{*}{\cellcolor{gray!15}\textbf{Model}} &
  \cellcolor{gray!15}UAR &
  \cellcolor{gray!15}WAR &
  UAR &
  WAR &
  UAR &
  WAR &
  \multirow{-2}{*}{\cellcolor{gray!15}\textbf{Model}} &
  UAR &
  WAR &
  UAR &
  WAR &
  UAR &
  WAR \\ \hhline{==============}
SVFAP-B &
  66.06 &
  76.51 &
  46.05 &
  54.10 &
  37.47 &
  48.45 &
  FRU-Adapter-B &
  65.69 &
  77.45 &
  44.45 &
  52.21 &
  38.71 &
  50.57 \\ \hline
\begin{tabular}[c]{@{}c@{}}SVFAP-B\\ (+FAUD)\end{tabular} &
  \begin{tabular}[c]{@{}c@{}}66.03\\ (-0.05\%)\end{tabular} &
  \begin{tabular}[c]{@{}c@{}}76.38\\ (-0.17\%)\end{tabular} &
  \begin{tabular}[c]{@{}c@{}}\cellcolor{gray!30}46.16\\ \cellcolor{gray!30}\textbf{(+0.24\%)}\end{tabular} &
  \begin{tabular}[c]{@{}c@{}}53.73\\ (-0.68\%)\end{tabular} &
  \begin{tabular}[c]{@{}c@{}}\cellcolor{gray!30}37.87\\ \cellcolor{gray!30}\textbf{(+1.07\%)}\end{tabular} &
  \begin{tabular}[c]{@{}c@{}}48.23\\ (-0.45\%)\end{tabular} &
  \begin{tabular}[c]{@{}c@{}}FRU-Adapter-B\\ (+FAUD)\end{tabular} &
  \begin{tabular}[c]{@{}c@{}}65.38\\ (-0.47\%)\end{tabular} &
  \begin{tabular}[c]{@{}c@{}}77.40\\ (-0.06\%)\end{tabular} &
  \begin{tabular}[c]{@{}c@{}}43.93\\ (-1.17\%)\end{tabular} &
  \begin{tabular}[c]{@{}c@{}}52.20\\ (-0.02\%)\end{tabular} &
  \begin{tabular}[c]{@{}c@{}}\cellcolor{gray!30}39.50\\ \cellcolor{gray!30}\textbf{(+2.04\%)}\end{tabular} &
  \begin{tabular}[c]{@{}c@{}}\cellcolor{gray!30}51.44\\ \cellcolor{gray!30}\textbf{(+1.72\%)}\end{tabular} \\ \hline
\begin{tabular}[c]{@{}c@{}}SVFAP-B\\ (+FAUE)\end{tabular} &
  \begin{tabular}[c]{@{}c@{}}\cellcolor{gray!30}66.31\\ \cellcolor{gray!30}\textbf{(+0.38\%)}\end{tabular} &
  \begin{tabular}[c]{@{}c@{}}76.21\\ (-0.39\%)\end{tabular} &
  \begin{tabular}[c]{@{}c@{}}\cellcolor{gray!30}46.18\\ \cellcolor{gray!30}\textbf{(+0.28\%)}\end{tabular} &
  \begin{tabular}[c]{@{}c@{}}54.05\\ (-0.09\%)\end{tabular} &
  \begin{tabular}[c]{@{}c@{}}\cellcolor{gray!30}38.27\\ \cellcolor{gray!30}\textbf{(+2.14\%)}\end{tabular} &
  \begin{tabular}[c]{@{}c@{}}\cellcolor{gray!30}49.27\\ \cellcolor{gray!30}\textbf{(+1.69\%)}\end{tabular} &
  \begin{tabular}[c]{@{}c@{}}FRU-Adapter-B\\ (+FAUE)\end{tabular} &
  \begin{tabular}[c]{@{}c@{}}65.58\\ (-0.17\%)\end{tabular} &
  \begin{tabular}[c]{@{}c@{}}\cellcolor{gray!30}77.53\\ \cellcolor{gray!30}\textbf{(+0.10\%)}\end{tabular} &
  \begin{tabular}[c]{@{}c@{}}43.94\\ (-1.15\%)\end{tabular} &
  \begin{tabular}[c]{@{}c@{}}52.20\\ (-0.02\%)\end{tabular} &
  \begin{tabular}[c]{@{}c@{}}\cellcolor{gray!30}39.05\\ \cellcolor{gray!30}\textbf{(+0.88\%)}\end{tabular} &
  \begin{tabular}[c]{@{}c@{}}\cellcolor{gray!30}51.11\\ \cellcolor{gray!30}\textbf{(+1.07\%)}\end{tabular} \\ \hline
\begin{tabular}[c]{@{}c@{}}SVFAP-B\\ (+SFER)\end{tabular} &
  \begin{tabular}[c]{@{}c@{}}\cellcolor{gray!30}66.73\\ \cellcolor{gray!30}\textbf{(+1.01\%)}\end{tabular} &
  \begin{tabular}[c]{@{}c@{}}\cellcolor{gray!30}77.02\\ \cellcolor{gray!30}\textbf{(+0.67\%)}\end{tabular} &
  \begin{tabular}[c]{@{}c@{}}\cellcolor{gray!30}46.19\\ \cellcolor{gray!30}\textbf{(+0.30\%)}\end{tabular} &
  \begin{tabular}[c]{@{}c@{}}53.70\\ (-0.74\%)\end{tabular} &
  \begin{tabular}[c]{@{}c@{}}\cellcolor{gray!30}39.90\\ \cellcolor{gray!30}\textbf{(+6.49\%)}\end{tabular} &
  \begin{tabular}[c]{@{}c@{}}\cellcolor{gray!30}50.03\\ \cellcolor{gray!30}\textbf{(+3.26\%)}\end{tabular} &
  \begin{tabular}[c]{@{}c@{}}FRU-Adapter-B\\ (+SFER)\end{tabular} &
  \begin{tabular}[c]{@{}c@{}}65.29\\ (-0.61\%)\end{tabular} &
  \begin{tabular}[c]{@{}c@{}}\cellcolor{gray!30}77.57\\ \cellcolor{gray!30}\textbf{(+0.15\%)}\end{tabular} &
  \begin{tabular}[c]{@{}c@{}}44.12\\ (-0.71\%)\end{tabular} &
  \begin{tabular}[c]{@{}c@{}}\cellcolor{gray!30}52.35\\ \cellcolor{gray!30}\textbf{(+0.27\%)}\end{tabular} &
  \begin{tabular}[c]{@{}c@{}}\cellcolor{gray!30}39.33\\ \cellcolor{gray!30}\textbf{(+1.60\%)}\end{tabular} &
  \begin{tabular}[c]{@{}c@{}}\cellcolor{gray!30}50.95\\ \cellcolor{gray!30}\textbf{(+0.75\%)}\end{tabular} \\ \hline
\begin{tabular}[c]{@{}c@{}}SVFAP-B\\ (+VAIE)\end{tabular} &
  \begin{tabular}[c]{@{}c@{}}\cellcolor{gray!30}67.04\\ \cellcolor{gray!30}\textbf{(+1.48\%)}\end{tabular} &
  \begin{tabular}[c]{@{}c@{}}\cellcolor{gray!30}76.72\\ \cellcolor{gray!30}\textbf{(+0.27\%)}\end{tabular} &
  \begin{tabular}[c]{@{}c@{}}\cellcolor{gray!30}46.44\\ \cellcolor{gray!30}\textbf{(+0.85\%)}\end{tabular} &
  \begin{tabular}[c]{@{}c@{}}53.52\\ (-1.07\%)\end{tabular} &
  \begin{tabular}[c]{@{}c@{}}\cellcolor{gray!30}38.07\\ \cellcolor{gray!30}\textbf{(+1.60\%)}\end{tabular} &
  \begin{tabular}[c]{@{}c@{}}\cellcolor{gray!30}48.83\\ \cellcolor{gray!30}\textbf{(+0.78\%)}\end{tabular} &
  \begin{tabular}[c]{@{}c@{}}FRU-Adapter-B\\ (+VAIE)\end{tabular} &
  \begin{tabular}[c]{@{}c@{}}65.10\\ (-0.90\%)\end{tabular} &
  \begin{tabular}[c]{@{}c@{}}77.40\\ (-0.06\%)\end{tabular} &
  \begin{tabular}[c]{@{}c@{}}44.07\\ (-0.85\%)\end{tabular} &
  \begin{tabular}[c]{@{}c@{}}52.20\\ (-0.02\%)\end{tabular} &
  \begin{tabular}[c]{@{}c@{}}\cellcolor{gray!30}39.11\\ \cellcolor{gray!30}\textbf{(+1.03\%)}\end{tabular} &
  \begin{tabular}[c]{@{}c@{}}\cellcolor{gray!30}50.95\\ \cellcolor{gray!30}\textbf{(+0.75\%)}\end{tabular} \\ \hline
  \begin{tabular}[c]{@{}c@{}}SVFAP-B\\ (+FPLE)\end{tabular} &
  \begin{tabular}[c]{@{}c@{}}66.03\\(-0.05\%)\end{tabular}&
  \begin{tabular}[c]{@{}c@{}}76.33\\(-0.24\%)\end{tabular}&
  \begin{tabular}[c]{@{}c@{}}45.85\\ (-0.43\%)\end{tabular} &
  \begin{tabular}[c]{@{}c@{}}53.60\\ (-0.92\%)\end{tabular} &
  \begin{tabular}[c]{@{}c@{}}\cellcolor{gray!30}37.63\\
  \cellcolor{gray!30}\textbf{(+0.43\%)}\end{tabular}&
  \begin{tabular}[c]{@{}c@{}}\cellcolor{gray!30}48.58\\
  \cellcolor{gray!30}\textbf{(+0.27\%)}\end{tabular}&
  \begin{tabular}[c]{@{}c@{}}FRU-Adapter-B\\ (+FPLE)\end{tabular} &
  \begin{tabular}[c]{@{}c@{}}64.11\\(-2.41\%)\end{tabular} &
  \begin{tabular}[c]{@{}c@{}}77.10\\(-0.45\%)\end{tabular} &
  \begin{tabular}[c]{@{}c@{}}42.64\\(-4.07\%)\end{tabular} &
  \begin{tabular}[c]{@{}c@{}}\cellcolor{gray!30}52.63\\
  \cellcolor{gray!30}\textbf{(+0.80\%)}\end{tabular}&
  \begin{tabular}[c]{@{}c@{}}38.21\\(-1.29\%)\end{tabular} &
  \begin{tabular}[c]{@{}c@{}}50.35\\(-0.44\%)\end{tabular} \\ \hline
  \begin{tabular}[c]{@{}c@{}}SVFAP-B\\ (+FDA)\end{tabular} &
  \begin{tabular}[c]{@{}c@{}}\cellcolor{gray!30}66.29\\
  \cellcolor{gray!30}\textbf{(+0.35\%)}\end{tabular}&
  \begin{tabular}[c]{@{}c@{}}76.51\\(+0.00\%)\end{tabular} &
  \begin{tabular}[c]{@{}c@{}}\cellcolor{gray!30}46.27\\
  \cellcolor{gray!30}\textbf{(+0.48\%)}\end{tabular}&
  \begin{tabular}[c]{@{}c@{}}53.81\\(-0.54\%)\end{tabular} &
  \begin{tabular}[c]{@{}c@{}}37.45\\(-0.05\%)\end{tabular} &
  \begin{tabular}[c]{@{}c@{}}48.17\\(-0.58\%)\end{tabular} &
  \begin{tabular}[c]{@{}c@{}}FRU-Adapter-B\\ (+FDA)\end{tabular} &
  \begin{tabular}[c]{@{}c@{}}63.98\\(-2.60\%)\end{tabular} &
  \begin{tabular}[c]{@{}c@{}}77.27\\(-0.23\%)\end{tabular} &
  \begin{tabular}[c]{@{}c@{}}43.06\\(-3.13\%)\end{tabular} &
  \begin{tabular}[c]{@{}c@{}}\cellcolor{gray!30}52.45\\
  \cellcolor{gray!30}\textbf{(+0.46\%)}\end{tabular}&
  \begin{tabular}[c]{@{}c@{}}37.48\\(-3.18\%)\end{tabular} &
  \begin{tabular}[c]{@{}c@{}}49.86\\(-1.40\%)\end{tabular}
  \\ \hline
  \begin{tabular}[c]{@{}c@{}}SVFAP-B\\ (+FAPR)\end{tabular} &
  \begin{tabular}[c]{@{}c@{}}\cellcolor{gray!30}66.25\\
  \cellcolor{gray!30}\textbf{(+0.29\%)}\end{tabular}&
  \begin{tabular}[c]{@{}c@{}}76.51\\(+0.00\%)\end{tabular} &
  \begin{tabular}[c]{@{}c@{}}46.01\\(-0.09\%)\end{tabular} &
  \begin{tabular}[c]{@{}c@{}}53.23\\(-1.61\%)\end{tabular} &
  \begin{tabular}[c]{@{}c@{}}\cellcolor{gray!30}37.53\\
  \cellcolor{gray!30}\textbf{(+0.16\%)}\end{tabular}&
  \begin{tabular}[c]{@{}c@{}}48.43\\(-0.04\%)\end{tabular} &
  \begin{tabular}[c]{@{}c@{}}FRU-Adapter-B\\ (+FAPR)\end{tabular} &
  \begin{tabular}[c]{@{}c@{}}64.23\\(-2.22\%)\end{tabular} &
  \begin{tabular}[c]{@{}c@{}}77.45\\(+0.00\%)\end{tabular} &
  \begin{tabular}[c]{@{}c@{}}39.75\\(-10.57\%)\end{tabular} &
  \begin{tabular}[c]{@{}c@{}}52.19\\(-0.04\%)\end{tabular} &
  \begin{tabular}[c]{@{}c@{}}38.52\\(-0.49\%)\end{tabular} &
  \begin{tabular}[c]{@{}c@{}}\cellcolor{gray!30}50.62 \\
  \cellcolor{gray!30}\textbf{(+0.10\%)}\end{tabular}
  \\ \hline
\begin{tabular}[c]{@{}c@{}}SVFAP-B\\ (+All)\end{tabular} &
  \begin{tabular}[c]{@{}c@{}}\cellcolor{gray!30}66.71\\ \cellcolor{gray!30}\textbf{(+0.98\%)}\end{tabular} &
  \begin{tabular}[c]{@{}c@{}}\cellcolor{gray!30}77.06\\ \cellcolor{gray!30}\textbf{(+0.72\%)}\end{tabular} &
  \begin{tabular}[c]{@{}c@{}}\cellcolor{gray!30}46.13\\ \cellcolor{gray!30}\textbf{(+0.17\%)}\end{tabular} &
  \begin{tabular}[c]{@{}c@{}}53.85\\ (-0.46\%)\end{tabular} &
  \begin{tabular}[c]{@{}c@{}}\cellcolor{gray!30}40.17\\ \cellcolor{gray!30}\textbf{(+7.21\%)}\end{tabular} &
  \begin{tabular}[c]{@{}c@{}}\cellcolor{gray!30}49.70\\ \cellcolor{gray!30}\textbf{(+2.58\%)}\end{tabular} &
  \begin{tabular}[c]{@{}c@{}}FRU-Adapter-B\\ (+All)\end{tabular} &
  \begin{tabular}[c]{@{}c@{}}65.28\\ (-0.62\%)\end{tabular} &
  \begin{tabular}[c]{@{}c@{}}\cellcolor{gray!30}77.66\\ \cellcolor{gray!30}\textbf{(+0.27\%)}\end{tabular} &
  \begin{tabular}[c]{@{}c@{}}43.56\\ (-2.00\%)\end{tabular} &
  \begin{tabular}[c]{@{}c@{}}52.01\\ (-0.38\%)\end{tabular} &
  \begin{tabular}[c]{@{}c@{}}38.41\\ (-0.77\%)\end{tabular} &
  \begin{tabular}[c]{@{}c@{}}50.57\\ (+0.00\%)\end{tabular} \\
  \Xhline{4.5px}
\end{tabular}
}
\label{tab:dfer_sec}
\end{table*}

\noindent\textbf{Enhancement of the best-performing DFER model.}
Table \ref{tab:dfer_sec} further presents the dataset-specific effects of different OpenAHBU functions on FRU-Adapter-B. Consistent with the observations in the main paper, predictions provided by all four frame-level functions improve the average WAR across the three datasets. FAUD provides the largest relative performance gain of 0.55\%, followed by SFER, FAUE, and VAIE with gains of 0.39\%, 0.38\%, and 0.22\%, respectively. These improvements are mainly observed on MAFW, where all four functions improve WAR, with FAUD and FAUE producing gains of 1.72\% and 1.07\%. This result further indicates that frame-level predictions describing local facial movements and affective states provide complementary information for recognizing dynamic facial expressions.

In comparison, the three video-level support functions provide limited benefits. FPLE and FDA reduce the average WAR by 0.03\% and 0.39\%, respectively, while FAPR produces only a marginal gain of 0.02\%. Jointly incorporating all seven support functions also decreases the average WAR by 0.04\%. This observation remains consistent with the explanation in the main paper: a binary pain state does not indicate a specific dynamic facial expression, while long-term depression and personality states may contain various short-term facial behaviours and therefore provide limited direct guidance for DFER. Jointly incorporating all seven support functions changes the average WAR by only 0.01\% but requires inference from every support model, providing no meaningful benefit over the most effective individual support.

\noindent\textbf{Enhancement of the second-best DFER model.} The SVFAP-B results provide further evidence that the benefits of frame-level predictions depend partly on the selected DFER model. SFER remains the most effective support function and improves the average WAR by 1.06\%, while FAUE provides a smaller gain of 0.40\%. Their improvements are again most evident on MAFW, where SFER and FAUE improve WAR by 3.26\% and 1.69\%, respectively. Although different DFER models exploit the support predictions differently, SFER consistently benefits both FRU-Adapter-B and SVFAP-B.

The video-level support functions remain ineffective for SVFAP-B: FPLE, FDA, and FAPR reduce the average WAR by 0.30\%, 0.37\%, and 0.55\%, respectively. Combining all support functions yields a gain of 0.95\%, which remains lower than the 1.06\% gain obtained using SFER alone. Considering the additional computation required by multiple support models, these results further support using the most effective single support function for DFER.

\subsubsection{FPLE enhancement results}

\begin{table*}[htbp]
\centering
\Huge
\setlength{\doublerulesep}{0.3em}
\renewcommand{\arraystretch}{1.4}
\caption{Class-specific and mean accuracy and F1-score results (\%) of the baseline and SMTE-enhanced binary FPLE models on BioVid and MIntPAIN. Results are reported for the best-performing MAE-DFER model and the second-best SVFAP-B model. FAUD, FAUE, SFER, VAIE, DFER, FDA, and FAPR denote the support functions incorporated through SMTE, while All denotes their joint use. Values in parentheses indicate the relative metric changes compared with the corresponding FPLE baseline, with positive changes highlighted in bold.}
\resizebox{1.0\linewidth}{!}{
 \\ \hline
  \Xhline{4.5px}
\end{tabular}
}
\label{tab:pain_sec_cls}
\end{table*}

\noindent\textbf{Enhancement of the best-performing FPLE model.} Table \ref{tab:pain_sec_cls} extends the SMTE results reported in the main paper by presenting the class-specific effects of different support functions on MAE-DFER. All four frame-level functions improve its average accuracy across BioVid and MIntPAIN. FAUD provides the largest relative performance gain of 2.44\%, while FAUE, SFER, and VAIE produce comparable gains of 1.81\%, 1.76\%, and 1.79\%, respectively. These results indicate that AU occurrence, AU intensity, and facial affect predictions provide complementary facial information for pain-level estimation. Among the video-level functions, DFER also improves the average accuracy by 2.11\%, supporting the observation in the main paper that dynamic facial expression predictions contain temporal facial changes associated with pain-related behaviours.

In contrast, the long-term predictions provided by FDA and FAPR reduce the average accuracy by 0.08\% and 0.28\%, respectively, showing that these predictions provide little direct support for FPLE. Jointly incorporating all support functions produces a gain of only 1.02\%, which is lower than those obtained using FAUD or DFER alone. Together with the additional computation required to infer all support models, this result again supports selecting the most effective single support function.

\noindent\textbf{Enhancement of the second-best FPLE model.} The results obtained with SVFAP-B show more pronounced improvements, particularly on MIntPAIN. DFER provides the largest average accuracy gain of 8.16\%, followed by FAUE, SFER, and FAUD with gains of 6.41\%, 6.21\%, and 6.13\%, respectively. VAIE also improves the average accuracy by 4.33\%. These gains mainly arise from the Pain class on MIntPAIN, for which the baseline accuracy of 20.51\% increases to 51.28\% with DFER and to above 40\% with FAUD, FAUE, and SFER. The magnitude of these improvements may partly reflect the substantial room for improvement left by the low baseline result on this class. Nevertheless, the consistent benefits from facial behaviour and expression predictions further demonstrate that these functions contain informative complementary cues for FPLE.

\begin{table*}[htbp]
\centering
\setlength{\doublerulesep}{0.3em}
\renewcommand{\arraystretch}{1.4}
\caption{RMSE, mean absolute error (MAE), Pearson correlation coefficient (PCC), and concordance correlation coefficient (CCC) results of the baseline and SMTE-enhanced FDA models on AVEC 2014. Results are reported for the MGNN (MFN) model used in the main paper and the second-best MAE-DFER-B model selected by jointly considering overall performance, number of parameters, computation, and inference speed. FAUD, FAUE, SFER, VAIE, DFER, FPLE, and FAPR denote the support functions incorporated through SMTE, while All denotes their joint use. Values in parentheses indicate the relative metric changes compared with the corresponding FDA baseline. Improvements, i.e., reductions in RMSE or MAE and increases in PCC or CCC, are highlighted in bold.}
\resizebox{0.9\linewidth}{!}{
\begin{tabular}{c|cccc|c|cccc}
\Xhline{1.2px}
\rowcolor{gray!15} 
\cellcolor{gray!15} &
  \multicolumn{4}{c|}{\cellcolor{gray!15}\textbf{AVEC 2014}} &
  \cellcolor{gray!15} &
  \multicolumn{4}{c}{\cellcolor{gray!15}\textbf{AVEC 2014}} \\ \hhline{~||----||~||----} 
\rowcolor{gray!15} 
\multirow{-2}{*}{\cellcolor{gray!15}\textbf{Methods}} &
  RMSE↓ &
  MAE↓ &
  PCC↑ &
  CCC↑ &
  \multirow{-2}{*}{\cellcolor{gray!15}\textbf{Methods}} &
  RMSE↓ &
  MAE↓ &
  PCC↑ &
  CCC↑ \\ \hhline{==========}
MGNN (MFN) &
  6.936 &
  5.279 &
  0.797 &
  0.778 &
  MAE-DFER-B &
  7.533 &
  5.678 &
  0.760 &
  0.739 \\ \hline
\begin{tabular}[c]{@{}c@{}}MGNN (MFN)\\ (+FAUD)\end{tabular} &
  \begin{tabular}[c]{@{}c@{}}\cellcolor{gray!30}6.896\\ \cellcolor{gray!30}\textbf{(+0.57\%)}\end{tabular} &
  \begin{tabular}[c]{@{}c@{}}5.301\\ (-0.42\%)\end{tabular} &
  \begin{tabular}[c]{@{}c@{}}\cellcolor{gray!30}0.799\\ \cellcolor{gray!30}\textbf{(+0.26\%)}\end{tabular} &
  \begin{tabular}[c]{@{}c@{}}\cellcolor{gray!30}0.787\\ \cellcolor{gray!30}\textbf{(+1.22\%)}\end{tabular} &
  \begin{tabular}[c]{@{}c@{}}MAE-DFER-B\\ (+FAUD)\end{tabular} &
  \begin{tabular}[c]{@{}c@{}}\cellcolor{gray!30}7.500\\ \cellcolor{gray!30}\textbf{(+0.44\%)}\end{tabular} &
  \begin{tabular}[c]{@{}c@{}}\cellcolor{gray!30}5.663\\ \cellcolor{gray!30}\textbf{(+0.26\%)}\end{tabular} &
  \begin{tabular}[c]{@{}c@{}}0.760\\ (+0.00\%)\end{tabular} &
  \begin{tabular}[c]{@{}c@{}}\cellcolor{gray!30}0.741\\ 
  \cellcolor{gray!30}\textbf{(+0.29\%)}\end{tabular} \\
  \hline
\begin{tabular}[c]{@{}c@{}}MGNN (MFN)\\ (+FAUE)\end{tabular} &
  \begin{tabular}[c]{@{}c@{}}\cellcolor{gray!30}6.919\\ \cellcolor{gray!30}\textbf{(+0.24\%)}\end{tabular} &
  \begin{tabular}[c]{@{}c@{}}5.304\\ (-0.48\%)\end{tabular} &
  \begin{tabular}[c]{@{}c@{}}\cellcolor{gray!30}0.798\\ \cellcolor{gray!30}\textbf{(+0.13\%)}\end{tabular} &
  \begin{tabular}[c]{@{}c@{}}\cellcolor{gray!30}0.786\\ \cellcolor{gray!30}\textbf{(+1.08\%)}\end{tabular} &
  \begin{tabular}[c]{@{}c@{}}MAE-DFER-B\\ (+FAUE)\end{tabular} &
  \begin{tabular}[c]{@{}c@{}}\cellcolor{gray!30}7.518\\ 
  \cellcolor{gray!30}\textbf{(+0.20\%)}\end{tabular} &
  \begin{tabular}[c]{@{}c@{}}\cellcolor{gray!30}5.673\\ \cellcolor{gray!30}\textbf{(+0.08\%)}\end{tabular} &
  \begin{tabular}[c]{@{}c@{}}0.758\\ (-0.22\%)\end{tabular} &
  \begin{tabular}[c]{@{}c@{}}0.739\\ (+0.00\%)\end{tabular} \\
  \hline
\begin{tabular}[c]{@{}c@{}}MGNN (MFN)\\ (+SFER)\end{tabular} &
  \begin{tabular}[c]{@{}c@{}}\cellcolor{gray!30}6.902\\ \cellcolor{gray!30}\textbf{(+0.49\%)}\end{tabular} &
  \begin{tabular}[c]{@{}c@{}}5.301\\ (+0.41\%)\end{tabular} &
  \begin{tabular}[c]{@{}c@{}}\cellcolor{gray!30}0.799\\ \cellcolor{gray!30}\textbf{(+0.26\%)}\end{tabular} &
  \begin{tabular}[c]{@{}c@{}}\cellcolor{gray!30}0.787\\ \cellcolor{gray!30}\textbf{(+1.22\%)}\end{tabular} &
  \begin{tabular}[c]{@{}c@{}}MAE-DFER-B\\ (+SFER)\end{tabular} &
  \begin{tabular}[c]{@{}c@{}}\cellcolor{gray!30}7.502\\ 
  \cellcolor{gray!30}\textbf{(+0.41\%)}\end{tabular} &
  \begin{tabular}[c]{@{}c@{}}\cellcolor{gray!30}5.653\\ \cellcolor{gray!30}\textbf{(+0.44\%)}\end{tabular} &
  \begin{tabular}[c]{@{}c@{}}\cellcolor{gray!30}0.761\\ 
  \cellcolor{gray!30}\textbf{(+0.15\%)}\end{tabular} &
  \begin{tabular}[c]{@{}c@{}}\cellcolor{gray!30}0.744\\ 
  \cellcolor{gray!30}\textbf{(+0.61\%)}\end{tabular} \\
  \hline
\begin{tabular}[c]{@{}c@{}}MGNN (MFN)\\ (+VAIE)\end{tabular} &
  \begin{tabular}[c]{@{}c@{}}\cellcolor{gray!30}6.905\\ \cellcolor{gray!30}\textbf{(+0.45\%)}\end{tabular} &
  \begin{tabular}[c]{@{}c@{}}5.287\\ (-0.16\%)\end{tabular} &
  \begin{tabular}[c]{@{}c@{}}\cellcolor{gray!30}0.799\\ \cellcolor{gray!30}\textbf{(+0.26\%)}\end{tabular} &
  \begin{tabular}[c]{@{}c@{}}\cellcolor{gray!30}0.787\\ \cellcolor{gray!30}\textbf{(+1.22\%)}\end{tabular} &
  \begin{tabular}[c]{@{}c@{}}MAE-DFER-B\\ (+VAIE)\end{tabular} &
  \begin{tabular}[c]{@{}c@{}}\cellcolor{gray!30}7.515\\ \cellcolor{gray!30}\textbf{(+0.24\%)}\end{tabular} &
  \begin{tabular}[c]{@{}c@{}}\cellcolor{gray!30}5.660\\ \cellcolor{gray!30}\textbf{(+0.32\%)}\end{tabular} &
  \begin{tabular}[c]{@{}c@{}}\cellcolor{gray!30}0.761\\ 
  \cellcolor{gray!30}\textbf{(+0.14\%)}\end{tabular} &
  \begin{tabular}[c]{@{}c@{}}\cellcolor{gray!30}0.740\\ 
  \cellcolor{gray!30}\textbf{(+0.13\%)}\end{tabular} \\
  \hline
\begin{tabular}[c]{@{}c@{}}MGNN (MFN)\\ (+DFER)\end{tabular} &
  \begin{tabular}[c]{@{}c@{}}\cellcolor{gray!30}6.926\\
  \cellcolor{gray!30}\textbf{(+0.14\%)}\end{tabular} &
  \begin{tabular}[c]{@{}c@{}}5.323\\ (-0.83\%)\end{tabular} &
  \begin{tabular}[c]{@{}c@{}}0.796\\ (-0.13\%)\end{tabular} &
  \begin{tabular}[c]{@{}c@{}}\cellcolor{gray!30}0.783\\
  \cellcolor{gray!30}\textbf{(+0.68\%)}\end{tabular} &
  \begin{tabular}[c]{@{}c@{}}MAE-DFER-B\\ (+DFER)\end{tabular} &
  \begin{tabular}[c]{@{}c@{}}\cellcolor{gray!30}7.495\\
  \cellcolor{gray!30}\textbf{(+0.50\%)}\end{tabular} &
  \begin{tabular}[c]{@{}c@{}}\cellcolor{gray!30}5.647\\
  \cellcolor{gray!30}\textbf{(+0.55\%)}\end{tabular} &
  \begin{tabular}[c]{@{}c@{}}\cellcolor{gray!30}0.762\\
  \cellcolor{gray!30}\textbf{(+0.28\%)}\end{tabular} &
  \begin{tabular}[c]{@{}c@{}}\cellcolor{gray!30}0.743\\ 
  \cellcolor{gray!30}\textbf{(+0.54\%)}\end{tabular} \\ 
  \hline
\begin{tabular}[c]{@{}c@{}}MGNN (MFN)\\ (+FPLE)\end{tabular} &
  \begin{tabular}[c]{@{}c@{}}\cellcolor{gray!30}6.910\\
  \cellcolor{gray!30}\textbf{(+0.37\%)}\end{tabular} &
  \begin{tabular}[c]{@{}c@{}}5.328\\ (-0.92\%)\end{tabular} &
  \begin{tabular}[c]{@{}c@{}}0.796\\ (-0.13\%)\end{tabular} &
  \begin{tabular}[c]{@{}c@{}}\cellcolor{gray!30}0.783\\
  \cellcolor{gray!30}\textbf{(+0.68\%)}\end{tabular} &
  \begin{tabular}[c]{@{}c@{}}MAE-DFER-B\\ (+FPLE)\end{tabular} &
  \begin{tabular}[c]{@{}c@{}}\cellcolor{gray!30}7.525\\ 
  \cellcolor{gray!30}\textbf{(+0.10\%)}\end{tabular} &
  \begin{tabular}[c]{@{}c@{}}\cellcolor{gray!30}5.664\\
  \cellcolor{gray!30}\textbf{(+0.24\%)}\end{tabular} &
  \begin{tabular}[c]{@{}c@{}}0.758\\ (-0.28\%)\end{tabular} &
  \begin{tabular}[c]{@{}c@{}}0.738\\ (-0.14\%)\end{tabular} \\
  \hline
\begin{tabular}[c]{@{}c@{}}MGNN (MFN)\\ (+FAPR)\end{tabular} &
  \begin{tabular}[c]{@{}c@{}}\cellcolor{gray!30}6.916\\
  \cellcolor{gray!30}\textbf{(+0.29\%)}\end{tabular} &
  \begin{tabular}[c]{@{}c@{}} 5.297\\ (-0.35\%)\end{tabular}&
  \begin{tabular}[c]{@{}c@{}}0.796\\ (-0.13\%)\end{tabular} &
  \begin{tabular}[c]{@{}c@{}}\cellcolor{gray!30}0.783\\
  \cellcolor{gray!30}\textbf{(+0.68\%)}\end{tabular} &
  \begin{tabular}[c]{@{}c@{}}MAE-DFER-B\\ (+FAPR)\end{tabular} &
  \begin{tabular}[c]{@{}c@{}}\cellcolor{gray!30}7.521\\
  \cellcolor{gray!30}\textbf{(+0.16\%)}\end{tabular} &
  \begin{tabular}[c]{@{}c@{}}\cellcolor{gray!30}5.674\\
  \cellcolor{gray!30}\textbf{(+0.07\%)}\end{tabular} &
  \begin{tabular}[c]{@{}c@{}}0.760\\ (+0.00\%)\end{tabular} &
  \begin{tabular}[c]{@{}c@{}}0.737\\ (-0.29\%)\end{tabular} \\ 
  \hline
\begin{tabular}[c]{@{}c@{}}MGNN (MFN)\\ (+All)\end{tabular} &
  \begin{tabular}[c]{@{}c@{}}6.945\\ (-0.13\%)\end{tabular} &
  \begin{tabular}[c]{@{}c@{}}5.319\\ (-0.76\%)\end{tabular} &
  \begin{tabular}[c]{@{}c@{}}0.795\\ (-0.26\%)\end{tabular} &
  \begin{tabular}[c]{@{}c@{}}\cellcolor{gray!30}0.784\\ \cellcolor{gray!30}\textbf{(+0.81\%)}\end{tabular} &
  \begin{tabular}[c]{@{}c@{}}MAE-DFER-B\\ (+All)\end{tabular} &
  \begin{tabular}[c]{@{}c@{}}\cellcolor{gray!30}7.519\\ \cellcolor{gray!30}\textbf{(+0.19\%)}\end{tabular} &
  \begin{tabular}[c]{@{}c@{}}\cellcolor{gray!30}5.669\\ \cellcolor{gray!30}\textbf{(+0.16\%)}\end{tabular} &
  \begin{tabular}[c]{@{}c@{}}0.760\\ (+0.00\%)\end{tabular} &
  \begin{tabular}[c]{@{}c@{}}0.738\\ (-0.14\%)\end{tabular} \\ 
  \Xhline{1.2px}
\end{tabular}
}
\label{tab:fda_sec}
\end{table*}

FDA and FAPR provide much smaller average gains of 0.88\% and 0.48\%, respectively, and both reduce the mean accuracy on BioVid. Their limited and inconsistent effects across the two datasets agree with the conclusion drawn from MAE-DFER. Moreover, jointly using all support functions improves the average accuracy by 4.24\%, which remains lower than the gain achieved by each individual frame-level support function and DFER.

\subsubsection{FDA enhancement results} 

\noindent\textbf{Enhancement of the best-performing FDA model.} 
Table \ref{tab:fda_sec} expands the FDA results reported in the main paper by presenting the effects of different support functions on MGNN (MFN). According to RMSE, the primary metric used for FDA, all four frame-level functions improve the baseline. FAUD provides the largest relative RMSE reduction of 0.57\%, followed by SFER, VAIE, and FAUE with reductions of 0.49\%, 0.45\%, and 0.24\%, respectively. These functions also improve CCC by 1.08-1.22\%. Although the improvements do not extend to MAE, the consistent RMSE reductions and CCC gains support the conclusion in the main paper that fine-grained facial movements, expressions, and affective states provide complementary information for FDA. 

The three video-level functions also reduce RMSE, with FPLE, FAPR, and DFER producing relative reductions of 0.37\%, 0.29\%, and 0.14\%, respectively. Each function further improves CCC by 0.68\%. These results provide additional evidence for the relationships discussed in the main paper: depression can be reflected by sustained changes in facial expressions and pain-related behaviours and is also associated with apparent personality. Nevertheless, the improvements are not consistent across MAE and PCC. Moreover, jointly using all support functions increases RMSE and MAE by 0.13\% and 0.76\%, respectively, showing that combining multiple support predictions does not provide further benefits. 

\noindent\textbf{Enhancement of the second-best FDA model.} 
MAE-DFER-B was further selected by jointly considering its overall performance, number of parameters, computation, and inference speed. Similar to MGNN (MFN), all individual support functions reduce its RMSE, confirming that the complementary relationships observed in the main paper are not limited to a single FDA model. Among them, DFER produces the most consistent improvement, reducing RMSE and MAE by 0.50\% and 0.55\%, respectively, while increasing PCC and CCC by 0.28\% and 0.54\%. This result further supports the contribution of dynamic facial expression information to FDA.

Frame-level predictions also remain beneficial. SFER reduces RMSE and MAE by 0.41\% and 0.44\% and provides the largest CCC gain of 0.61\%, while FAUD and VAIE improve at least three of the four metrics. In comparison, FPLE and FAPR provide smaller RMSE reductions of 0.10\% and 0.16\% and decrease the correlation metrics, indicating less consistent support. Jointly incorporating all functions reduces RMSE by only 0.19\% and decreases CCC by 0.14\%, remaining inferior to DFER or SFER alone.

\subsubsection{FAPR enhancement results} 

\begin{table*}[htbp]
\centering
\Huge
\setlength{\doublerulesep}{0.3em}
\renewcommand{\arraystretch}{1.4}
\caption{Trait-specific and mean Pearson correlation coefficient (PCC), concordance correlation coefficient (CCC), and mean squared error (MSE) results of the baseline and SMTE-enhanced FAPR models on First Impression. Results are reported for the best-performing MGNN (MFN) model and the second-best MAE-DFER-B model. FAUD, FAUE, SFER, VAIE, DFER, FPLE, and FDA denote the support functions incorporated through SMTE, while All denotes their joint use. Values in parentheses indicate the relative metric changes compared with the corresponding FAPR baseline. Improvements, i.e., increases in PCC or CCC and reductions in MSE, are highlighted in bold.}
\resizebox{1.0\linewidth}{!}{
 \\ 
  \Xhline{4.5px}
\end{tabular}
 }
\label{tab:fapr_sec}
\end{table*}

\noindent\textbf{Enhancement of the best-performing FAPR model.} Table \ref{tab:fapr_sec} expands the FAPR results reported in the main paper by presenting the trait-specific effects of different support functions on MGNN (MFN). Among the four frame-level functions, only SFER improves the mean CCC, producing a marginal relative performance gain of 0.14\%. The other frame-level predictions provide no improvement or slightly reduce the overall results. This observation supports the explanation in the main paper that the best-performing MGNN (MFN) baseline already captures sufficiently rich behavioural representations, leaving limited complementary information for SMTE to exploit.

A similar pattern is observed for the video-level support functions. DFER and FPLE both reduce the mean CCC by 0.28\%, while FDA leaves it unchanged. Jointly incorporating all support functions also provides no improvement in mean PCC or CCC and increases MSE by 9.09\%. Thus, the additional computation required to infer multiple support models is not justified for the best-performing FAPR model.

\noindent\textbf{Enhancement of the second-best FAPR model.} The results obtained with MAE-DFER-B further show that the limited improvements on MGNN (MFN) do not indicate an absence of complementary information from frame-level functions. FAUD, FAUE, and SFER each improve the mean CCC by 0.56\%, while VAIE produces a gain of 0.42\%. Moreover, all four functions consistently improve CCC across the five personality traits. SFER provides the largest mean PCC gain of 0.42\%, while FAUD, SFER, and VAIE each reduce MSE by 7.69\%. These results indicate that predictions describing local facial movements, categorical expressions, and continuous affective states can provide informative complementary cues when the FAPR baseline leaves greater room for improvement.

In contrast, the video-level support functions remain ineffective. DFER and FPLE reduce the mean CCC by 0.42\% and 0.28\%, respectively, while FDA provides no mean CCC improvement. This result is consistent with that obtained using MGNN (MFN) and further supports the conclusion in the main paper that DFER, FPLE, and FDA predictions provide little direct benefit to FAPR. Although jointly using all support functions yields the largest mean CCC gain of 0.70\%, it provides only a further 0.14\% gain over the best individual support functions, without additional improvements in PCC or MSE. Such a limited benefit does not compensate for the substantially increased computation required by multiple support models.

\subsection{Statistical significance analysis}

\begin{figure*}[htbp]
    \centering
    \includegraphics[width=1.0\linewidth]{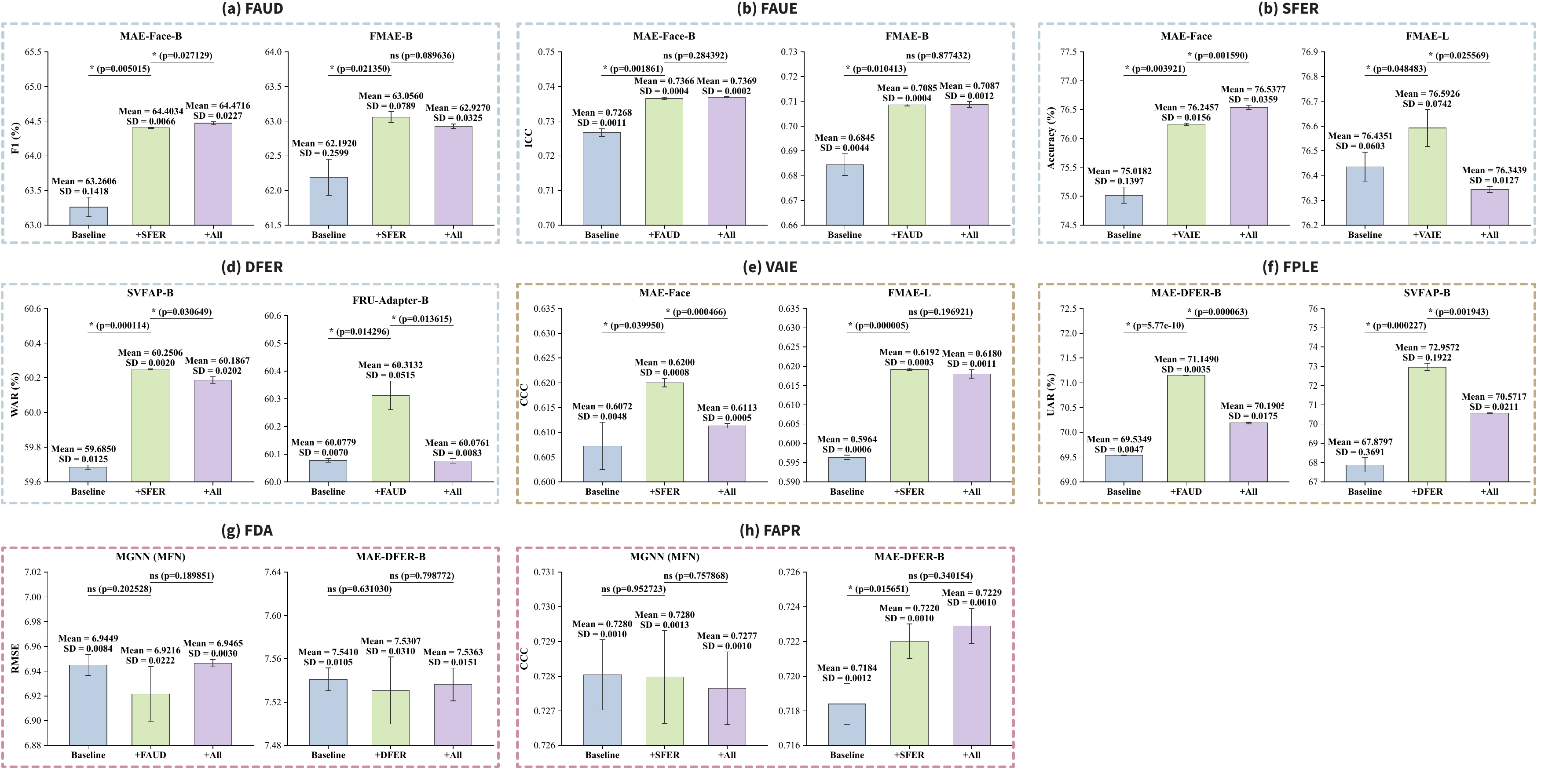}
    \caption{Statistical significance analysis of the baseline and SMTE-enhanced models across the eight re-developed OpenAHBU functions. For each model, Baseline denotes the original model without SMTE, +[function] denotes SMTE using predictions from the most effective single support function indicated on the horizontal axis, and +All denotes SMTE jointly using all support functions evaluated for that target. Bars and error bars represent the mean performance and sample standard deviation (SD) across three independent experiments, respectively. The metrics are F1 score for FAUD, ICC(3,1) for FAUE, accuracy for SFER, WAR for DFER, CCC for VAIE and FAPR, class-averaged accuracy (UAR) for binary FPLE, and RMSE for FDA. Higher values indicate better performance except for FDA. Brackets indicate the compared settings, with p-values obtained using a two-sided Welch’s t-test. An asterisk ($*$) denotes $p < 0.05$, while $ns$ denotes no statistically significant difference ($p \geq 0.05$).}
    \label{fig:sig}
\end{figure*}

To further examine the statistical significance of the SMTE results in Tables \ref{tab:faud_sec}–\ref{tab:fapr_sec}, we select two representative models for each function following the selection criteria described in the preceding subsections. For each model, three independent experiments are conducted under three settings: (1) the original baseline without SMTE; (2) SMTE incorporating predictions from the most effective single support function; and (3) SMTE jointly incorporating predictions from all evaluated support functions. Fig. \ref{fig:sig} reports the mean performance and sample standard deviation across the three experiments. We apply a two-sided Welch’s t-test to compare the baseline with the single-support setting, and the single-support setting with the all-support setting. A difference is considered statistically significant when $p < 0.05$, while ns indicates that no statistically significant difference is detected.

Compared with the corresponding baselines, SMTE using the most effective single support function produces statistically significant improvements in 13 of the 16 evaluated model configurations, including both models for FAUD, FAUE, SFER, DFER, VAIE, and FPLE. These results provide further evidence that predictions from related AFBU functions contain informative complementary facial cues that can enhance the target models. The exceptions are the two FDA models and the MGNN (MFN) model for FAPR. For FDA, the mean RMSE reductions remain small relative to the variation across repeated experiments and do not reach statistical significance for either model. For FAPR, SFER predictions provide no statistically significant improvement for MGNN (MFN), but significantly improve MAE-DFER-B. This model-dependent difference is consistent with the explanation in the main paper: the stronger MGNN (MFN) baseline may already capture most of the relevant facial behavioural information, leaving limited complementary cues for further enhancement.

Compared with using the most effective single support function, jointly incorporating all support functions produces no statistically significant difference in eight configurations and significantly reduces performance in six configurations. The decreases occur for FMAE-L on SFER, MAE-Face on VAIE, and both evaluated models on DFER and FPLE, suggesting that additional support predictions may introduce redundant or less relevant information that limits the benefit of the most informative support function. Nevertheless, MAE-Face-B on FAUD and MAE-Face on SFER remain exceptions, with all support functions providing statistically significant additional improvements. Their relative performance gains over the single-support setting are approximately 0.11\% and 0.38\%, respectively, while requiring inference from all three support models. Therefore, jointly incorporating multiple support functions does not provide consistent additional improvements across target models, despite increasing the end-to-end computational cost. These results further support selecting the most effective single support function to achieve a more favourable balance between predictive performance and computation.

\section{OpenAHBU GUI Visualisation}
This section provides further details of the stand-alone OpenAHBU GUI introduced in Sec. 3.1 of the main paper, including function and input-mode selection, prediction visualisation, and structured result export.

\begin{figure*}[htbp]
\centering
\begin{subfigure}[t]{1.0\textwidth}
    \centering
    \includegraphics[width=1.0\linewidth]{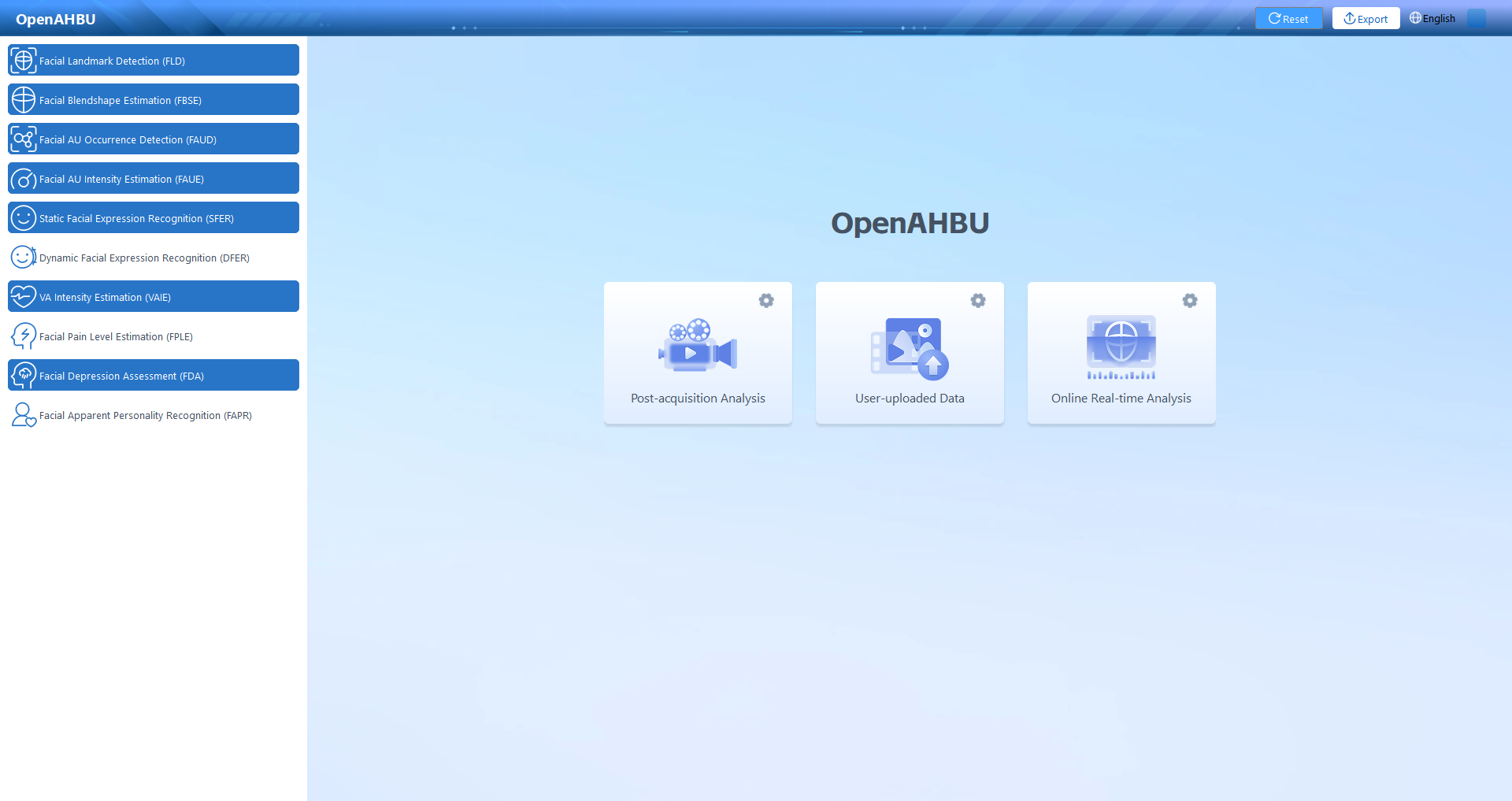}
    \caption{Opening page of the stand-alone OpenAHBU GUI. The left panel provides function selection, while the central area provides Post-acquisition Analysis, User-uploaded Data, and Online Real-time Analysis modes.}
    \label{fig:gui_1}
\end{subfigure}

\vspace{0.5cm}

\begin{subfigure}[t]{1.0\textwidth}
    \centering
    \includegraphics[width=1.0\linewidth]{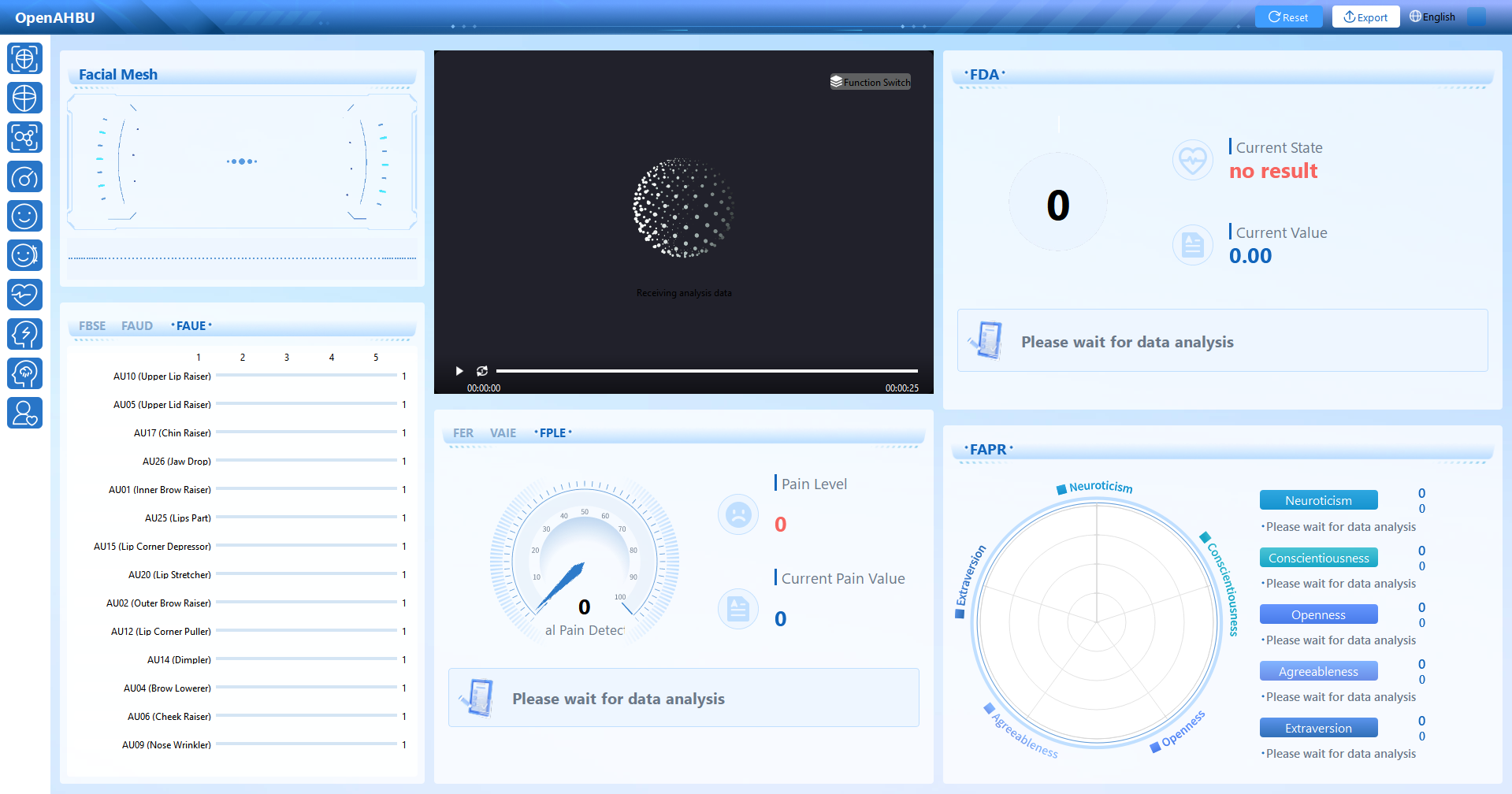}
    \caption{OpenAHBU analysis page while waiting for the prediction results of the user-selected functions.}
    \label{fig:gui_2}
\end{subfigure}

\caption{Opening page and analysis page while waiting for the prediction results of the stand-alone OpenAHBU GUI.}
\label{fig:gui_1_2}
\end{figure*}

\begin{figure*}[htbp]
\centering
\begin{subfigure}[t]{1.0\textwidth}
    \centering
    \includegraphics[width=1.0\linewidth]{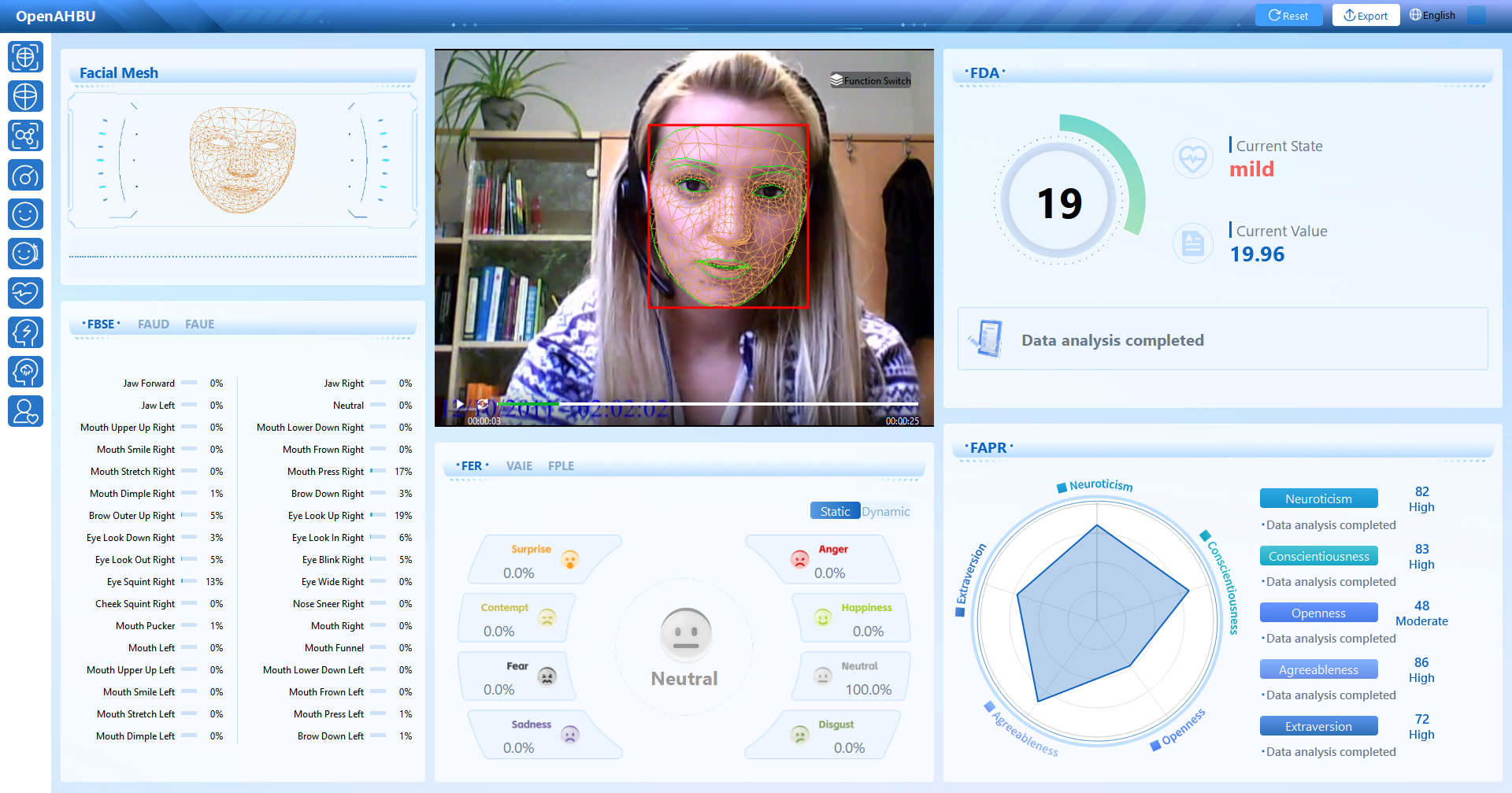}
    \caption{OpenAHBU GUI visualisation of FLD, FBSE, SFER, FDA, and FAPR prediction results.}
    \label{fig:gui_3}
\end{subfigure}

\vspace{0.5cm}

\begin{subfigure}[t]{1.0\textwidth}
    \centering
    \includegraphics[width=1.0\linewidth]{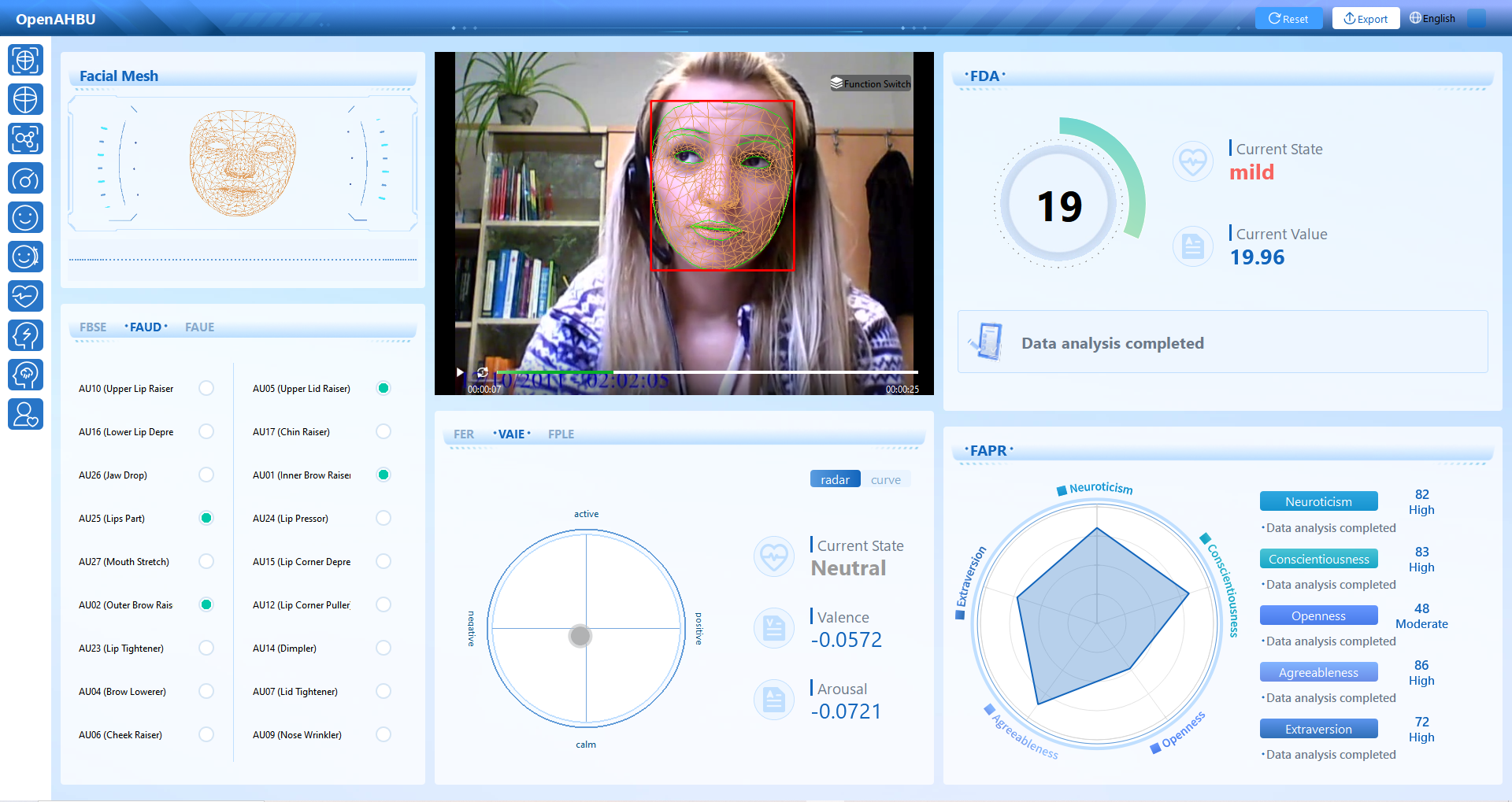}
    \caption{OpenAHBU GUI visualisation of FLD, FAUD, VAIE, FDA, and FAPR prediction results, with VAIE displayed in the radar view.}
    \label{fig:gui_4}
\end{subfigure}

\caption{OpenAHBU GUI visualisation.}
\label{fig:gui_3_4}
\end{figure*}

\begin{figure*}[htbp]
\centering
\begin{subfigure}[t]{1.0\textwidth}
    \centering
    \includegraphics[width=1.0\linewidth]{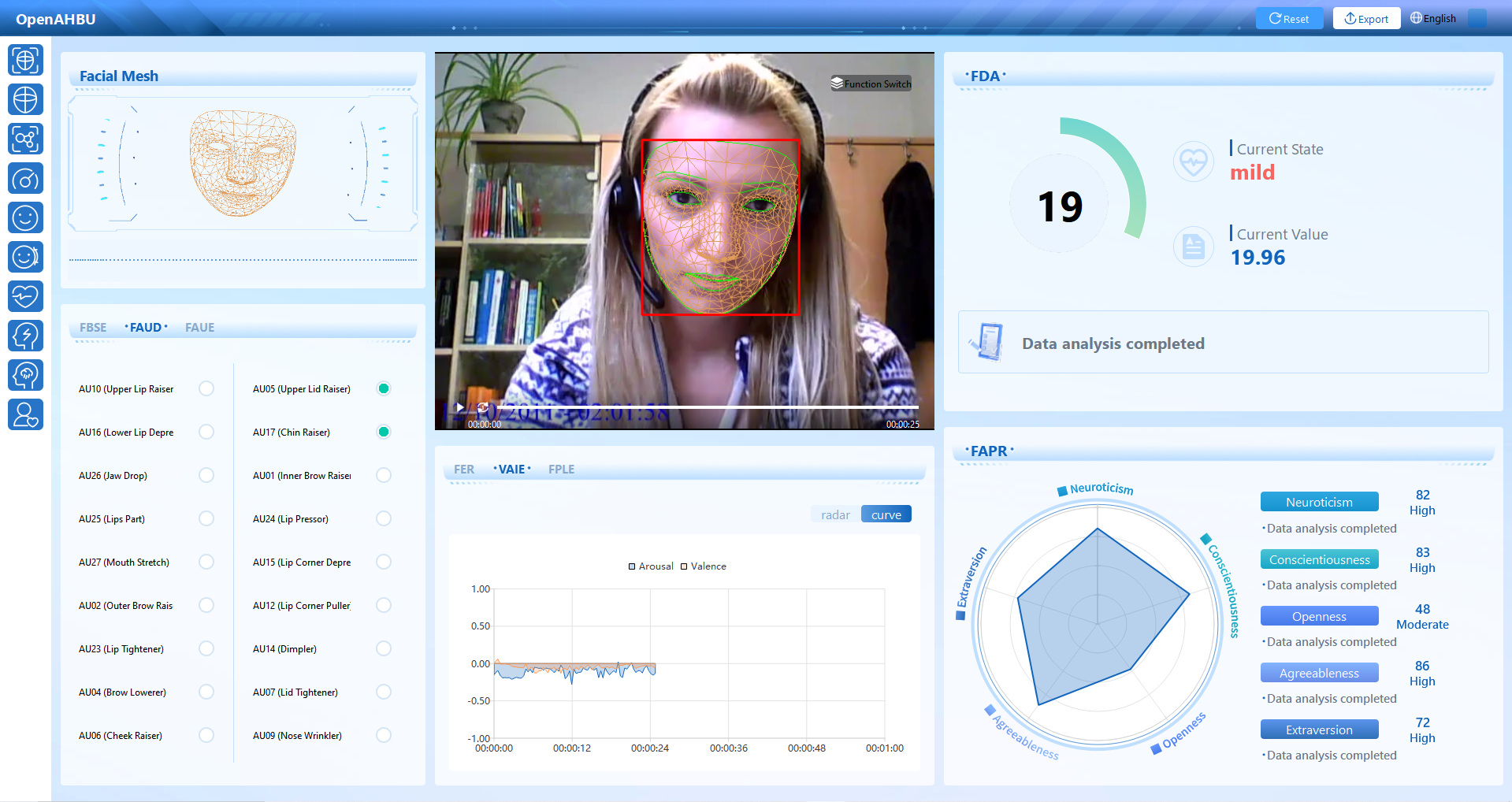}
    \caption{OpenAHBU GUI visualisation of FLD, FAUD, VAIE, FDA, and FAPR prediction results, with the temporal valence and arousal predictions displayed in the curve view.}
    \label{fig:gui_5}
\end{subfigure}

\vspace{0.5cm}

\begin{subfigure}[t]{1.0\textwidth}
    \centering
    \includegraphics[width=1.0\linewidth]{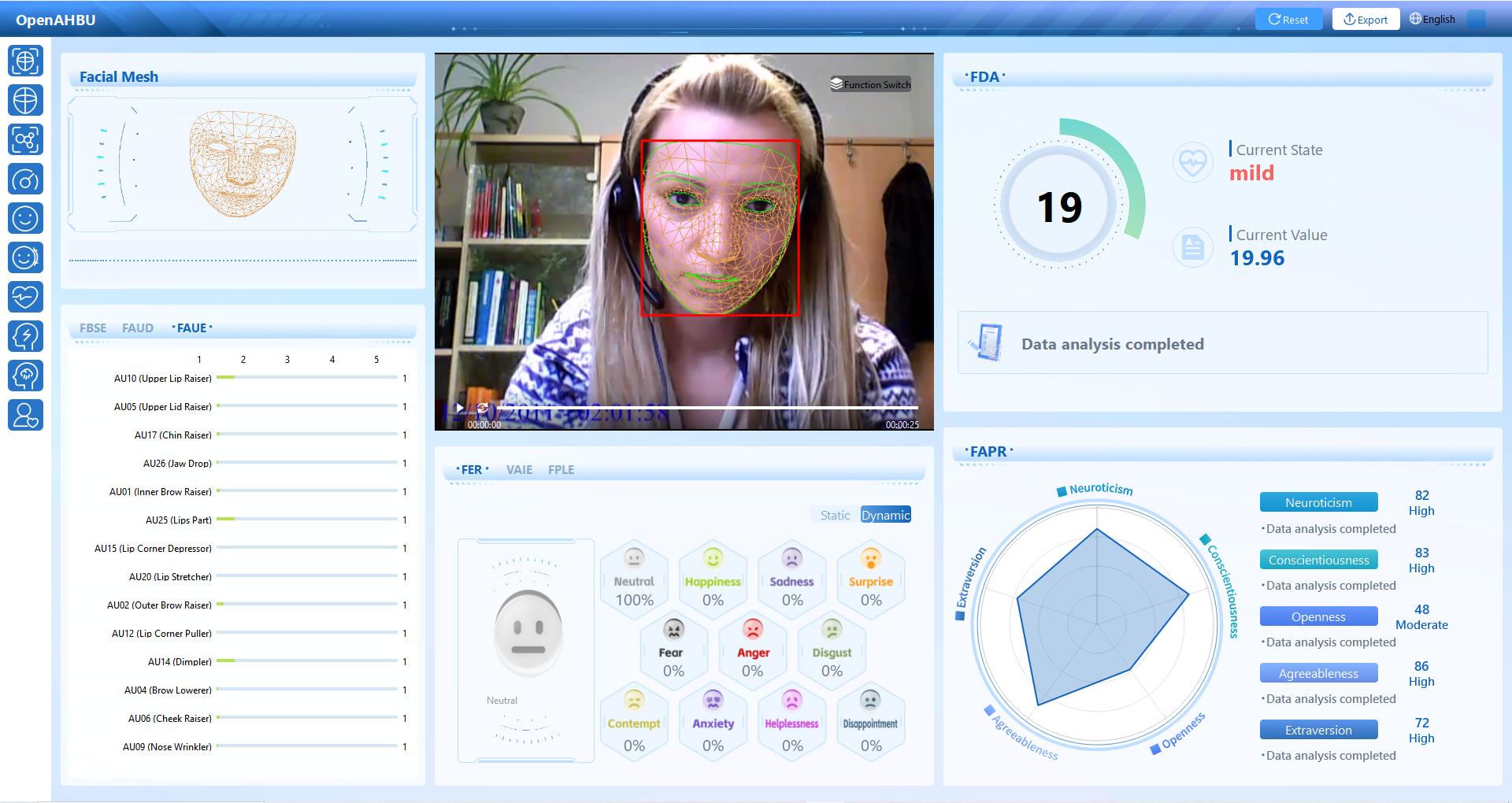}
    \caption{OpenAHBU GUI visualisation of FLD, FAUE, DFER, FDA, and FAPR prediction results.}
    \label{fig:gui_6}
\end{subfigure}

\caption{OpenAHBU GUI visualisation.}
\label{fig:gui_5_6}
\end{figure*}

\begin{figure*}[htbp]
\centering
\begin{subfigure}[t]{1.0\textwidth}
    \centering
    \includegraphics[width=1.0\linewidth]{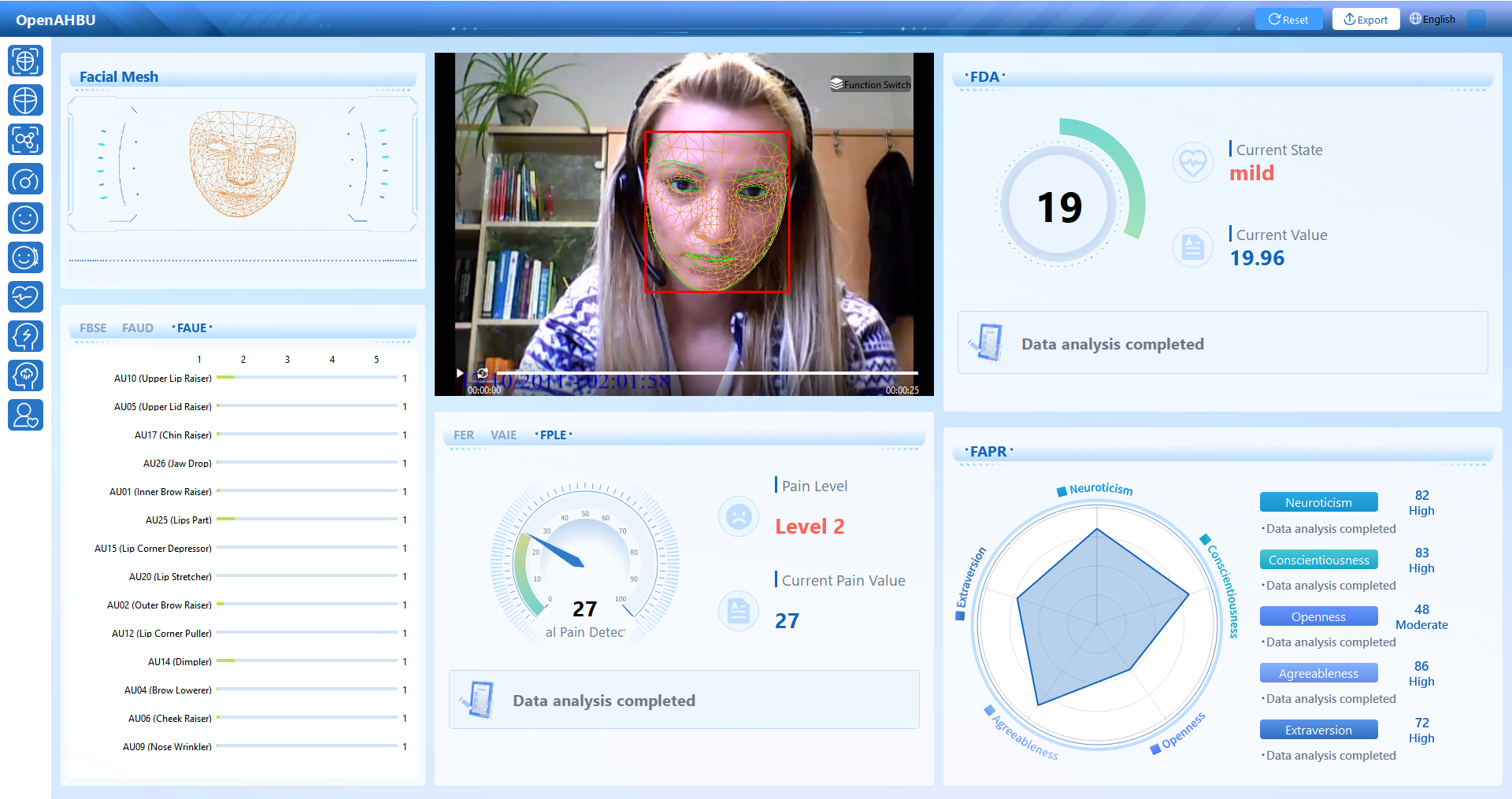}
    \caption{OpenAHBU GUI visualisation of FLD, FAUE, FPLE, FDA, and FAPR prediction results.}
    \label{fig:gui_7}
\end{subfigure}

\vspace{0.5cm}

\begin{subfigure}[t]{1.0\textwidth}
    \centering
    \includegraphics[width=1.0\linewidth]{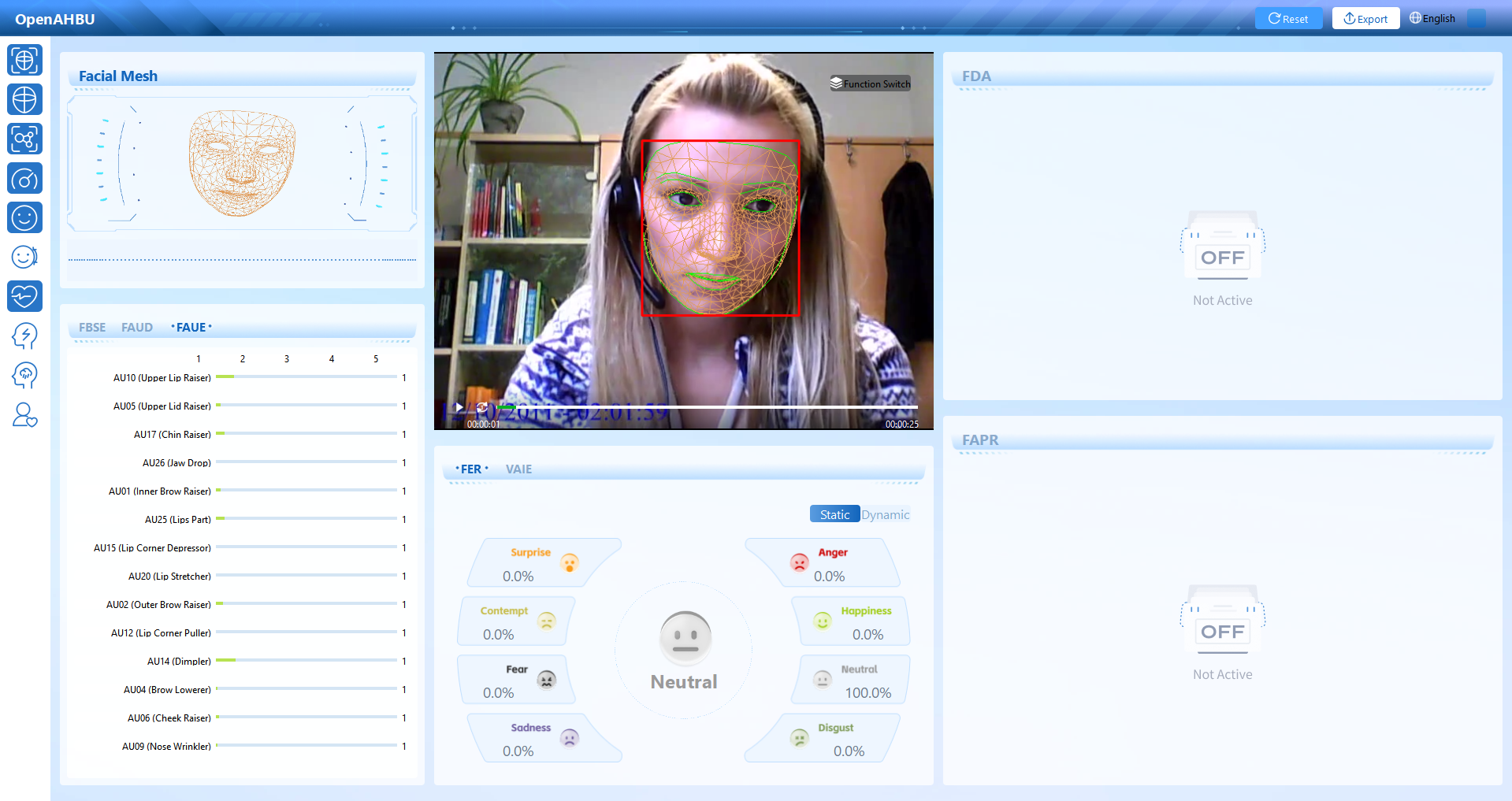}
    \caption{OpenAHBU GUI state when FDA and FAPR are not selected. The corresponding function panels are marked as inactive.}
    \label{fig:gui_8}
\end{subfigure}

\caption{OpenAHBU GUI visualisation.}
\label{fig:gui_7_8}
\end{figure*}

\begin{figure*}[htbp]
\centering
\begin{subfigure}[t]{1.0\textwidth}
    \centering
    \includegraphics[width=1.0\linewidth]{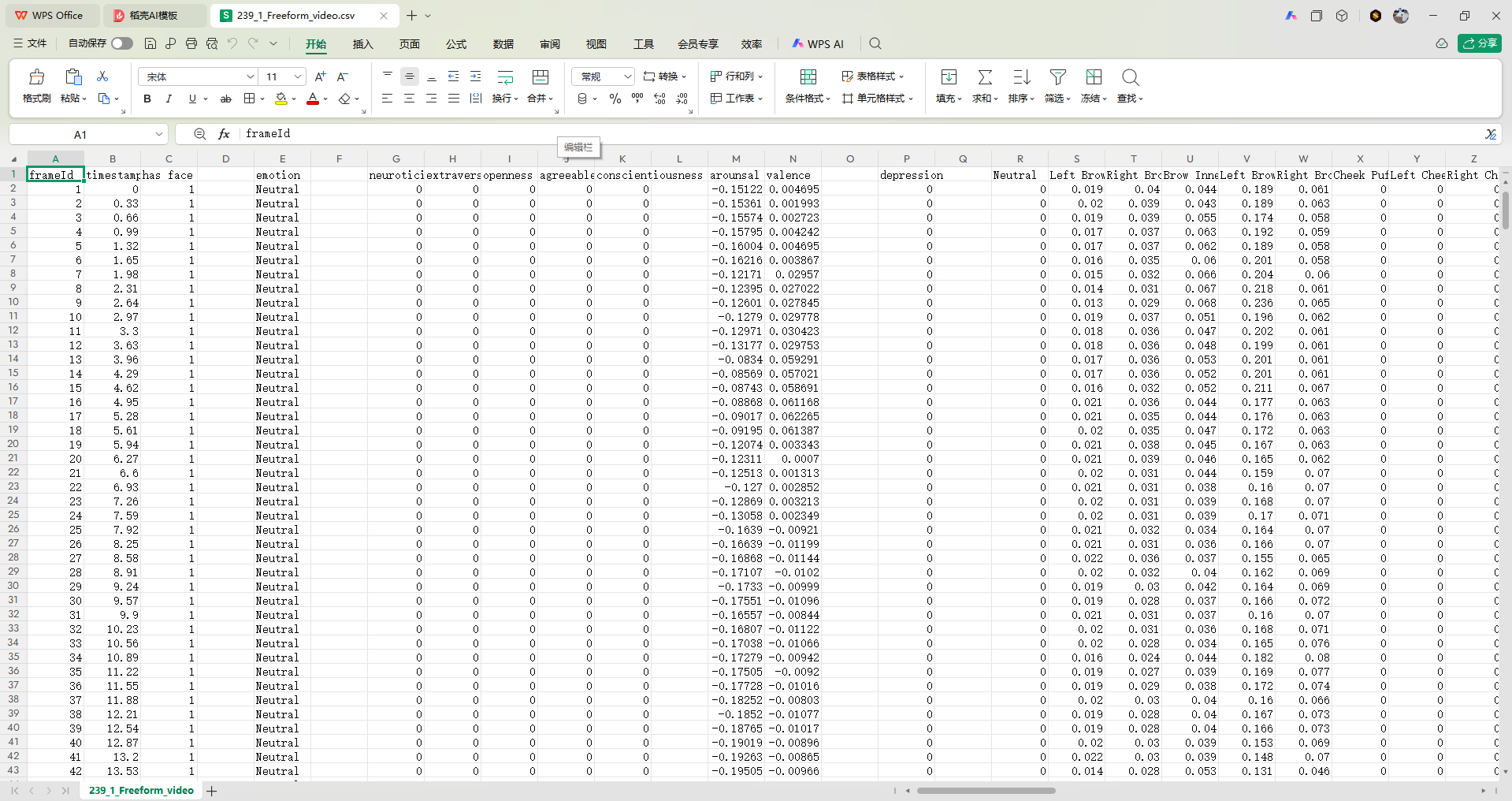}
    \caption{Frame-wise prediction records at the beginning of the input video.}
    \label{fig:report1}
\end{subfigure}

\vspace{0.5cm}

\begin{subfigure}[t]{1.0\textwidth}
    \centering
    \includegraphics[width=1.0\linewidth]{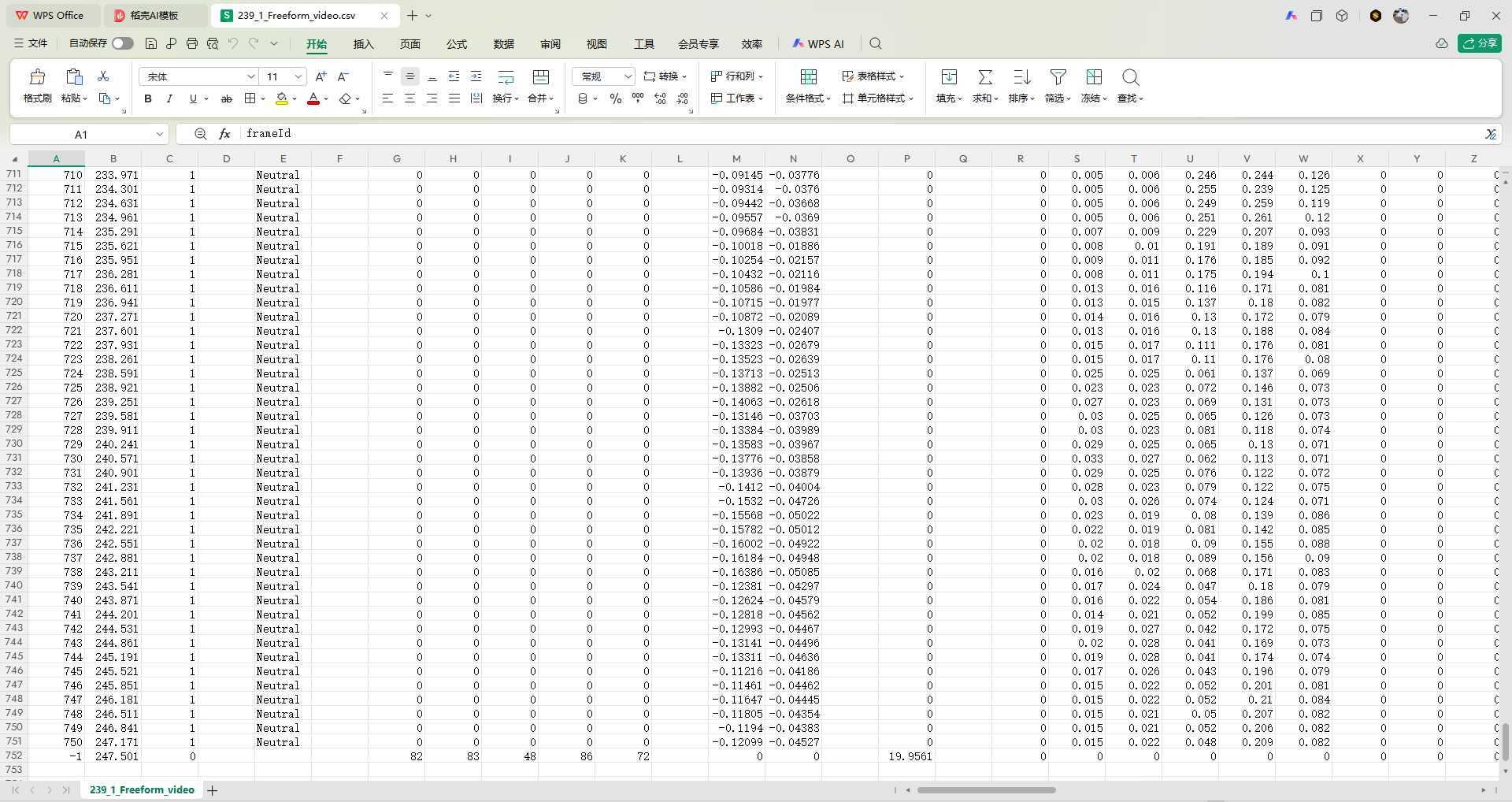}
    \caption{Frame-wise prediction records at the end of the input video and the completed video-level outputs.}
    \label{fig:report2}
\end{subfigure}

\caption{Example of the structured CSV file exported by OpenAHBU. The exported file uses the same base filename as the input video.}
\label{fig:report}
\end{figure*}

\textbf{Function and input mode selection.} As shown in Fig. \ref{fig:gui_1}, the opening page consists of a function-selection panel on the left and three input modes in the central area. Users first activate the functions required by their applications, with the selected functions highlighted in blue. The current GUI integrates the ten functions available for direct deployment, including FLD, FBSE, FAUD, FAUE, SFER, DFER, VAIE, FPLE, FDA, and FAPR. After function selection, users can choose one of the three input modes described in Sec. 3.1 of the main paper: (1) Post-acquisition Analysis records video using a connected camera and subsequently performs offline analysis; (2) User-uploaded Data analyses an existing image or video selected from the local disk; and (3) Online Real-time Analysis directly processes the live stream captured by a connected camera.

\textbf{Analysis process.} 
After the input data and analysis mode are determined, OpenAHBU enters the analysis page shown in Fig. \ref{fig:gui_2}. Before the predictions are returned, the input image or video is displayed in the central area, while the corresponding function panels remain in the waiting state. Once the analysis is completed, the detected face, facial mesh, and function-specific predictions are synchronously displayed with the input data. Multiple functions sharing the same display region can be accessed through the function tabs, allowing users to switch between different prediction results without restarting the analysis.

\textbf{Short-term observable facial behaviour visualisation.}
As shown in Figs. \ref{fig:gui_3}, \ref{fig:gui_4}, \ref{fig:gui_5}, \ref{fig:gui_6}, and \ref{fig:gui_7}, FLD displays the detected facial mesh both in an independent panel and as an overlay on the input face. FBSE reports the estimated percentages of individual facial blendshapes, as illustrated in Fig. \ref{fig:gui_3}. FAUD indicates the occurrence state of each AU using binary status markers (Fig. \ref{fig:gui_4}), while FAUE displays the estimated intensity of each AU on a five-level scale (Fig. \ref{fig:gui_6}). For facial expression analysis, SFER presents the predicted static facial expression together with the scores of all expression categories (Fig. \ref{fig:gui_3}), whereas DFER provides the corresponding video-level dynamic facial expression prediction and category scores (Fig. \ref{fig:gui_7}).

\textbf{Short-term and long-term human state visualisation.}
VAIE provides both radar and curve views. The radar view displays the current valence-arousal state and the corresponding intensity values, as shown in Fig. \ref{fig:gui_4}, while the curve view presents the temporal trajectories of valence and arousal over the input video (Fig. \ref{fig:gui_5}). FPLE displays the estimated pain level and its current prediction value using a gauge-based interface (Fig. \ref{fig:gui_7}). For long-term facial behaviour-related human psychological state understanding, FDA reports the predicted depression severity state and score, while FAPR visualises the predicted neuroticism, conscientiousness, openness, agreeableness, and extraversion scores using a radar chart and the corresponding trait levels (Fig. \ref{fig:gui_3}).

\textbf{Function activation and result export.}
OpenAHBU only performs inference and displays predictions for the functions activated on the opening page. As illustrated in Fig. \ref{fig:gui_8}, when FDA and FAPR are not selected, their display regions are marked as inactive, while the selected functions remain available. After analysis, users can export the predictions using the Export button. As shown in Fig. \ref{fig:report}, the exported structured CSV file uses the same base filename as the input video. Each frame-wise record contains the frame index, timestamp, face-presence state, and the predictions produced by the selected frame-level functions, while the outputs of the video-level functions are stored together with the completed analysis results. This structured output allows the predictions to be directly used for subsequent statistical analysis, storage, or application development.

\ifCLASSOPTIONcaptionsoff
  \newpage
\fi

\bibliographystyle{IEEEtran}

\bibliography{ref}





\end{document}